\documentclass[10pt,conference,a4paper]{IEEEtran}
\IEEEoverridecommandlockouts

\usepackage{cite}
\usepackage{amsmath,amssymb,amsfonts}
\usepackage{algorithmic}
\usepackage{graphicx}
\usepackage{subfigure}
\usepackage[font=scriptsize]{caption}
\usepackage{xcolor}

\usepackage{alphalph}
\usepackage{multirow}
\usepackage{textcomp}
\usepackage{xcolor}
\usepackage[ruled,vlined]{algorithm2e}
\DeclareMathOperator*{\argmin}{Arg min} 
\def\BibTeX{{\rm B\kern-.05em{\sc i\kern-.025em b}\kern-.08em
    T\kern-.1667em\lower.7ex\hbox{E}\kern-.125emX}}
\begin{document}

\title{Unidentified Floating Object detection in maritime environment using dictionary learning\\
}

\author{\IEEEauthorblockN{Darshan Venkatrayappa}
\IEEEauthorblockA{\textit{Centre Borelli, ENS Paris-Saclay} \\
Cachan, France \\
darshan.venkatrayappa@cmla.ens-cachan.fr}
\and
\IEEEauthorblockN{Agn\`es Desolneux}
\IEEEauthorblockA{\textit{Centre Borelli, ENS Paris-Saclay}  \\
Cachan, France \\
agnes.desolneux@math.cnrs.fr}
\and
\IEEEauthorblockN{Jean-Michel Hubert}
\IEEEauthorblockA{\textit{iXblue} \\
Saint-Germain-en-Laye, France \\
jean-michel.hubert@ixblue.com }
\and
\hspace{8cm}\IEEEauthorblockN{Josselin Manceau}
\hspace{8cm}\textit{iXblue}\\
\hspace{8cm}Saint-Germain-en-Laye, France \\
\hspace{8cm}josselin.manceau@ixblue.com 
}
%
\maketitle

\begin{abstract}
Maritime domain is one of the most challenging scenarios for object detection due to the complexity of the observed scene. In this article, we present a new approach to detect unidentified floating objects in the maritime environment. The proposed approach is capable of detecting floating objects without any prior knowledge of their visual appearance, shape or location. The input image from the video stream is denoised using a visual dictionary learned from a K-SVD algorithm. The denoised image is made of self-similar content. Later, we extract the residual image, which is the difference between the original image and the denoised (self-similar) image. Thus, the residual image contains noise and salient structures (objects). These salient structures can be extracted using an a contrario model. We demonstrate the capabilities of our algorithm by testing it on videos exhibiting varying maritime scenarios.
\end{abstract}


\section{Introduction}
Maritime transport via waterways is proven to be an important and integral part of the worldwide economics. Maritime transportation can be used for commerce, recreation or for military purpose by means of tankers, boat, ship, sailboat etc. The economics involved in maritime transportation has increased the threats emanating from pirates. Additionally, a wide variety of objects such as small buoys, cargo ships, tankers and debris (broken ship, cargo boxes etc) in the maritime domain can be an obstacle to small ships or fishing trolleys. These threats need to be detected and tracked in order to access the risk. Detecting and tracking objects in maritime scenario is more challenging compared to objects on land. Reflection from the sun, boat wakes, waves on water surface contribute to generate a highly dynamic background. Weather related issues (heavy rain or fog), gradual and sudden illumination variations (clouds), motion changes (e.g., camera jitter due to winds), and modifications of the background geometry can increase false detection \cite{FOBD_1}. More details about the challenges in the maritime scenario can be found in \cite{FOBD_11}.

\begin{figure*}
\label{Work_flow}
\centering
    \includegraphics[width=1\textwidth]{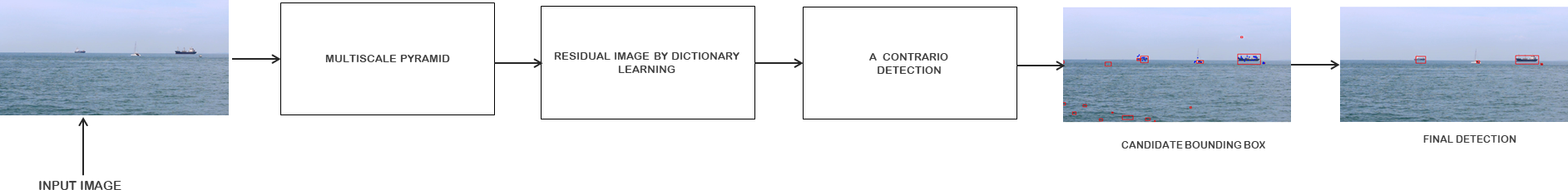}
\caption{Workflow}    
\end{figure*} 

From a century ago until recent past, radar \cite{FOBD_8} and sonar \cite{FOBD_9} have been used to overcome the above mentioned risky scenarios. On the contrary, radar data is sensitive to the variation in the climate, the shape, size, and material of the targets. As a result, radar has to be supplemented by other sensors for automatic floating object detection. Thus, video cameras (infrared and color) are used to complement the ranging devices. The camera used for object detection can be mounted on-board a ship or off-board on the harbor, port, mobile surveillance platform and also to an aircraft or a drone. The authors of \cite{FOBD_3},\cite{FOBD_4} have used thermal cameras to address the above mentioned problems. Since thermal cameras are expensive in comparison to visible cameras, the later can be used as an alternative and also provides surveillance coverage of a larger region as well as complement other available sensors such as radar. This presents a challenge for further improving the performance of visible camera based computer vision algorithms for operating in maritime environments \cite{FOBD_5}. In this paper we address this challenge by proposing a maritime specific salient object detection algorithm. 

This work mainly concentrates on detecting unidentified floating objects in the far sea using camera mounted on a ship. That being said, our algorithm performs well near the shore with a camera on shore.  Unidentified objects are usually random, diverse and rare, like drifting containers, drifting iceberg, drifting cargo boxes, driftwood, debris, etc. and can be encountered anywhere in the ocean. Our work can also detect commonly known floating objects like ships and boats. These floating objects can have large motion (ship, boats) or float with very less motion (wooden logs, debris). It is to be noted that here, we are not concerned with labeling or classifying the object as boat, ship, cargo container, etc. rather, we assume that any floating object is an obstacle and needs to be detected. Our work is more of an early warning system.

In this far sea scenario, the background (sea and sky) have their respective color and texture. The floating objects can be considered as salient and can easily be detected by separating the background from the object.
Our algorithm works on residual image at different scales. Residual image is the difference between the denoised image(self-similar image) and the original image (noisy image). Thus, the self-similar content of the image is removed and the residual image is left with the noise and salient structures. We use K-SVD based dictionary learning \cite{FOBD_7} to achieve denoising. The noise and the salient structures can be separated by statistical tests based on a contrario approach \cite{FOBD_34}.

%

\section{Related Work}

The economic and military importance of the maritime domain has encouraged the research community to propose a plethora of algorithms related to object detection and tracking in the maritime domain. Initial maritime navigation systems made use of different sensors such as radar \cite{FOBD_8}, Sonar \cite{FOBD_9}, Lidar \cite{FOBD_10} and different ranging device. Most of these range scanners fail to detect small objects and have limited range. To tackle these limitations the computer vision community has resorted to camera based object detection and tracking. Several researchers \cite{FOBD_1}, \cite{FOBD_4},\cite{FOBD_9},\cite{FOBD_10},\cite{FOBD_11} have used camera or a combination of camera and ranging device for floating object detection. Different sensors used in maritime scenario and their characterstics can be found in \cite{FOBD_11}.

In \cite{FOBD_1}, the authors propose a three-step approach involving online clustering, background update and noise removal for object detection and to deal with variation of intensity and clutter in the maritime environment. The authors of \cite{FOBD_15} combine the Mixtures of Gaussians (MOG) background subtraction technique with the Visual Attention Map to detect threats in the sea. \cite{FOBD_16} proposes a Bayesian decision framework based hybrid foreground object detection algorithm in the maritime domain. Kristan et. al \cite{FOBD_22} use Gaussian Mixture Model (GMM) to segment water, land and sky regions. The GMM relies on the availability of priors that are pre-computed using training data. The major drawback of this method is the expensive pre-training step. The problem with these methods is that most of them fail with varying background or lightning conditions and the ones that succeed are computationally very expensive. A review of different background subtraction algorithms used in maritime object detection can be found in \cite{FOBD_11}.


Boracchi et.al \cite{FOBD_28} have used sparsity based dictionary learning for detecting anomalies in texture and SEM images depicting a nanofibrous material. Authors in \cite{FOBD_29} use sparse coding with anomaly detection for specular reflectance and shadows removal from color images. The above methods use normal images to learn dictionary unlike the approach used in the proposed paper. Authors of \cite{FOBD_30} and \cite{FOBD_31} use sparsity based dictionary for unusual event detection. Authors of \cite{FOBD_26} use  Robust Principal Components Analysis (RPCA) to detect sailboats in maritime scene from the separated sparse and low-rank matrices. Authors of \cite{FOBD_33} have used RPCA based approach for dolphine detection and tracking. Most of the methods based on low rank and sparse representation are not suitable for real time applications as they are highly complex and require a collection of frames to detect the object \cite{FOBD_32}.

With the advent of deep learning, many new models have been proposed for floating object detection in general and ship detection in particular. The authors of \cite{FOBD_17} train deep learning models on the publically available Singapore Maritime Dataset (SMD) \cite{FOBD_11} to come up with a benchmark for deep learning based object detection in the maritime environment. In \cite{FOBD_18}, the authors have trained the existing deep learning models to detect maritime objects in the long wavelength infrared images. The authors of \cite{FOBD_19} propose a new deep encoder-decoder architecture for obstacle detection by semantic segmentation. \cite{FOBD_20} and \cite{FOBD_21} have also trained deep learning models for ship detection and classification. An evaluation of different deep semantic segmentation networks for object detection in maritime surveillance can be found in \cite{FOBD_25}. Most of the deep learning methods require a large amount of annotated data and in our case data is sparse as we have no prior knowledge about the unidentified floating objects encountered in the ocean.

\section{Workflow}	
The workflow of the proposed floating object detection is shown in Fig.1.

\subsection{Residual image generation by Dictionary learning}
The noise added to the image during image acquisition process can be modeled as an additive Gaussian noise model. This image is denoised using a method based on sparse and redundant representations over trained dictionaries. The K-SVD algorithm is used to obtain a dictionary that describes the image content effectively. The atoms of the learnt dictionary captures the self similar content of the image. Thus, the reconstructed(denoised) image is free from both: noise and the dissimilar content. Image denoising using K-SVD based dictionary learning was proposed by \cite{FOBD_26}. We follow the implementation of \cite{FOBD_7} for K-SVD based image denoising.

\begin{figure*}
\label{Dict_learn}
\centering
\subfigure[Input $y$]{\label{f2a}\includegraphics[height=25mm,width=38mm]{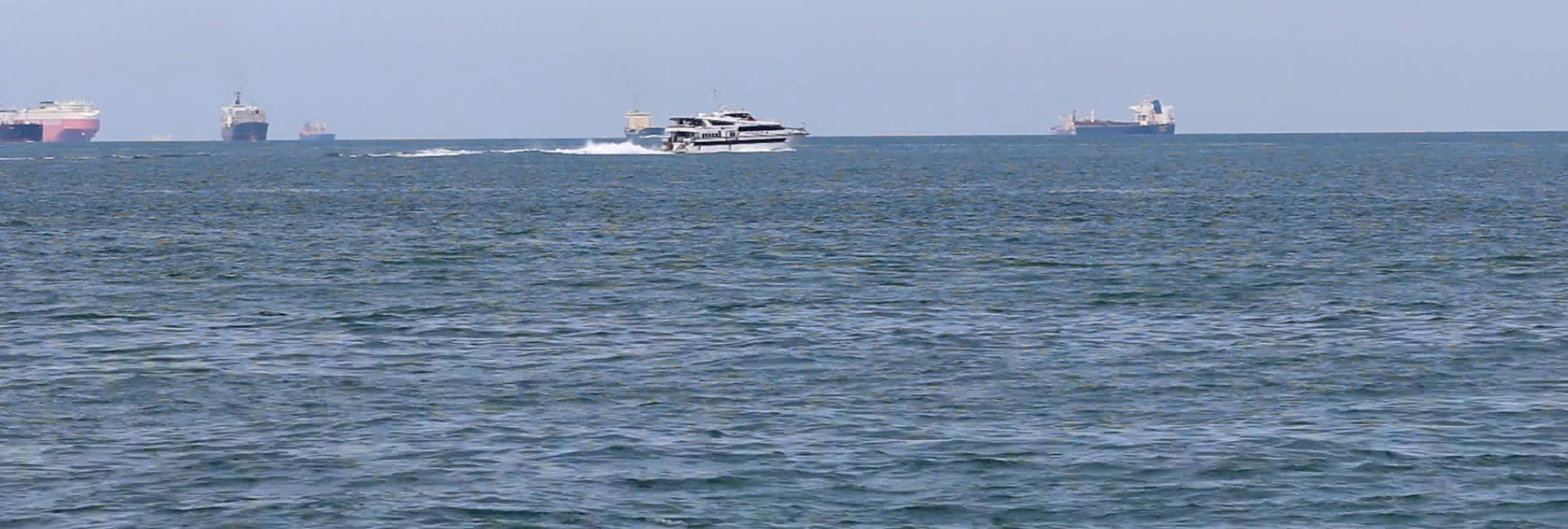} }
\subfigure[Denoised $\hat{x}$]{\label{f2b}\includegraphics[height=25mm,width=38mm]{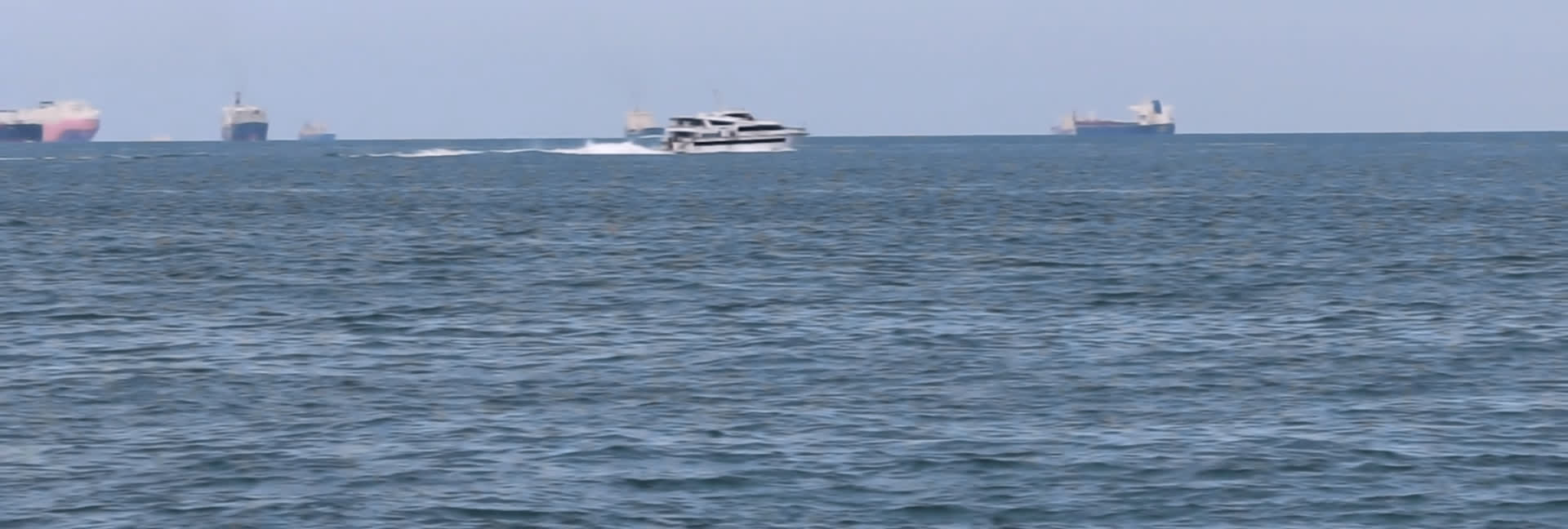}}
\subfigure[Residual $r$]{\label{f2c}\includegraphics[height=25mm,width=38mm]{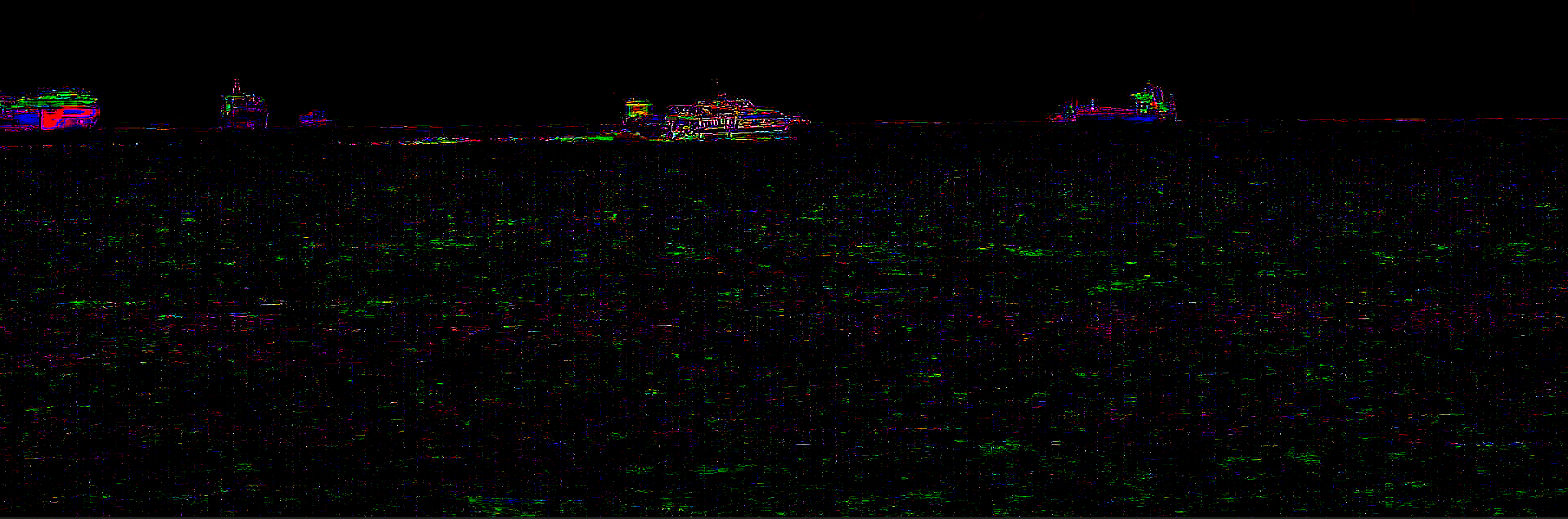}}
\subfigure[$\hat{D}_{init}$]{\label{f2d}\includegraphics[width=23mm]{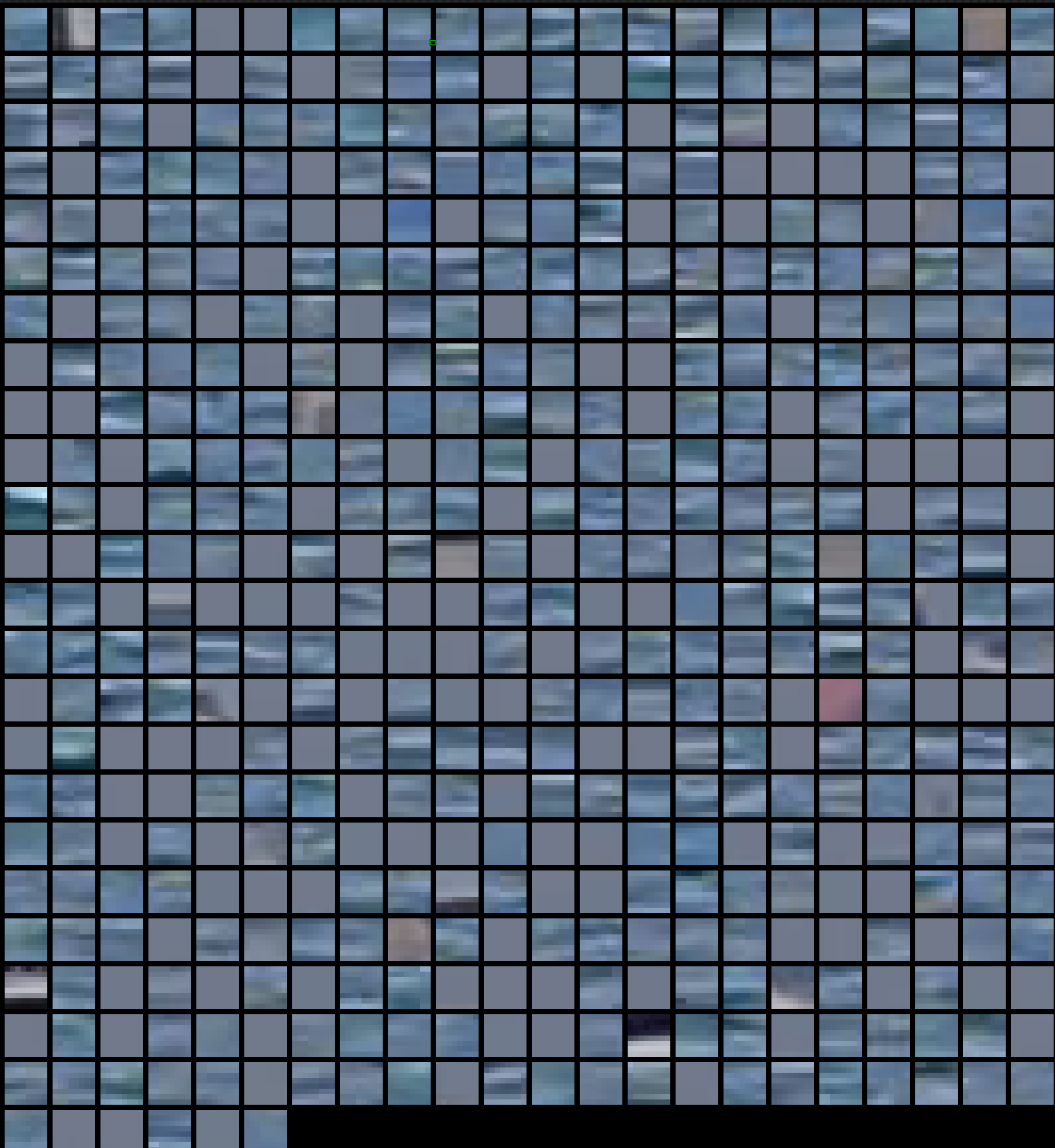}}
\subfigure[Learnt $\hat{D}$]{\label{f2e}\includegraphics[width=23mm]{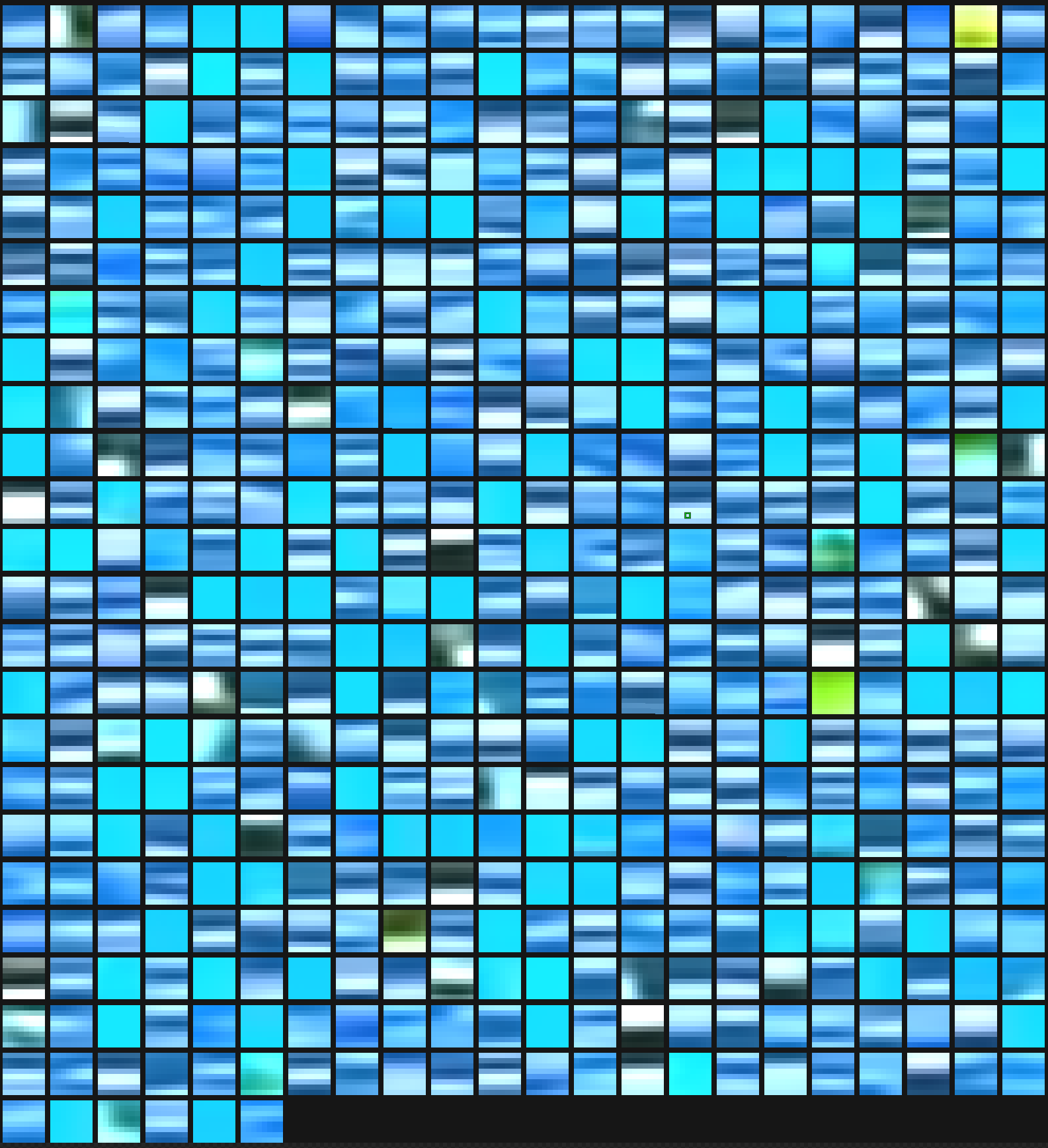}}

\caption{
(a) Shows the input image from which the dictionary is learnt, (b) Denoised image obtained from dictionary learning algorithm, (c) Residual image (contrast and brightness adjusted), (d) Random patches from input image used to learn the dictionary, (e) Final learnt dictionary. These image are obtained by using the dictionary of size 512, patch size 64 (8x8). K\_iter (the number of iterations of the K-SVD algorithm)  is fixed at 15. Note that this figure is best viewed in color.}
\end{figure*}

Let $x$ denote an unknown clean gray scale image (i.e. ideal image). It is represented as a $N$ column vector. Let $y=x+w$ be an image obtained during the image acquisition process (i.e. noisy image Fig. \ref{f2a}), where $w$ is the additive Gaussian noise with zero mean and standard deviation $\sigma$. We are interested in reconstructing an image $\hat{x}$ (Fig. \ref{f2b}) that is close to the initial image $x$, and such that each of its patches admit a sparse representation in terms of a learned dictionary \cite{FOBD_26}. For each pixel position $(i,j)$ of the image $y$, we denote by $R_{ij}y$ the size $n$ column vector formed by the gray-scale levels of the squared $\sqrt{n}\times\sqrt{n}$ patch of the image $y$ and the top-left corner of the patch is represented by the coordinates $(i,j)$. The goal is to learn a dictionary $\hat{D}$ (Fig. \ref{f2e}) of size $n \times k$, with $k \geq n$ and whose columns are normalized. Here, an initialization of the dictionary denoted by $D_{init}$ (Fig.  \ref{f2d}) is required which can be done using a pre-learned dictionary like (DCT) or the random patches from the noisy image or from a clean image. The proposed method uses the patches from image $y$ to initialize dictionary. The K-SVD based denoising algorithm has three important steps. To start, we set $\hat{D}= D_{init}$.

\begin{table}

\centering
\resizebox{.45\textwidth}{!}{%
\begin{tabular}{|c |c | c | c |c | c | c |} 

\hline
 Seq & Name & Number of frames & Resolution & Time taken(sec) & Details & Camera \\ [0.5ex] 
 \hline\hline
 1 & Ship-wreckage & 257 & 1276 x 546 & 10 & Wreckage from a broken ship & On-board, motion\\ 
 \hline
 2 & Floating-Container & 300 & 640 x 352 & 3 & Containers floating in sea & On-board, motion \\
 \hline
 3 & Sea-Truck & 299 &1280 x 720 & 15 & Water vehicle &  On-board, motion \\
 \hline
 4 & Space-capsule & 257 & 1276 x 438 & 9 & Space capsule being retrieved back & On-board \\  
 \hline
 5 & Floating-Cargo & 119 & 848 x 480 & 6 & cargo boxes floating near shore & On-shore, motion \\
 \hline
 6 & MVI\_1644\_VIS & 252 & 1920 x 1080 & 28 & From SMD \cite{FOBD_11}, contains big ships & On-shore, no motion \\ 
 \hline

\end{tabular}
}
\caption{Dataset and its properties} 
\label{table1}
\end{table}

\begin{itemize}
\item \textit{Sparse coding} : Here we use the fixed dictionary $\hat{D}$ to compute the sparse approximation $\hat{\alpha}$ of all the patches $R_{ij}y$ of the image in $\hat{D}$. i.e. for each patch $R_{ij}y$ a column vector $\hat{\alpha}_{i,j}$ of size $k$ is built such that it has only few non-zero coefficients and such that the distance between $R_{ij}y$ and its sparse approximation $\hat{D}\hat{\alpha}_{i,j}$ is very small. 

\begin{equation} \label{eq:1}
\argmin_{\hat{\alpha}_{ij}} \| \hat{\alpha}_{ij}\|_{0} \; \texttt{such that}  \;  \|R_{ij}y - \hat{D}\hat{\alpha}_{ij} \|_{2}^{2}\leq \epsilon^{2} 
\end{equation}  

Where $\| \hat{\alpha}_{ij}\|_{0} $ refers to the number of non-zero coefficients of $\hat{\alpha}_{ij}$ also known as $l^0$ norm of $\hat{\alpha}_{ij}$. This is a NP hard problem and we make use of Orthogonal Recursive Matching Pursuit (ORMP) as implemented in \cite{FOBD_26} to get an approximate solution. In Eq. (\ref{eq:1}), $\epsilon$ is used during the break condition of the ORMP, $\hat{D}$ is of size  $n \times k$, $\hat{\alpha}_{ij}$ is a column vector of size $k \times 1$ and $R_{ij}y$ is a column vector of size $n \times 1$. \\

\item \textit{Dictionary update} : Here we update the columns of the dictionary $\hat{D}$ one by one, to reduce the quantity in Eq. (\ref{eq:2}) without increasing the sparsity penalty $\hat{\alpha}_{ij}$ such that all the patches in the image $y$ are efficient. This is achieved using the K-SVD algorithm. More details about K-SVD can be found in \cite{FOBD_27} and \cite{FOBD_26}.  
\begin{equation} \label{eq:2}
\sum_{i,j} \|\hat{D}\hat{\alpha}_{ij} - R_{ij}y\|_{2}^{2}
\end{equation}

\item \textit{Reconstruction}: We repeat the above two steps for some iterations say $K\_iter$. Once these $K\_iter$ iterations are done, each patch $R_{ij}y$ of the image $y$ corresponds to the denoised version $\hat{D}\hat{\alpha}_{ij}$. In the third and final step
we reconstruct the final denoised version of the image from all the denoised patches by solving the minimization problem in Eq. (\ref{eq:3}).

\begin{equation} \label{eq:3}
	\hat{x} = \argmin_{x\in{\rm I\!R}^{N}} \;\lambda\Vert x- y \Vert_{2}^{2} + \sum_{i,j} \|\hat{D}\hat{\alpha}_{ij} - R_{ij}y\|_{2}^{2}
\end{equation}

First term in Eq. (\ref{eq:3}) represents a fidelity term which controls the global proximity to our reconstruction $\hat{x}$ with the noisy image $y$ \cite{FOBD_26}. The second term controls the proximity of the patch $R_{ij}\hat{x}$ of our reconstruction to the denoised patch $D\alpha_{ij}$. Eq. (\ref{eq:3}) is quadratic, coercive, and differentiable \cite{FOBD_26}. It has a closed-form solution
of the form given in Eq. (\ref{eq:4}).

\begin{equation} \label{eq:4}
\hat{x} =  \left( \lambda\textbf{I} +  \sum_{i,j} R_{ij}^{T} R_{ij} \right)^{-1} \left( \lambda y +  \sum_{i,j} R_{ij}^{T} \hat{D} \hat{\alpha}_{ij} \right)
\end{equation}

In Eq. (\ref{eq:4}), $\textbf{I}$ represents an identity matrix of size $N \times N$.  This equation also implies that the value of a pixel in the denoised image is computed by averaging the value of this pixel in the noisy image (weighted by $\lambda$) and the values of this pixel on the patch to which it belongs
(weighted by 1).

\end{itemize} 

Our problem statement deals with color images. So, we extend the above algorithm to color images
by concatenating the R,G,B values of the patch to a single column. Thus the algorithm learns the correlation between the color channels resulting in a better update of the dictionary \cite{FOBD_26}. We obtain the residual image $r$ Fig.\ref{f2c} by taking the pixel-wise difference between the noisy image $y$ and the reconstructed image $\hat{x}$ as in Eq. (\ref{eq:5}):

\begin{equation} \label{eq:5}
r(i,j,ch) = y(i,j,ch) - \hat{x}(i,j,ch)
\end{equation}

In Eq. (\ref{eq:5}) above, $i$, $j$ and $ch$ represents the pixel coordinates and the channel number respectively. Thus obtained residual image contains only noise and salient structures as it is free from self-similar content. The intuition is that it is very
easy to detect salient regions/objects in the residual image compared to detection in the original image
$y$.

We use a multi-scale approach to detect the salient structures of different sizes. We follow \cite{FOBD_35} to obtain images at different scales and learn individual dictionary for each of the scaled version of the original image. Finally, we construct the residual image at different scales as previously explained. 

\subsection{A contrario model for object detection}

The a contrario detection theory was primarily proposed by \cite{FOBD_38} and has been successfully employed in many computer vision applications such as shape matching \cite{FOBD_39}, 
vanishing point detection \cite{FOBD_42}, anomaly detection  \cite{FOBD_6}, spot detection \cite{FOBD_41} etc. The a contrario framework is based on the probabilistic formalization of the Helmholtz perceptual grouping principle.  According to this principle, perceptually meaningful structures represent large deviations form randomness/naive model. Here, structures to be detected are the co-occurrence of several local observations \cite{FOBD_38}. 

We define a naive model by assuming that all local observations are independent. By using this a contrario assumption, we can compute the probability that a given structure occurs. More precisely, we call the number of false alarms (NFA) of a structure configuration, its expected number of occurrences in the naive model. We say that a structure is $\varepsilon$-meaningful if its NFA is smaller than $ \varepsilon$. The smaller the $\varepsilon$ the more meaningful the event. Given a set of random variables $(X_{i})_{i\in\left[ \mid 1, N\mid\right] }$, a function $f$ is called a NFA if it guarantees a bound on the expectation of its number of false alarms under the naive model/null-hypothesis, namely
\begin{equation}
\forall \varepsilon > 0, \quad \mathbb{E}\left[\# \left\lbrace i, f(i,X_{i})  \leq \varepsilon \right\rbrace  \right]\leq \varepsilon 
\end{equation}
Here we consider NFA as :
\begin{equation} \label{nfa}
f(i,x) = N \, \mathbb{P}(\mid X_{i} \mid \geq \mid x_{i} \mid) ,
\end{equation}

Where $N$ is the total number of tests, which is nothing but the total number of pixels in all the images at different channels and scales. $i$ is the index of the pixels used in the tests, $X_{i}$ is a random variable distributed as the residual at position $i$, and $x_{i}$ the actual measured value at position $i$. Here, the naive model is constructed in such a way that each pixel  of the residual image follows a standard normal distribution.

Our aim is to detect salient structures in the residual image $r$. We follow the procedure proposed by \cite{FOBD_41} and \cite{FOBD_6}. The residual image $r$ is unstructured, similar to a colored noise and not necessarily Gaussian. A careful study of the residual distribution shows that it follows a generalized Gaussian distribution \cite{FOBD_6}. A non-linear transform is used to re-scale the residual image in order to fit a centered Gaussian distribution with unit variance. This centered Gaussian distribution with unit variance is considered as the naive model. The naive model doesn't require the noise to be uncorrelated. Since structures are expected to deviate from this naive model, this amounts to checking the tails of the Gaussian and to retain high values as significant if their tail has a very small area. 

We convolve the residual image with a kernel $K_c$ of given radius which results in a new image $\overline{r} = r * K_c$. Thus obtained $ \overline{r}$ is normalized to have a unit variance. As we have assumed the residual to be stationary Gaussian field, the result after filtering is also Gaussian. We use $N_{k}=3$ number of disks with radius $1, 2$ and $3$ at each scale to detect the salient structures in the residual image. We detect the salient structures on both the tail of the distribution using the NFA given in Eq. (\ref{nfa}). The number of tests in our case is given by Eq. (\ref{nfa1}). Here, $N_{k}$ refers to the number of disk kernels, $N_{ch}$ refers to the number of channels, $N_{scale}$ is the number of scales used and $\Omega$ is the set of pixels in the residual image at a given scale  

\begin{equation} \label{nfa1}
N = N_{k}\times N_{ch}\times\sum_{0}^{N_{scale}-1} \mid \Omega \mid
\end{equation}

\begin{figure*}

\centering
\subfigure[Seq.1, Frame 52]{\label{fig5m}\includegraphics[height=13mm,width=35mm]{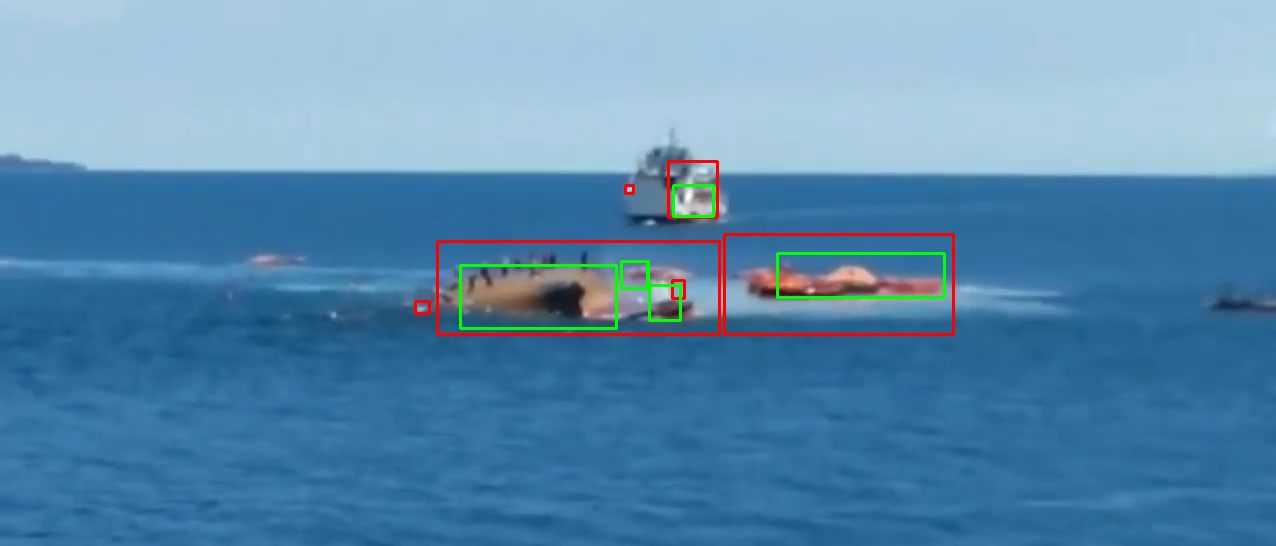}}
\subfigure[Seq.1, Frame 110]{\label{fig5n}\includegraphics[height=13mm,width=35mm]{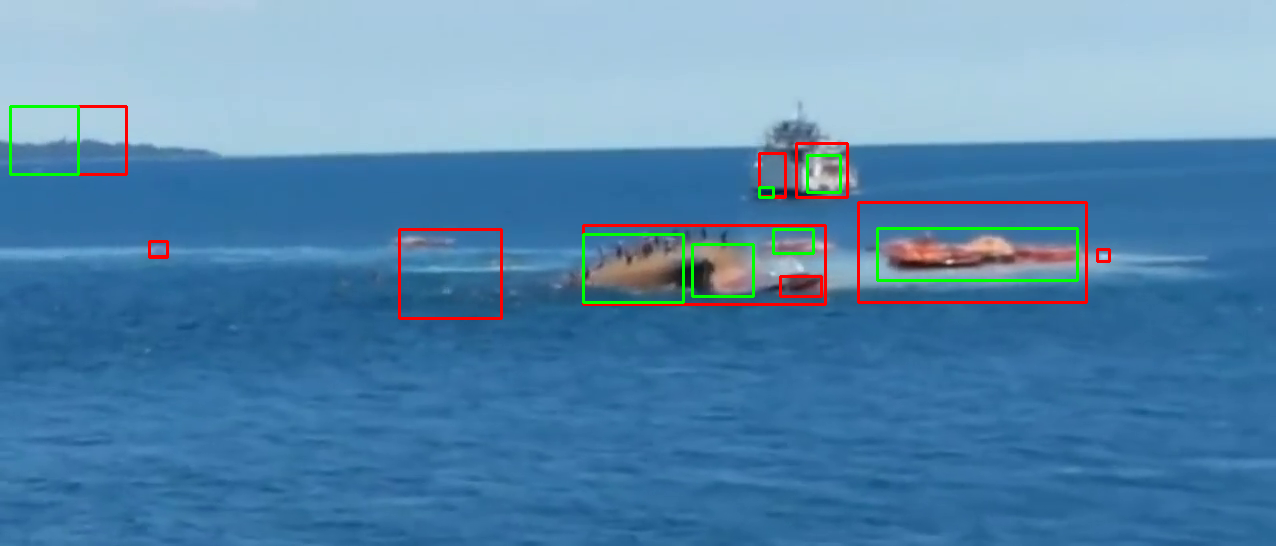}}
\subfigure[Seq.2, Frame 132]{\label{fig5e}\includegraphics[height=13mm,width=35mm]{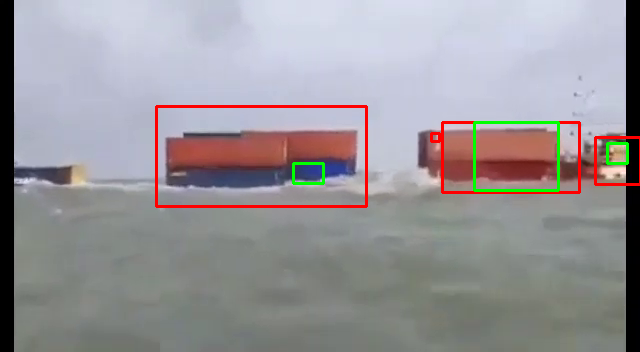}}
\subfigure[Seq.2, Frame 270]{\label{fig5f}\includegraphics[height=13mm,width=35mm]{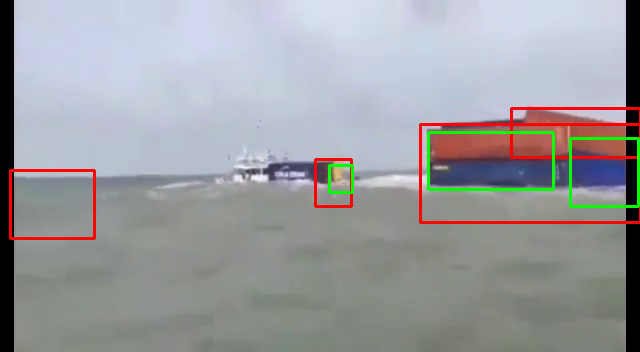}}
\subfigure[Seq.3, Frame 37]{\label{fig5c}\includegraphics[height=13mm,width=35mm]{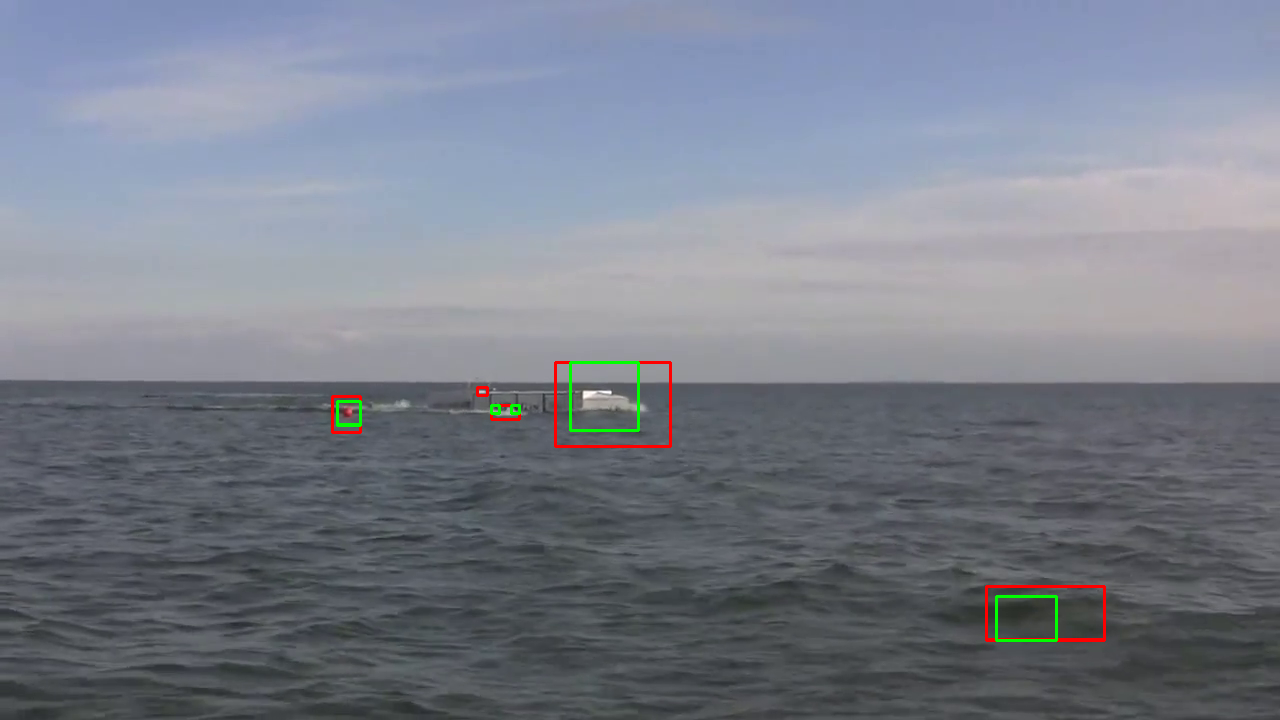}}
\subfigure[Seq.3, Frame 65]{\label{fig5d}\includegraphics[height=13mm,width=35mm]{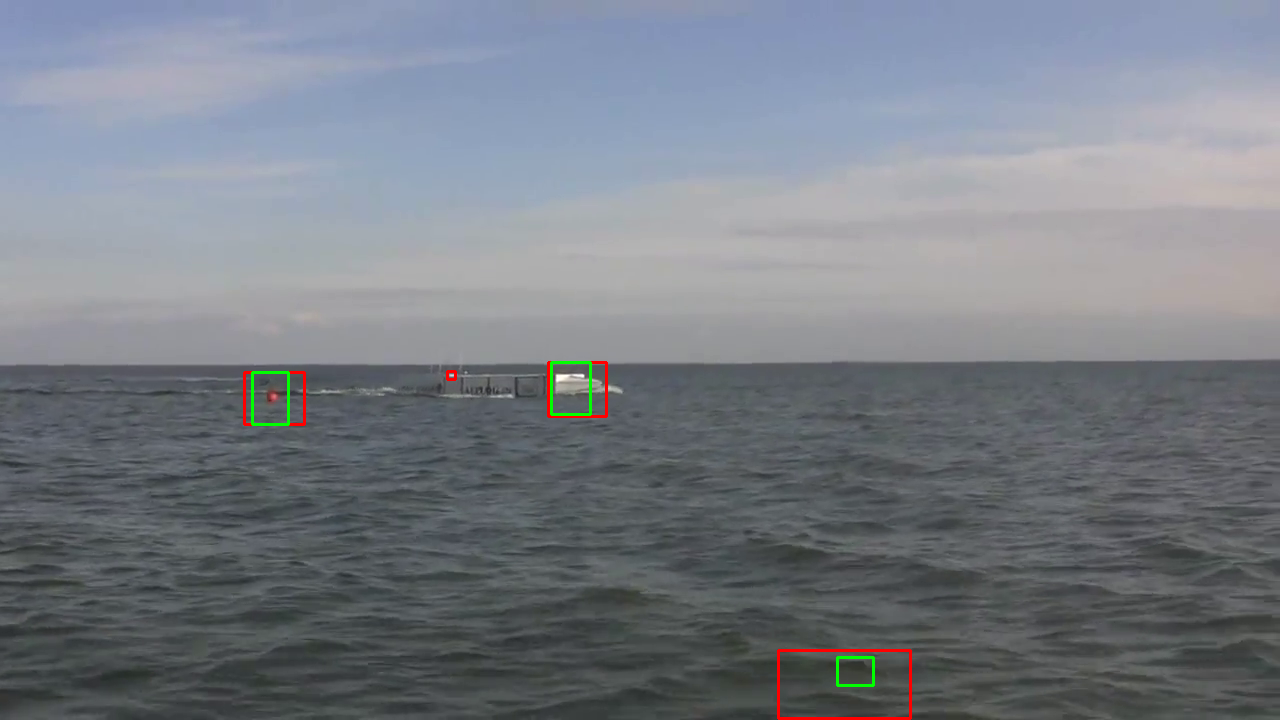}}
\subfigure[Seq.4, Frame 64]{\label{fig5a}\includegraphics[height=13mm,width=35mm]{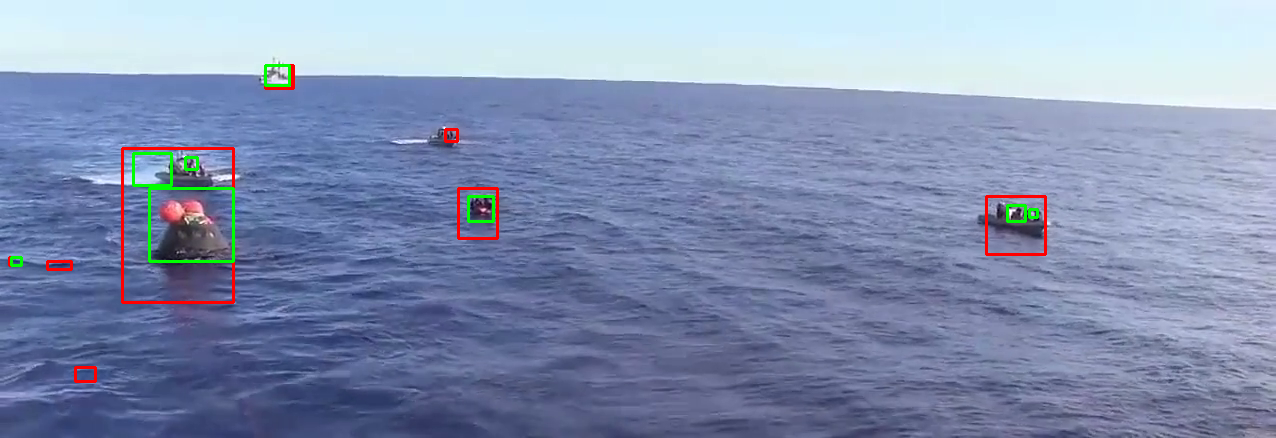}}
\subfigure[Seq.4, Frame 129]{\label{fig5b}\includegraphics[height=13mm,width=35mm]{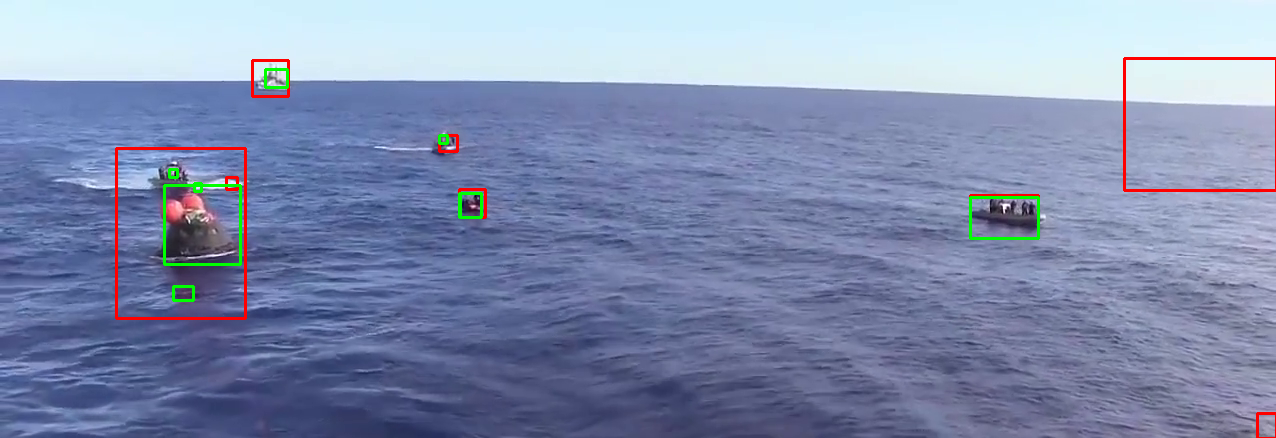}}
\subfigure[Seq.5, Frame 25]{\label{fig5k}\includegraphics[height=13mm,width=35mm]{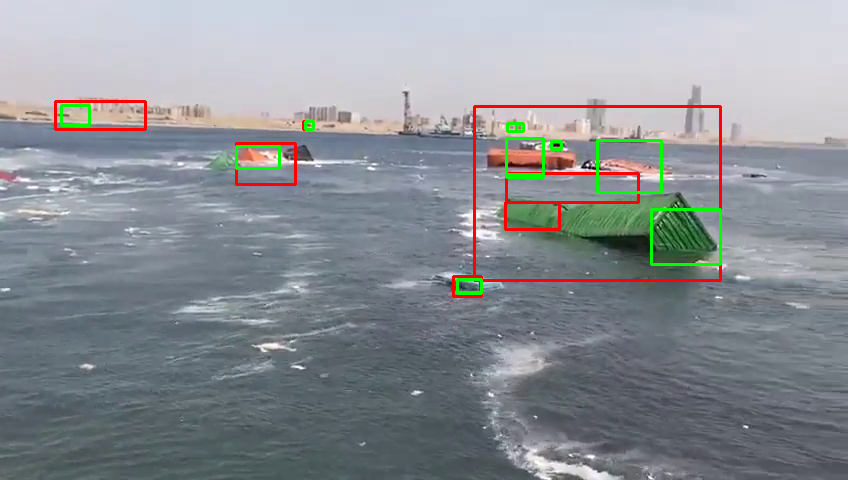}}
\subfigure[Seq.5, Frame 65]{\label{fig5l}\includegraphics[height=13mm,width=35mm]{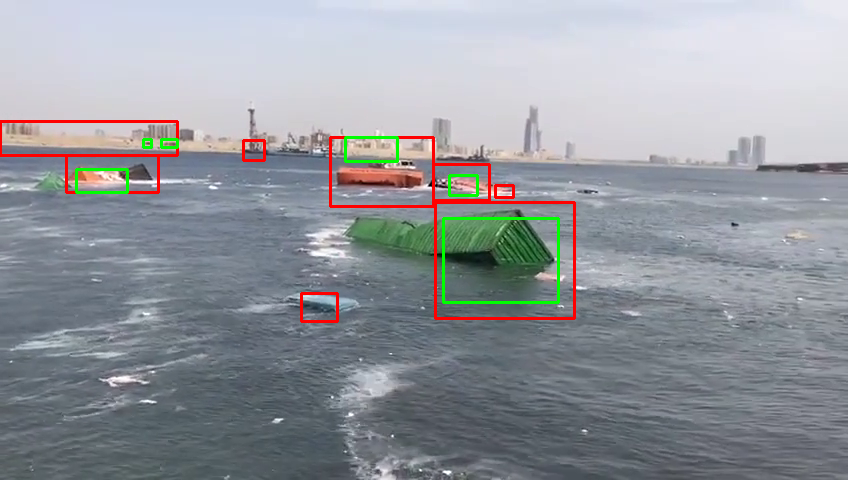}}
\subfigure[Seq.6, Frame 127]{\label{fig5g}\includegraphics[height=13mm,width=35mm]{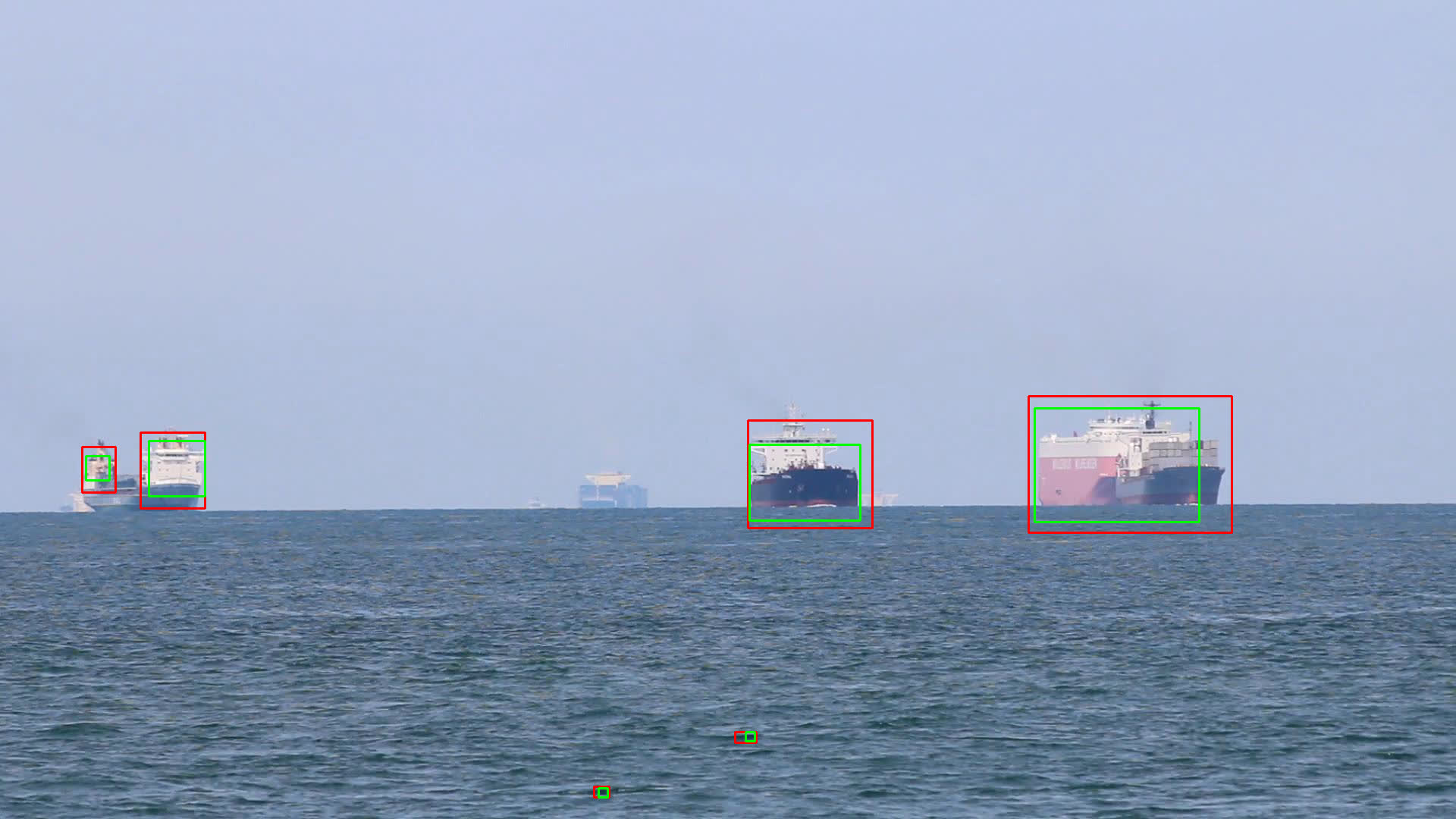}}
\subfigure[Seq.6, Frame 157]{\label{fig5h}\includegraphics[height=13mm,width=35mm]{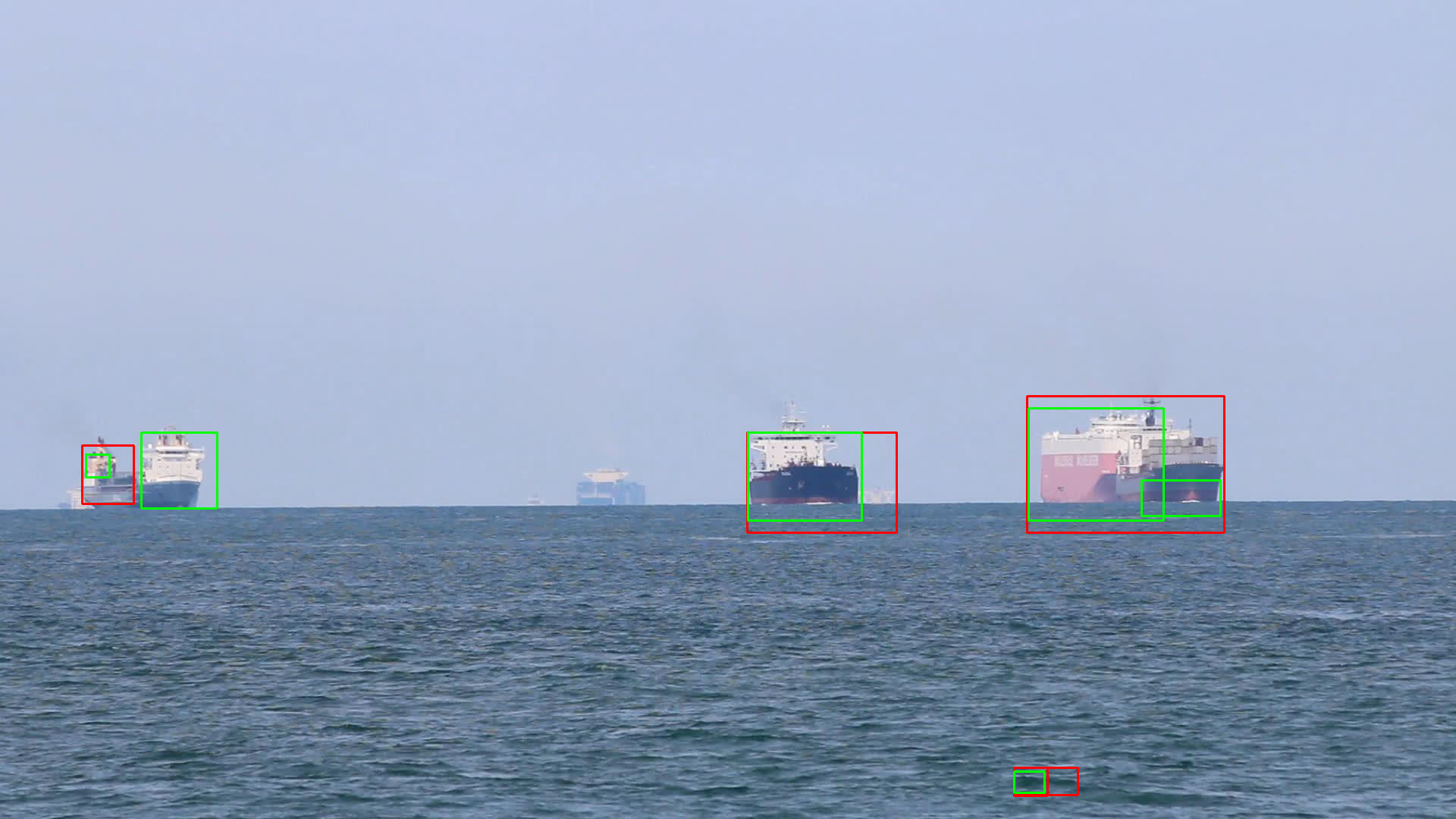}}

\caption{Qualitative results : False detections resulting from our approach.}
\label{fig555}
\end{figure*}

%

\begin{table*}

\centering
\resizebox{\linewidth}{!}{%
\begin{tabular}{ |l|l|l|l|l|l|l|l|l||l|l|l|l|l|l|l|l|l|l|l|l| }
  \hline
  \multicolumn{1}{|c|}{} & \multicolumn{5}{|c|}{logNFA=-2 } & \multicolumn{5}{|c|}{logNFA=2} & \multicolumn{5}{|c|}{IITI} & \multicolumn{5}{|c|}{SRA}\\
  \hline
   Seq & TP & FP & FN & DR & FAR & TP & FP & FN & DR & FAR &TP & FP & FN & DR & FAR & TP & FP & FN & DR & FAR \\ 
  \hline
  1 & 1218 & 9   & 424 & 0.741 & \textbf{0.007} &1480 &177 &162 &\textbf{0.901}  &0.106 &796 &808  &846 &0.484 &0.503 &1169 
  &667 &473 &0.711 &0.363  \\\hline
  
  2 & 751  & 10  & 106 & 0.876 & 0.013 &843  &7   &14  &\textbf{0.983}  &\textbf{0.008} &564 &20  &293 &0.655 &0.034 &536 
  &146 &321 &0.624 &0.214 \\\hline
  
  3 & 479  & 55  & 1   & \textbf{0.991} & \textbf{0.103} &466  &270 &14  &0.970  &0.366 &439 &642  &41  &0.914 &0.593 &317 
  &1745 &163 &0.66 &0.846 \\\hline
  
  4 & 1349 & 80  & 193 & 0.874 & \textbf{0.056}  &1527 &354 &65  &\textbf{0.959}  &0.188 &557 &1029 &985 &0.361 &0.648 &1281 
  &942 &261 &0.83 &0.423 \\\hline
  
  5 & 554  & 133 & 522 & 0.515 & \textbf{0.193} &815  &216 &261 &\textbf{0.757}  &0.209 &455 &132  &579 &0.442 &0.217 &517 
  &371 &559 &0.48 &0.38\\\hline
  
  6 & 781  & 32  & 731 & 0.516 & \textbf{0.039} &865  &201 &647 &0.572  &0.188 &911 &256  &601 &0.602 &0.219 &1008 
  &1145 &504 &\textbf{0.66} &0.531 \\
  
  \hline
  
\end{tabular}}
\caption{Quantitative Evaluation : TP: True Positive, FP: False Positive, FN: False Negative, DR: Detection Rate, FAR: False Alarm Rate} 

\label{table2}
\end{table*}

\subsection{Bounding Box detection and refinement}
  By using the above approach, structures are detected at a certain radius of the kernel. Using the center and the radius of detection, we construct a square bounding box. Many of these Bounding Boxe(BB) overlap. We use the opencv function such as "findContours" and "approxPolyDP" to fuse the overlapping BB and get a single big rectangular BB. Thus obtained BB has many false detection. One of the easiest way to refine false detection is to ignore the BB which has no key-points present in it. A combination of key-points from SIFT\cite{FOBD_35}, SURF\cite{FOBD_36} and ORB\cite{FOBD_37} detectors are used as they provide a good coverage of the image space, including corners, edges and textured areas. In our case we make use of 200 most dominant key-points from each of the detectors to refine false detections.

\section{Experiments and Results} 
  Here we present the results of the proposed algorithm. Initially, we describe the parameters used by the algorithm followed by the dataset and the results. 
\subsection{Parameters}
 In the dictionary learning part, the patch size $n$ is set to 4 and the size of the dictionary $k$ is fixed to 64. The value of $n$ and $k$ are empirically chosen as it gives a good trade-off between the size of the detected object and the speed of the algorithm. The break condition for ORMP $\epsilon$ is set to $10^{-6}$. The number of iterations in the K-SVD process $K\_iter$ is fixed to 7. More details about these parameters can be found in \cite{FOBD_7}. In the a contrario detection part, we have experimented with both NFA=$10^{2}$  (or logNFA = $2$) and NFA = $10^{-2}$ (or logNFA= $-2$) where logNFA is the logarithm of NFA. $N_{scales}$ represents the number of scales used in the multi-scale approach and it is set to 4.The Radius of the circular kernel $K_c$= 1,2,3. 
 
\subsection{Dataset}
 	In the maritime object detection scenario, only a few data-sets have been proposed and most of these data-sets are intended for ship detection. As the unidentified floating objects are random and sporadic, it is very difficult to come up with a database/dataset and there is hardly any dataset available. Here, we introduce a small dataset by extracting portion of videos from YouTube. To demonstrate the capabilities of our algorithm to detect boats in the sea, we make use of the Singapore maritime dataset. The features of our dataset are tabulated in Table \ref{table1}.

\subsection{Results and Explaination}

The algorithm was coded in C++ using OpenCV library on a laptop with 8 cores. The average time taken (in seconds) for each frame of the sequence is given in the 5th column of Table 1. In our experiments, the dictionary is learnt for every frame. Real time performance can be achieved by initially using a pre-learnt dictionary and then learning the dictionary (online and in parallel) for every $M$ duration of time. This is possible as the maritime environment in the far sea scenario does not change rapidly. This learnt dictionary can be used for object detection for that particular duration of interval. Additionally, we can speed up the detection process by detecting the horizon line and then cropping a part of the sky region. Depending on the camera setup on-board a ship, we can crop a part of the sea surface near the bottom of the ship (As our work is part of an early warning system, we are trying to detect floating objects which are at least 500-750 meters away from the ship).

\begin{figure*}

\centering
\includegraphics[height=16mm,width=35mm]{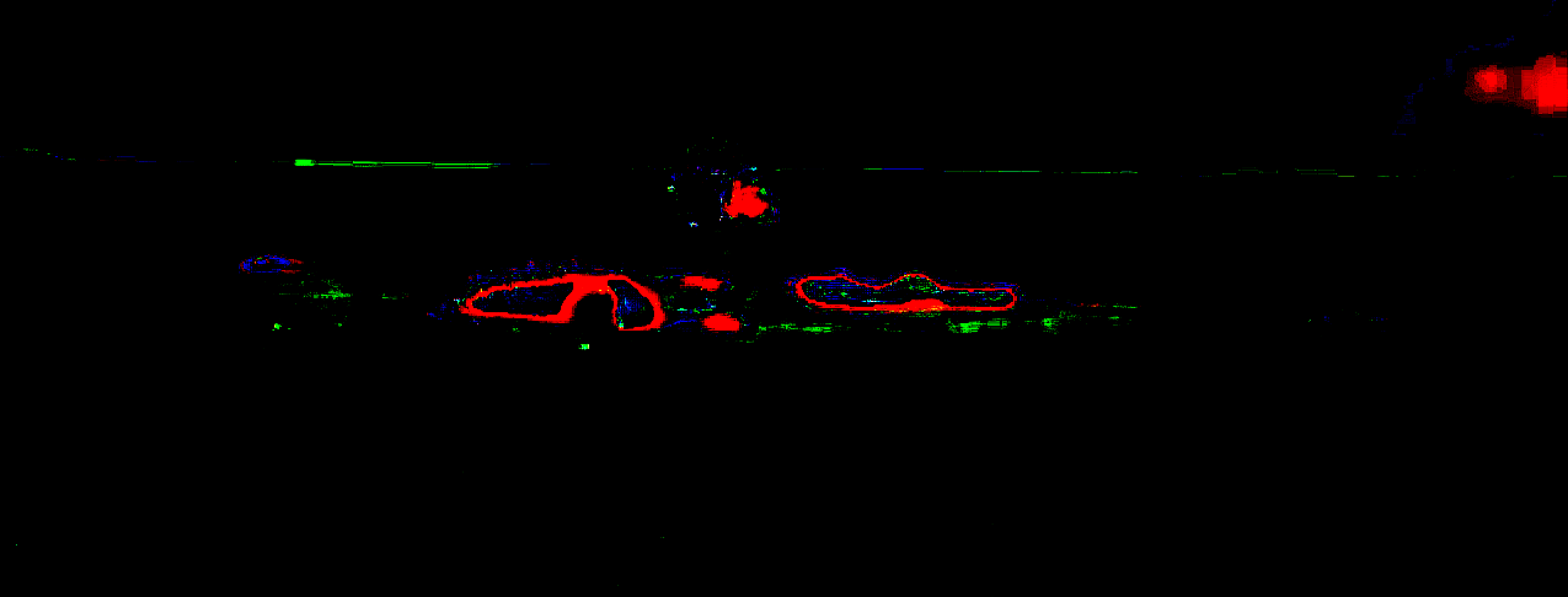}
\includegraphics[height=16mm,width=35mm]{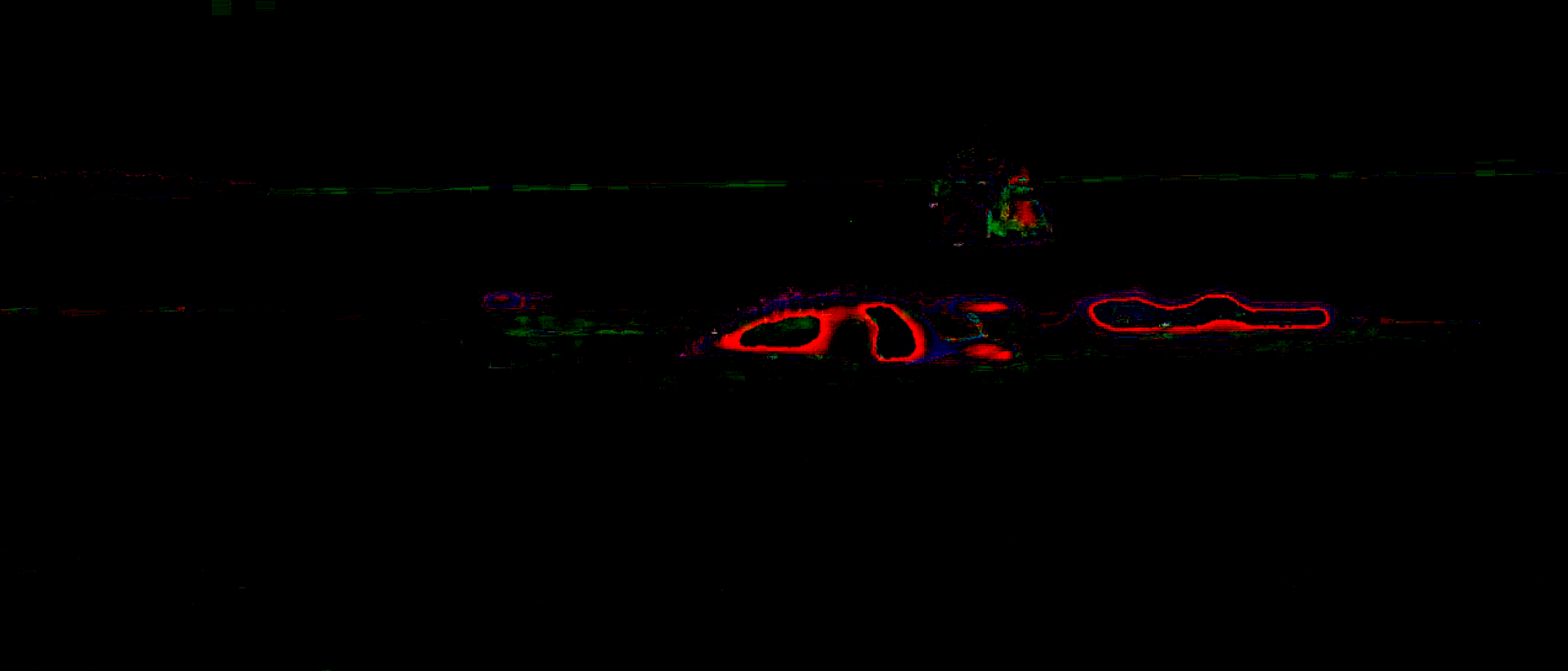}
\includegraphics[height=16mm,width=35mm]{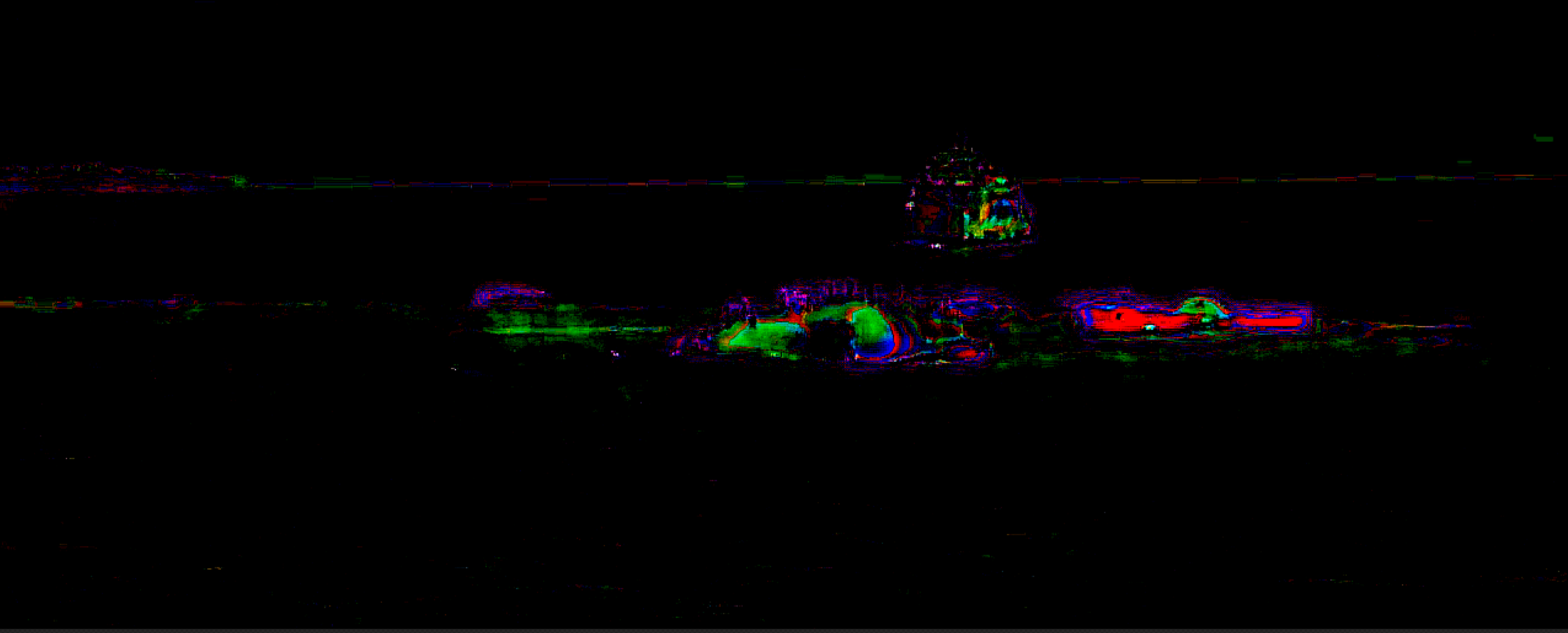}
\includegraphics[height=16mm,width=35mm]{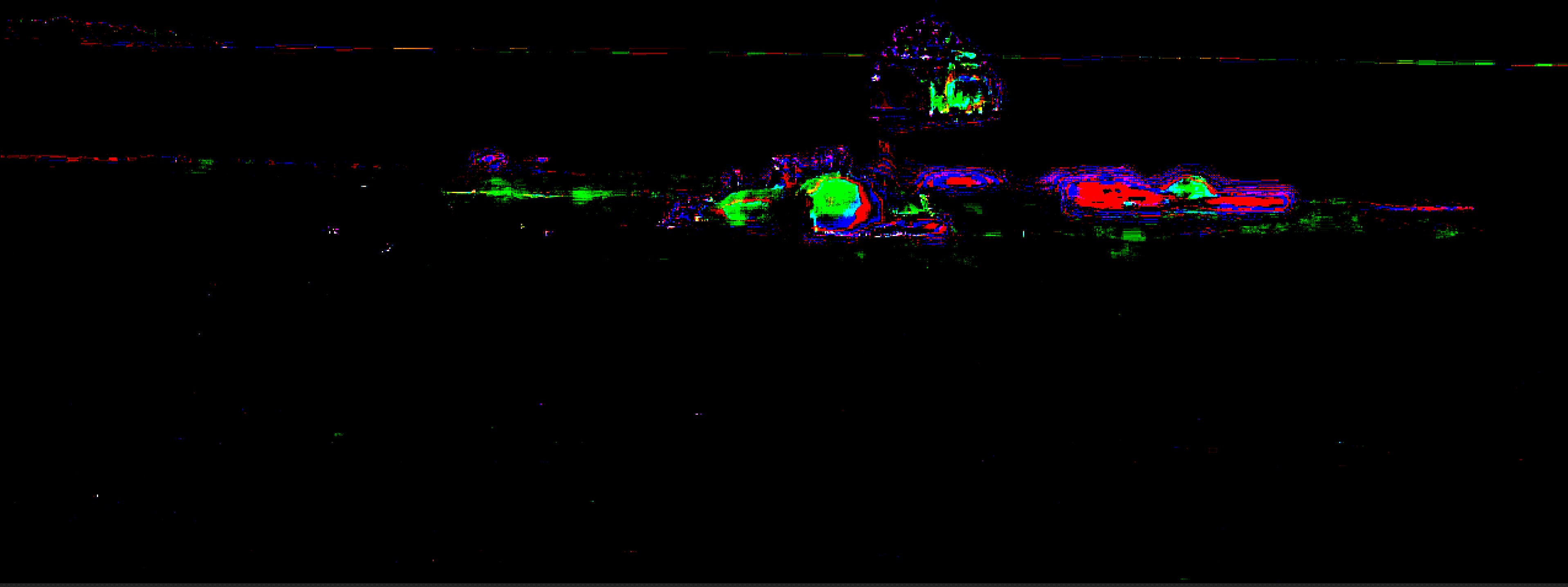}
\includegraphics[height=16mm,width=35mm]{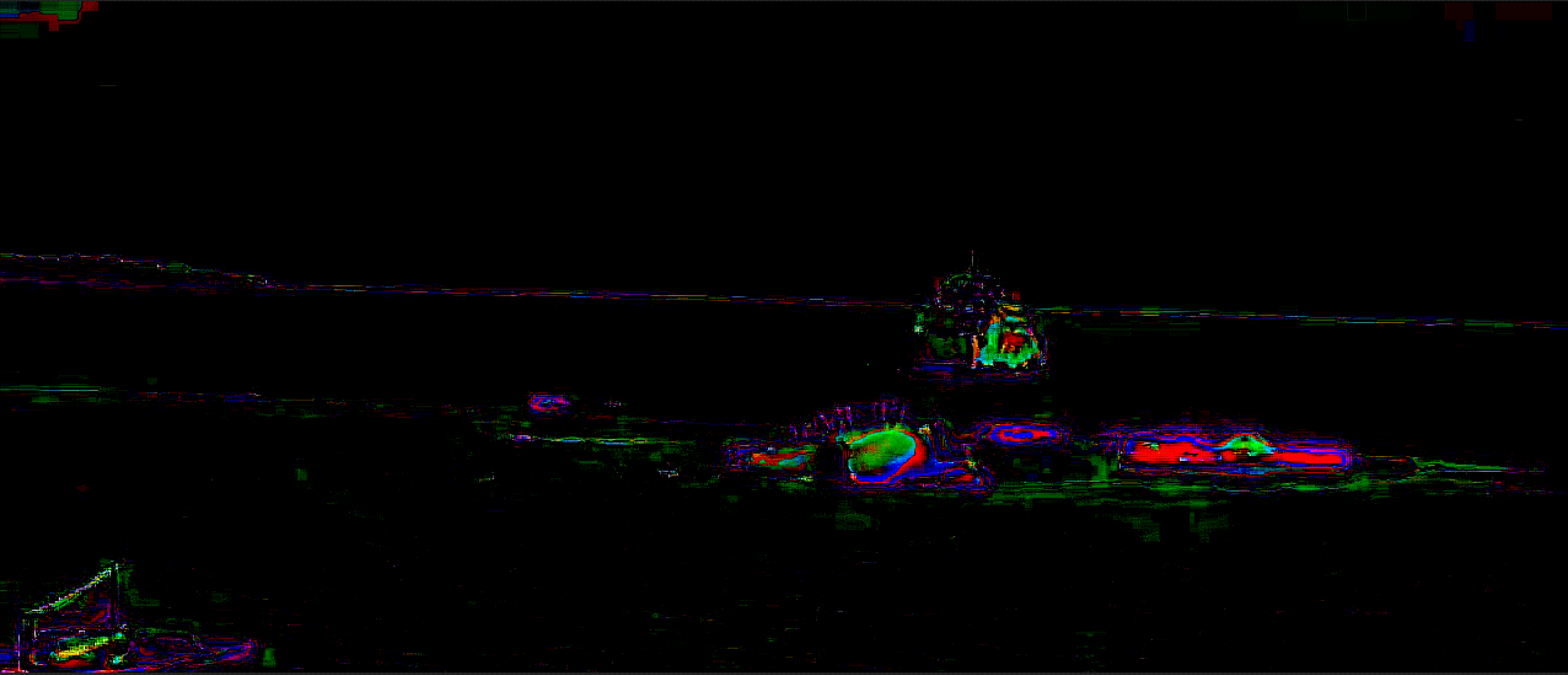}
\includegraphics[height=16mm,width=35mm]{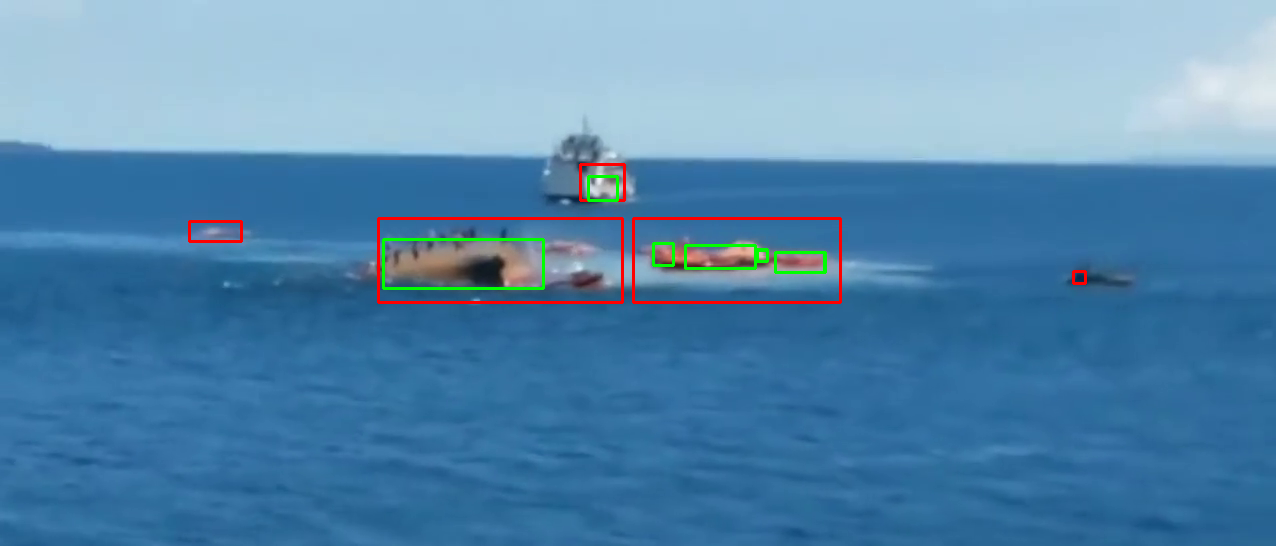}
\includegraphics[height=16mm,width=35mm]{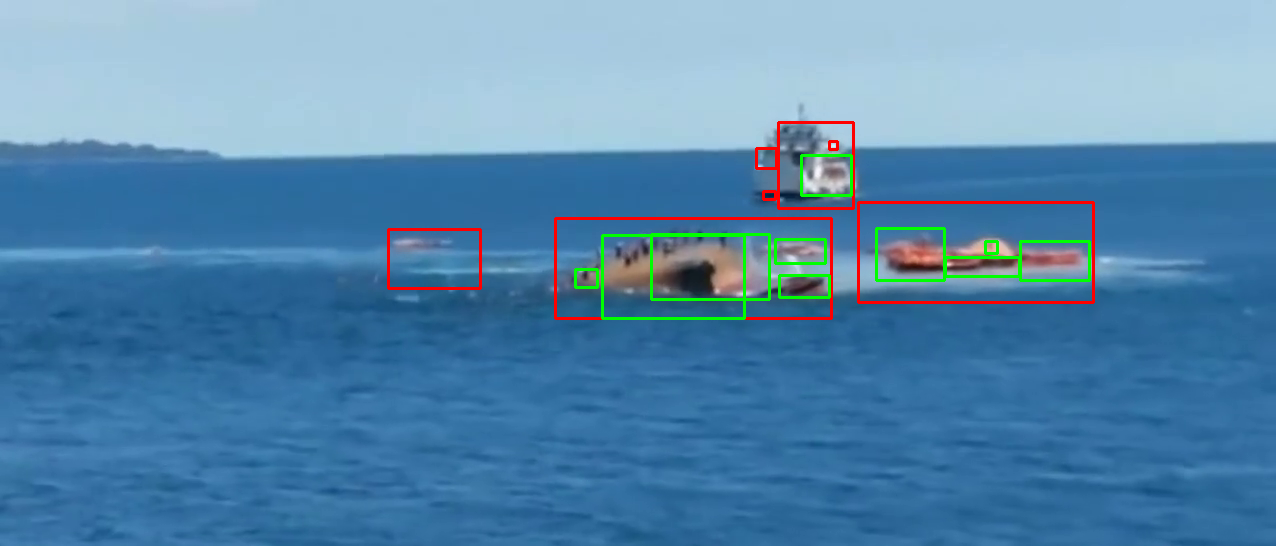}
\includegraphics[height=16mm,width=35mm]{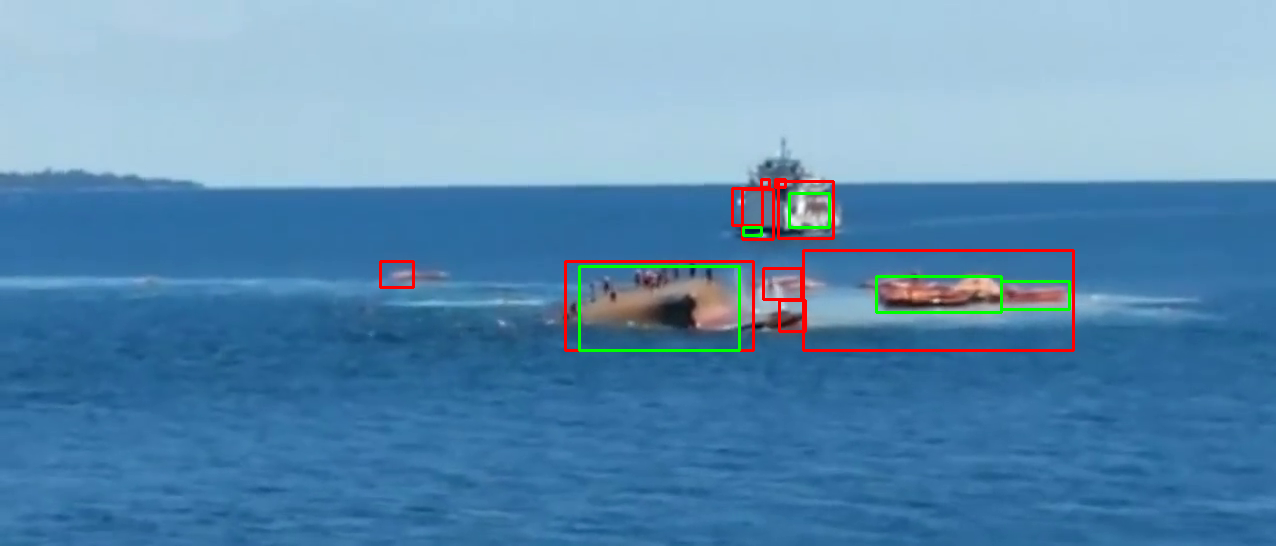}
\includegraphics[height=16mm,width=35mm]{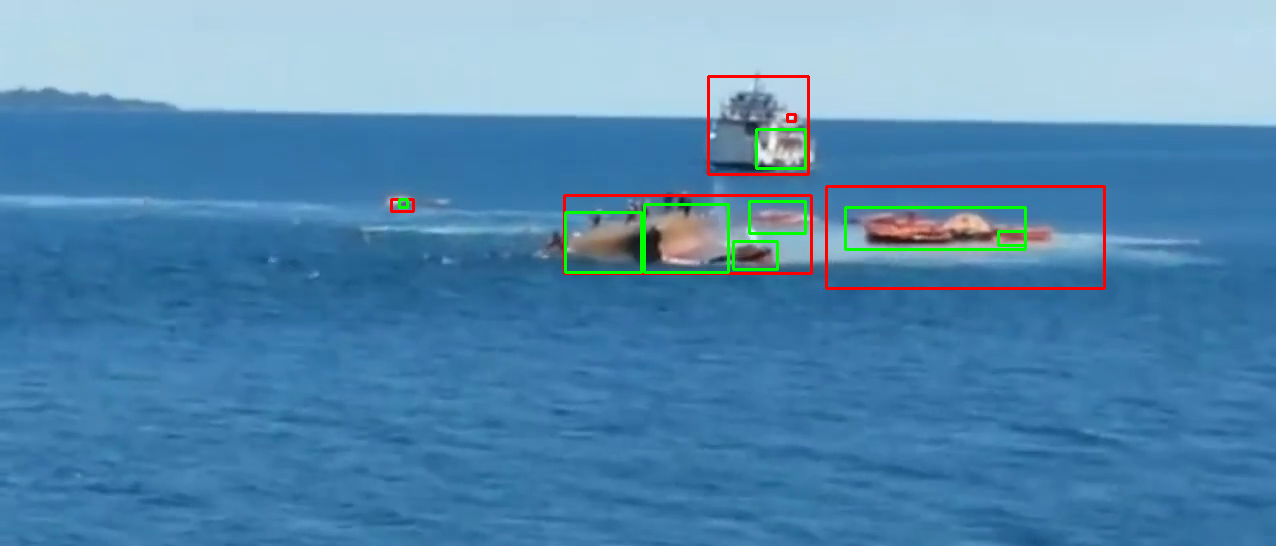}
\includegraphics[height=16mm,width=35mm]{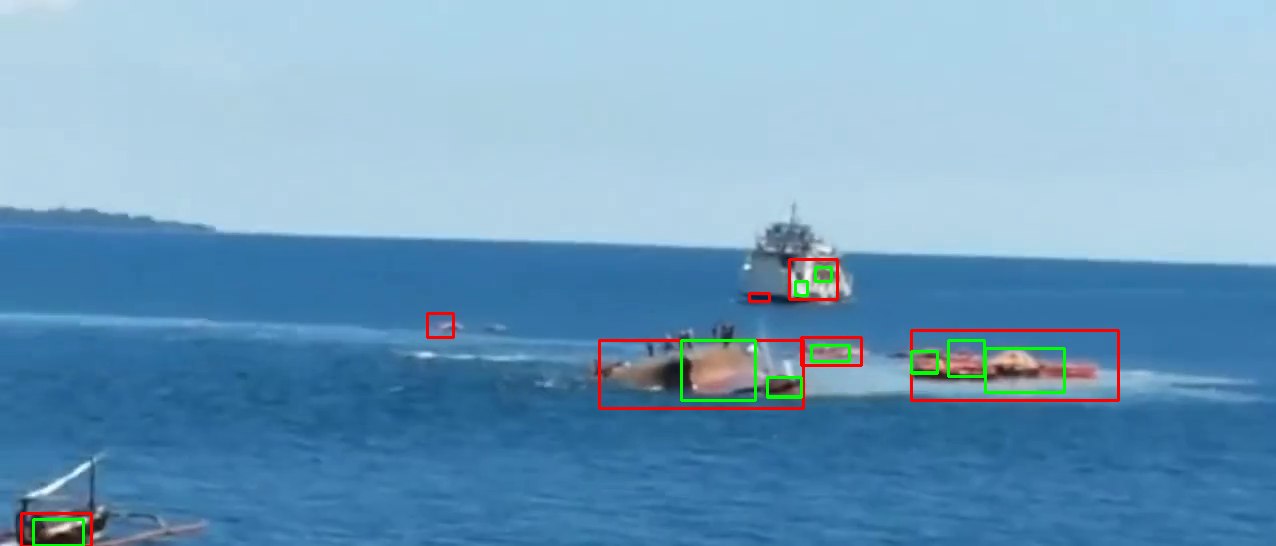}
\subfigure[Seq.1, Frame 39]{\label{fig3a}\includegraphics[height=16mm,width=35mm]{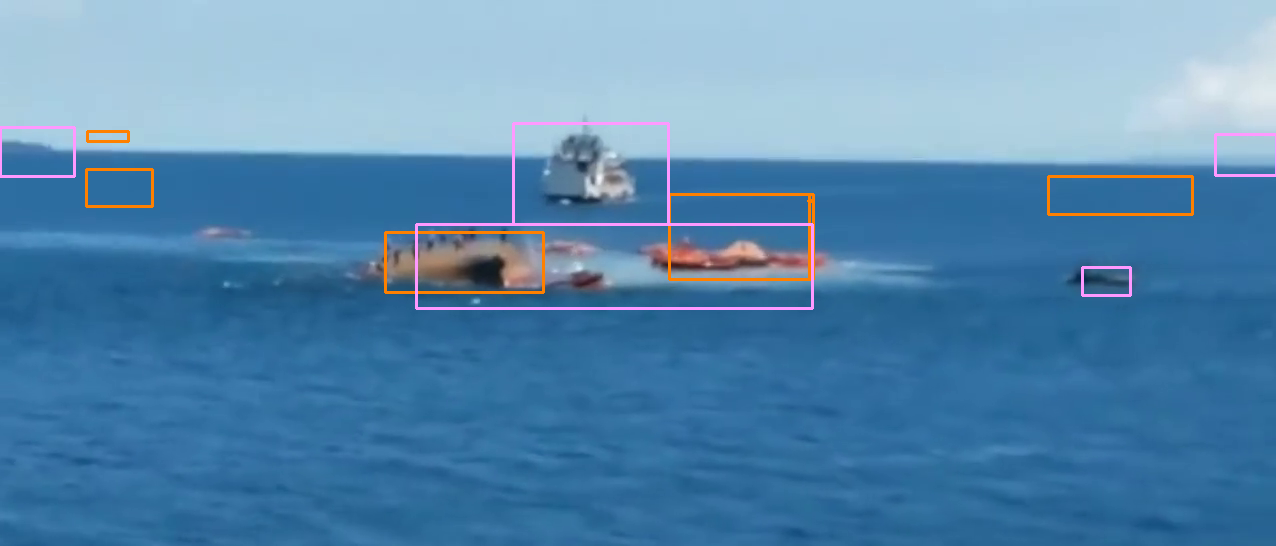}}
\subfigure[Seq.1, Frame 97]{\label{fig3b}\includegraphics[height=16mm,width=35mm]{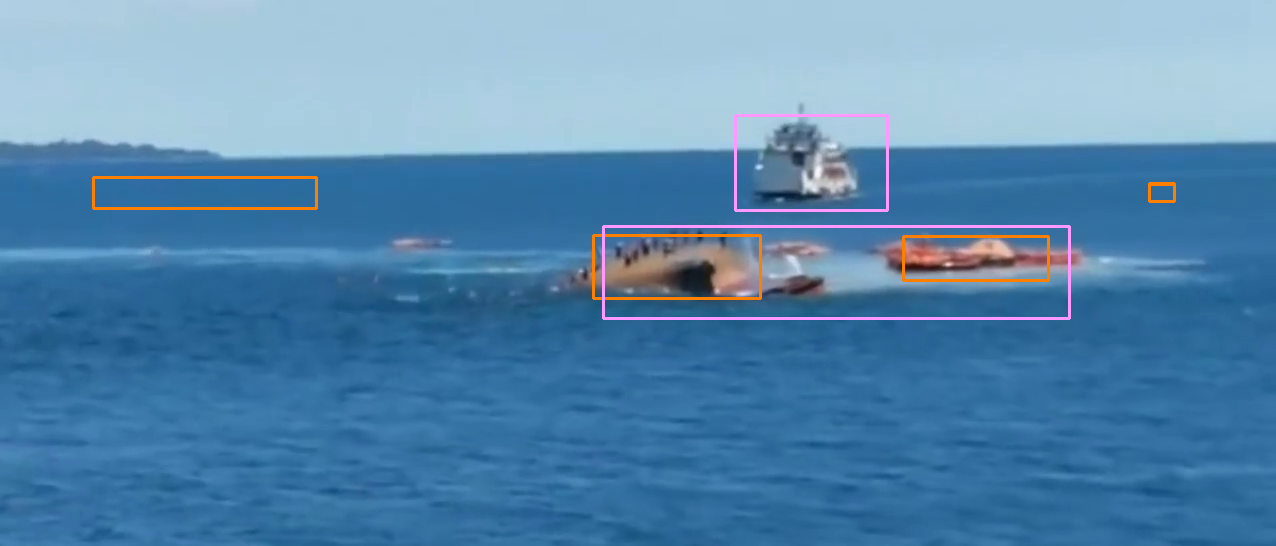}}
\subfigure[Seq.1, Frame 149]{\label{fig3c}\includegraphics[height=16mm,width=35mm]{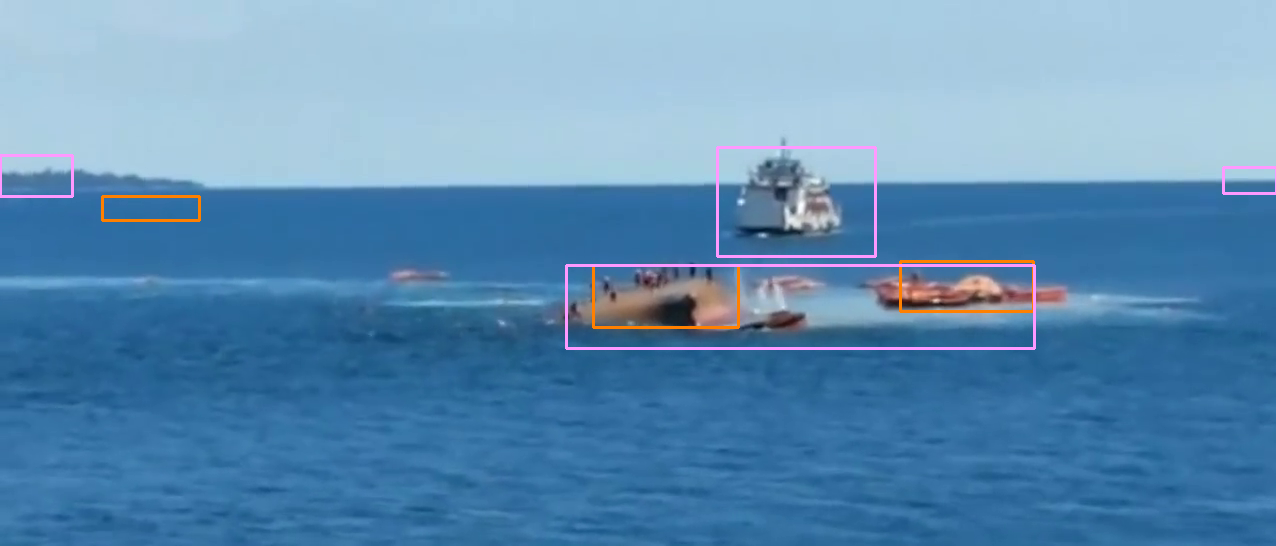}}
\subfigure[Seq.1, Frame 273]{\label{fig3d}\includegraphics[height=16mm,width=35mm]{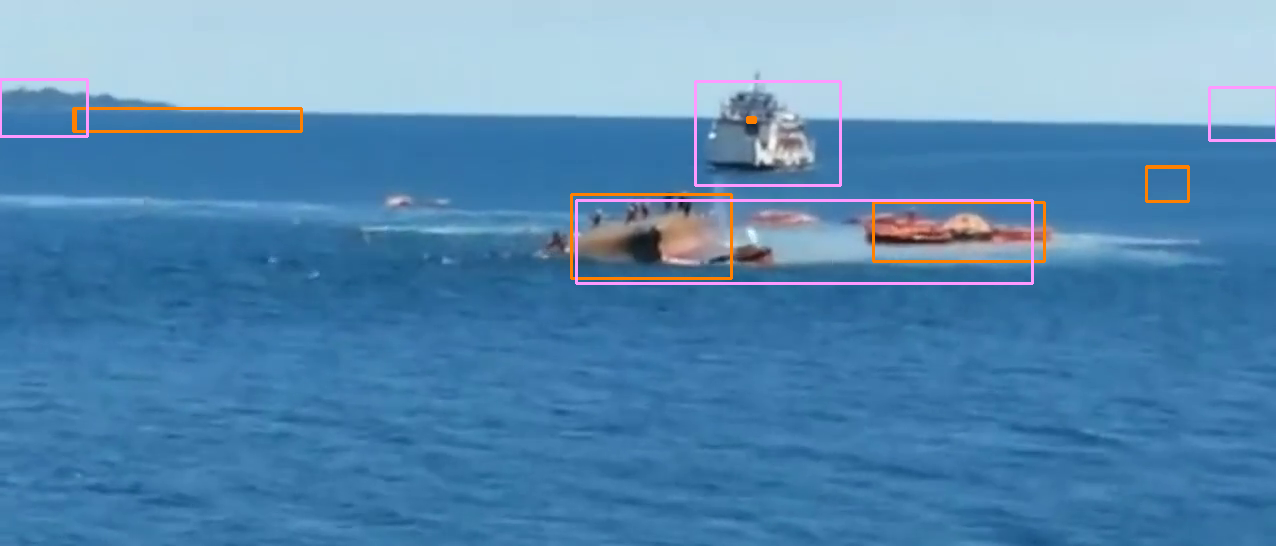}}
\subfigure[Seq.1, Frame 360]{\label{fig3e}\includegraphics[height=16mm,width=35mm]{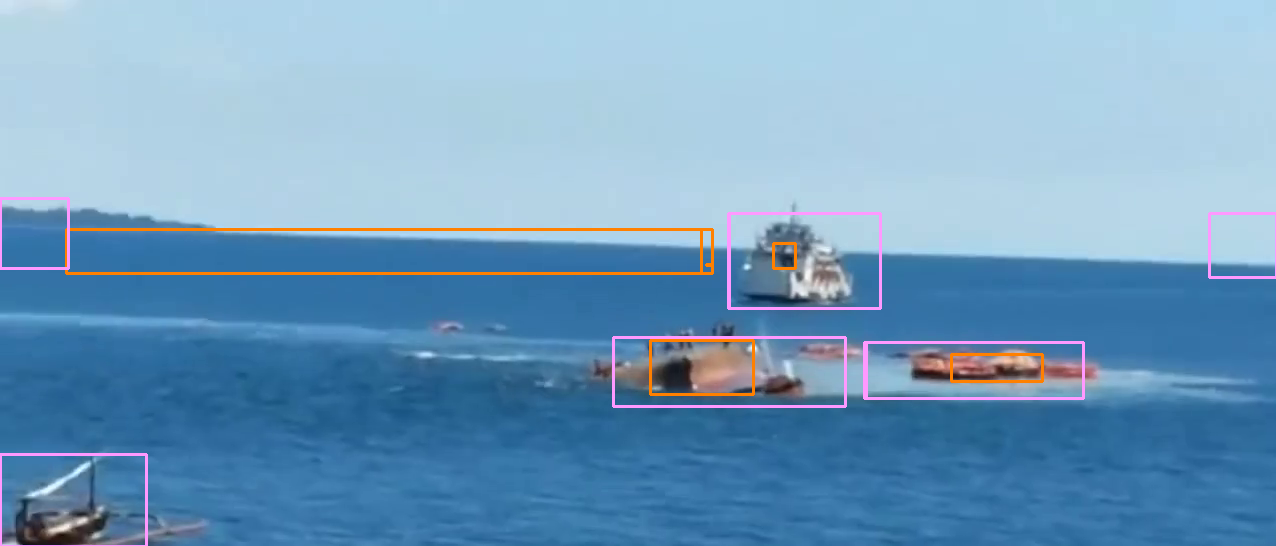}}

\includegraphics[height=16mm,width=35mm]{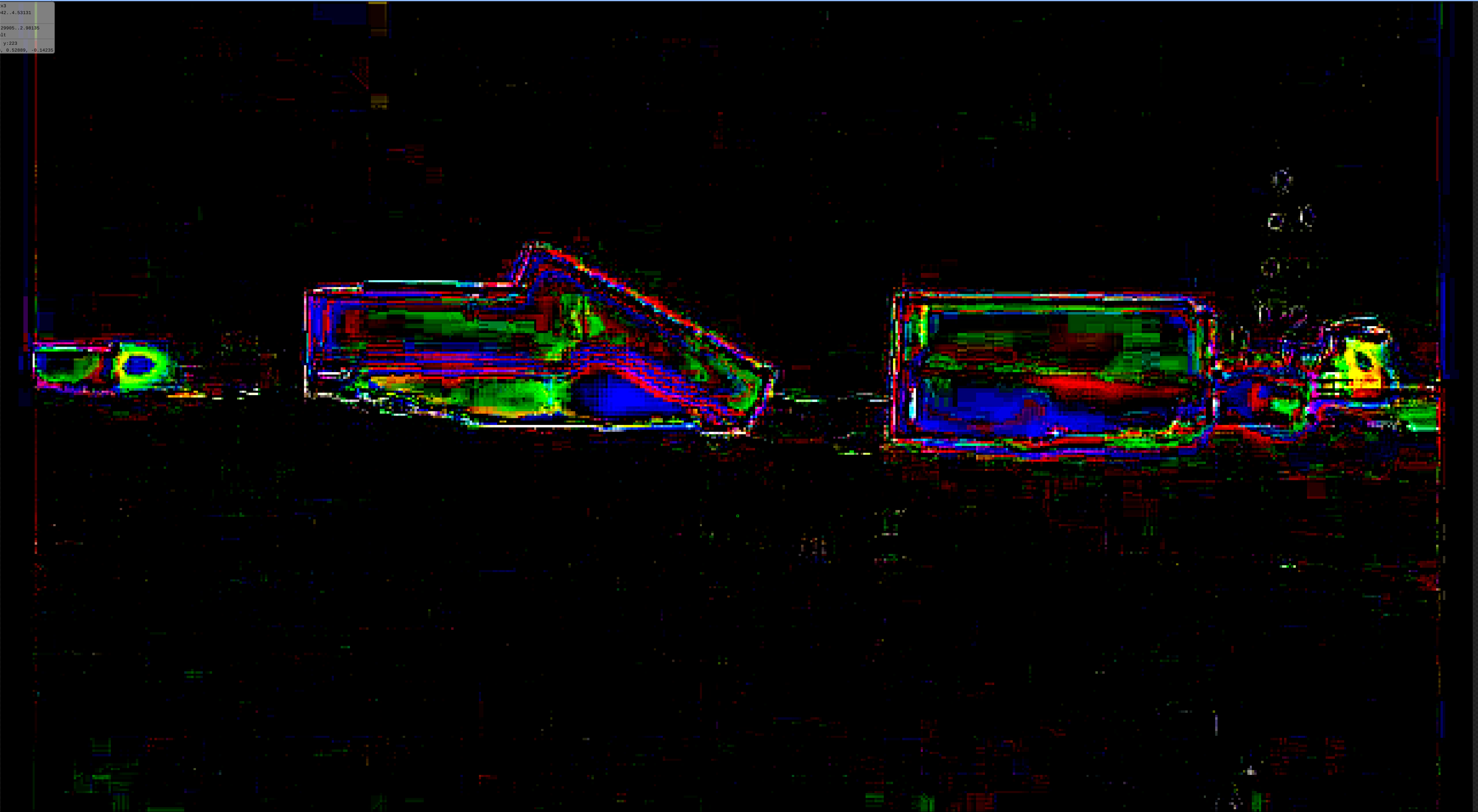}
\includegraphics[height=16mm,width=35mm]{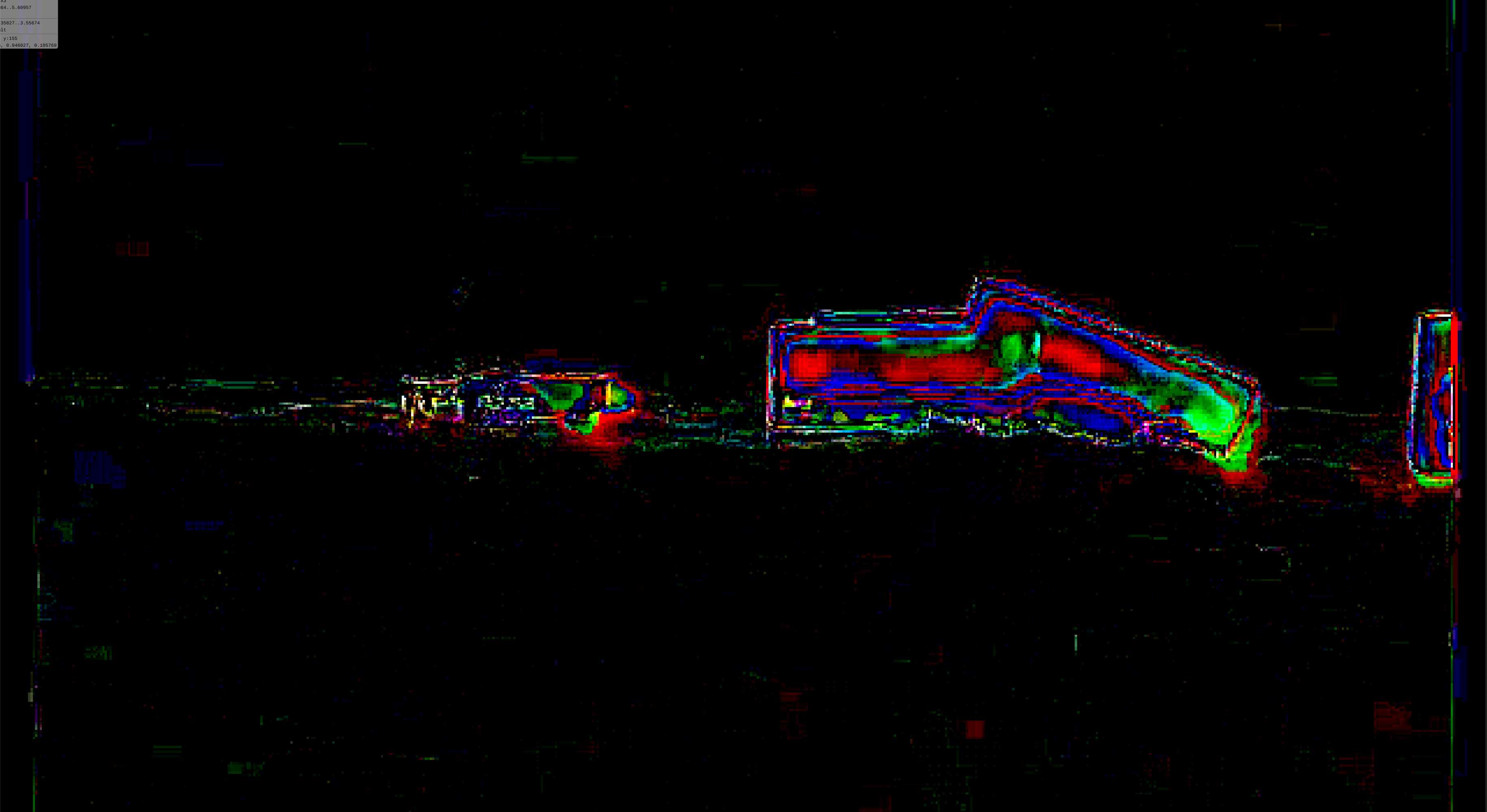}
\includegraphics[height=16mm,width=35mm]{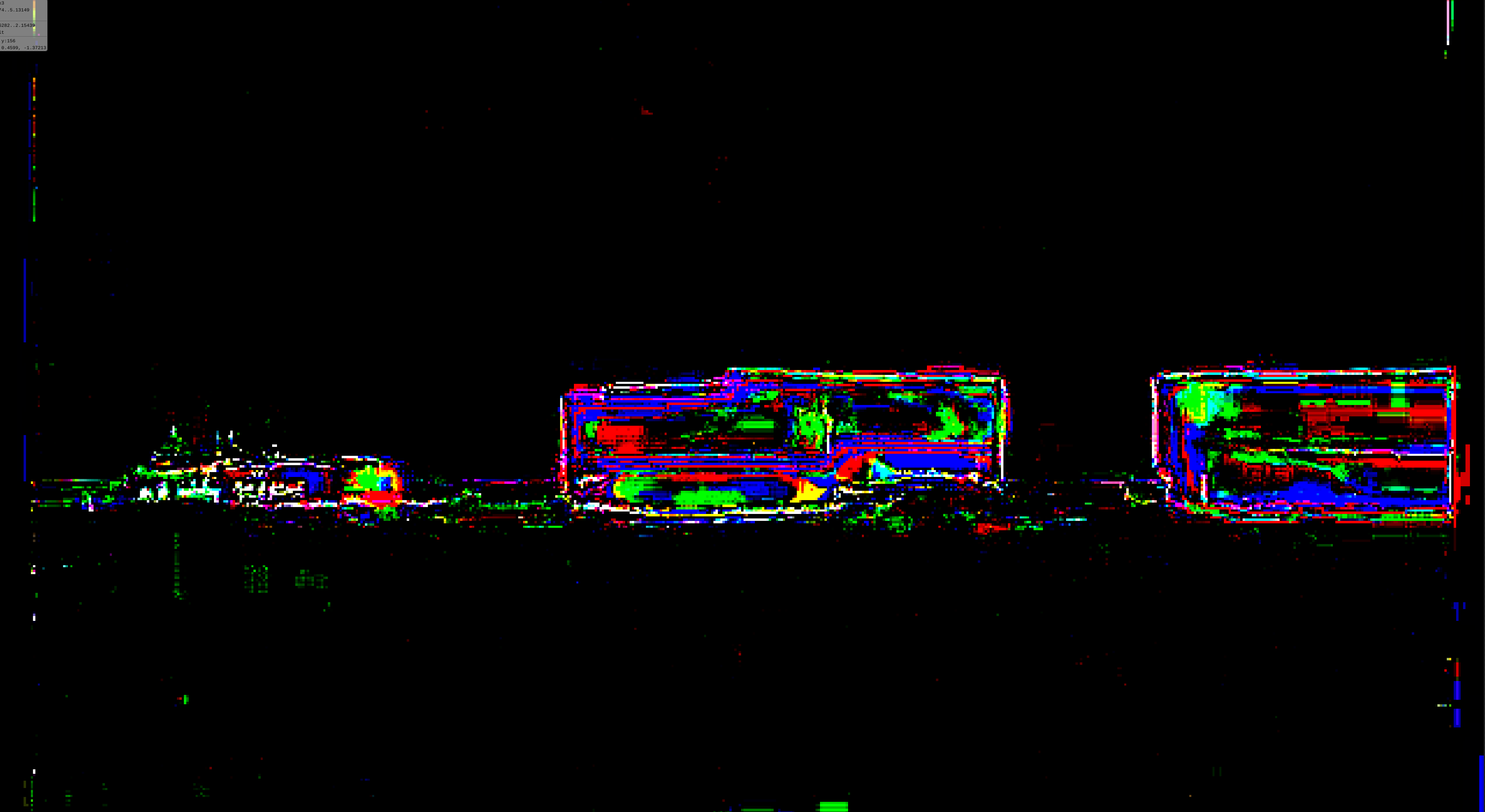}
\includegraphics[height=16mm,width=35mm]{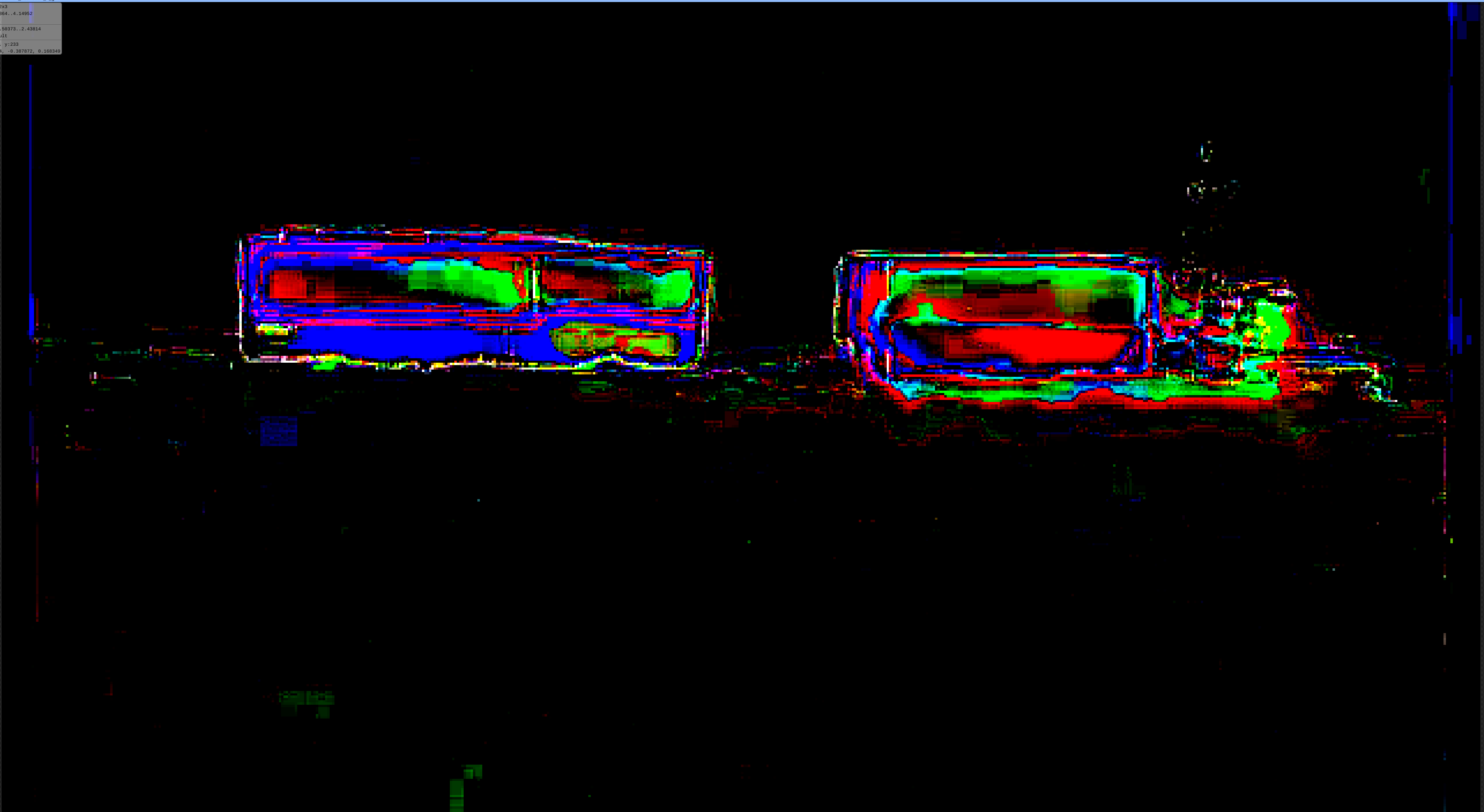}
\includegraphics[height=16mm,width=35mm]{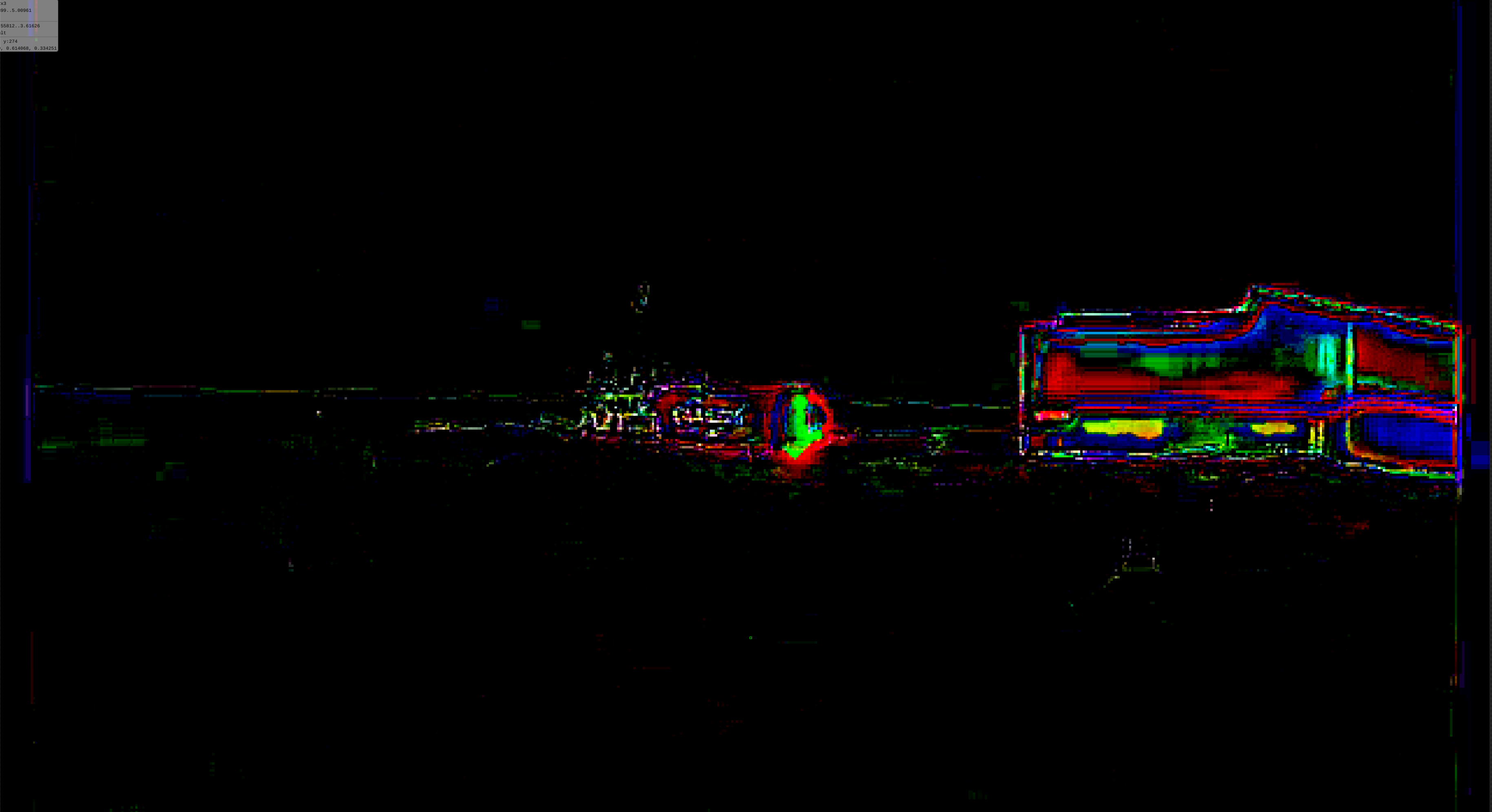}
\includegraphics[height=16mm,width=35mm]{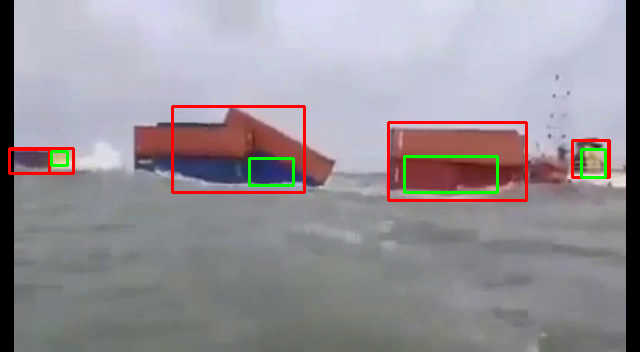}
\includegraphics[height=16mm,width=35mm]{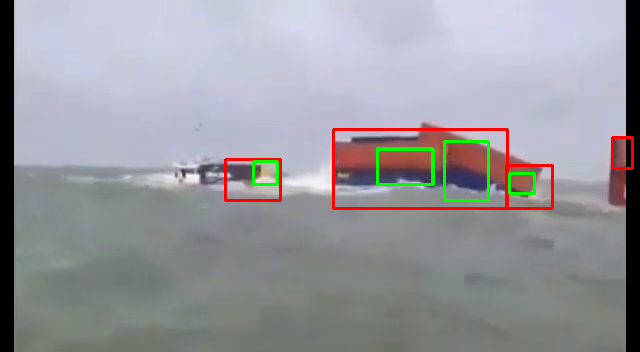}
\includegraphics[height=16mm,width=35mm]{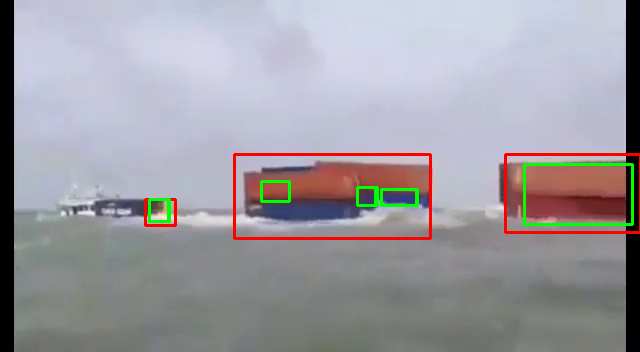}
\includegraphics[height=16mm,width=35mm]{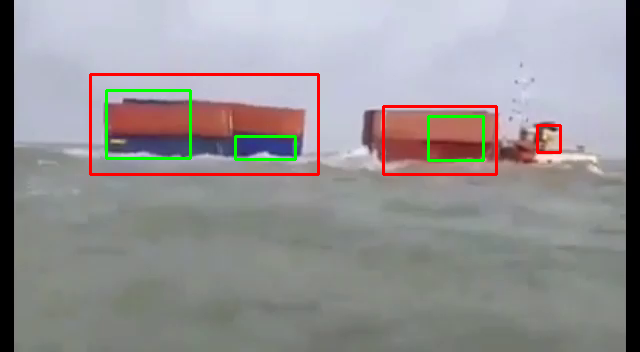}
\includegraphics[height=16mm,width=35mm]{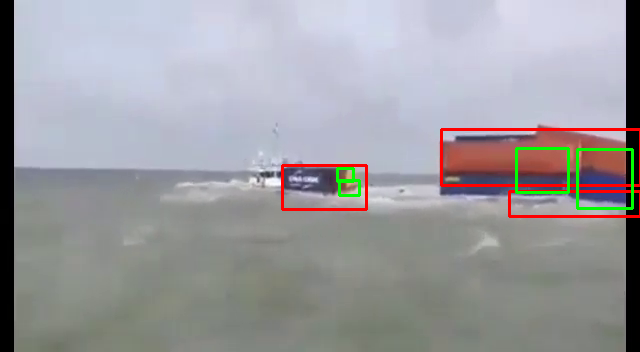}
\subfigure[Seq.2, Frame 0]{\label{fig3f}\includegraphics[height=16mm,width=35mm]{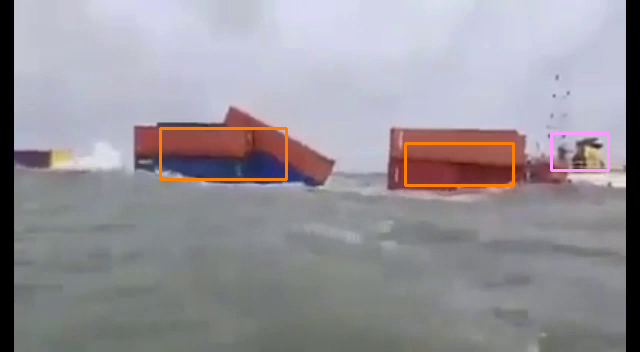}}
\subfigure[Seq.2, Frame 50]{\label{fig3g}\includegraphics[height=16mm,width=35mm]{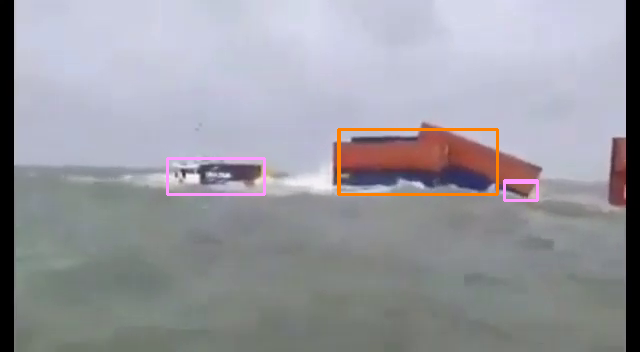}}
\subfigure[Seq.2, Frame 105]{\label{fig3h}\includegraphics[height=16mm,width=35mm]{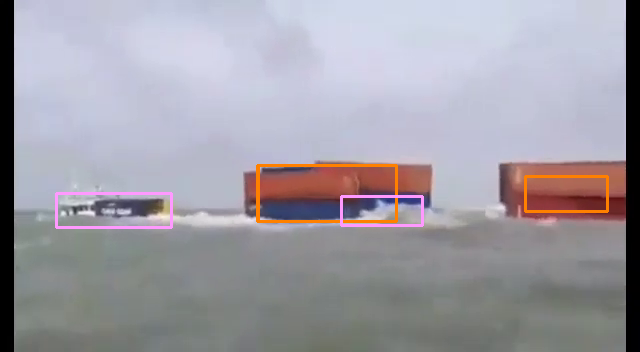}}
\subfigure[Seq.2, Frame 150]{\label{fig3i}\includegraphics[height=16mm,width=35mm]{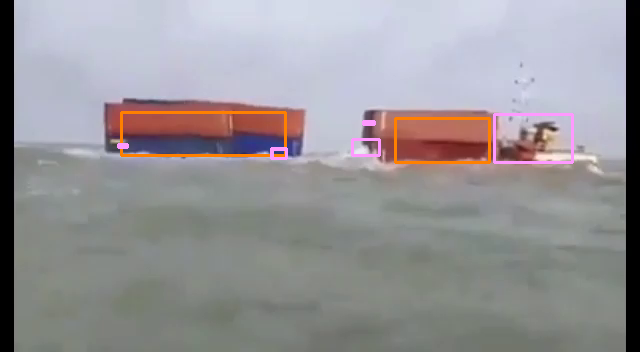}}
\subfigure[Seq.2, Frame 239]{\label{fig3j}\includegraphics[height=16mm,width=35mm]{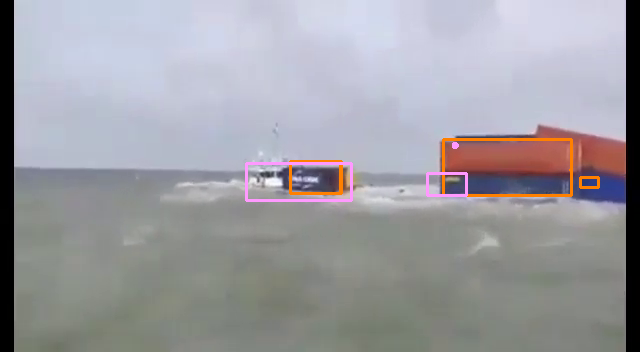}}

\includegraphics[height=16mm,width=35mm]{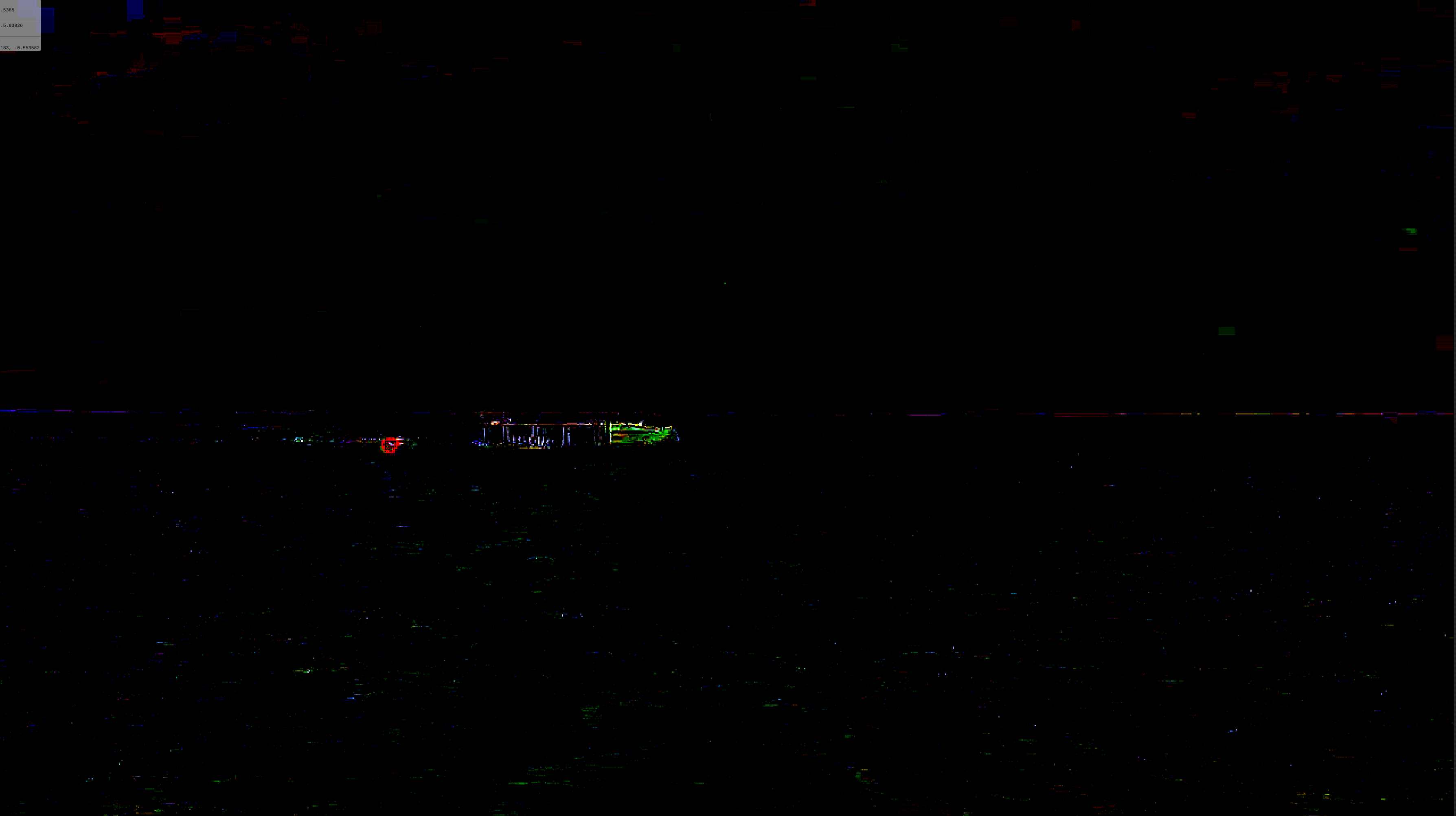}
\includegraphics[height=16mm,width=35mm]{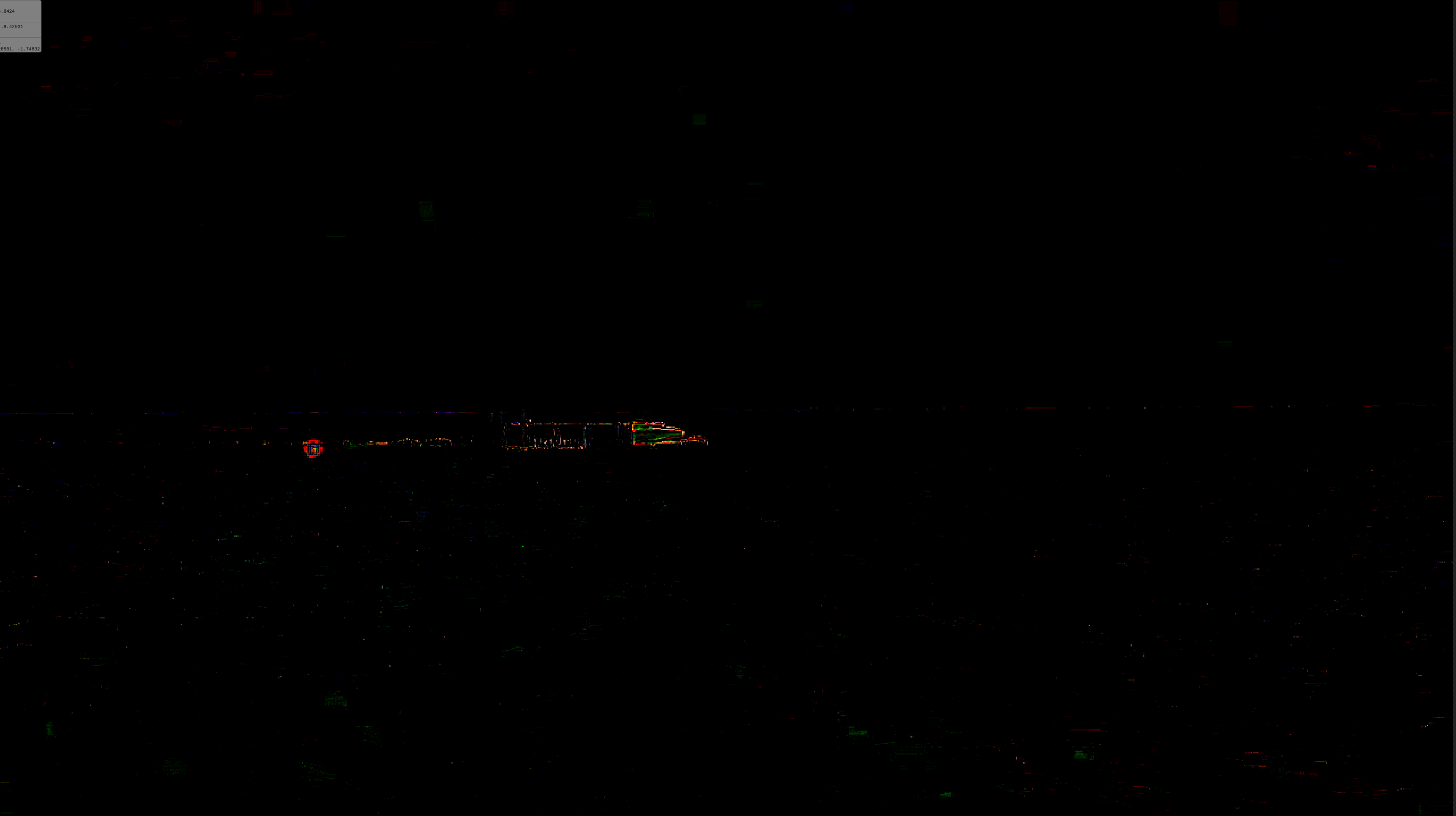}
\includegraphics[height=16mm,width=35mm]{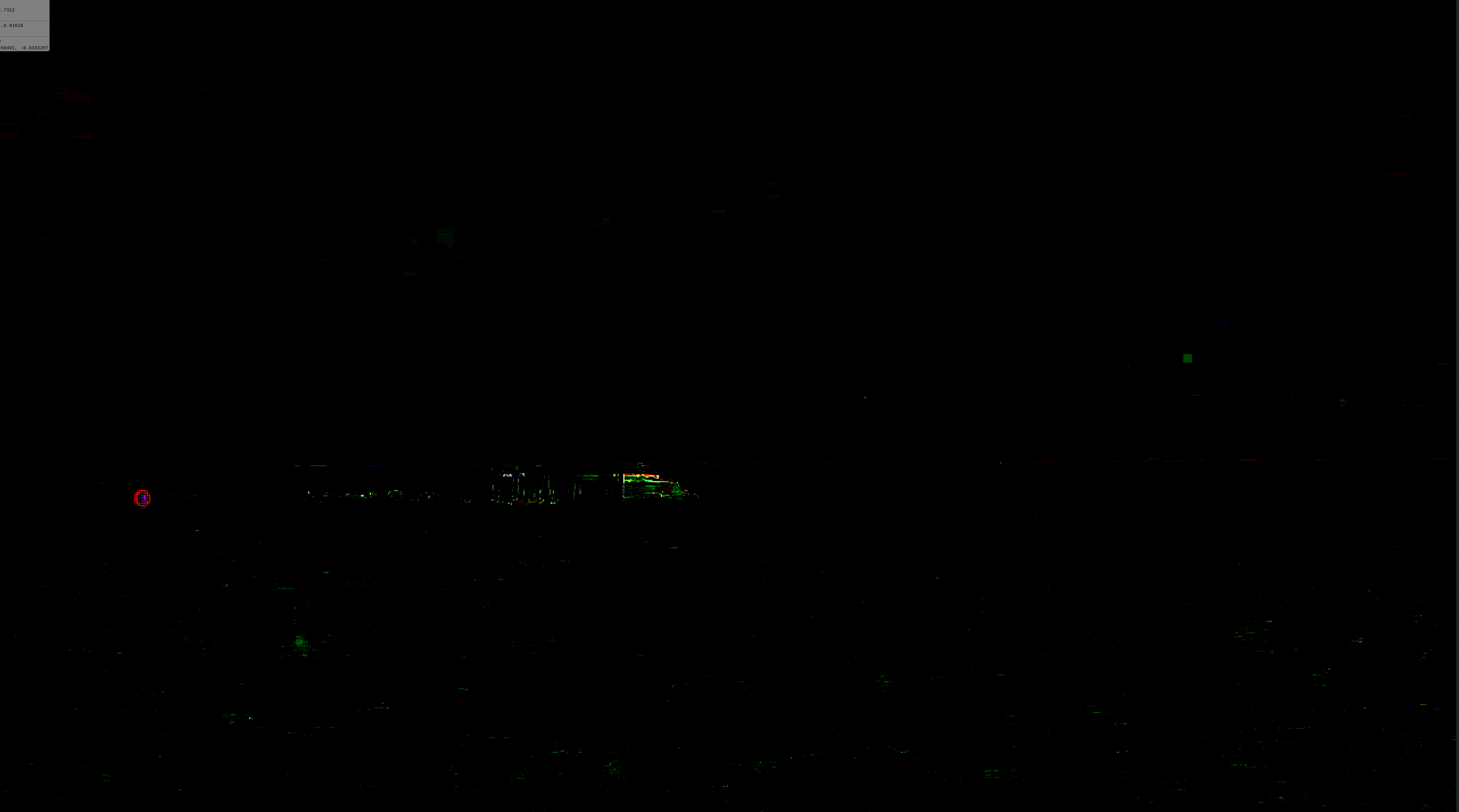}
\includegraphics[height=16mm,width=35mm]{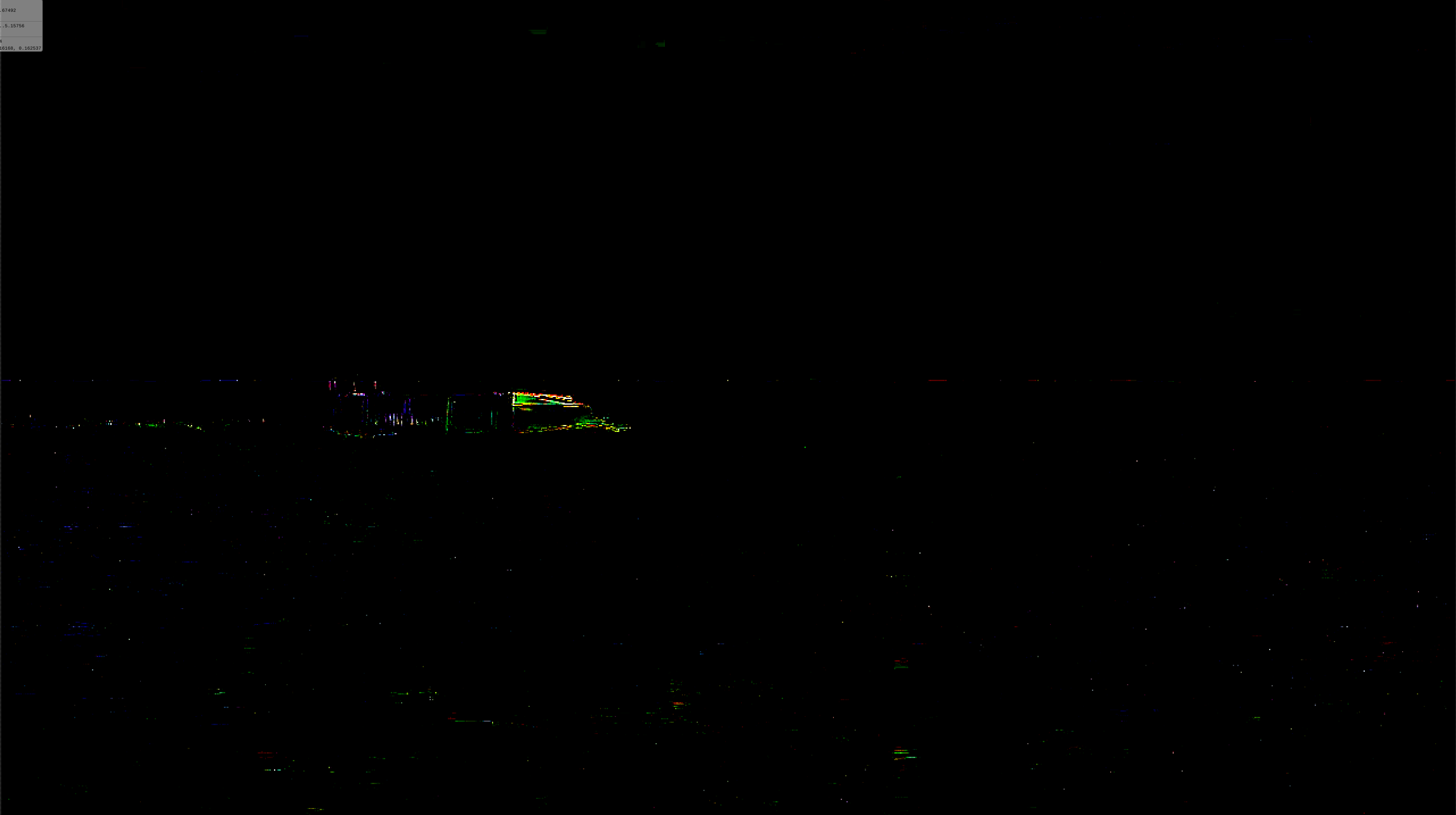}
\includegraphics[height=16mm,width=35mm]{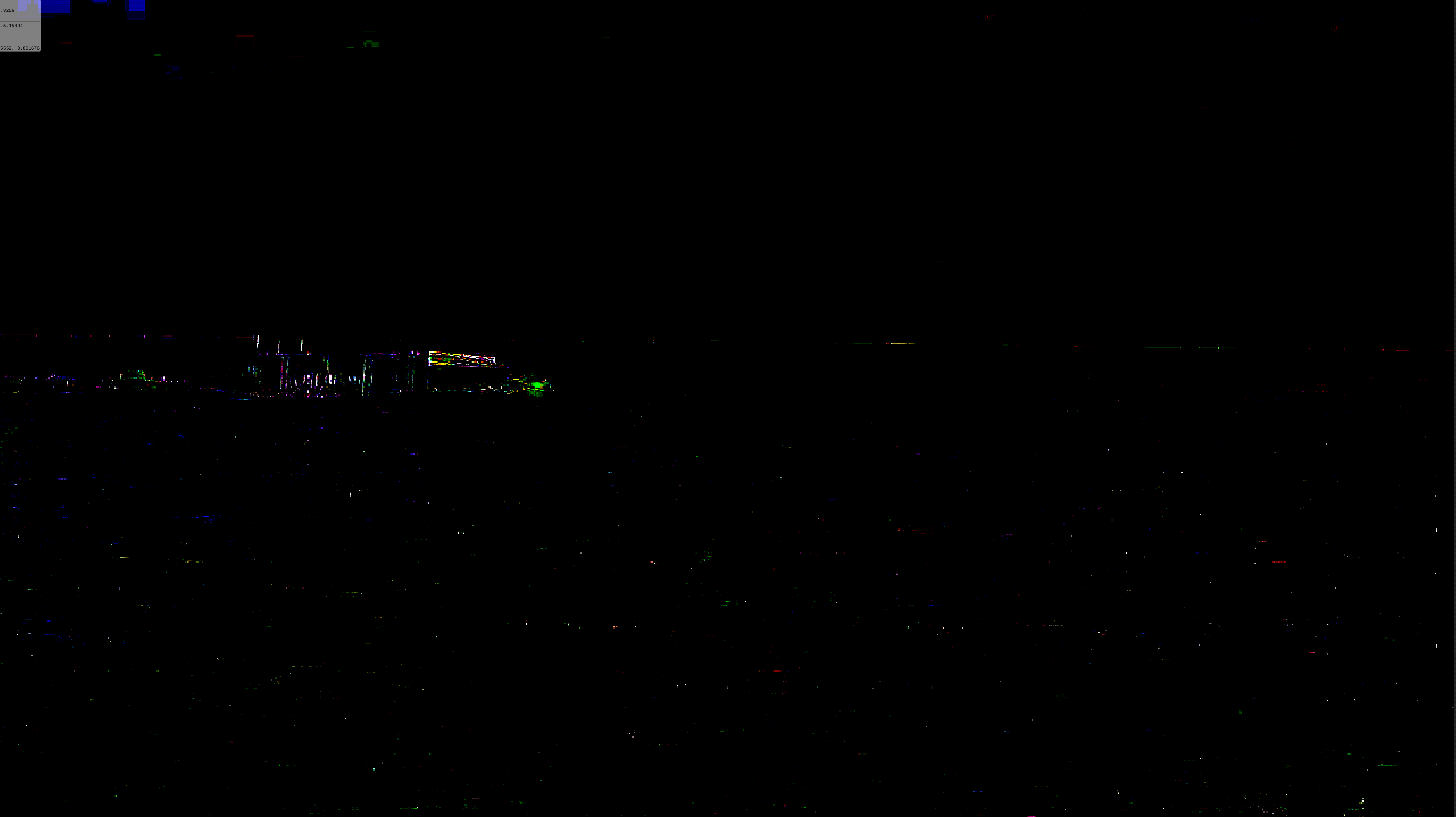}
\includegraphics[height=16mm,width=35mm]{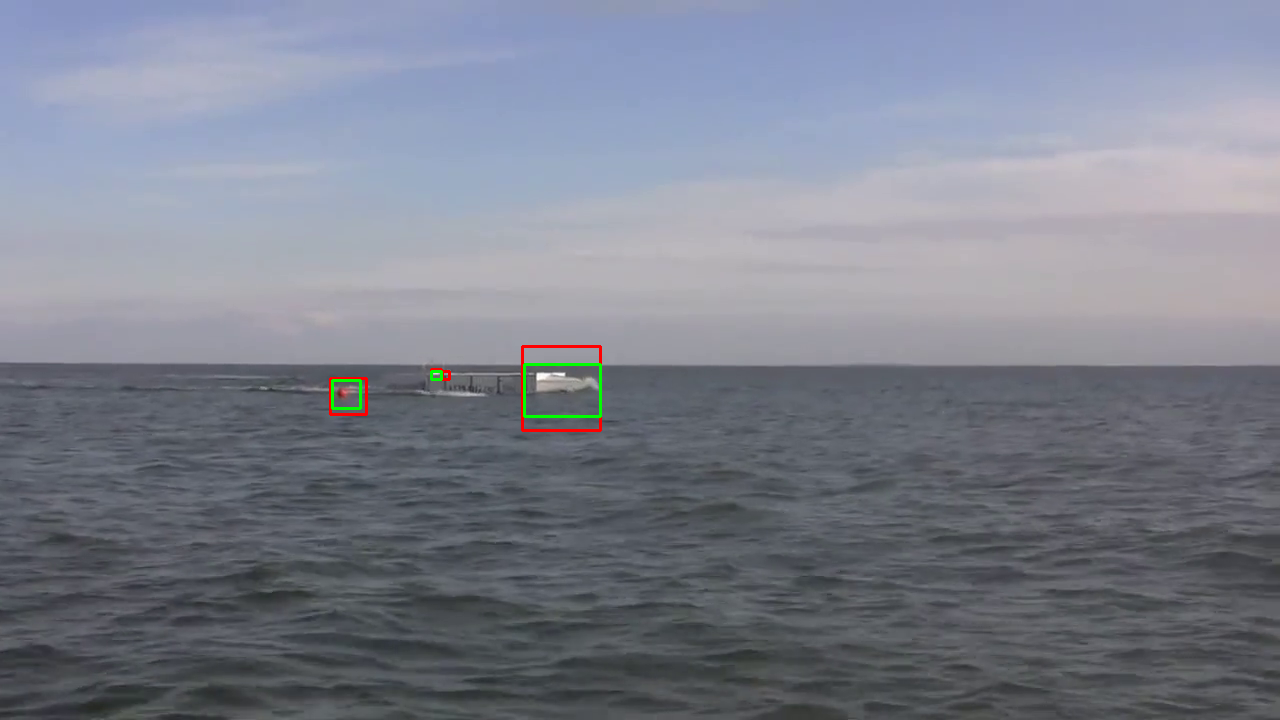}
\includegraphics[height=16mm,width=35mm]{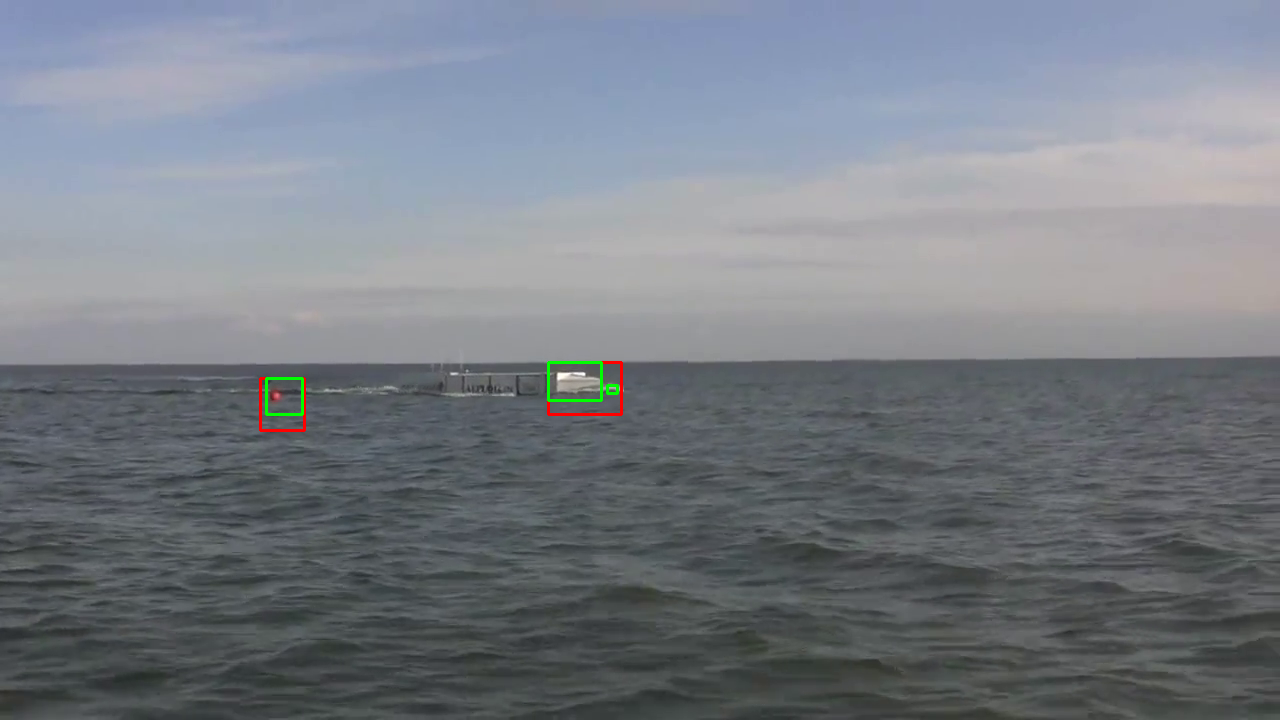}
\includegraphics[height=16mm,width=35mm]{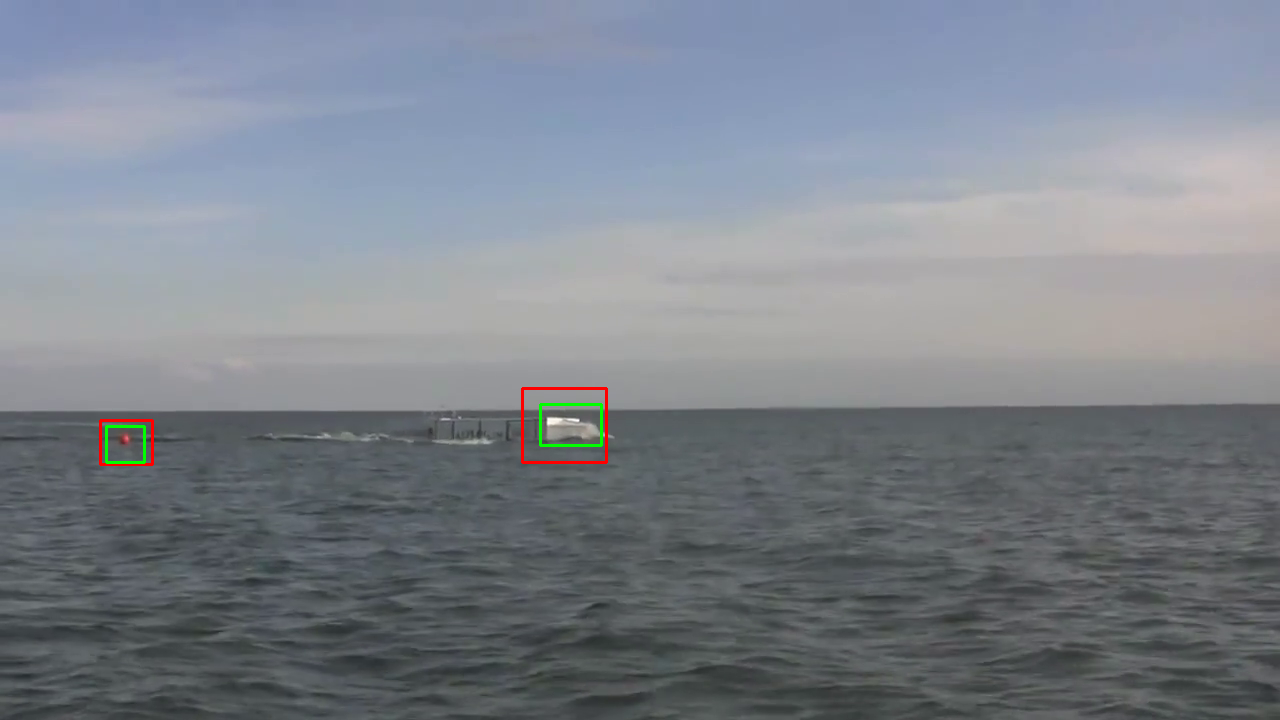}
\includegraphics[height=16mm,width=35mm]{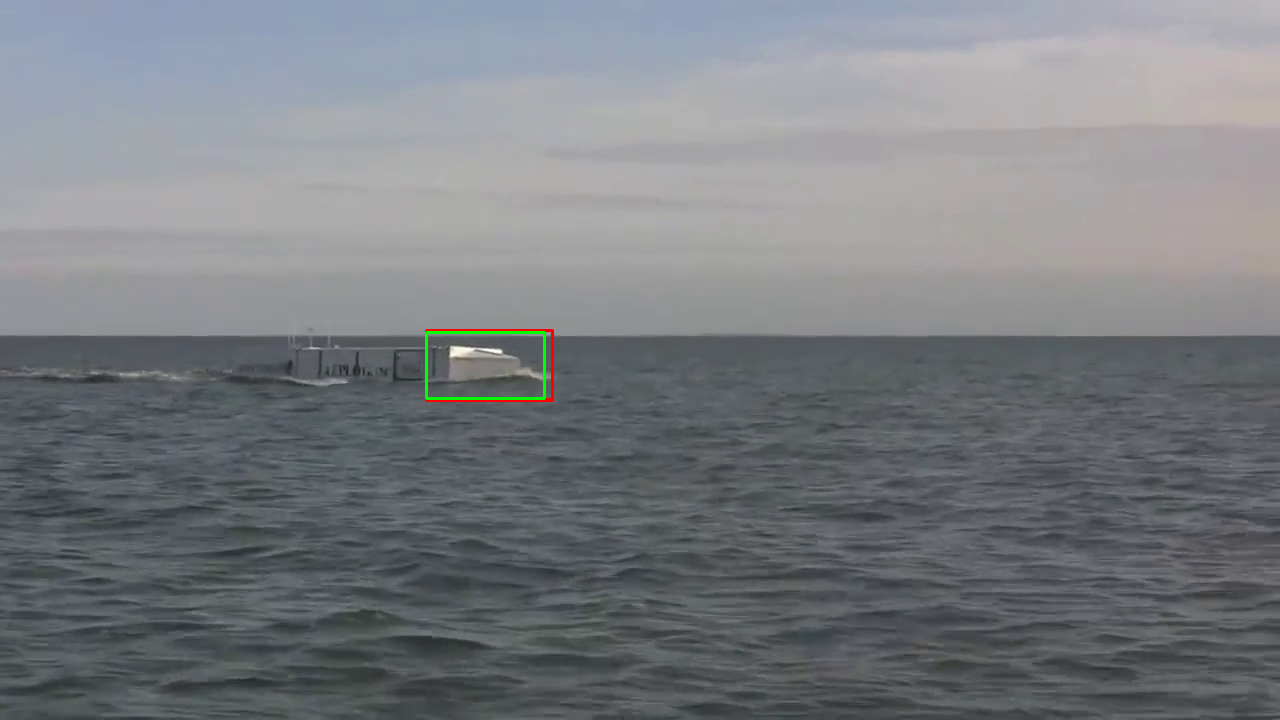}
\includegraphics[height=16mm,width=35mm]{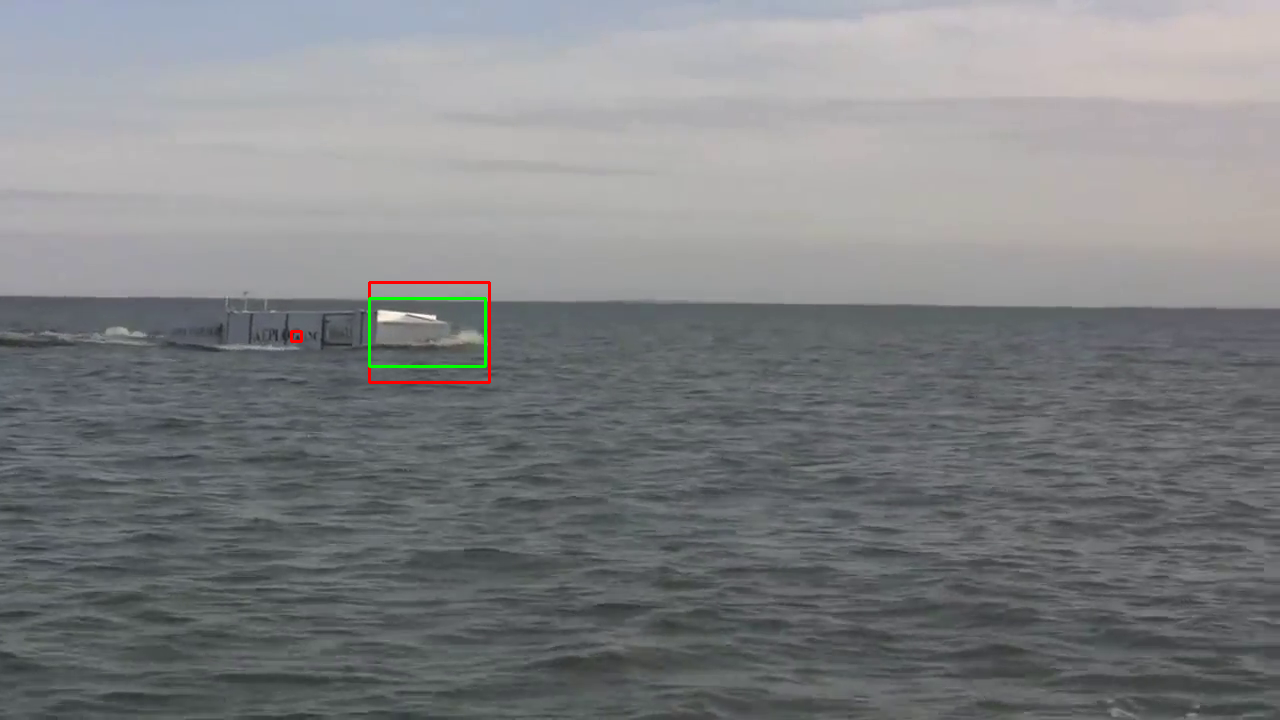}
\subfigure[Seq.3, Frame 13]{\label{fig3k}\includegraphics[height=16mm,width=35mm]{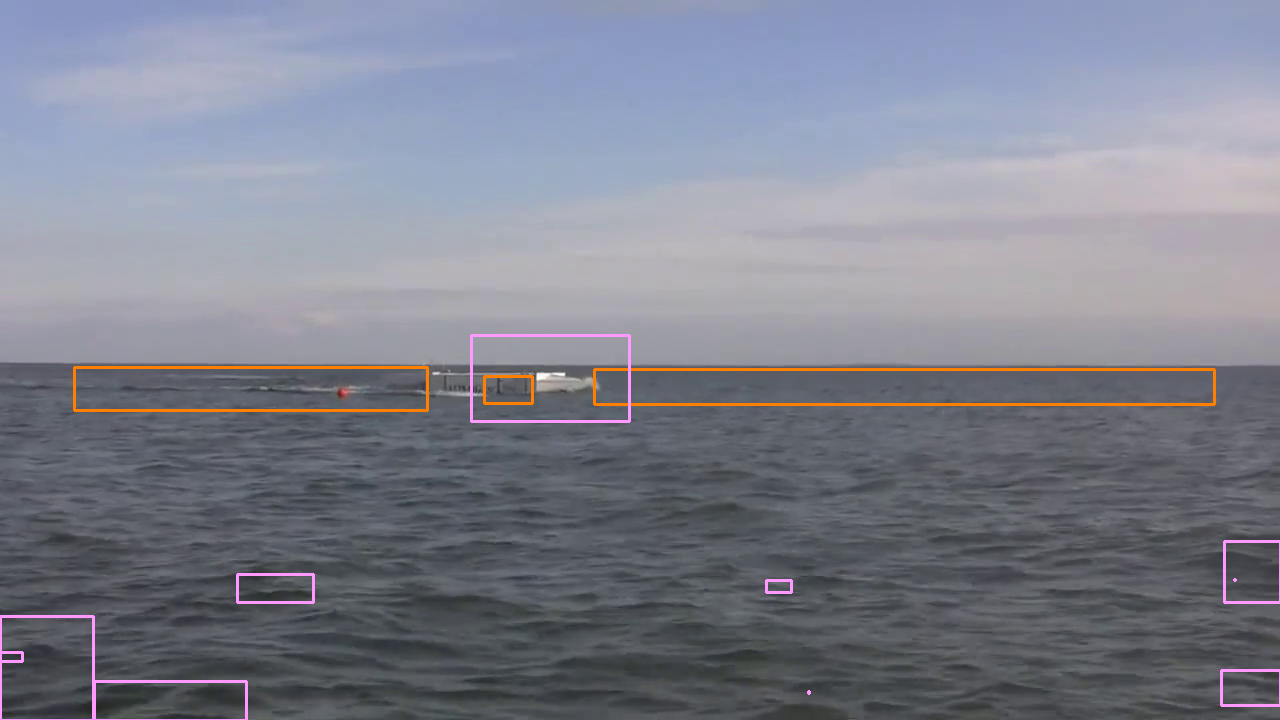}}
\subfigure[Seq.3, Frame 64]{\label{fig3l}\includegraphics[height=16mm,width=35mm]{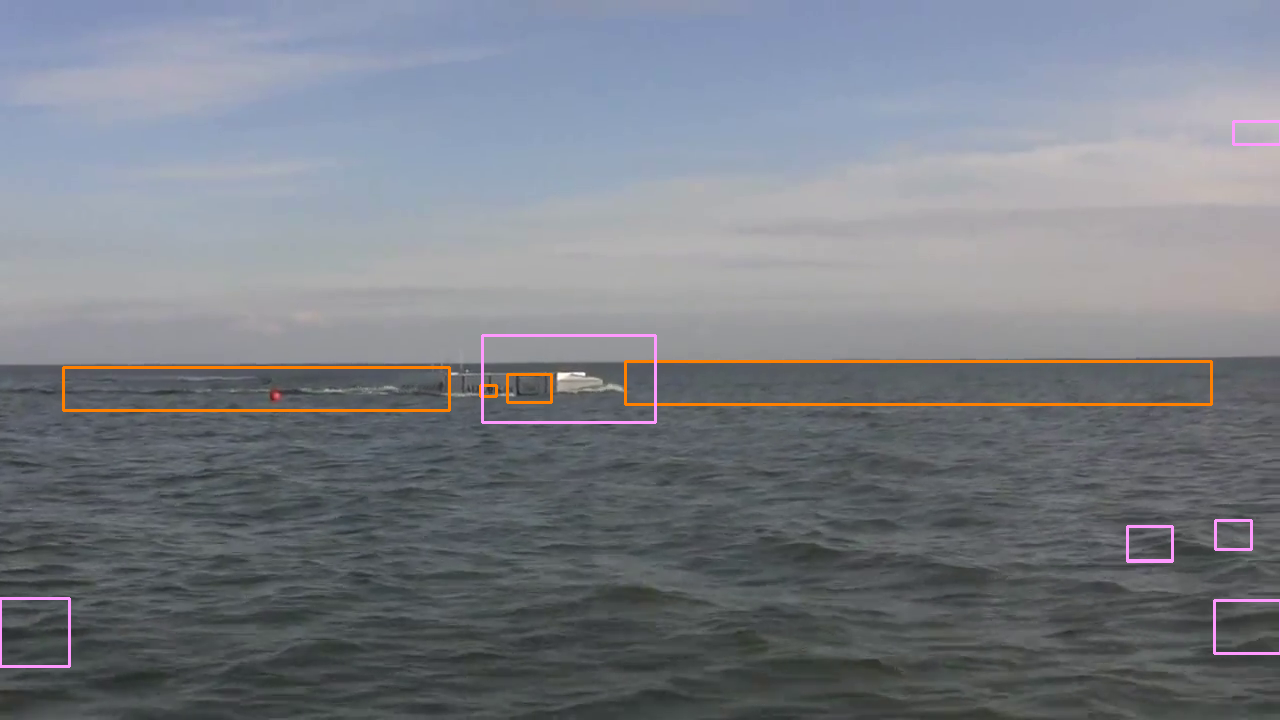}}
\subfigure[Seq.3, Frame 149]{\label{fig3m}\includegraphics[height=16mm,width=35mm]{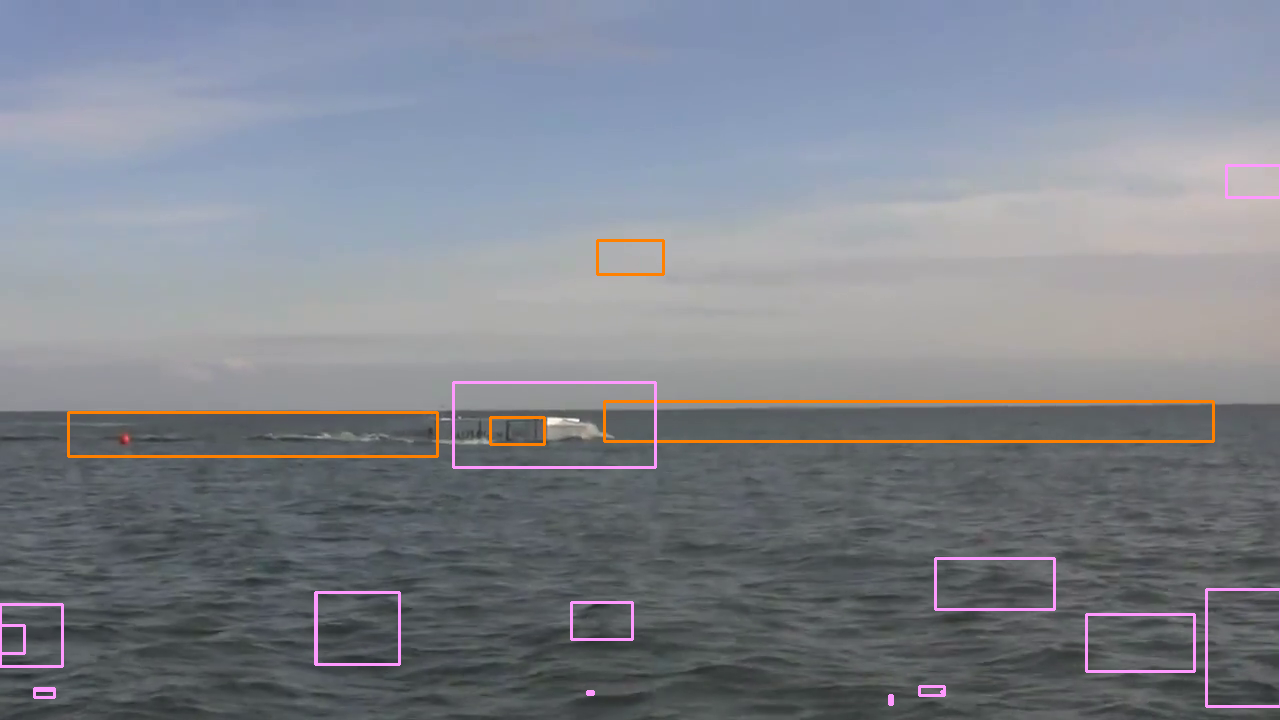}}
\subfigure[Seq.3, Frame 227]{\label{fig3n}\includegraphics[height=16mm,width=35mm]{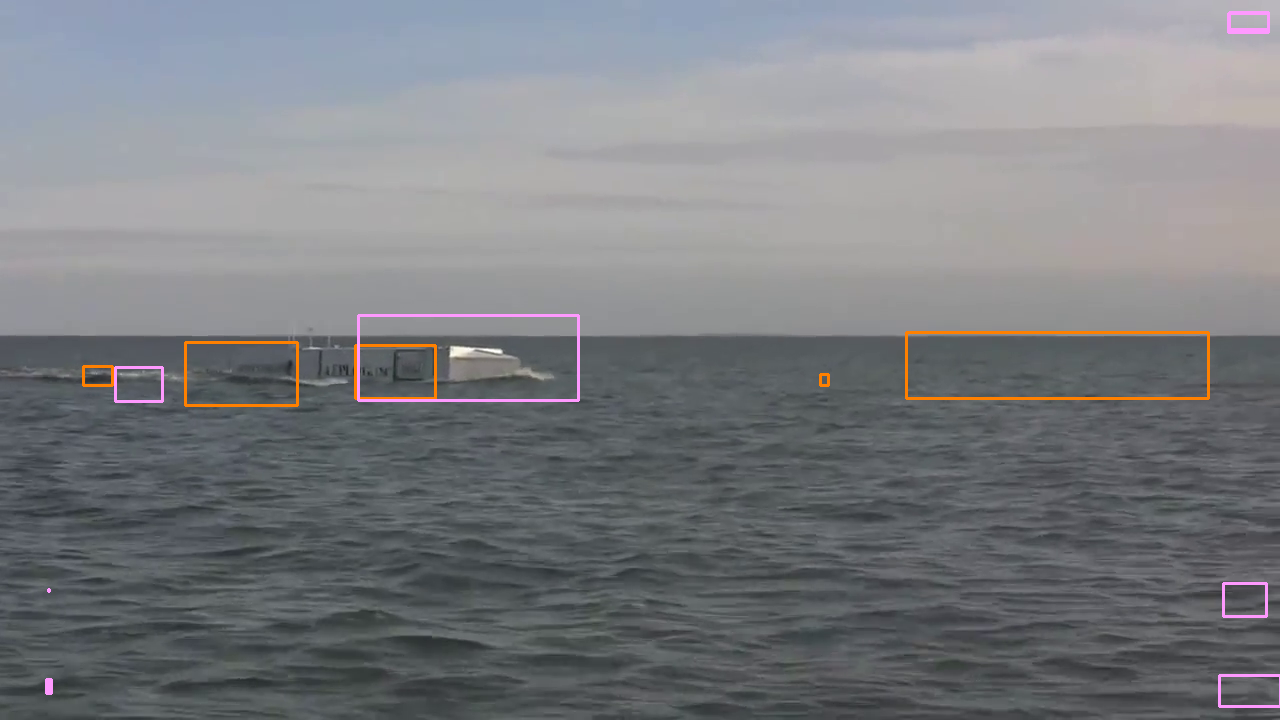}}
\subfigure[Seq.3, Frame 288]{\label{fig3o}\includegraphics[height=16mm,width=35mm]{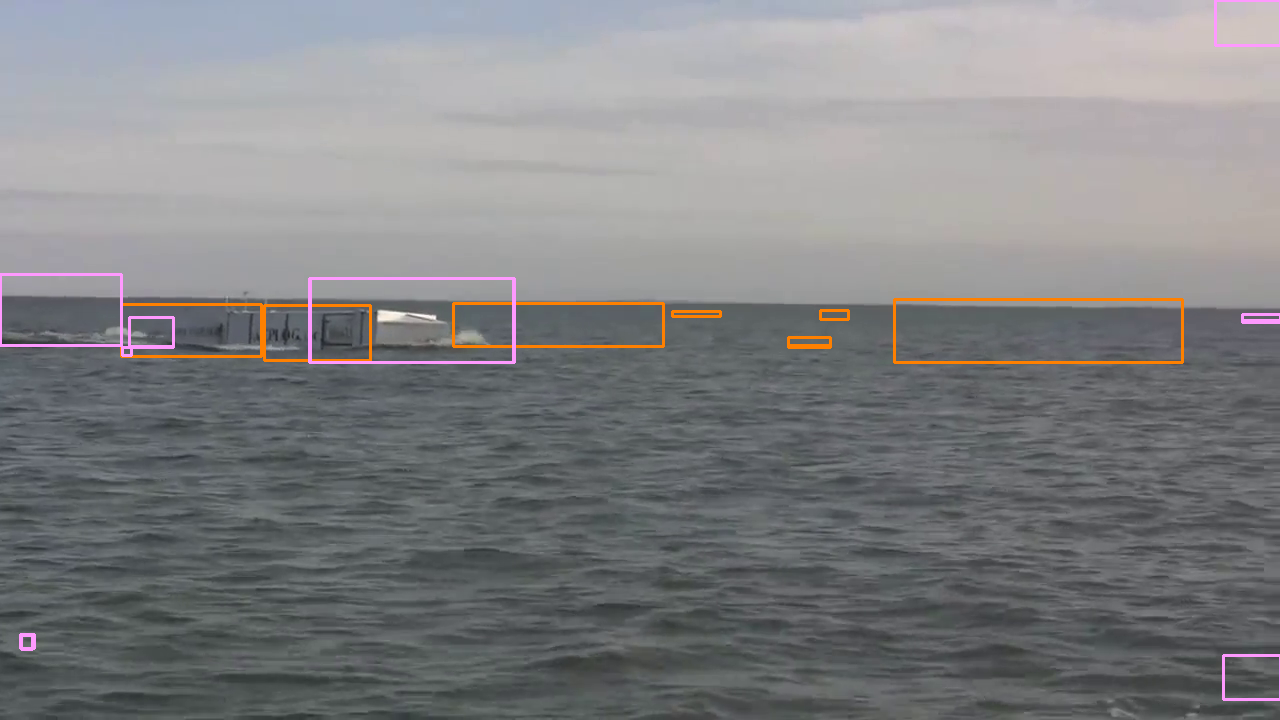}}

\includegraphics[height=16mm,width=35mm]{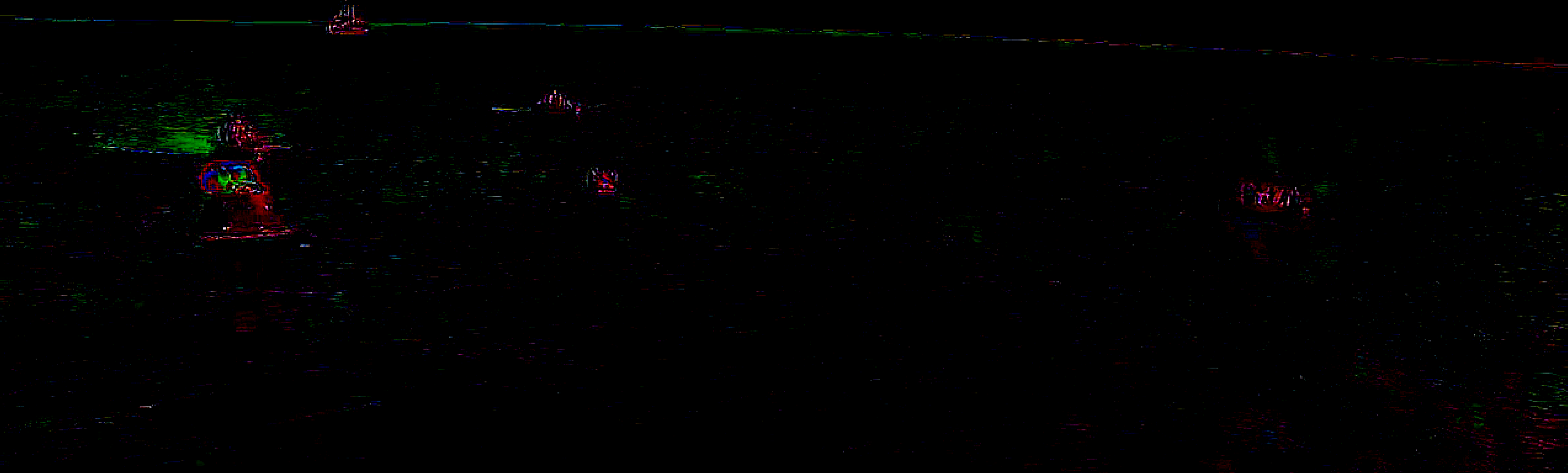}
\includegraphics[height=16mm,width=35mm]{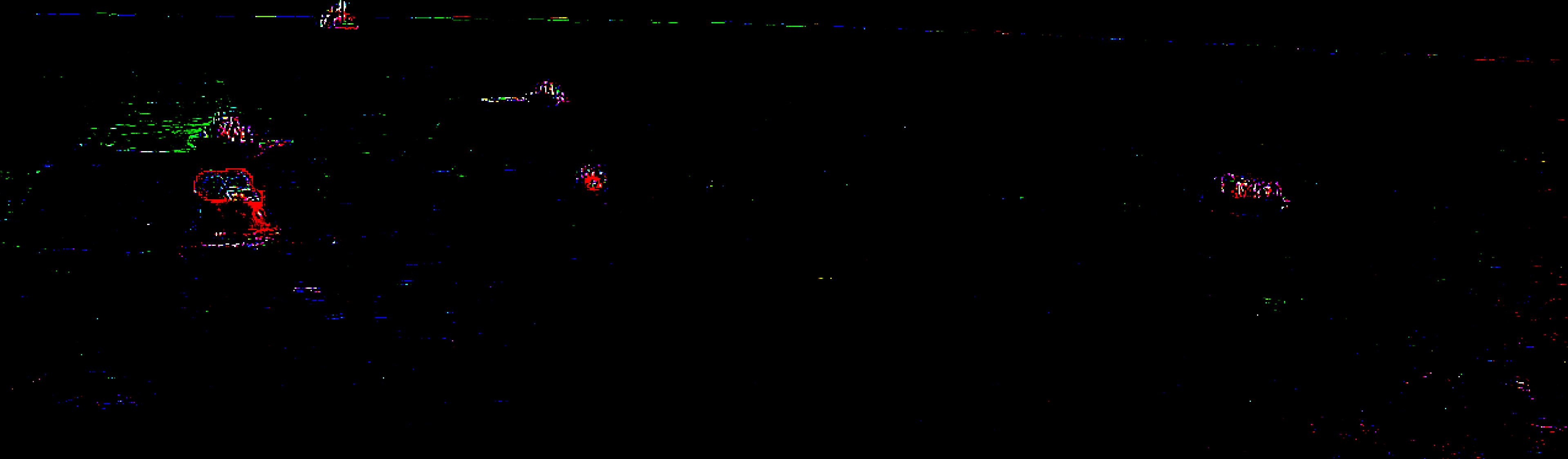}
\includegraphics[height=16mm,width=35mm]{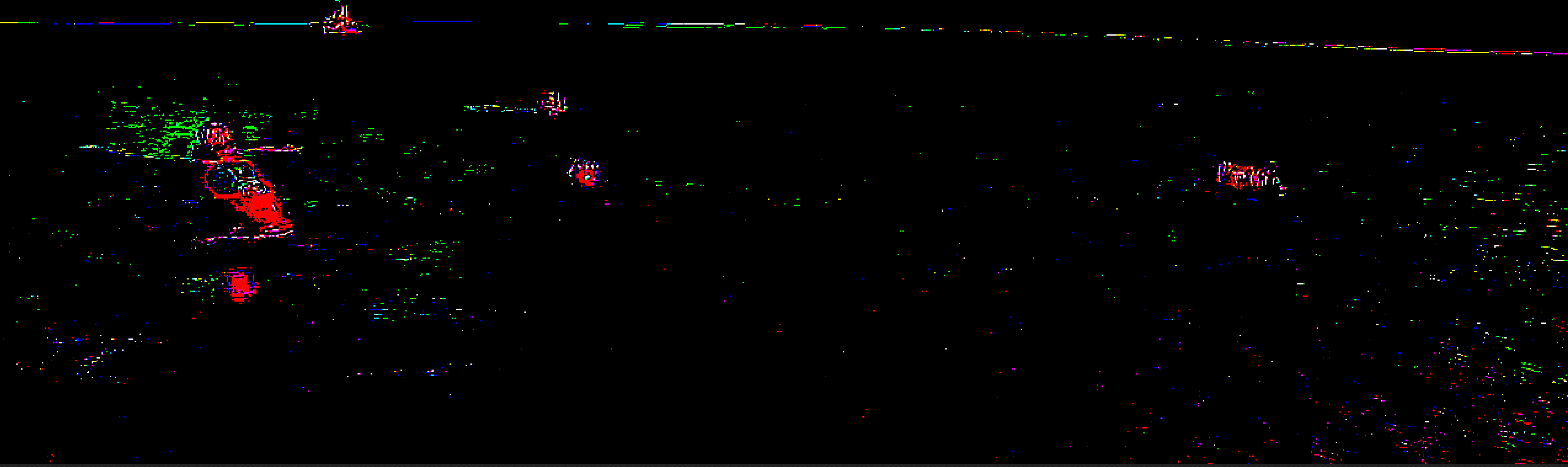}
\includegraphics[height=16mm,width=35mm]{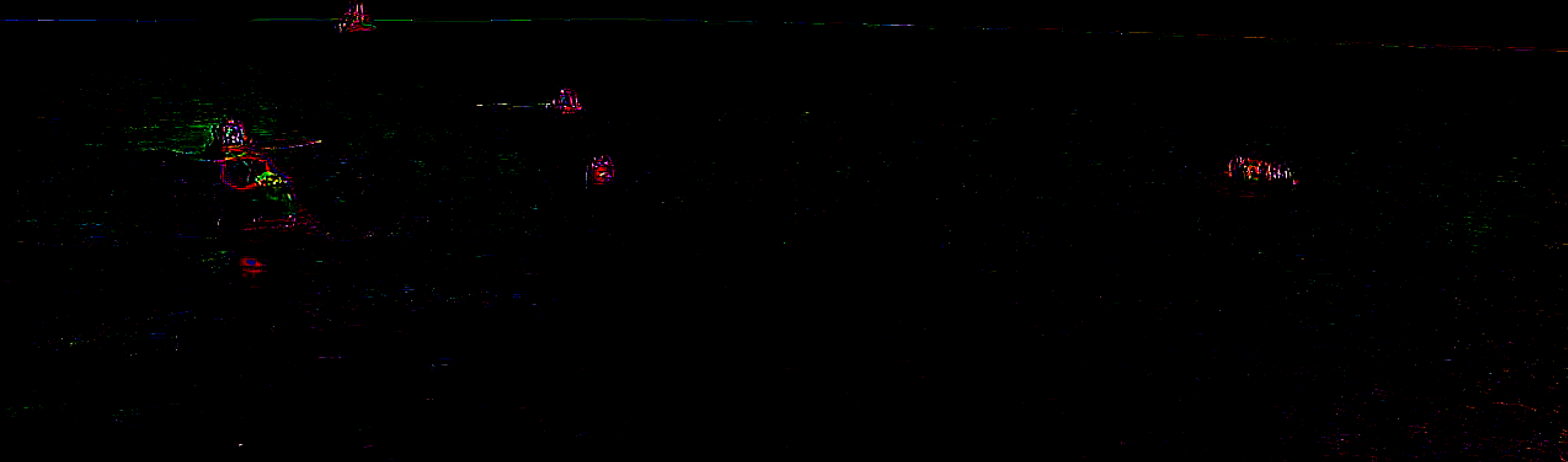}
\includegraphics[height=16mm,width=35mm]{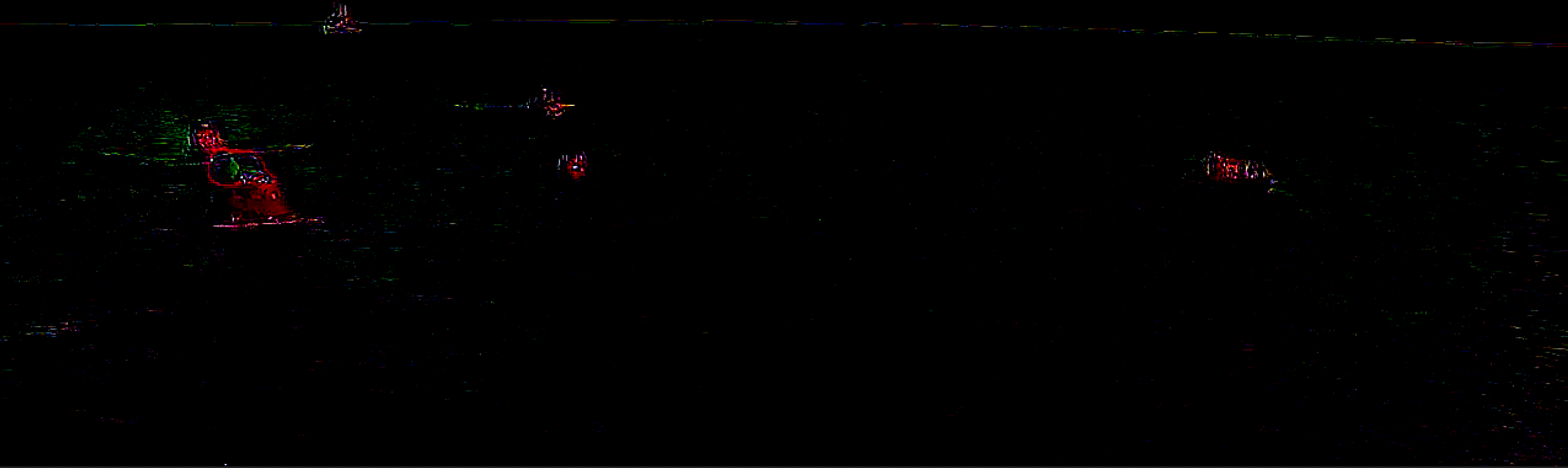}
\includegraphics[height=16mm,width=35mm]{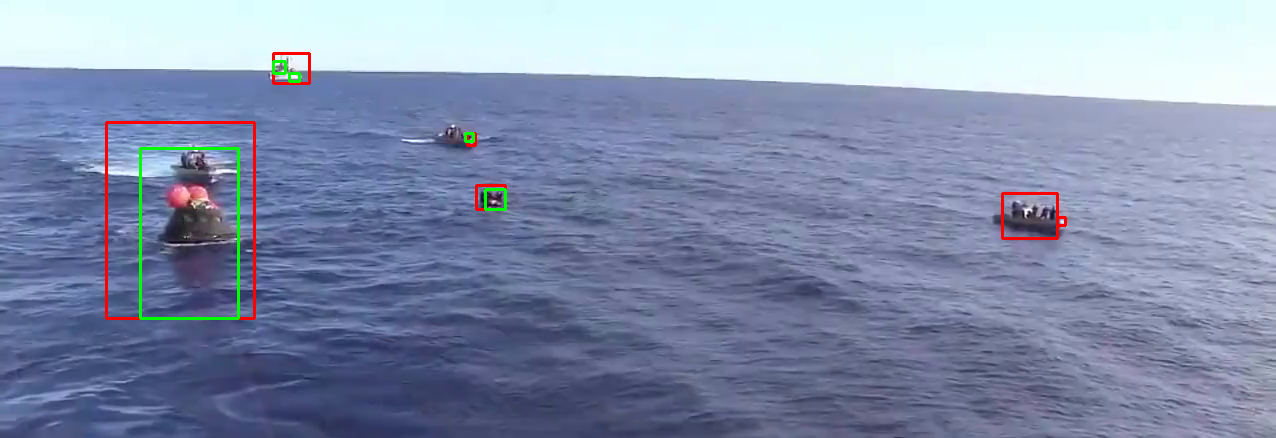}
\includegraphics[height=16mm,width=35mm]{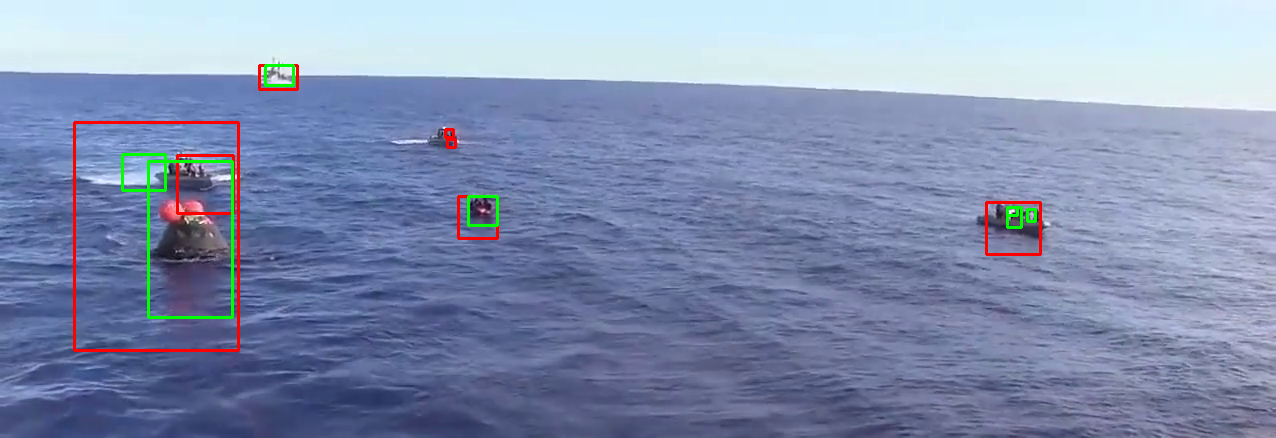}
\includegraphics[height=16mm,width=35mm]{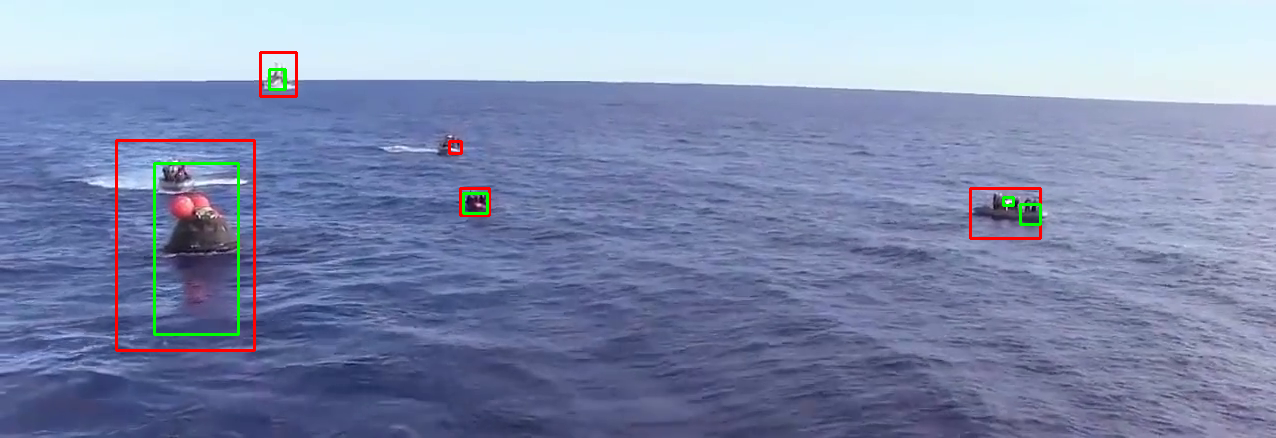}
\includegraphics[height=16mm,width=35mm]{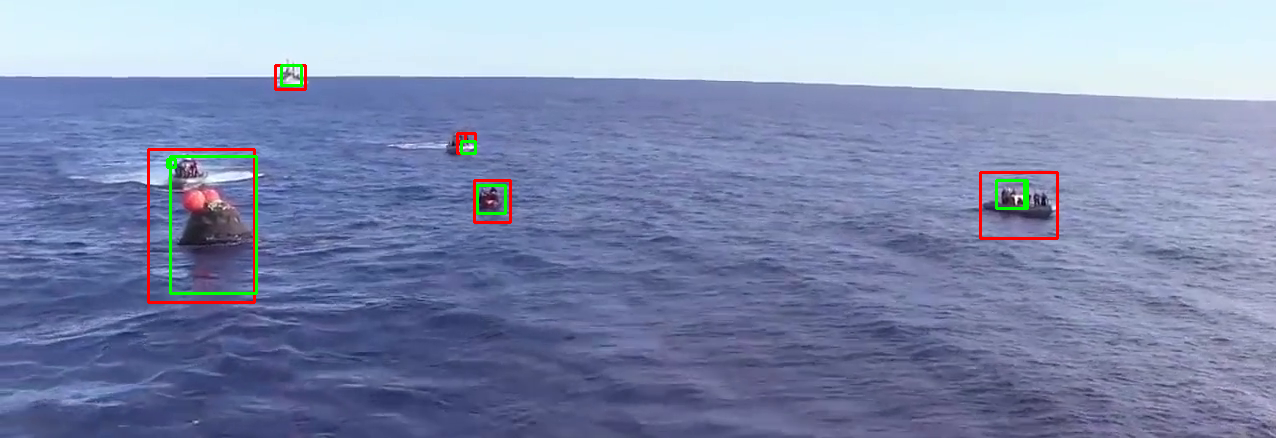}
\includegraphics[height=16mm,width=35mm]{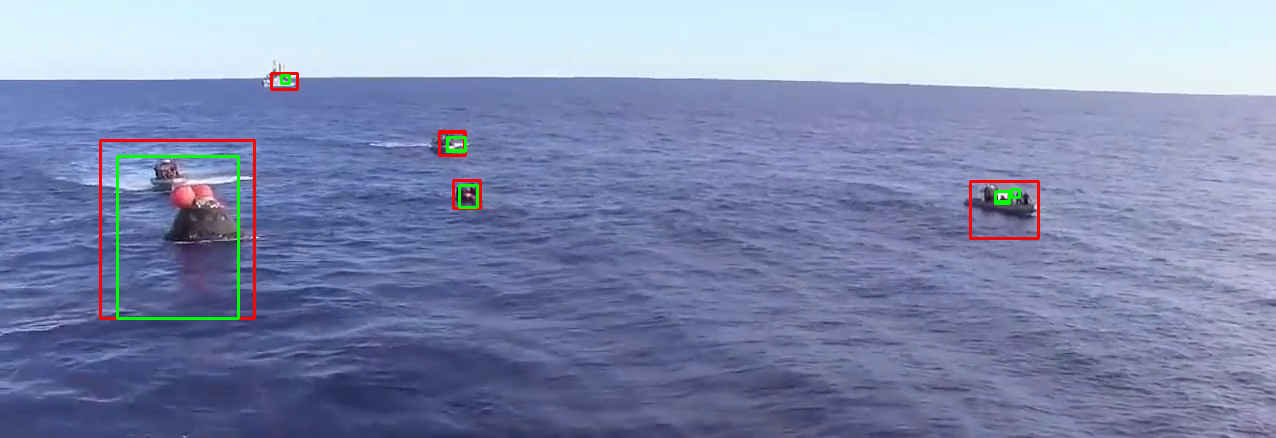}
\subfigure[Seq.4, Frame 3]{\label{fig3p}\includegraphics[height=16mm,width=35mm]{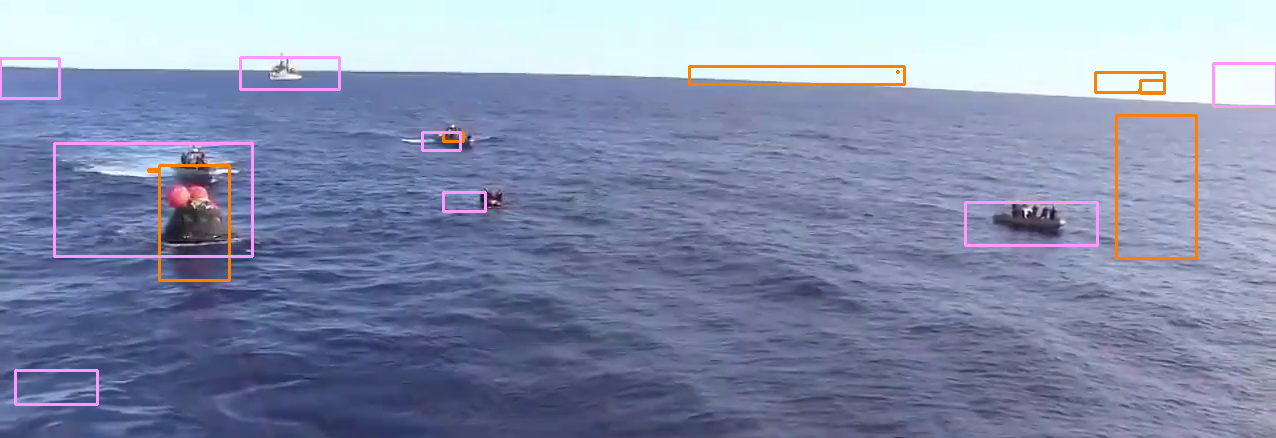}}
\subfigure[Seq.4, Frame 67]{\label{fig3q}\includegraphics[height=16mm,width=35mm]{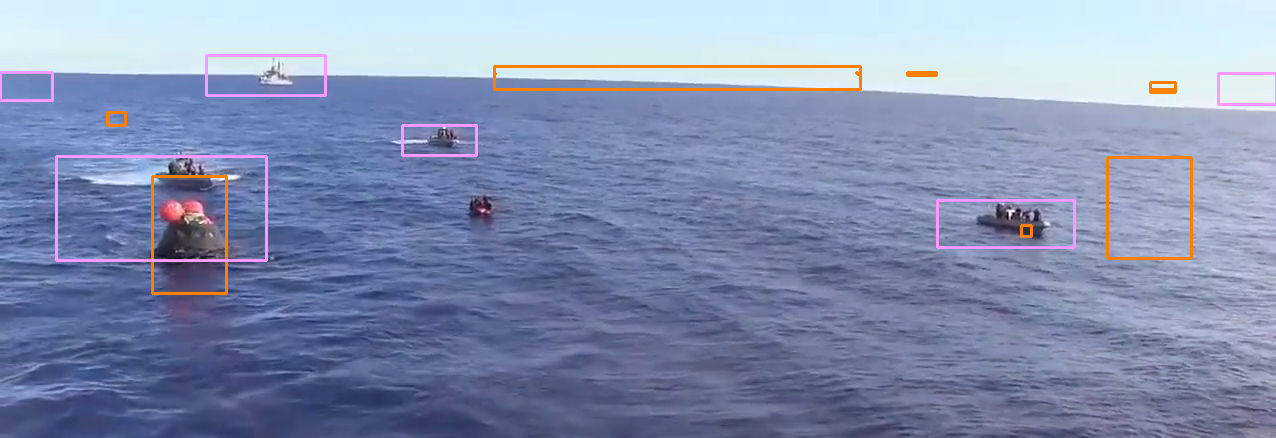}}
\subfigure[Seq.4, Frame 158]{\label{fig3r}\includegraphics[height=16mm,width=35mm]{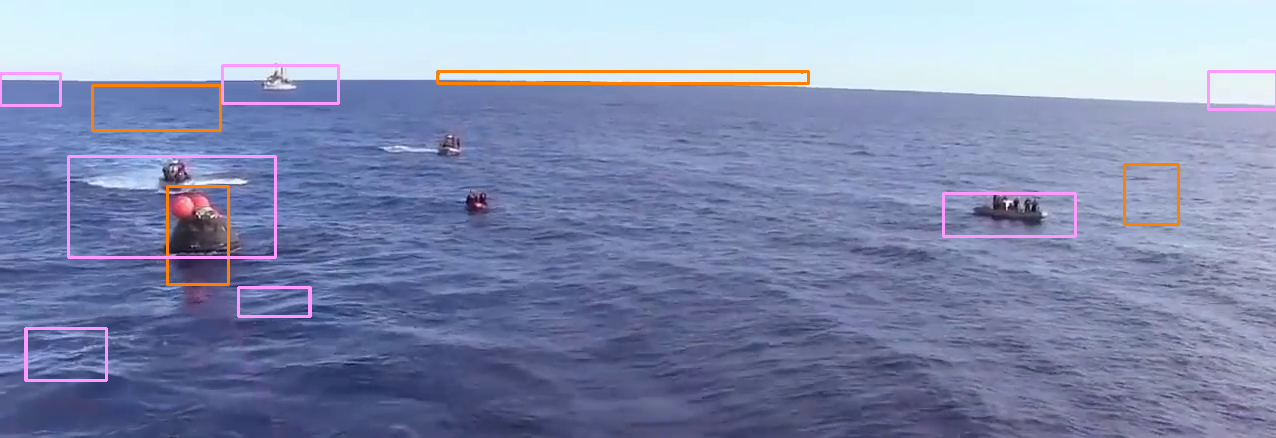}}
\subfigure[Seq.4, Frame 193]{\label{fig3s}\includegraphics[height=16mm,width=35mm]{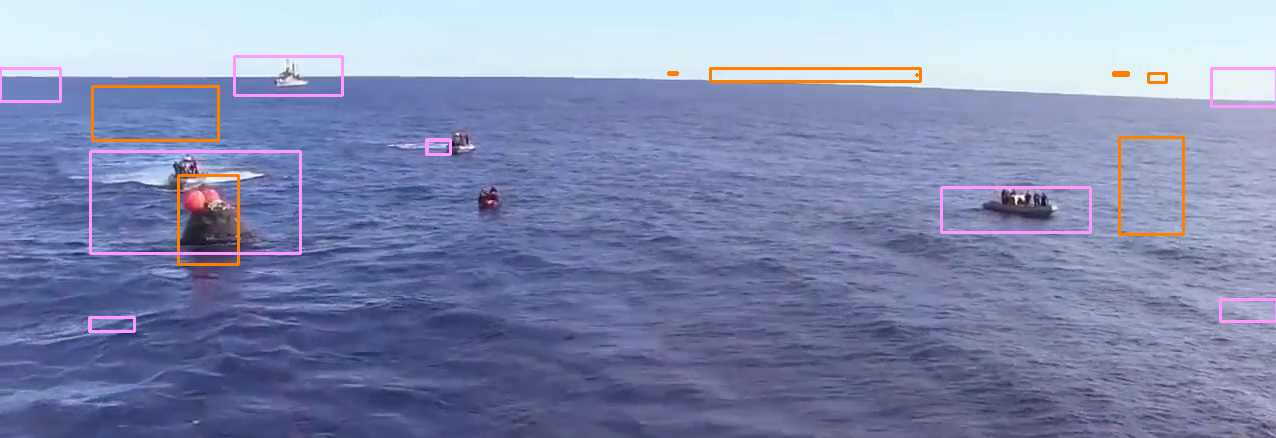}}
\subfigure[Seq.4, Frame 248]{\label{fig3t}\includegraphics[height=16mm,width=35mm]{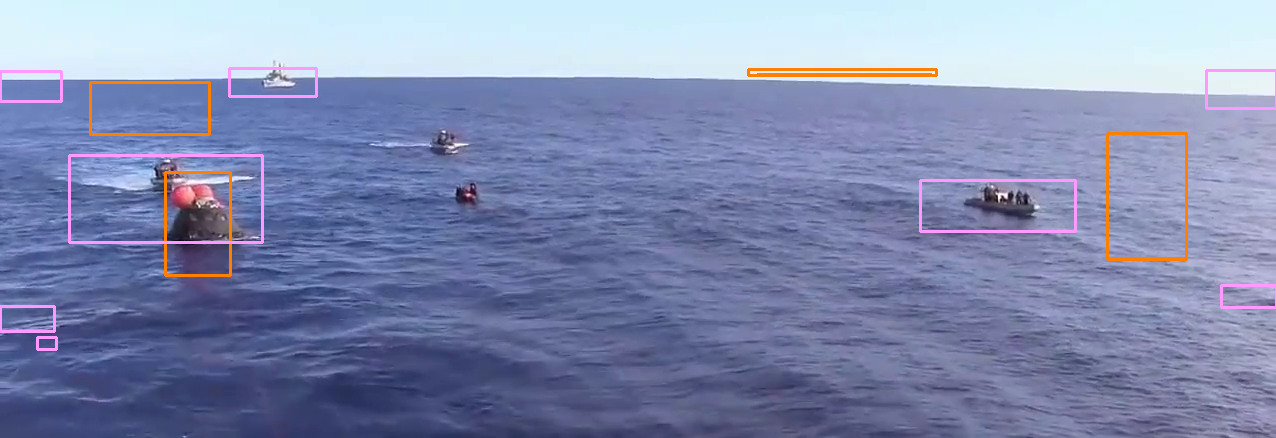}}

\caption{Qualitative results : Images in each row belongs to a single video sequence. The sequence number and frame numbers are indicated below each of the image. Top layer of each row represents the Residual image(contrast and brightness adjusted). Middle layer of each row indicates the detection made by our algorithm with logNFA=$2$ (Red BB) and logNFA=$-2$ (Green BB). Detections in the bottom layer of each row belongs to ITTI (Orange BB) and SRA (Pink BB). }
\label{fig333}
\end{figure*}

\begin{figure*}

\centering 
\includegraphics[height=17mm,width=35mm]{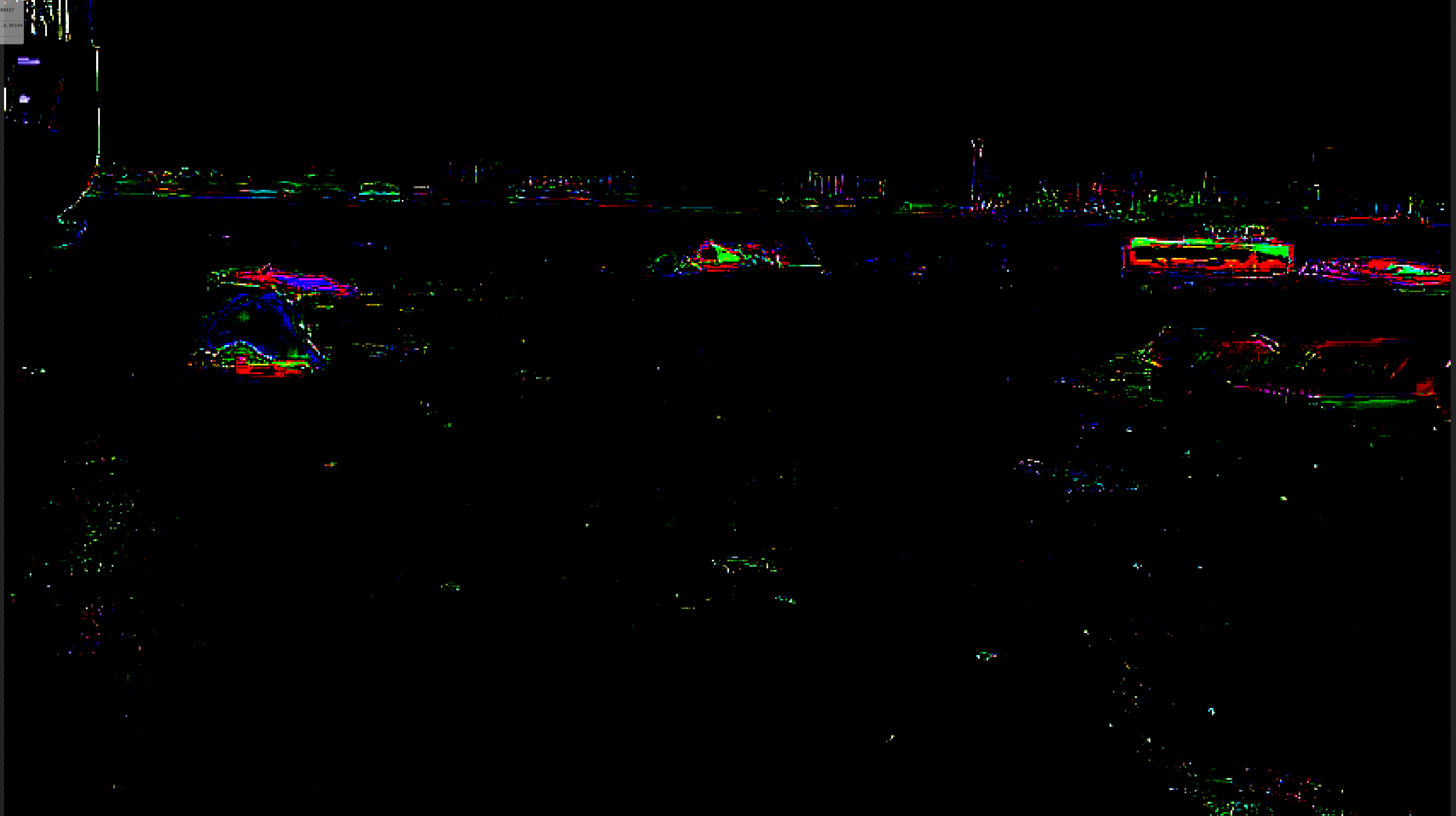}
\includegraphics[height=17mm,width=35mm]{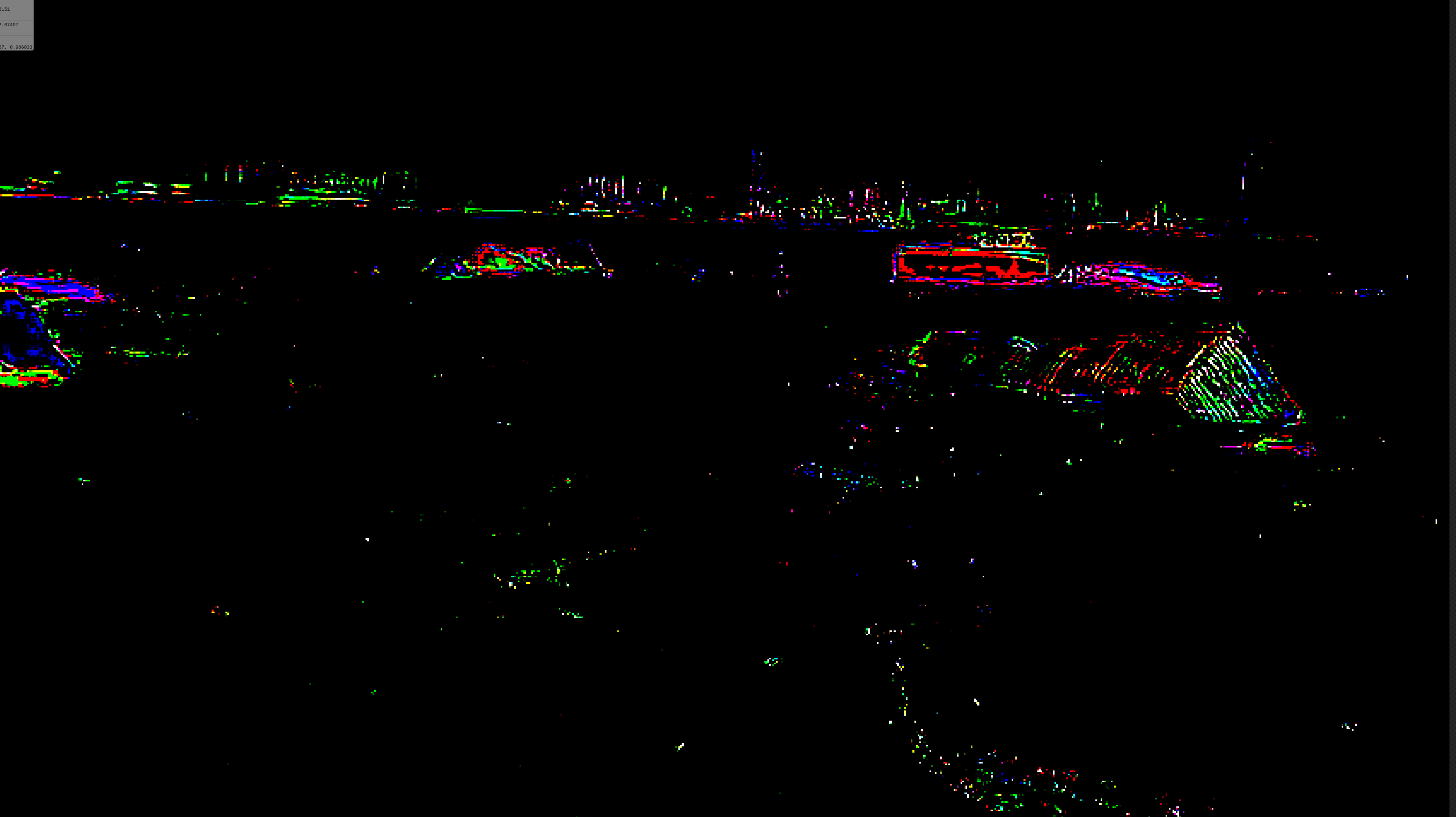}
\includegraphics[height=17mm,width=35mm]{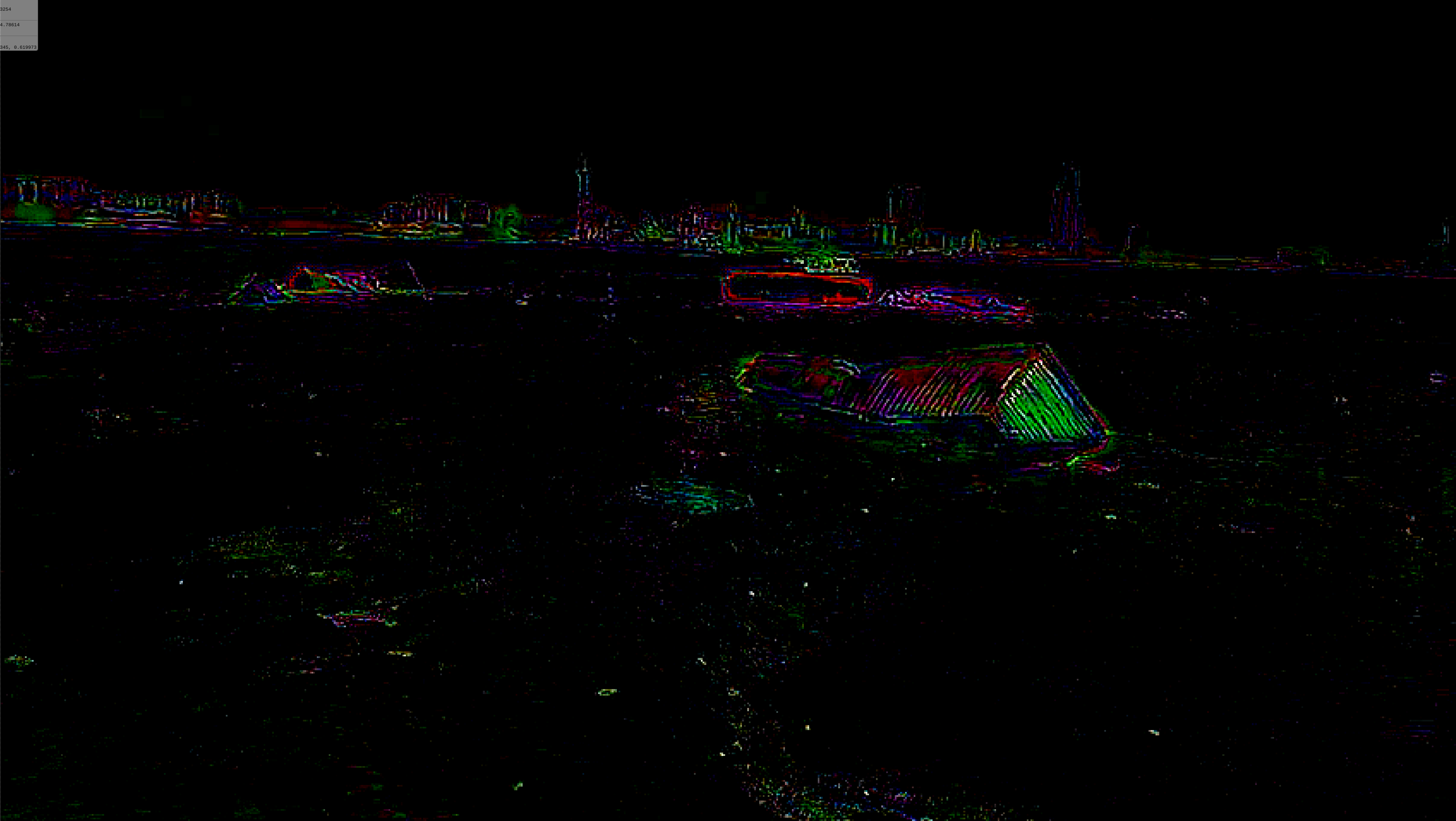}
\includegraphics[height=17mm,width=35mm]{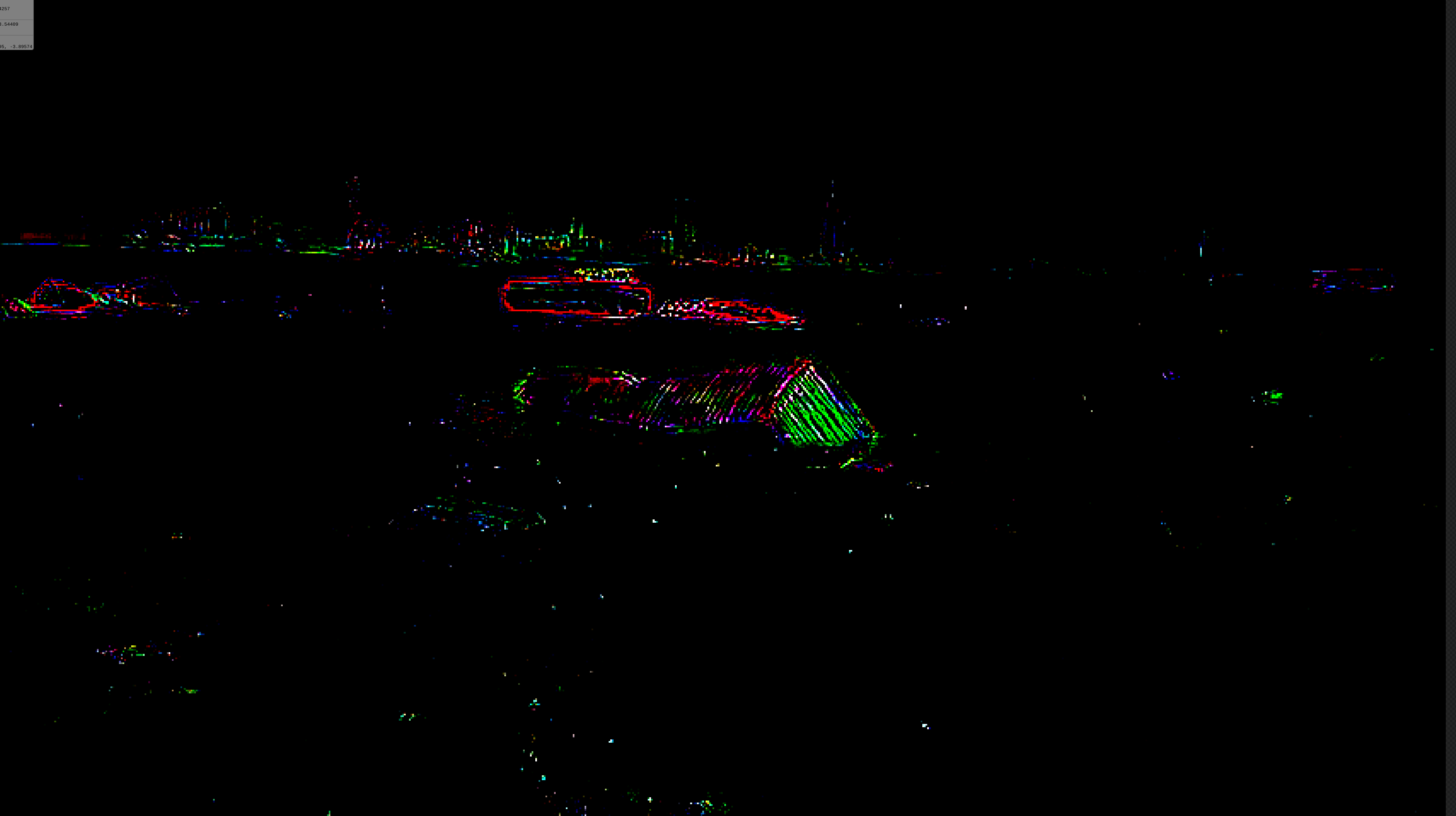}
\includegraphics[height=17mm,width=35mm]{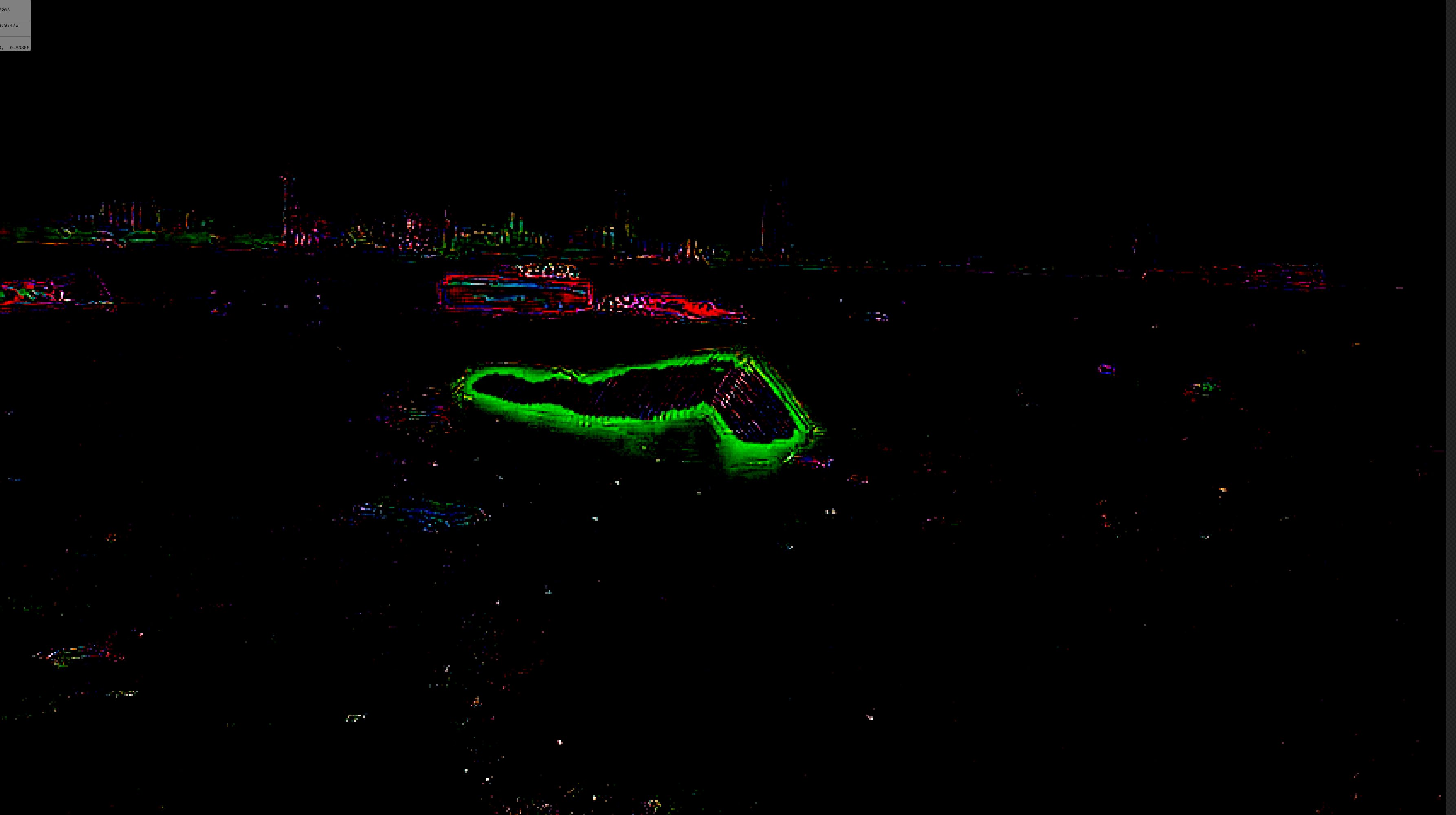}
\includegraphics[height=17mm,width=35mm]{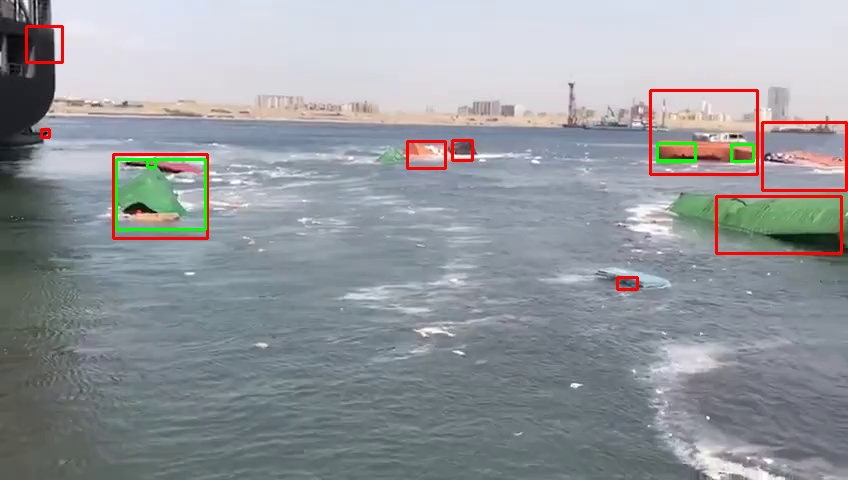}
\includegraphics[height=17mm,width=35mm]{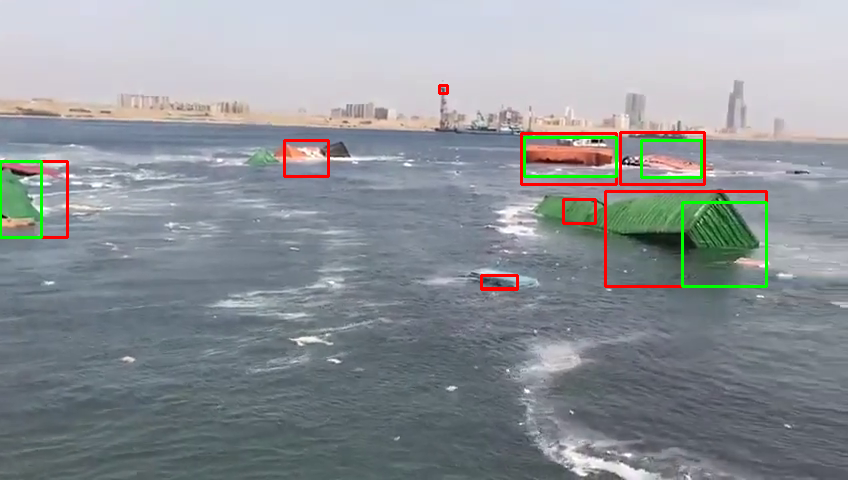}
\includegraphics[height=17mm,width=35mm]{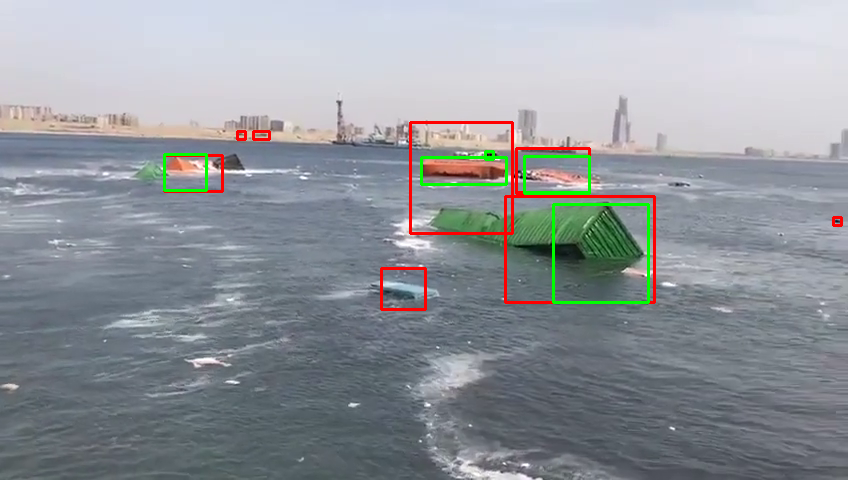}
\includegraphics[height=17mm,width=35mm]{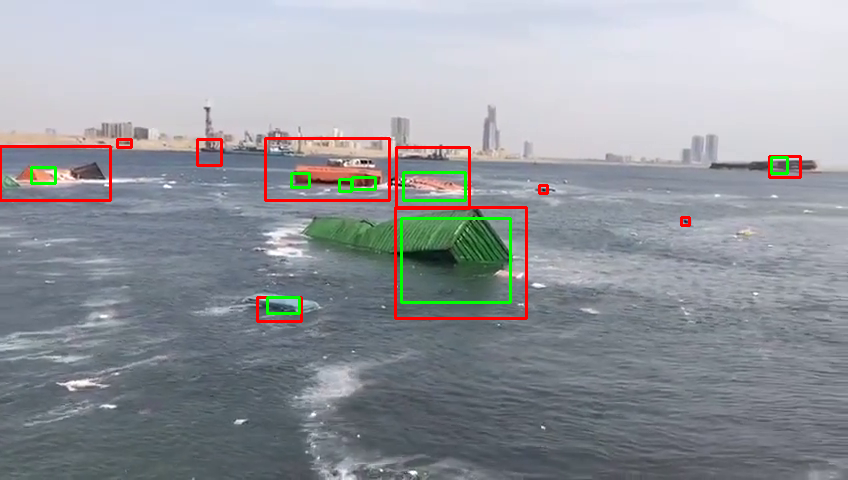}
\includegraphics[height=17mm,width=35mm]{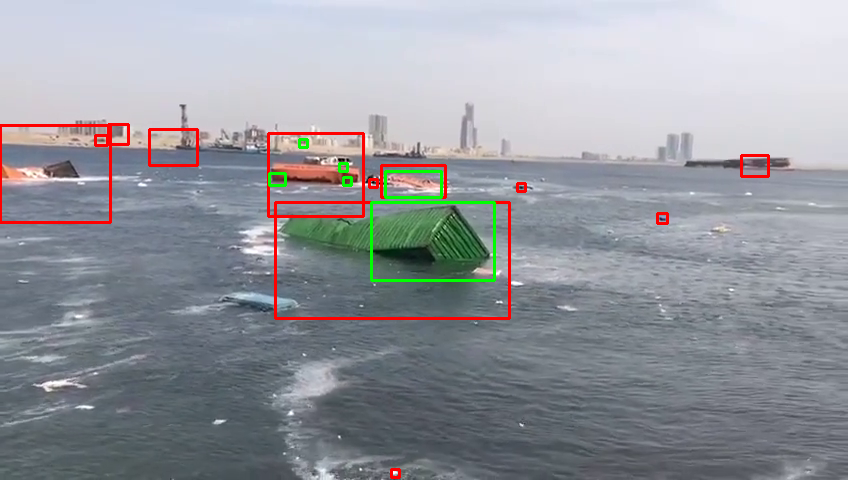}
\subfigure[Seq.5, Frame 0]{\label{fig4a}\includegraphics[height=17mm,width=35mm]{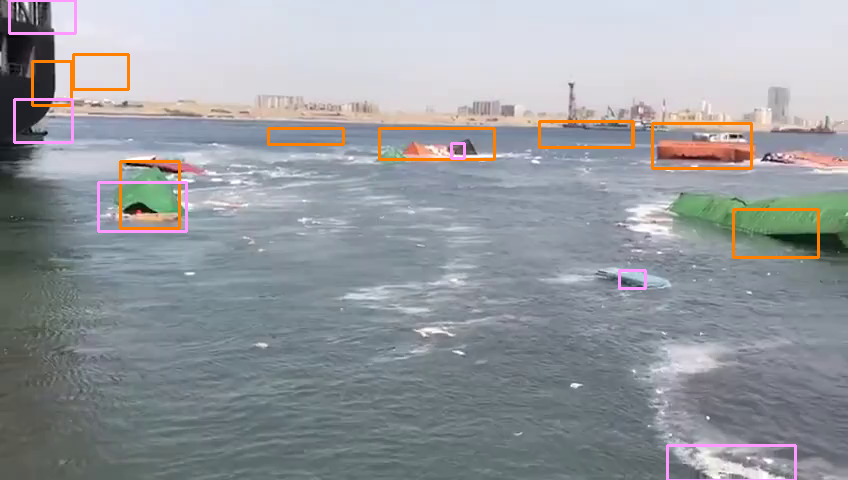}}
\subfigure[Seq.5, Frame 19]{\label{fig4b}\includegraphics[height=17mm,width=35mm]{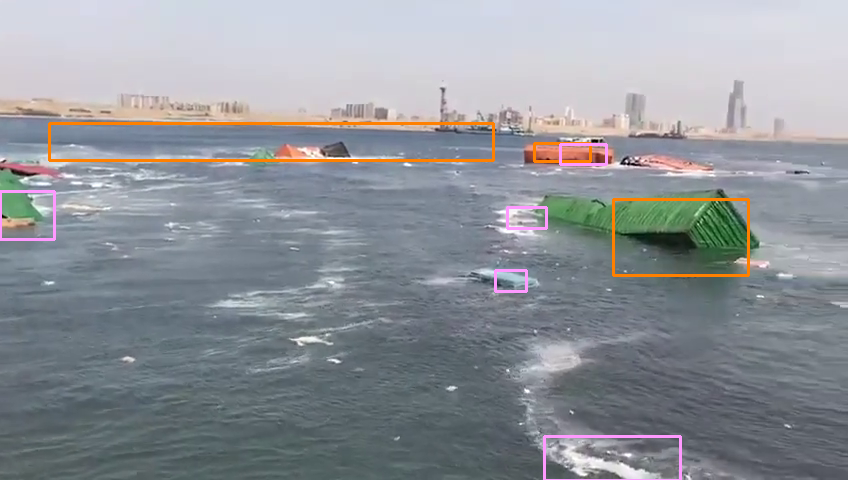}}
\subfigure[Seq.5, Frame 38]{\label{fig4c}\includegraphics[height=17mm,width=35mm]{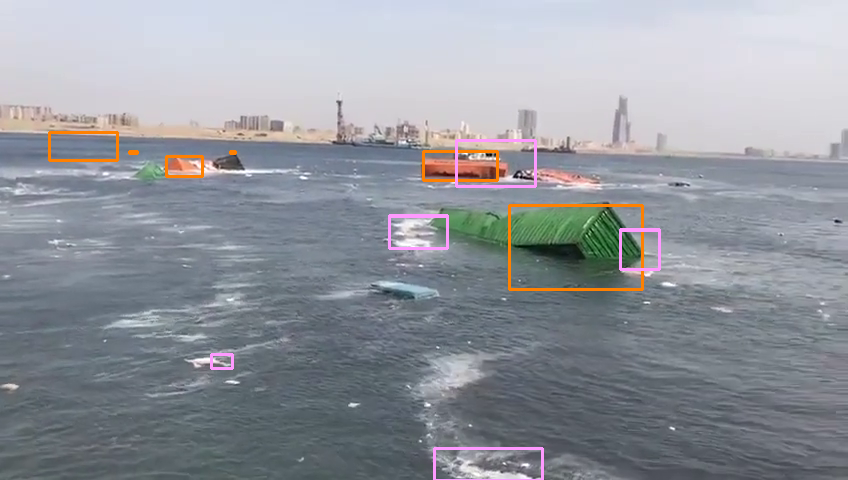}}
\subfigure[Seq.5, Frame 81]{\label{fig4d}\includegraphics[height=17mm,width=35mm]{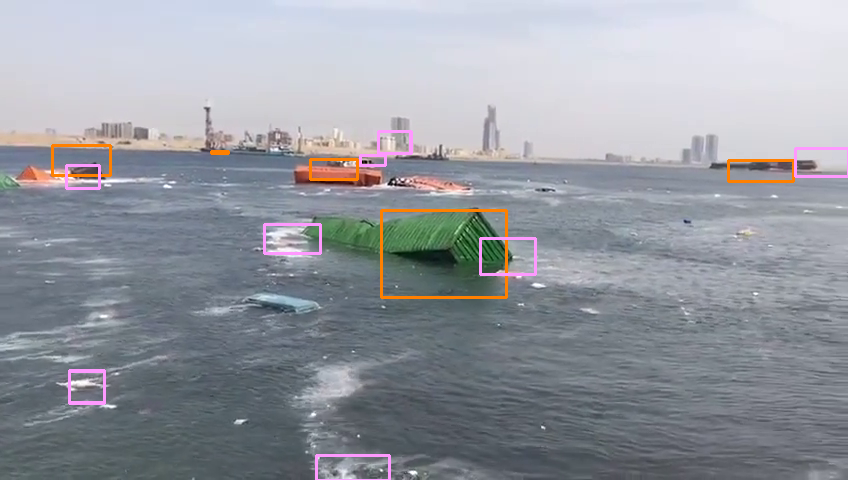}}
\subfigure[Seq.5, Frame 93]{\label{fig4e}\includegraphics[height=17mm,width=35mm]{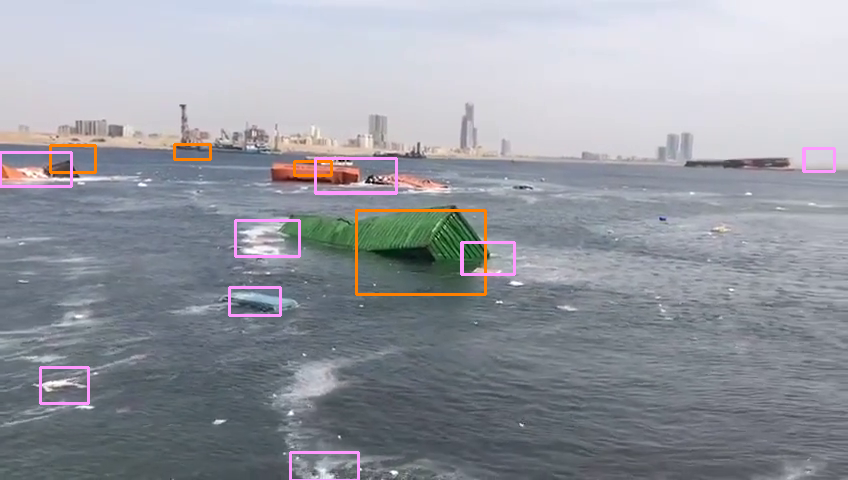}}

\includegraphics[height=17mm,width=35mm]{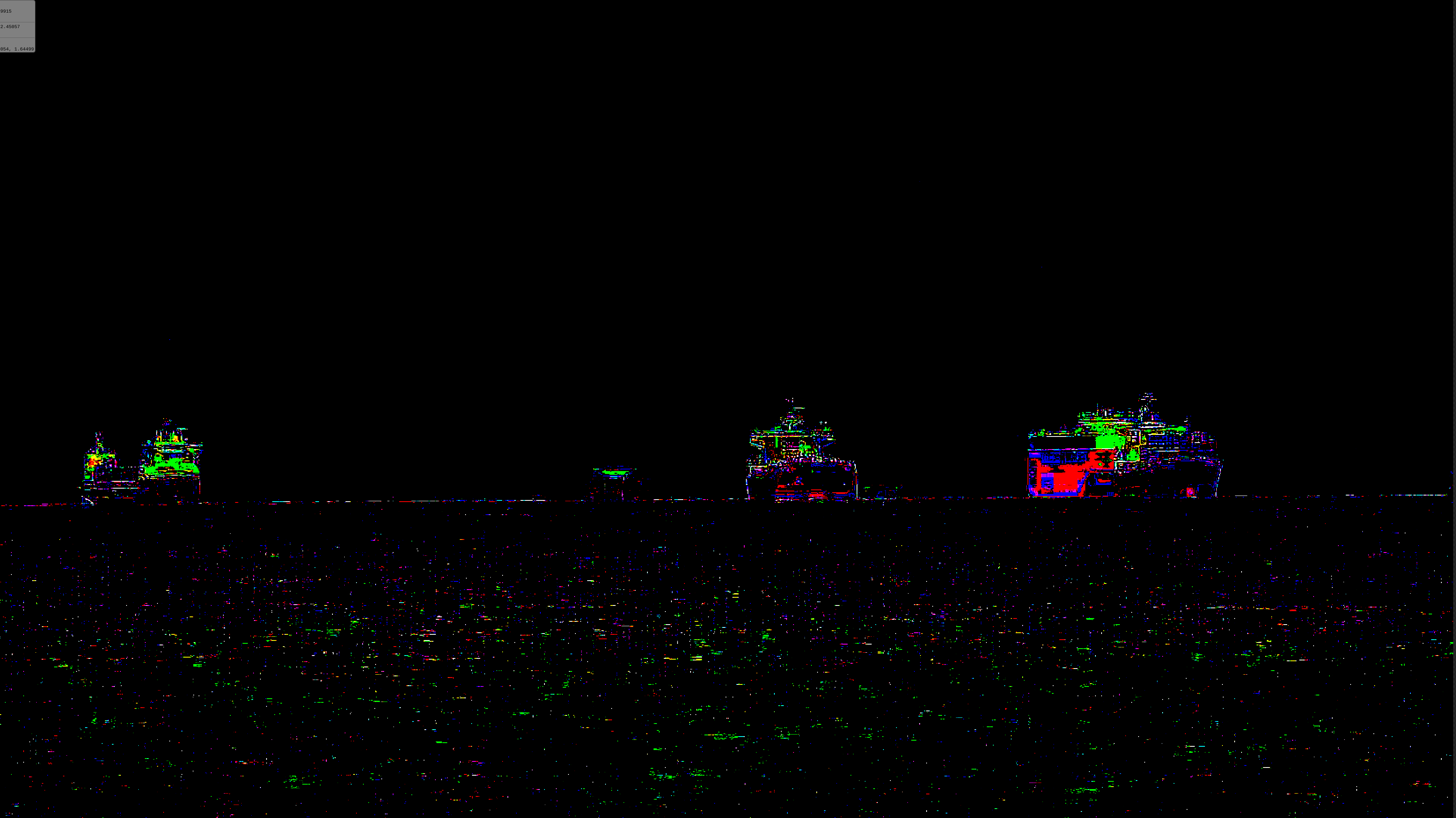}
\includegraphics[height=17mm,width=35mm]{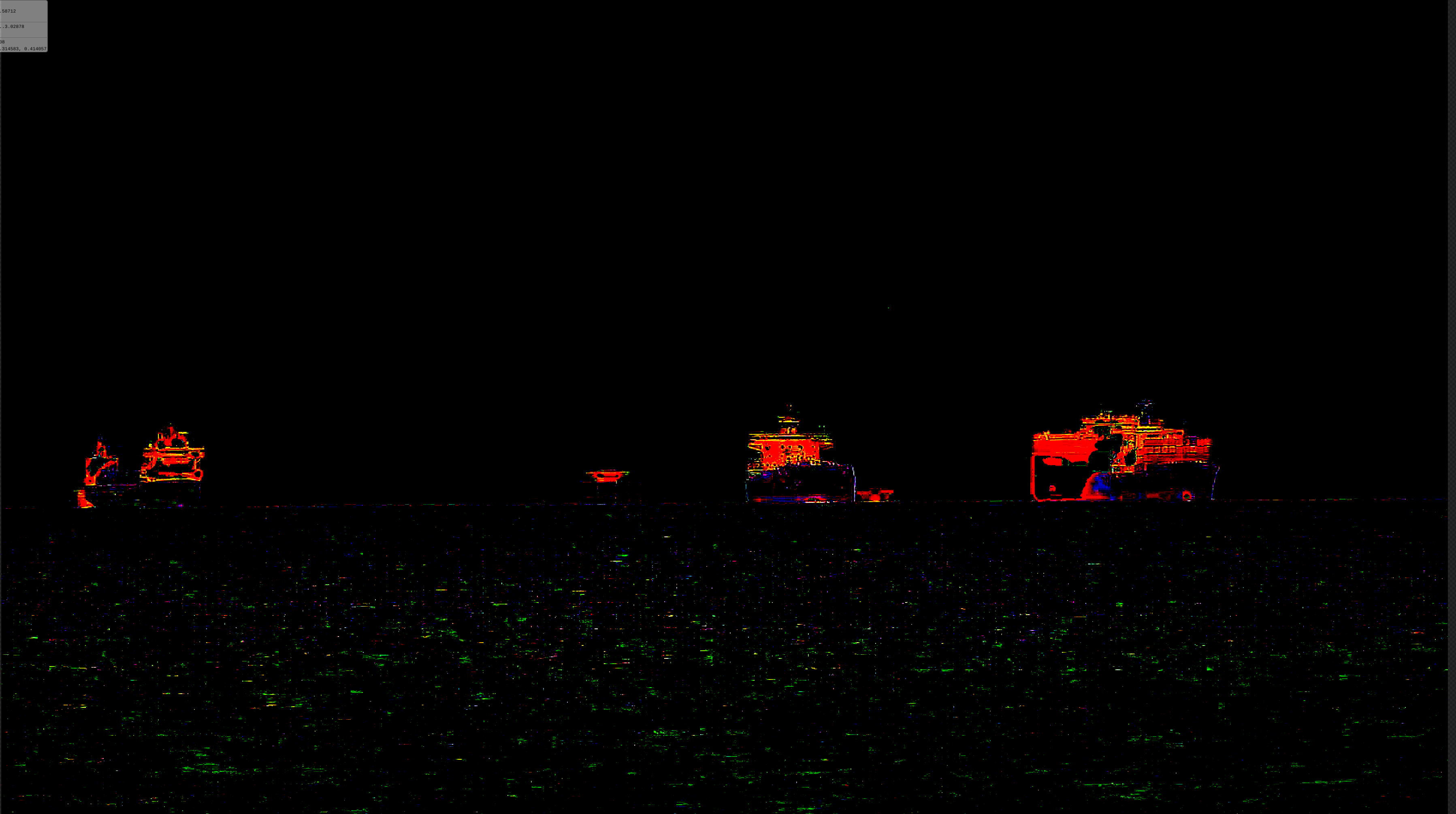}
\includegraphics[height=17mm,width=35mm]{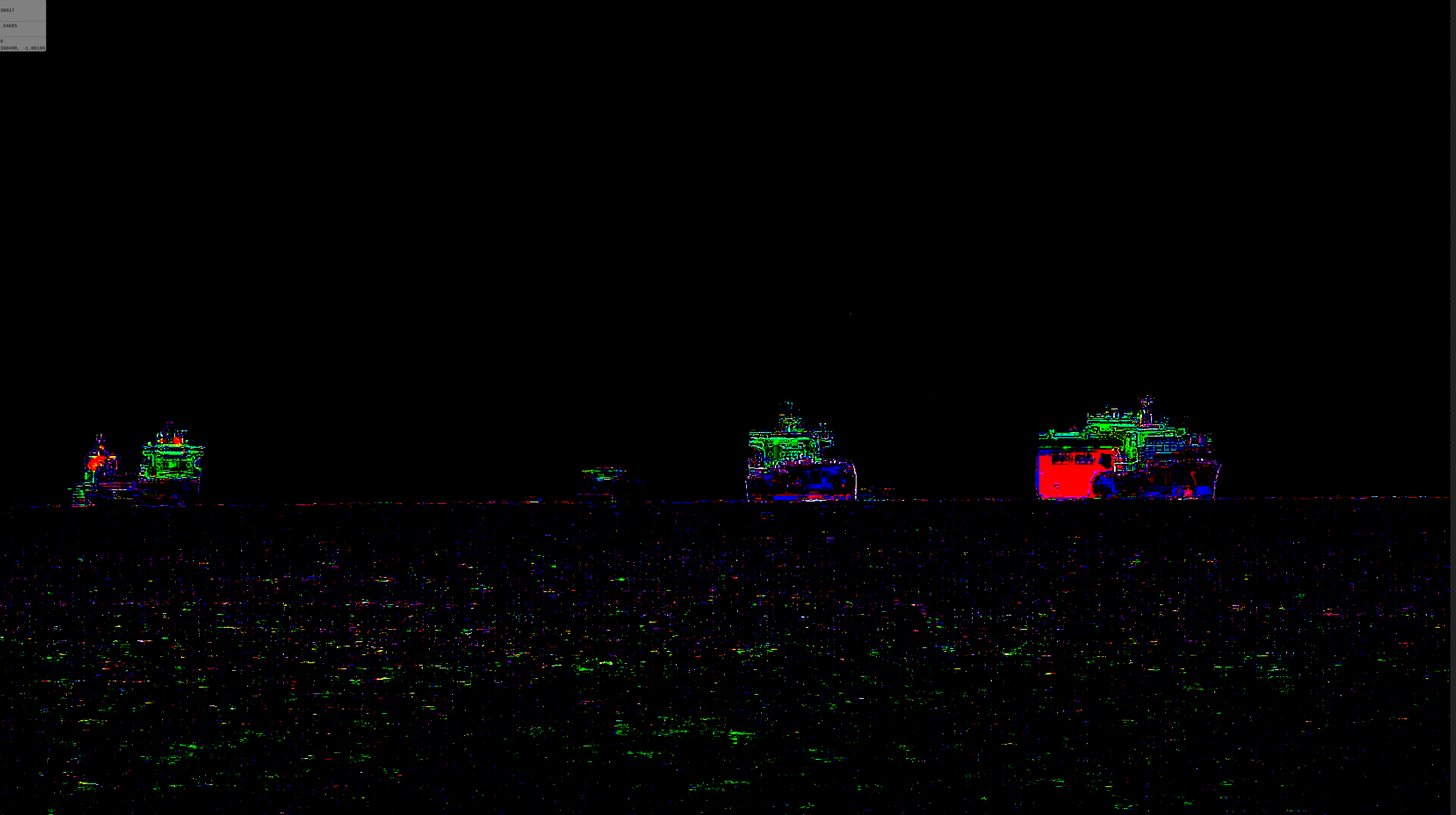}
\includegraphics[height=17mm,width=35mm]{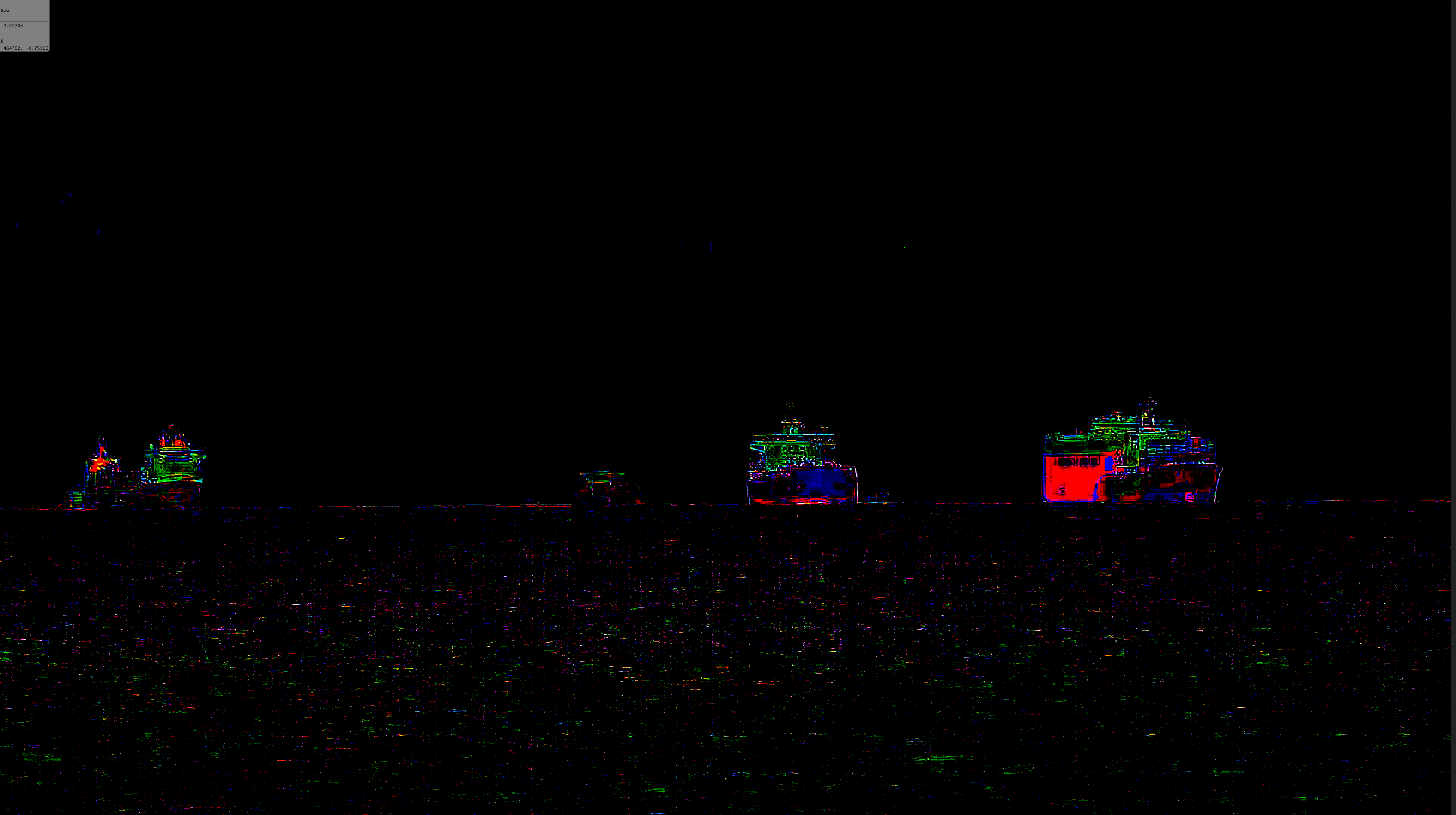}
\includegraphics[height=17mm,width=35mm]{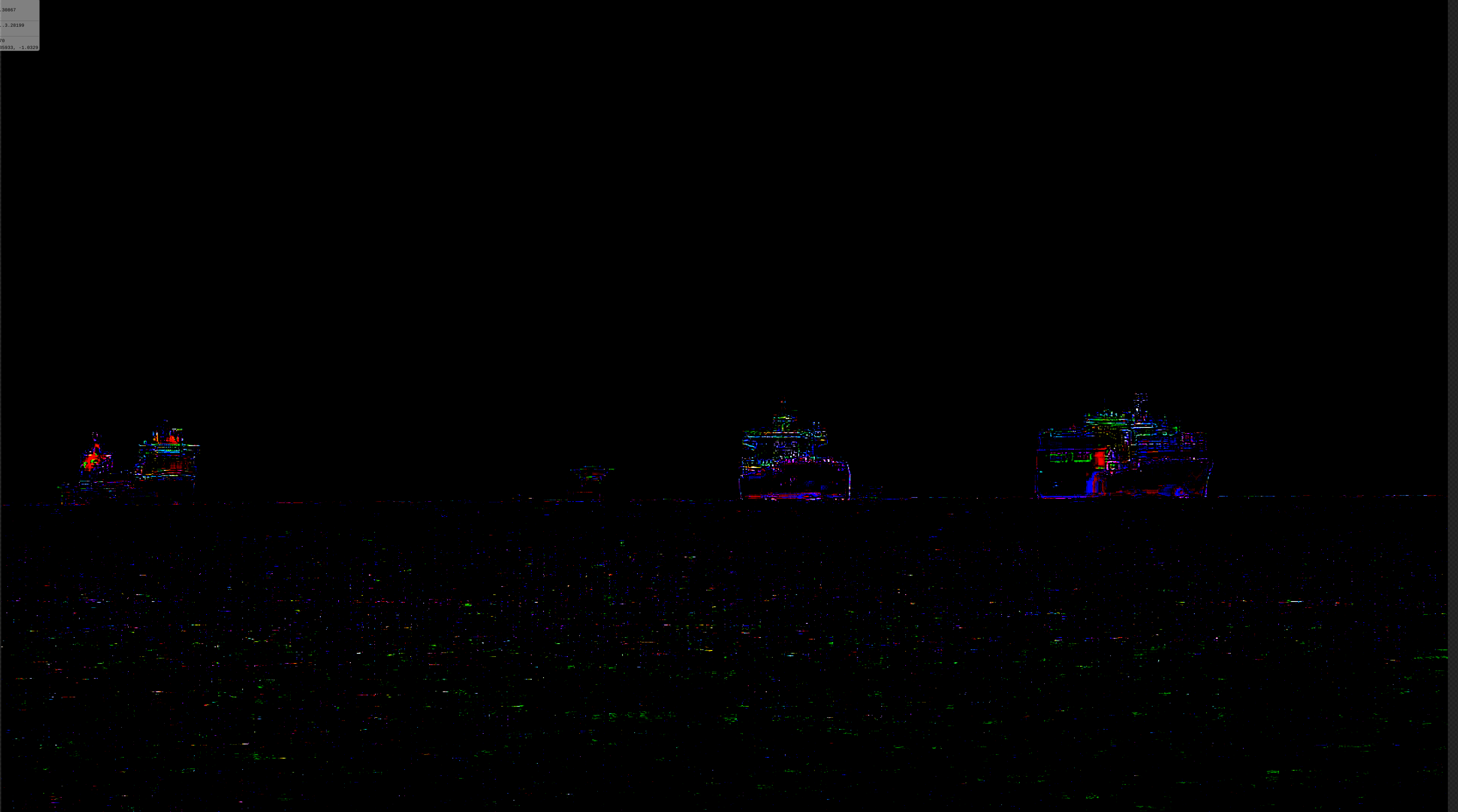}
\includegraphics[height=17mm,width=35mm]{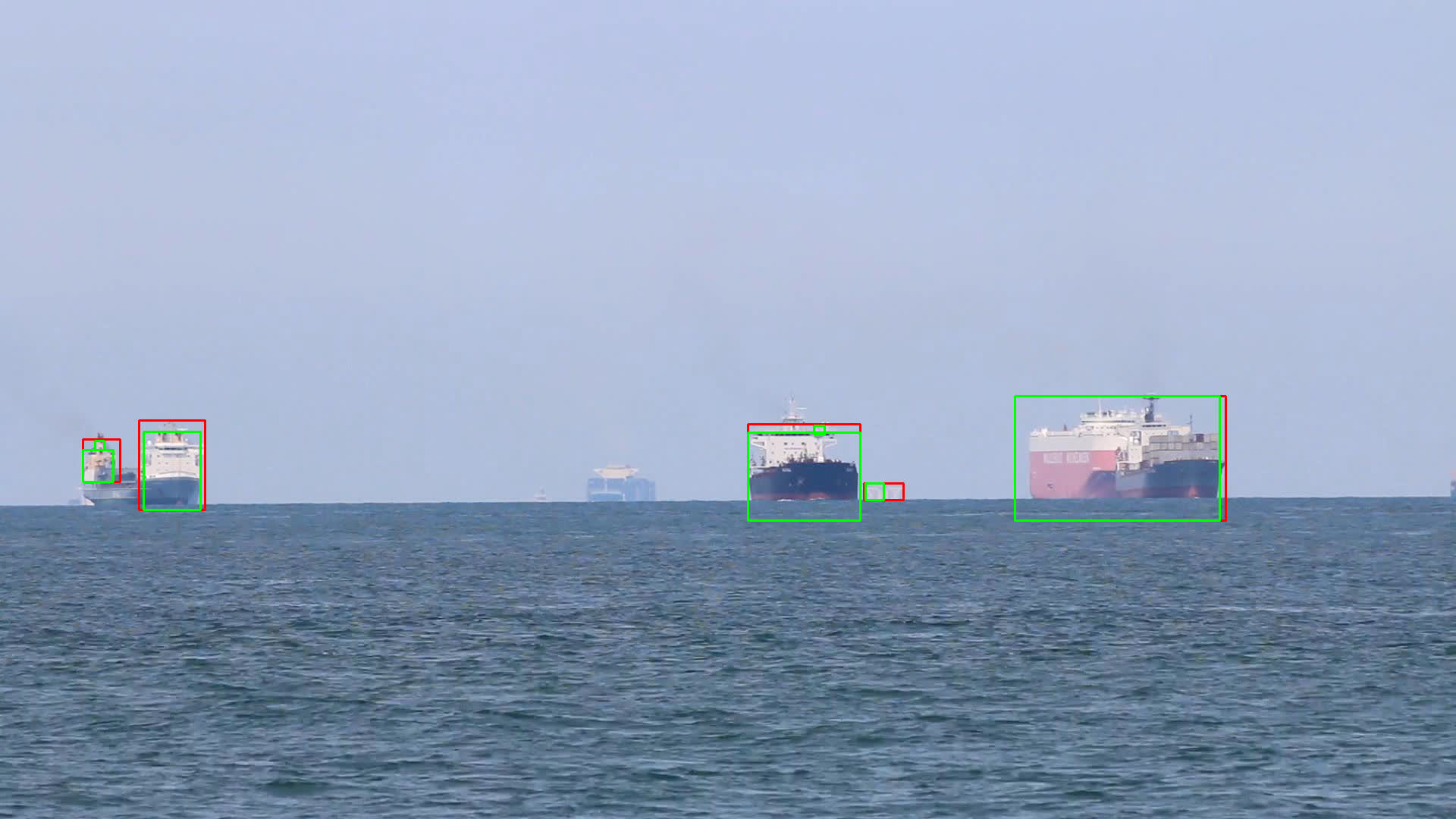}
\includegraphics[height=17mm,width=35mm]{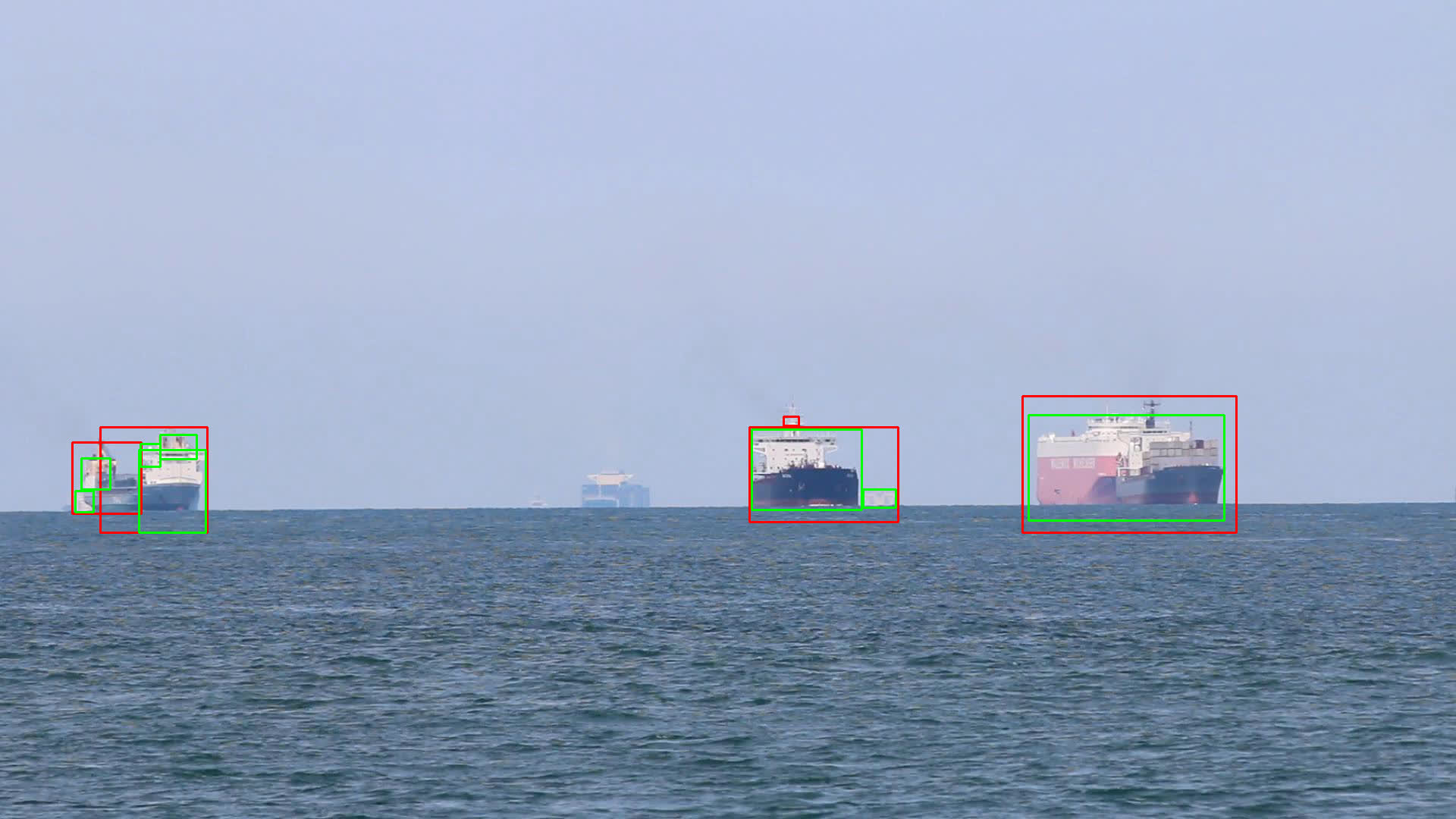}
\includegraphics[height=17mm,width=35mm]{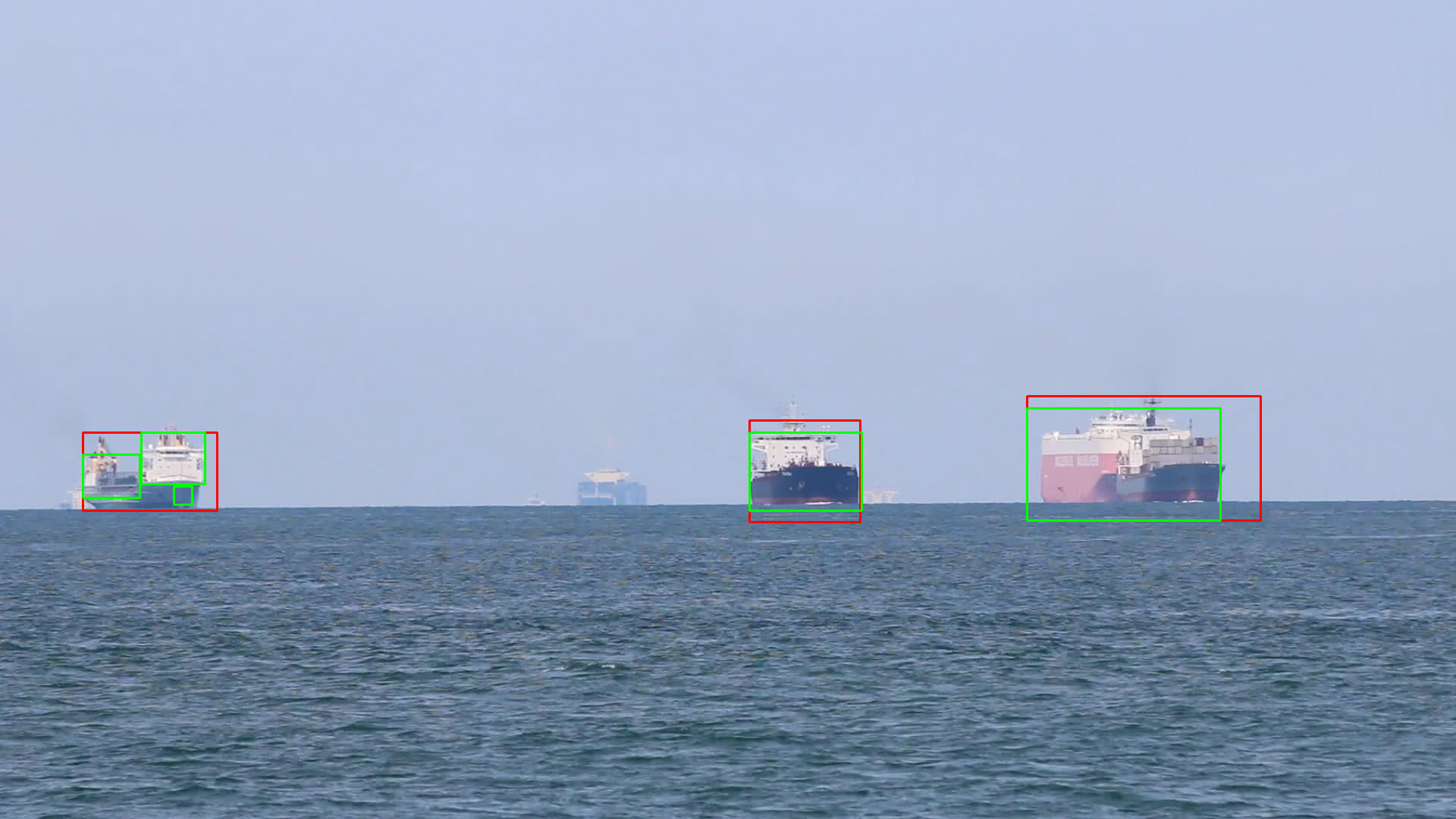}
\includegraphics[height=17mm,width=35mm]{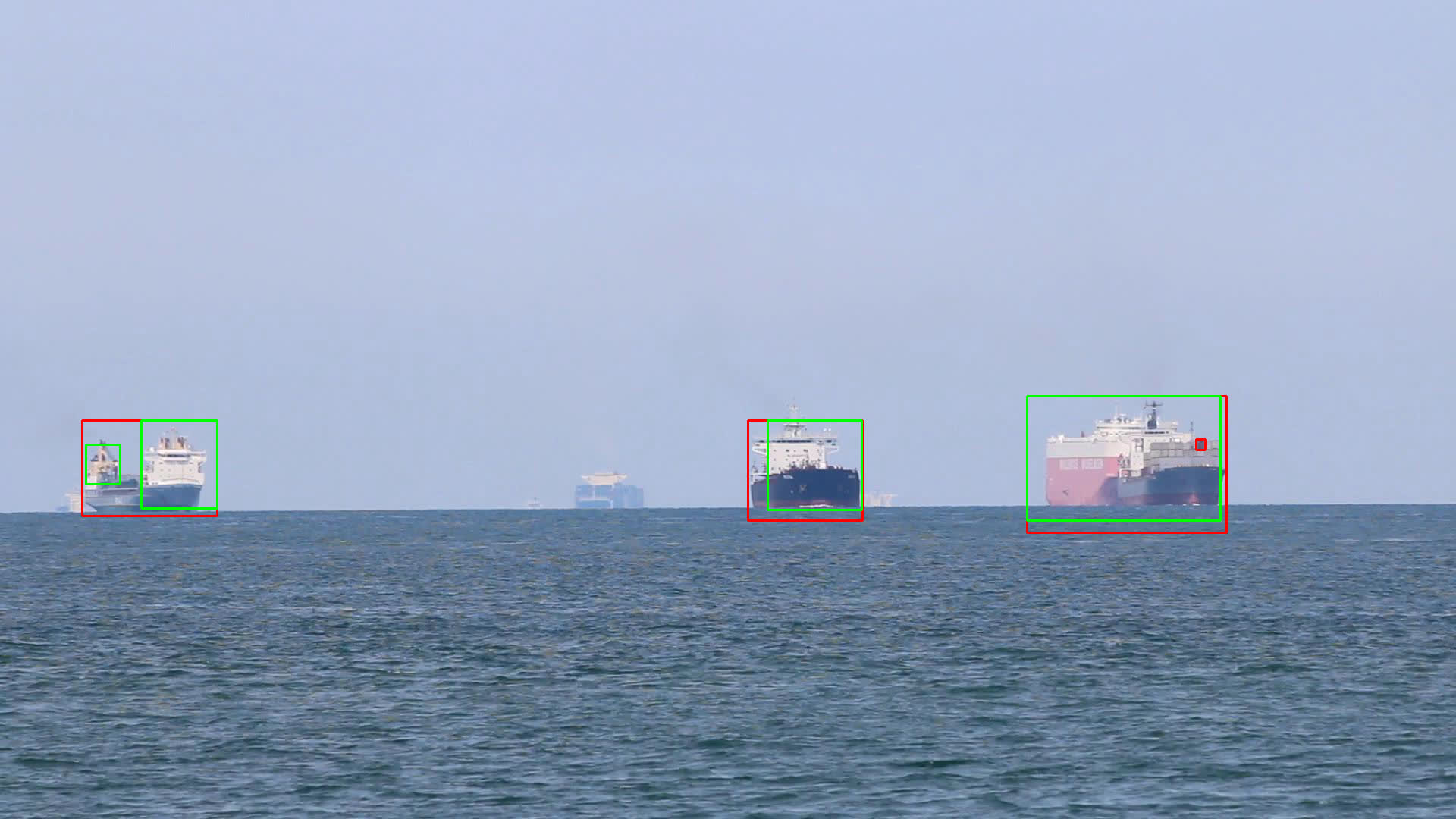}
\includegraphics[height=17mm,width=35mm]{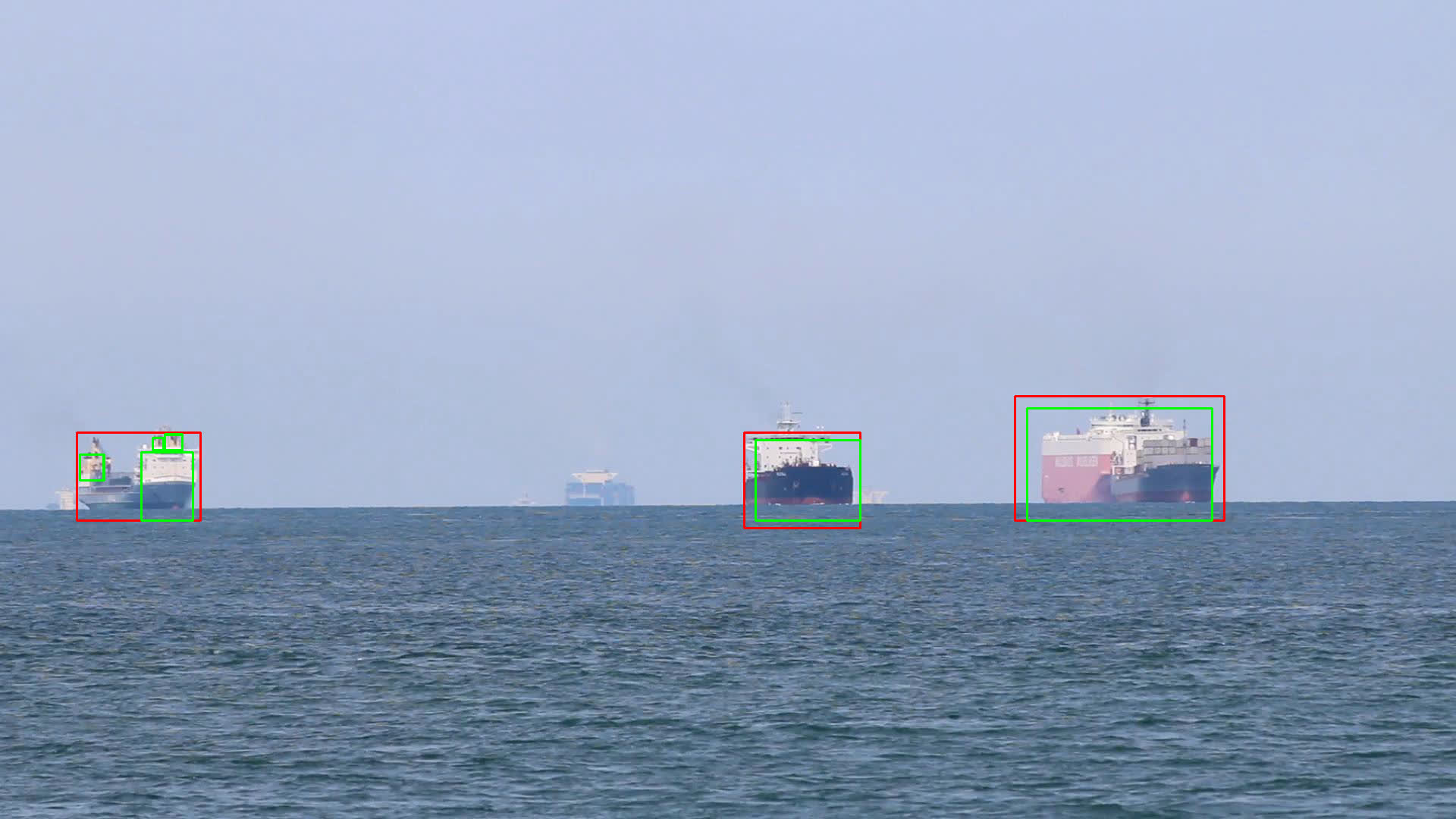}
\subfigure[Seq.6, Frame 8]{\label{fig4f}\includegraphics[height=17mm,width=35mm]{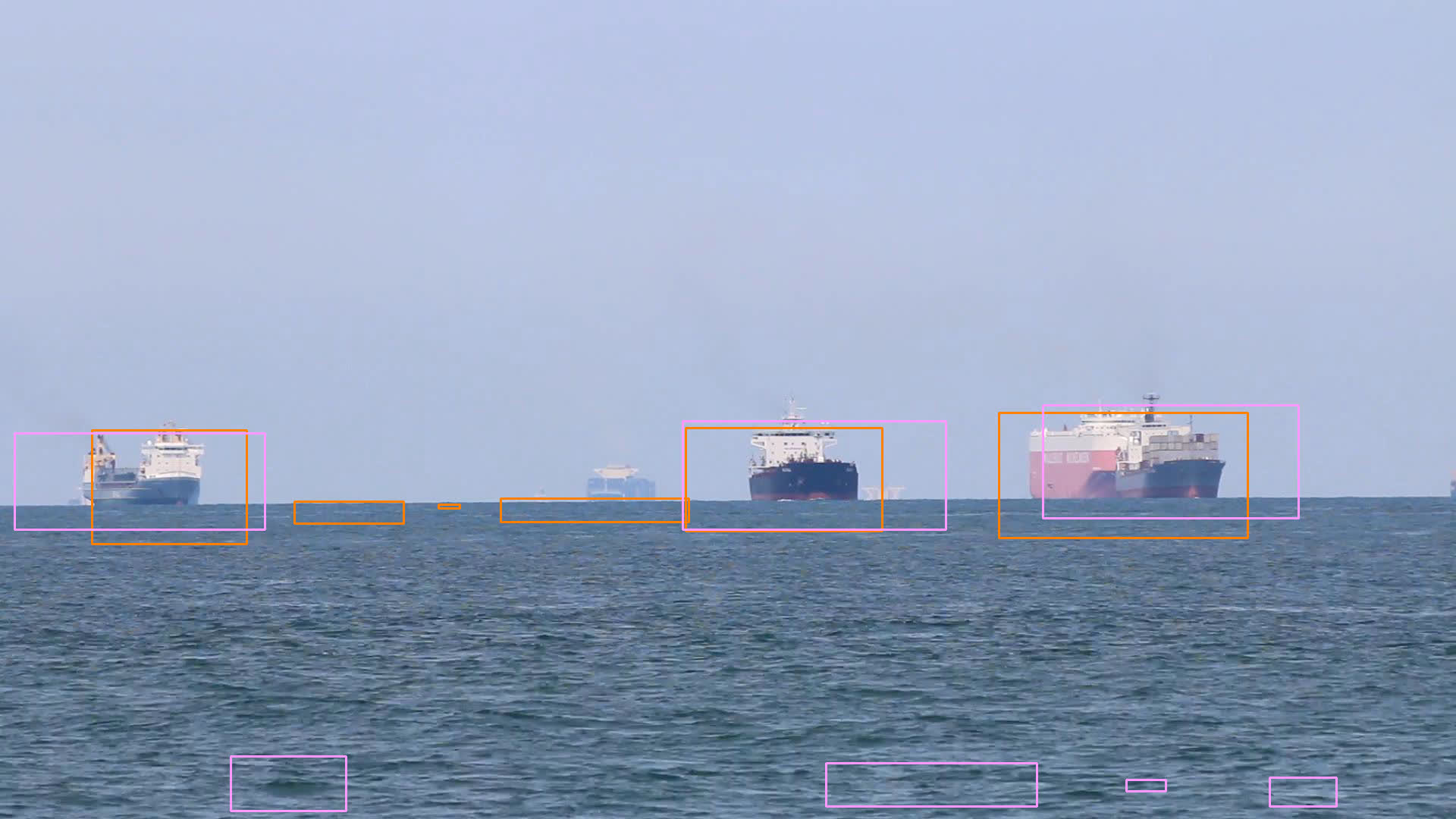}}
\subfigure[Seq.6, Frame 99]{\label{fig4g}\includegraphics[height=17mm,width=35mm]{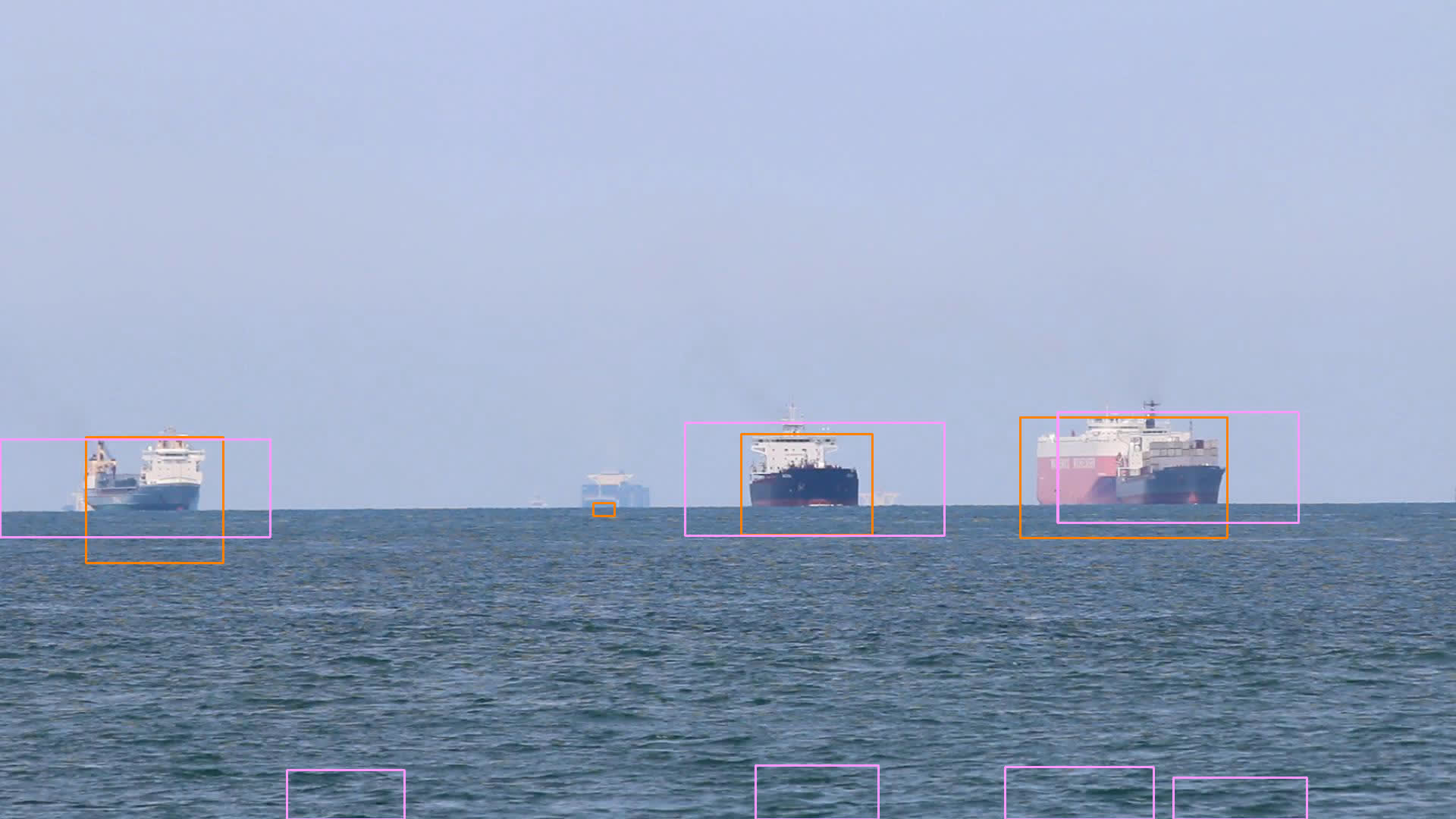}}
\subfigure[Seq.6, Frame 152]{\label{fig4h}\includegraphics[height=17mm,width=35mm]{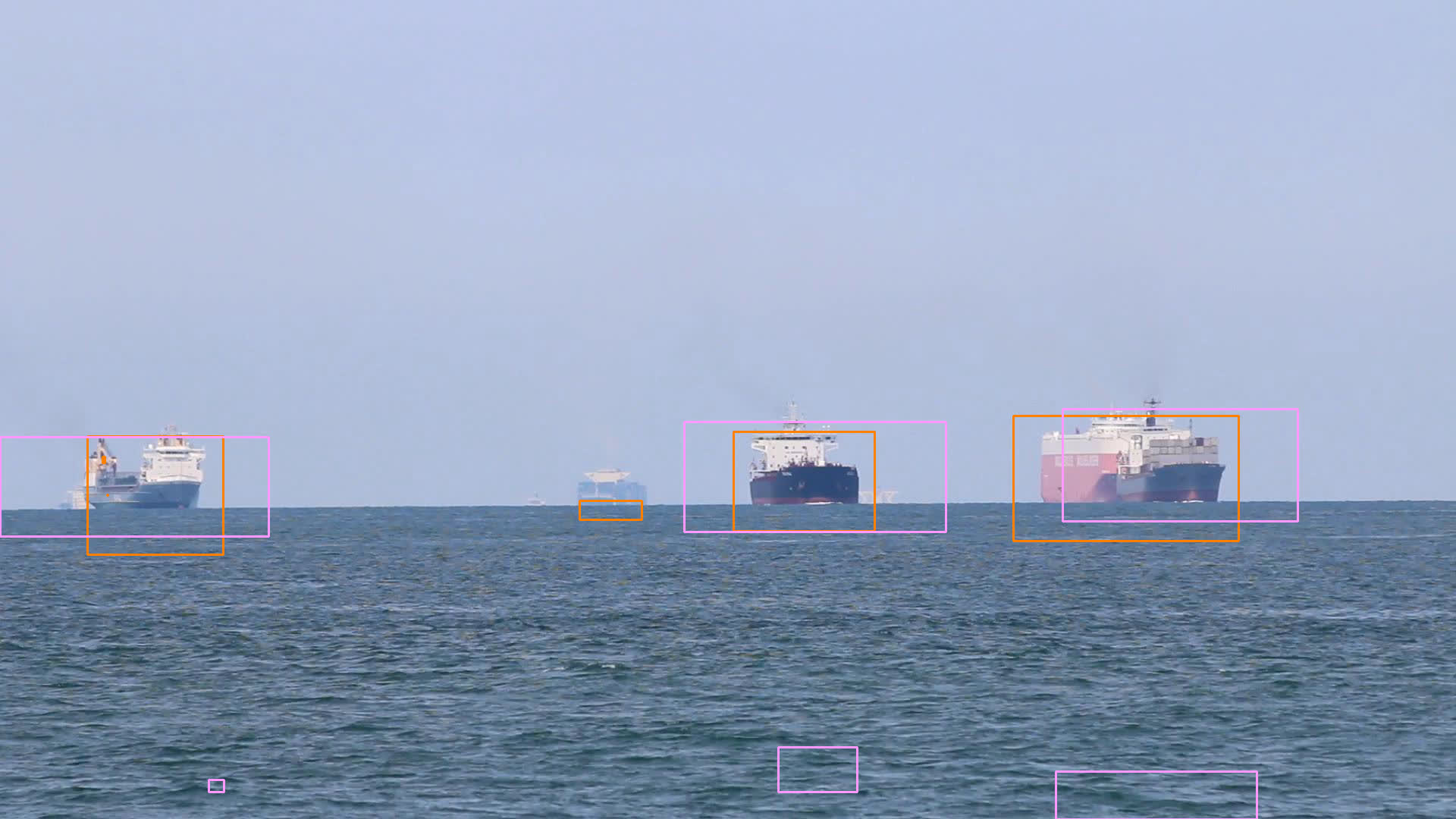}}
\subfigure[Seq.6, Frame 209]{\label{fig4i}\includegraphics[height=17mm,width=35mm]{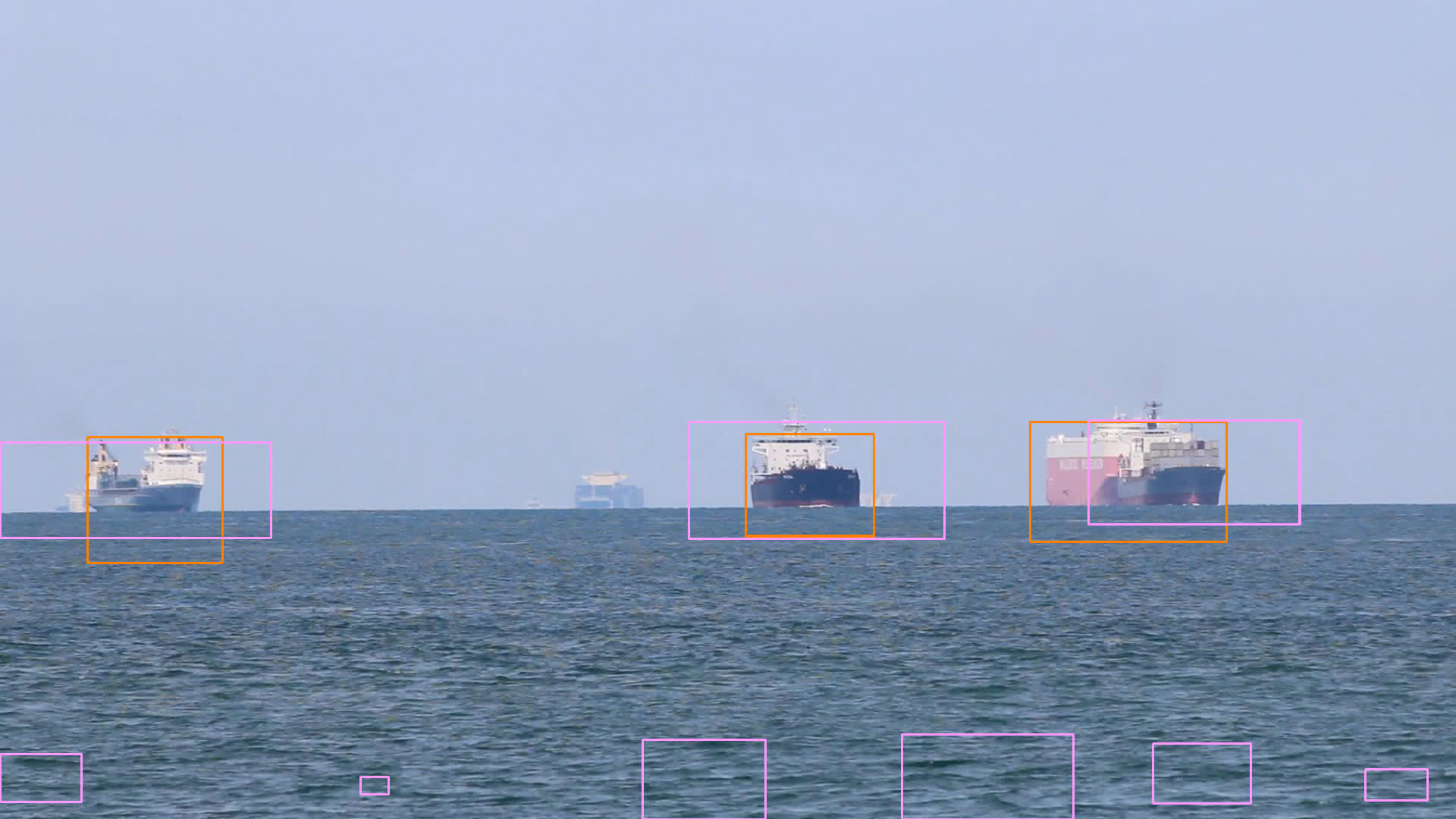}}
\subfigure[Seq.6, Frame 244]{\label{fig4k}\includegraphics[height=17mm,width=35mm]{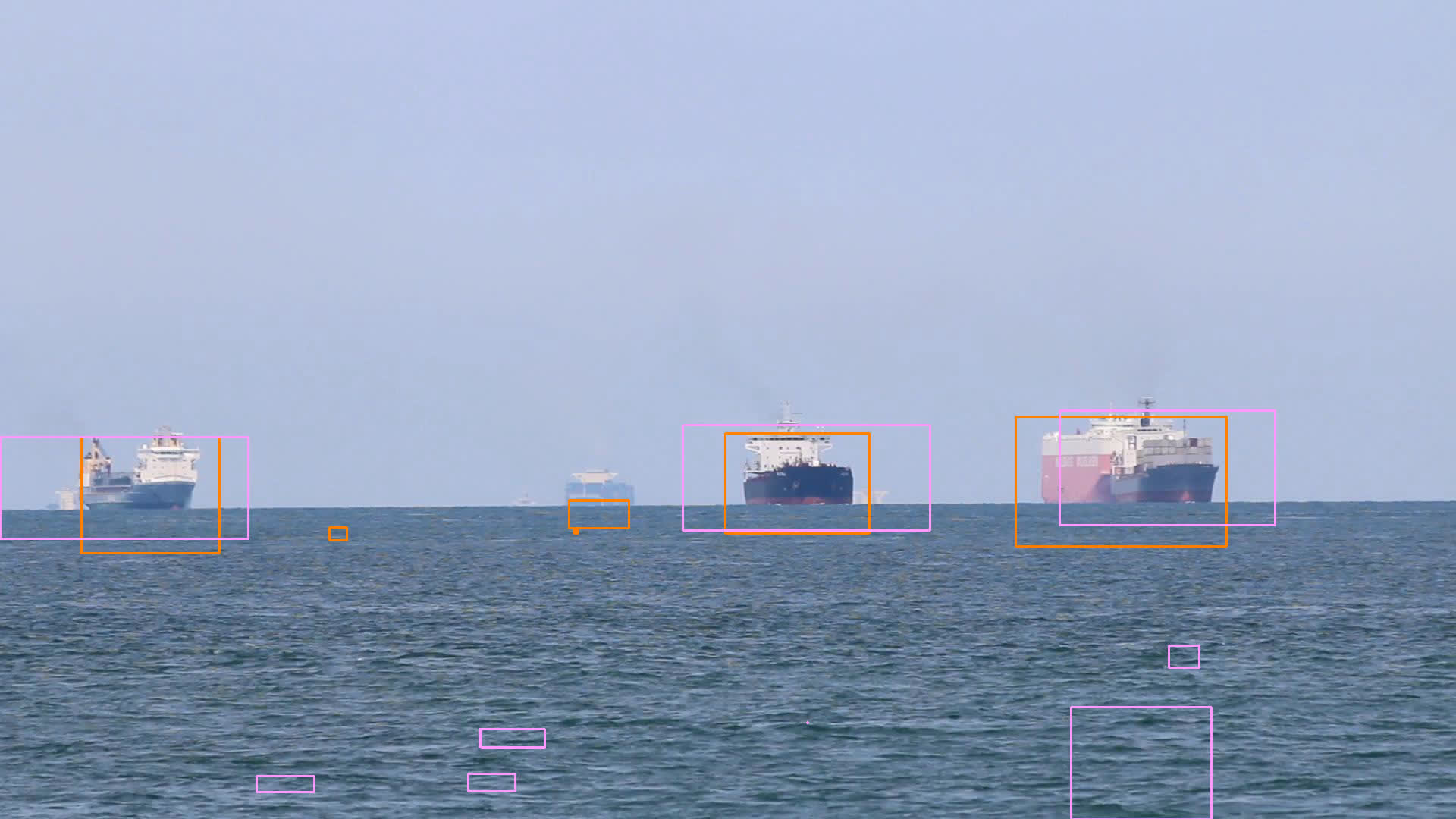}}

\caption{Qualitative results : Images in each row belongs to a single video sequence. The Sequence number and frame numbers are indicated below each of the image. Top layer of each row represents the Residual images(contrast and brightness adjusted).Middle layer of each row indicates the detection made by our algorithm with logNFA=$2$ (Red BB) and logNFA=$-2$ (Green BB). Detections in the bottom layer of each row belongs to ITTI (Orange BB) and SRA (Pink BB).}
\label{fig444}
\end{figure*}

The results of our algorithm is compared with the results of Spectral Residual Approach (SRA) \cite{FOBD_43} and ITTI \cite{FOBD_44}. The code for SRA and ITTI can be found in \cite{FOBD_43} and \cite{FOBD_45}. Fig. \ref{fig333} and Fig. \ref{fig444} presents the qualitative comparison of the three methods. The red and green Bounding Boxes(BB) present in the middle layer of each row in Fig. \ref{fig333} and Fig. \ref{fig444} represent the detections obtained by our algorithm with logNFA = $2$ and logNFA = $-2$ respectively. The Bounding Boxes(BB) in orange and pink in the bottom layer of each row in Fig. \ref{fig333} and Fig. \ref{fig444} belong to ITTI and SRA, respectively. False detections from our approach are shown in Fig. \ref{fig555}. 

\begin{figure*}

\centering

\includegraphics[height=15mm,width=42mm]{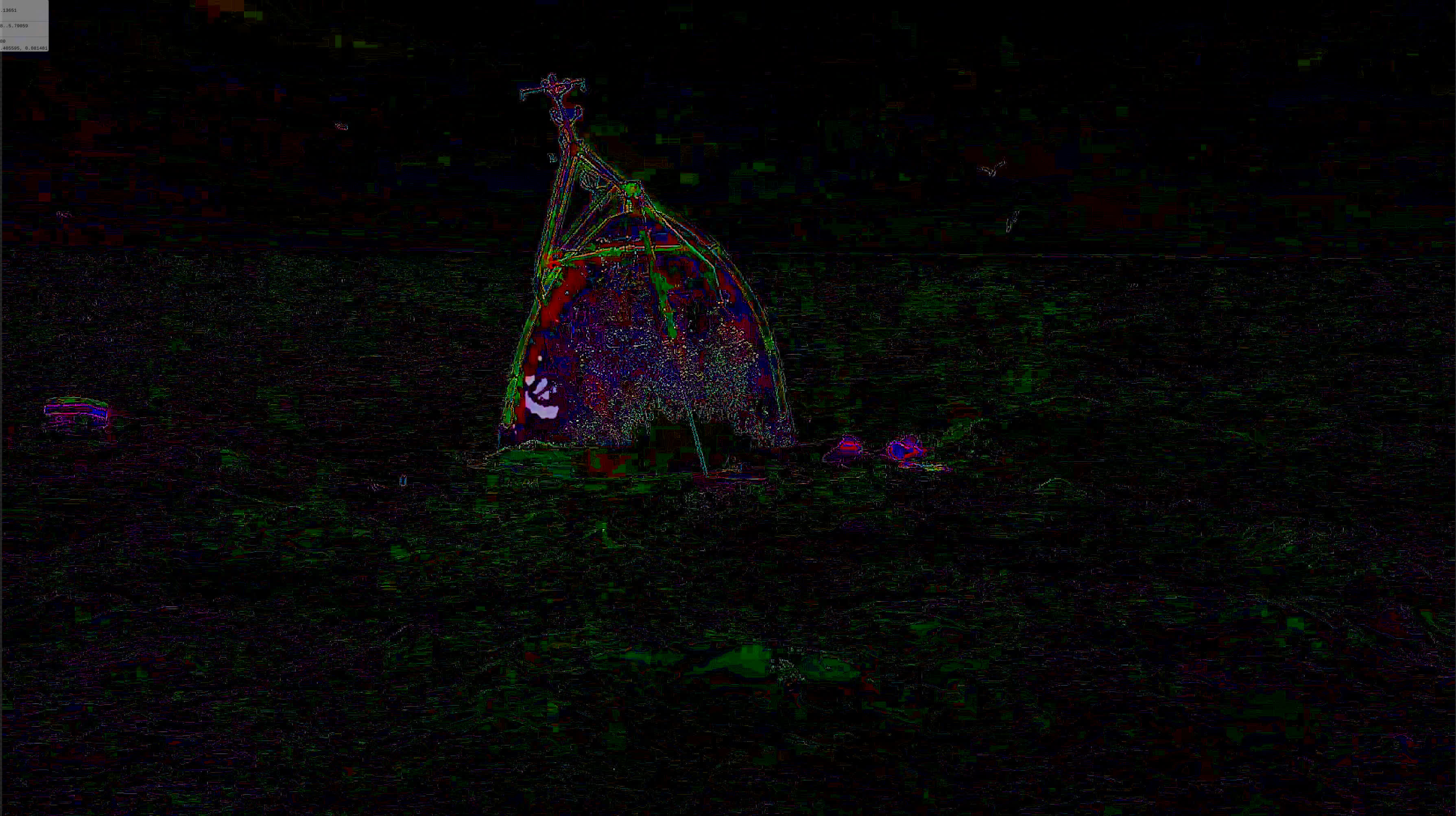}
\includegraphics[height=15mm,width=42mm]{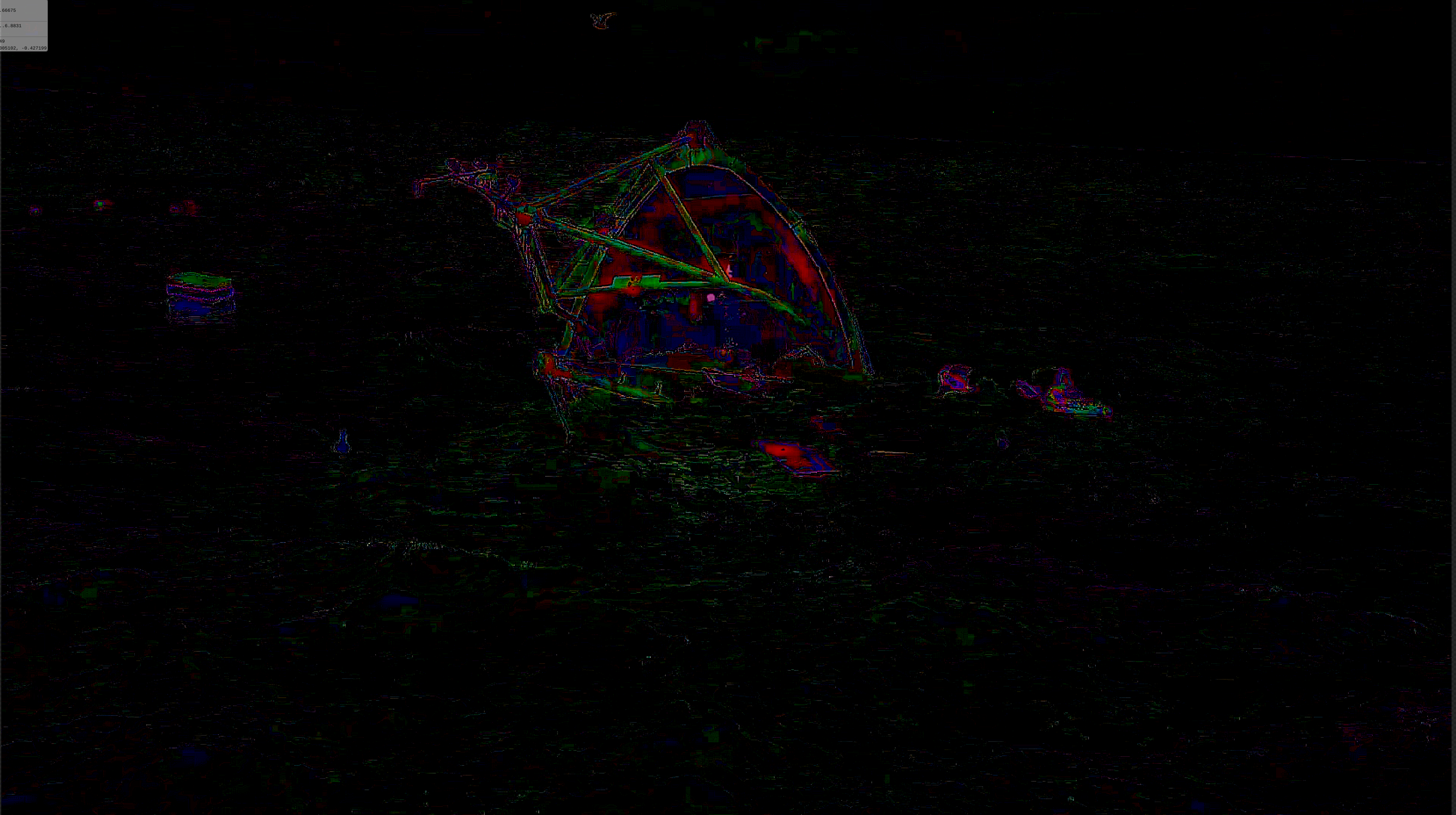}
\includegraphics[height=15mm,width=42mm]{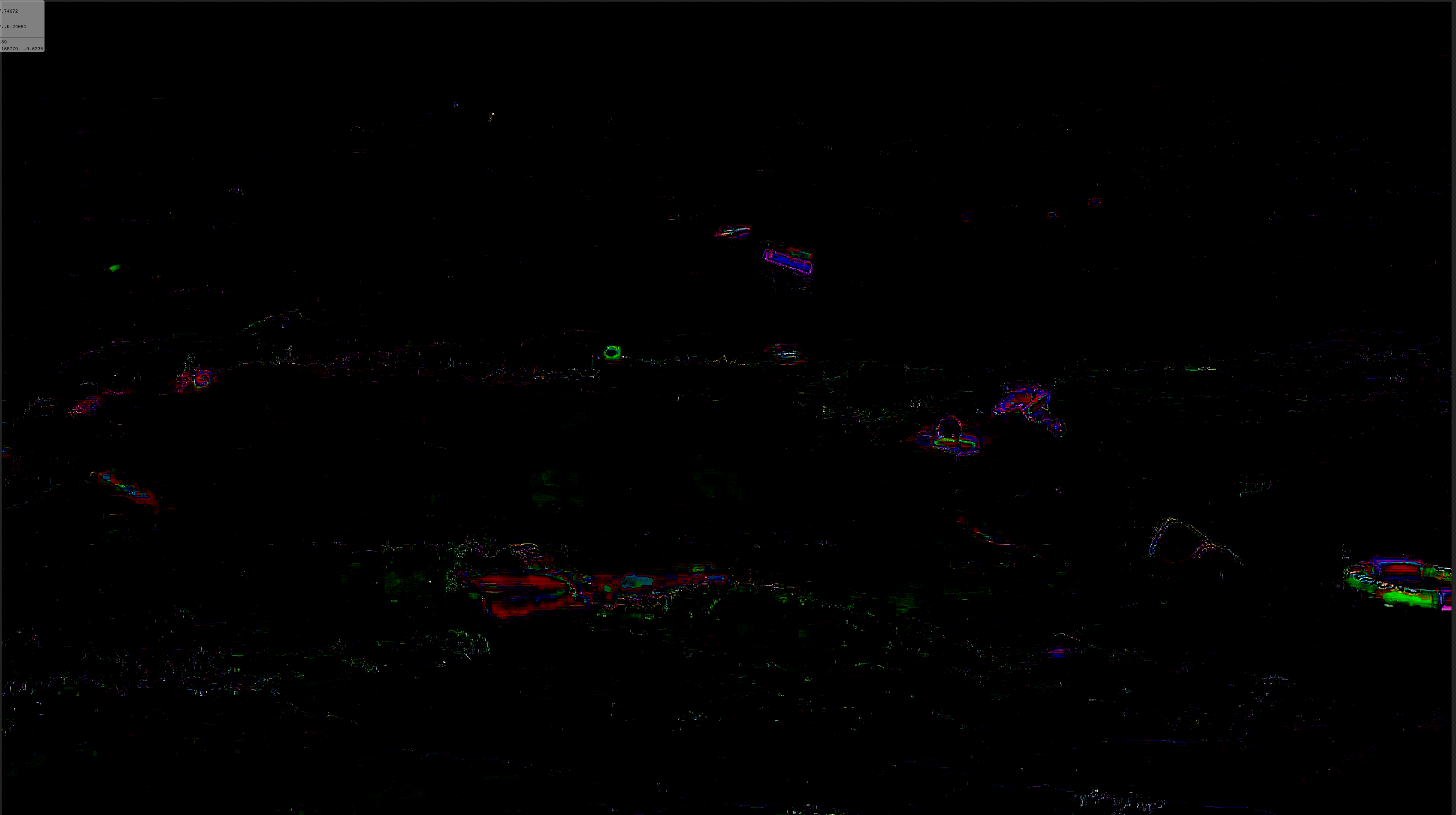}
\includegraphics[height=15mm,width=42mm]{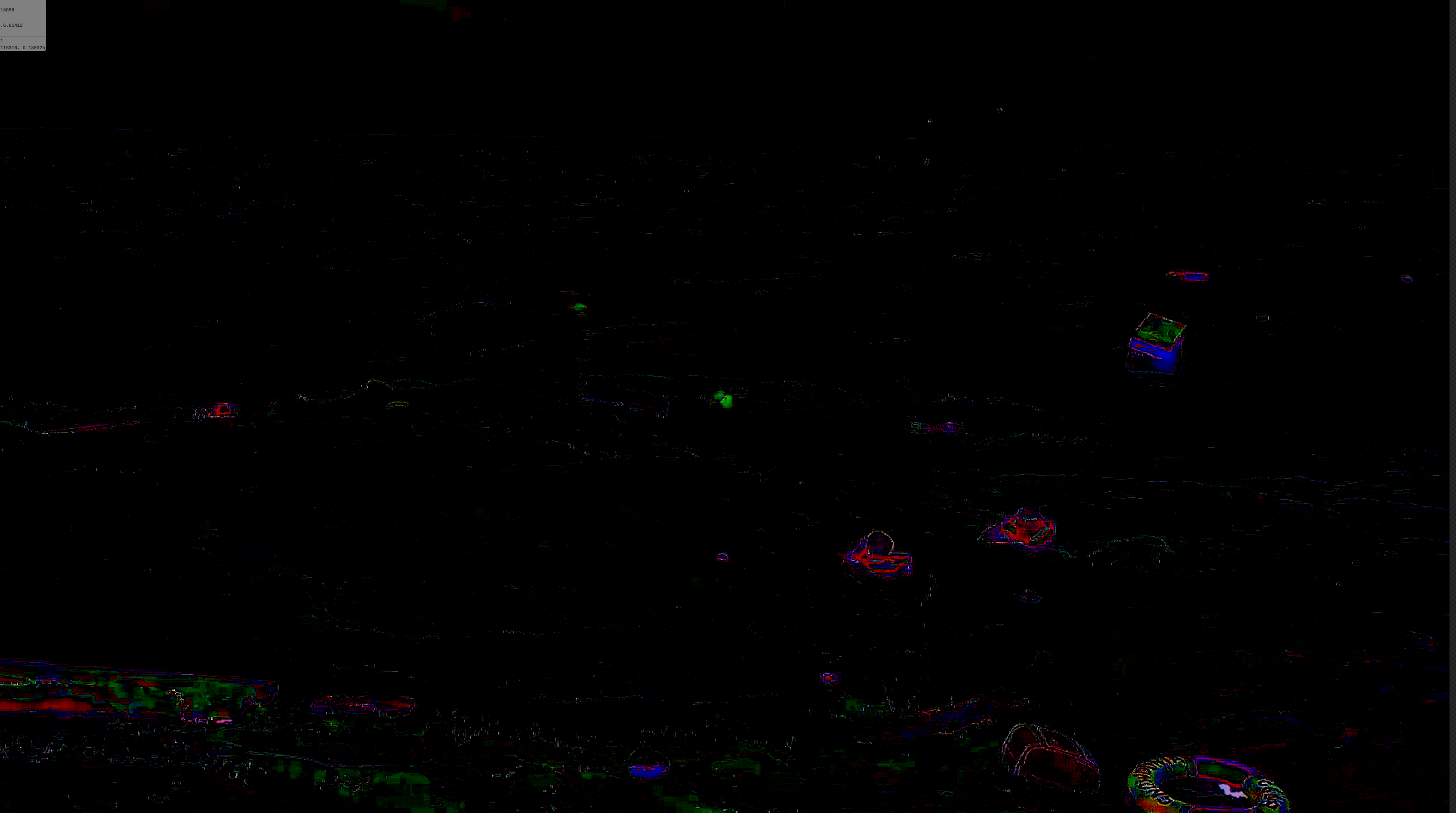}
\includegraphics[height=15mm,width=42mm]{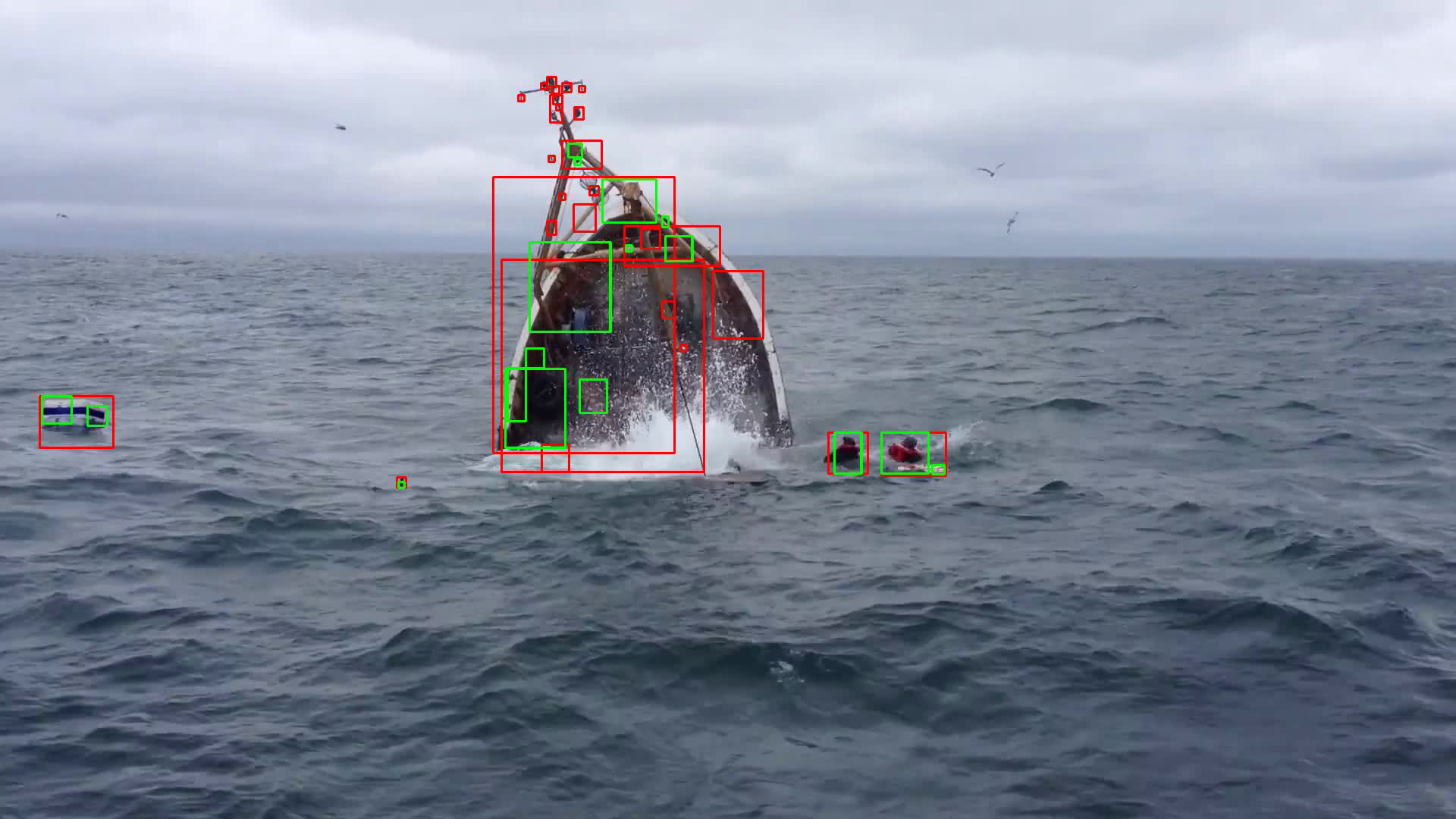}
\includegraphics[height=15mm,width=42mm]{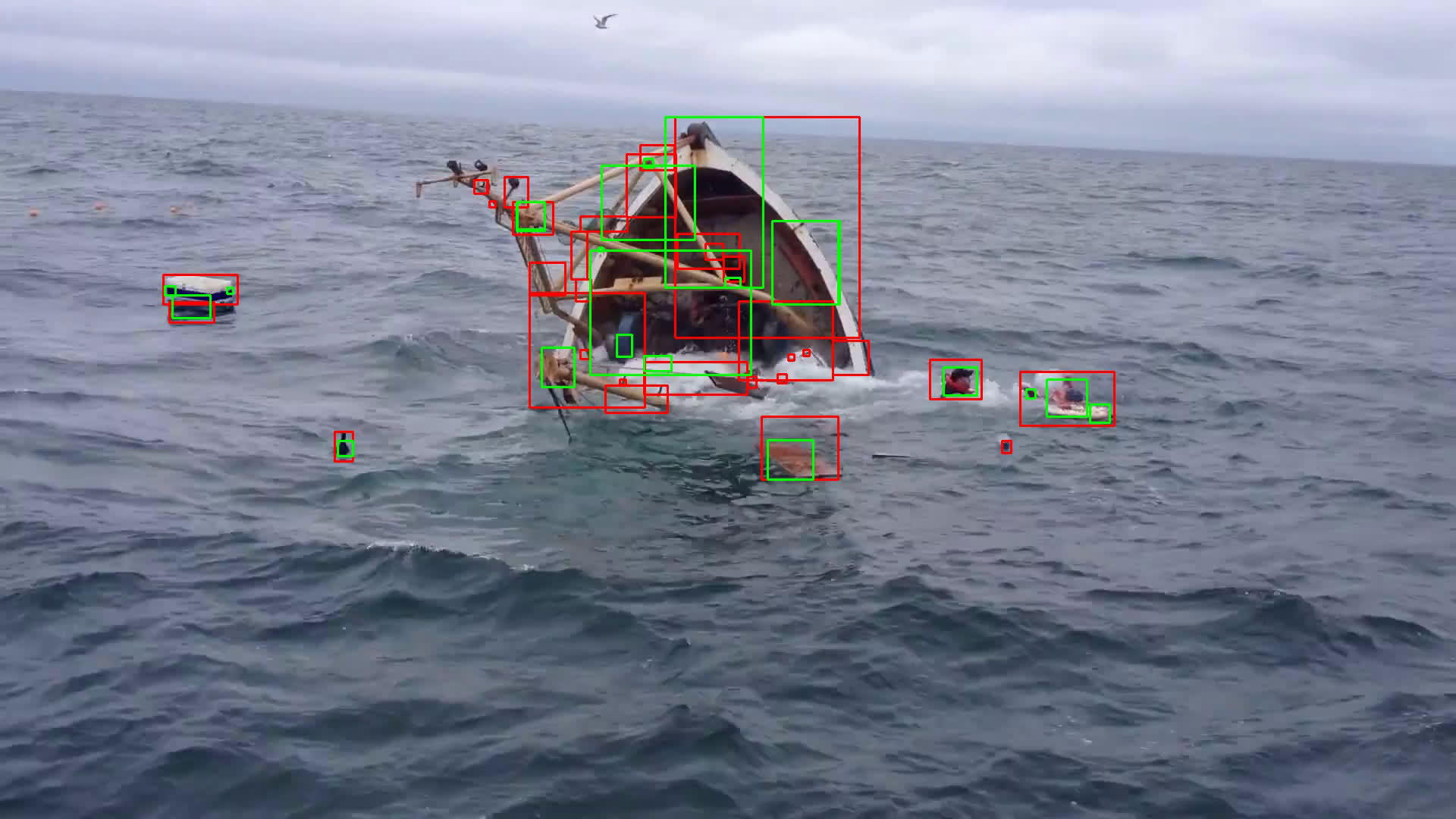}
\includegraphics[height=15mm,width=42mm]{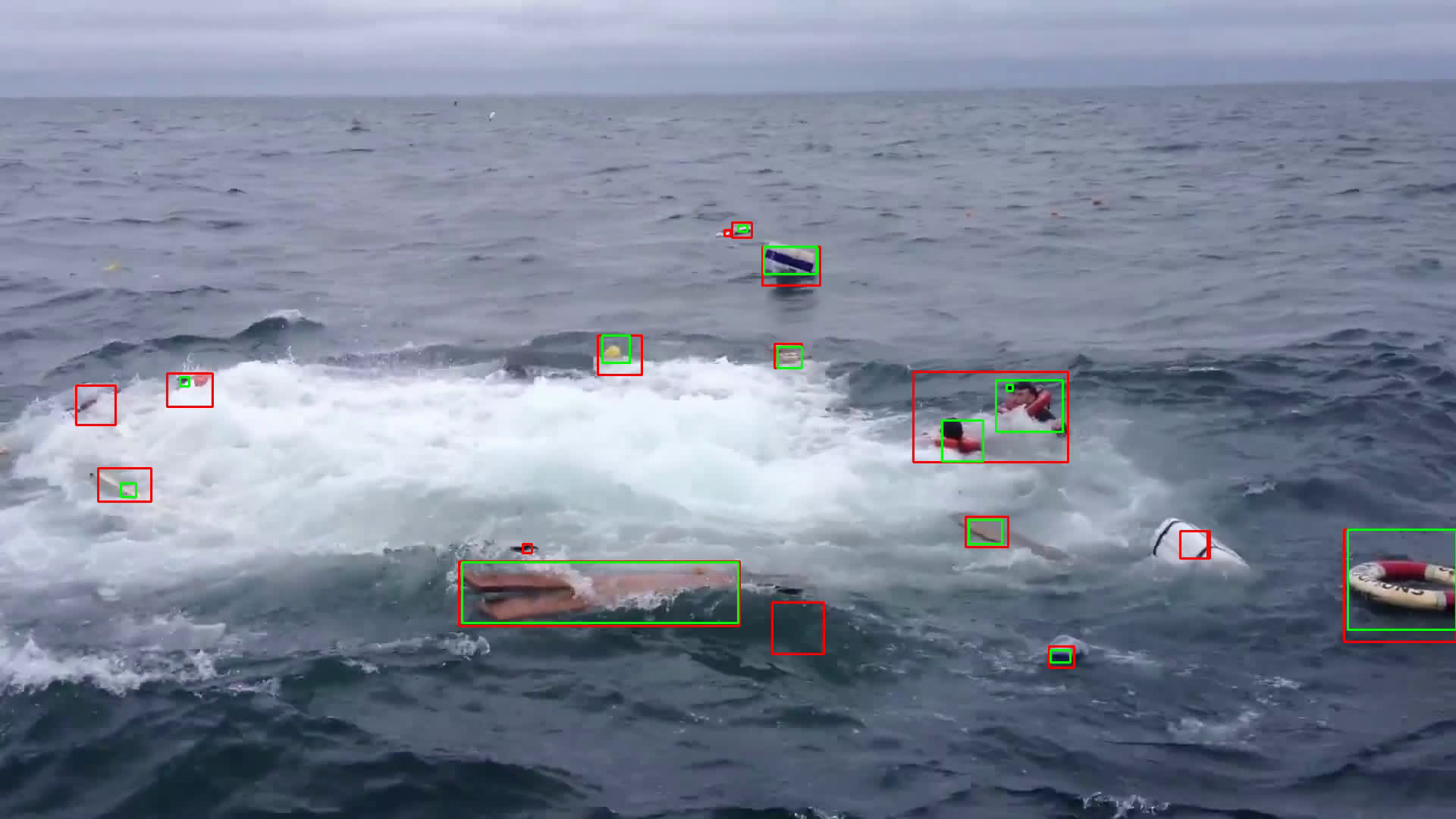}
\includegraphics[height=15mm,width=42mm]{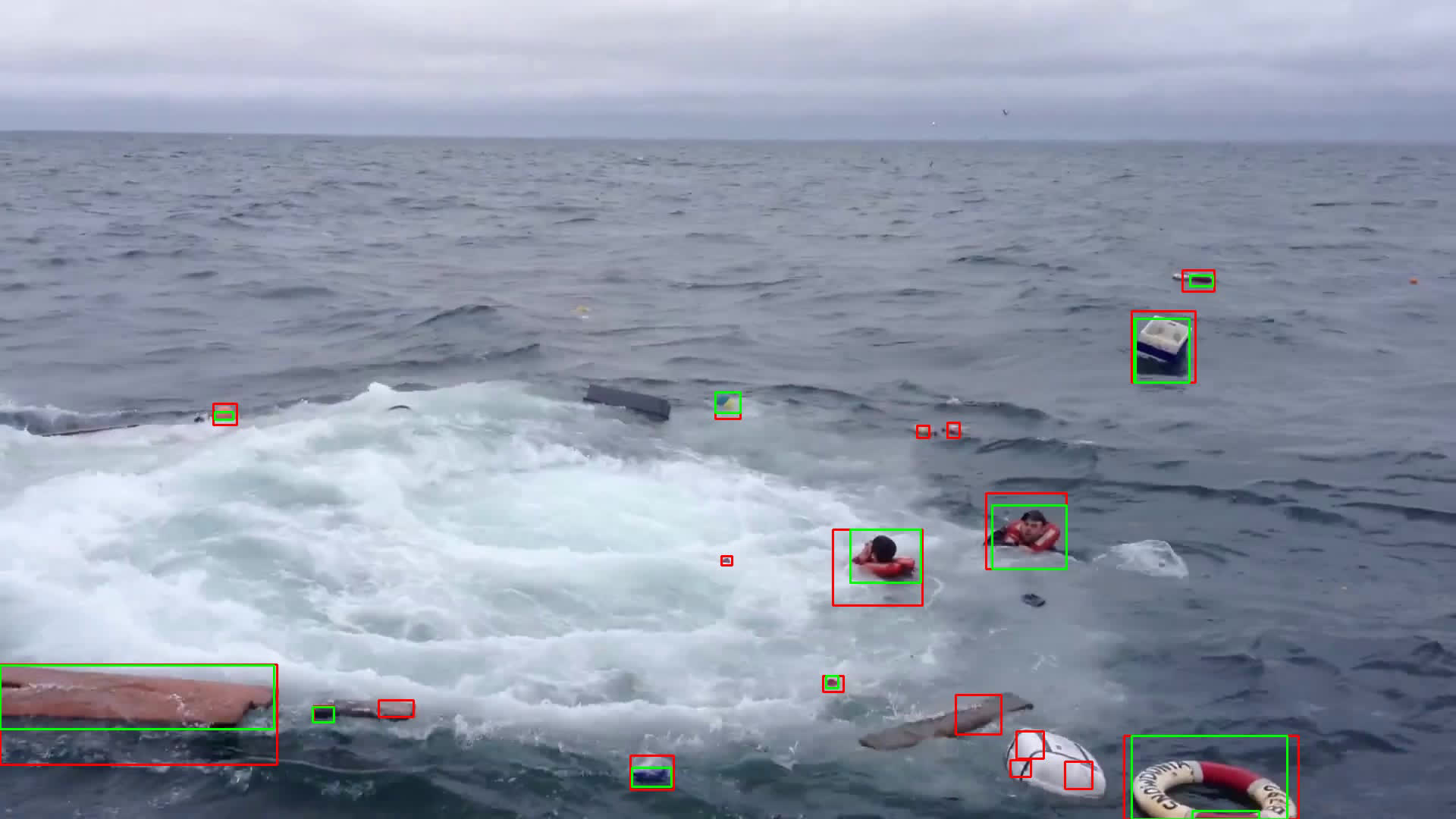}
\subfigure[Seq.A, Frame 1]{\label{fig4f}\includegraphics[height=15mm,width=42mm]{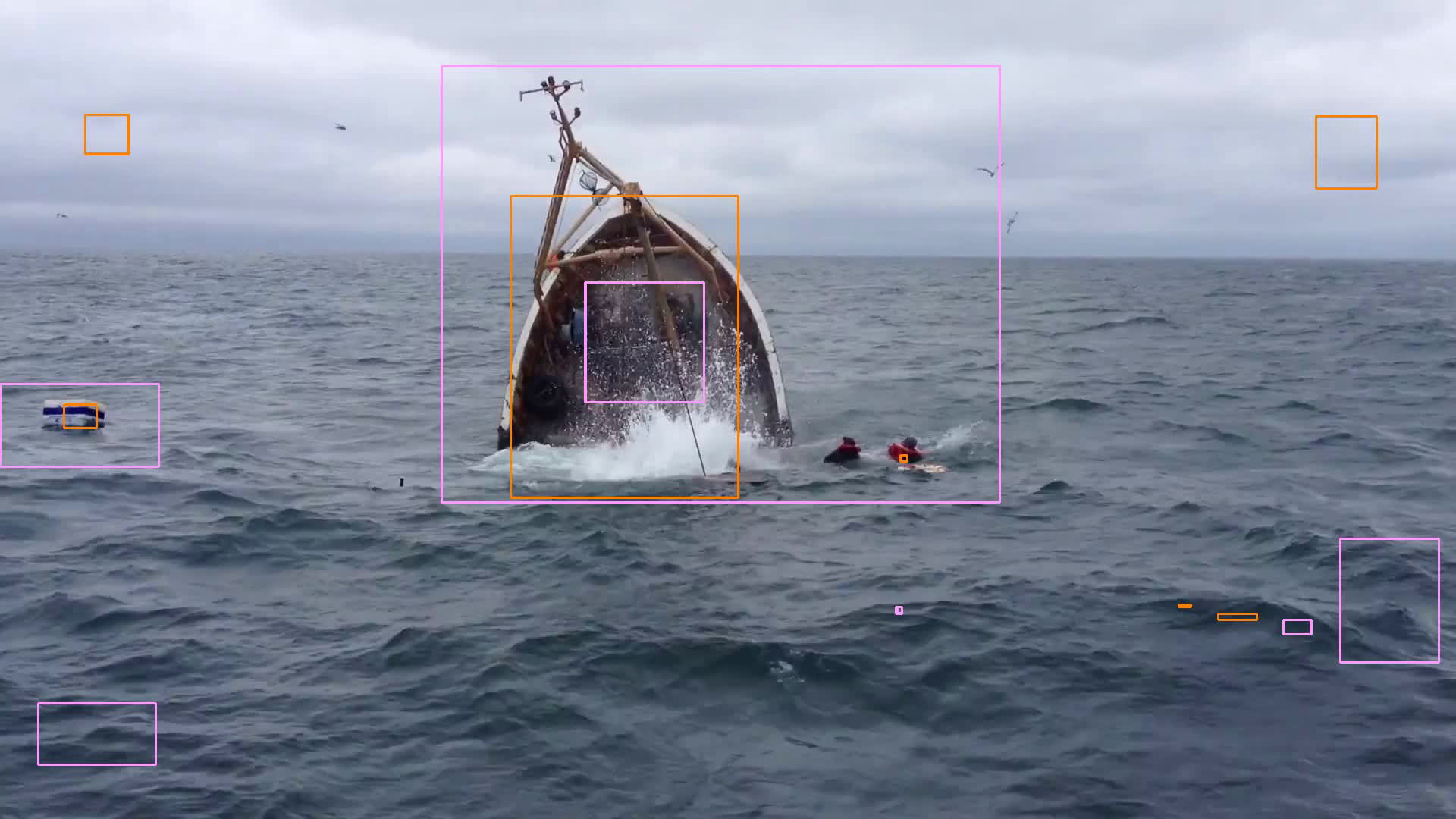}}
\subfigure[Seq.A, Frame 88]{\label{fig4f}\includegraphics[height=15mm,width=42mm]{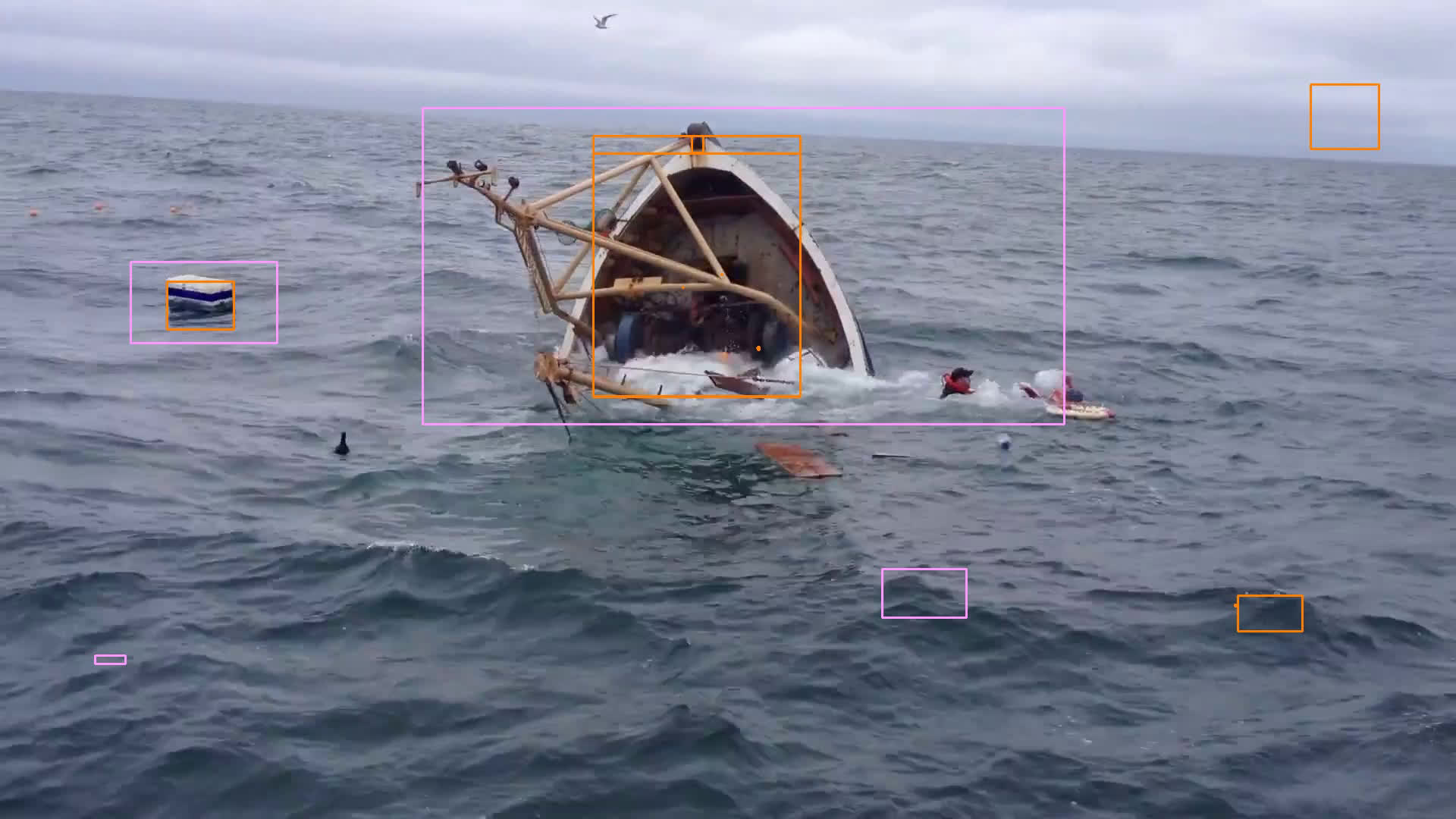}}
\subfigure[Seq.A, Frame 273]{\label{fig4f}\includegraphics[height=15mm,width=42mm]{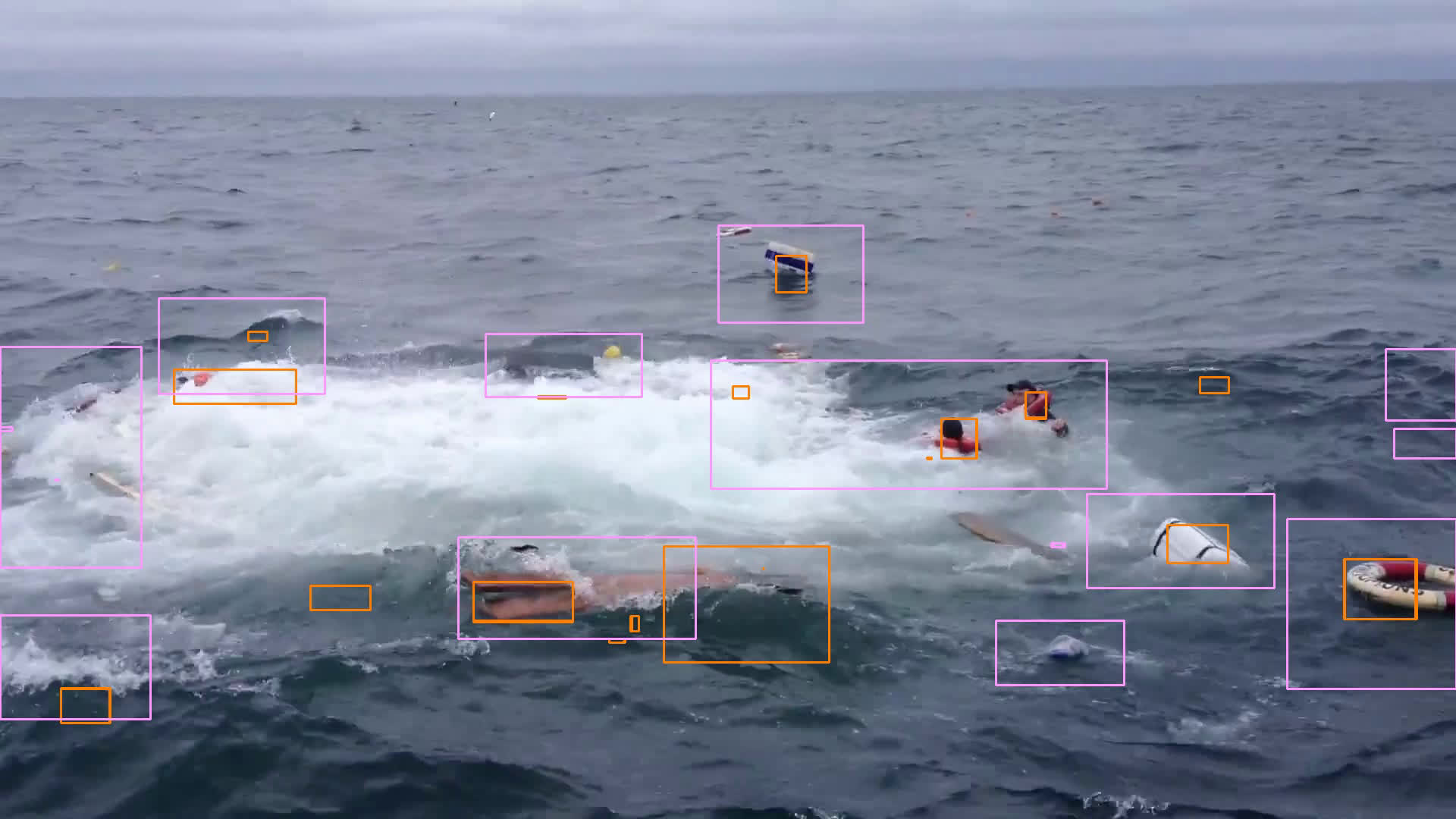}}
\subfigure[Seq.A, Frame 358]{\label{fig4f}\includegraphics[height=15mm,width=42mm]{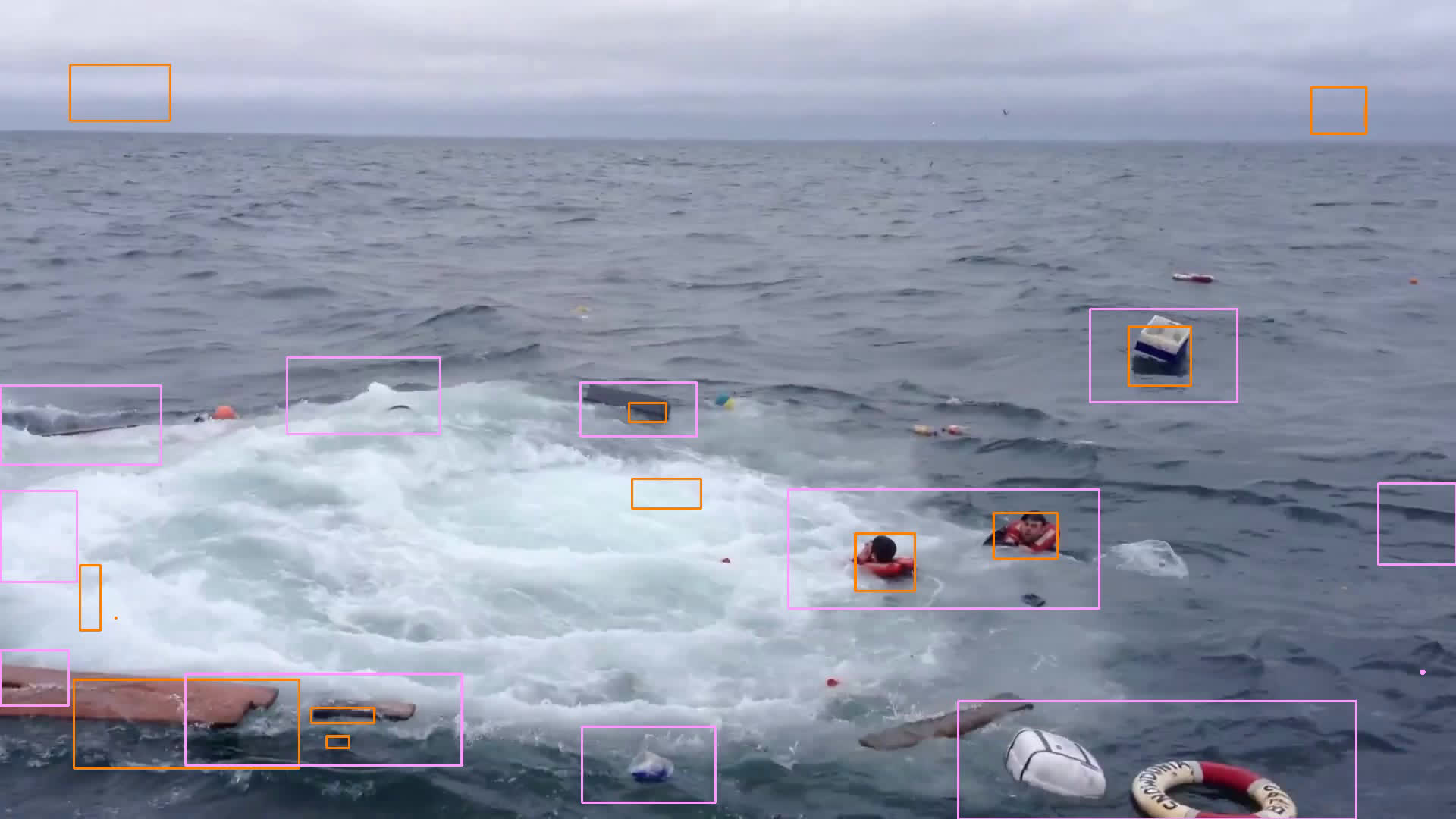}}

\includegraphics[height=15mm,width=42mm]{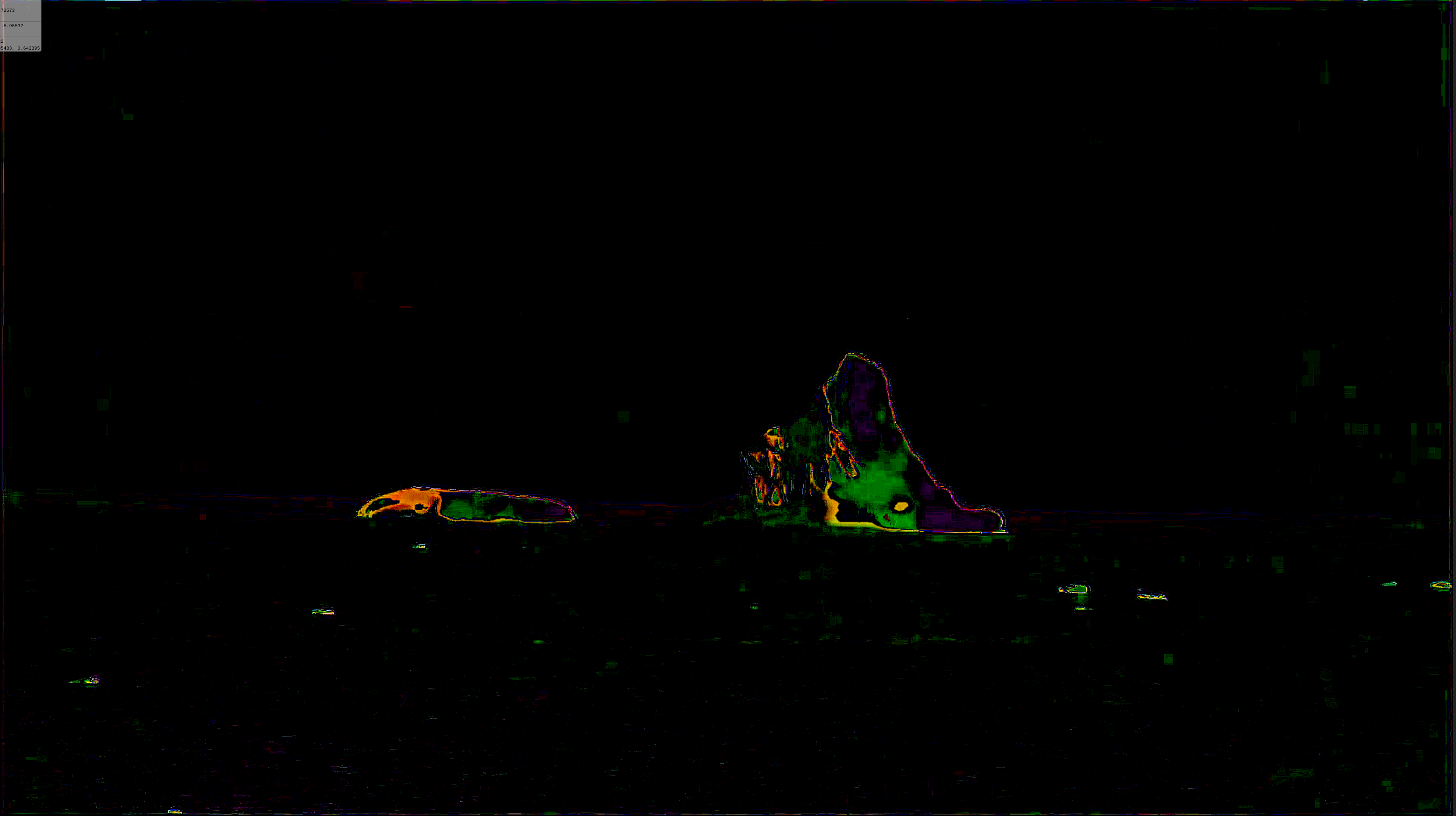}
\includegraphics[height=15mm,width=42mm]{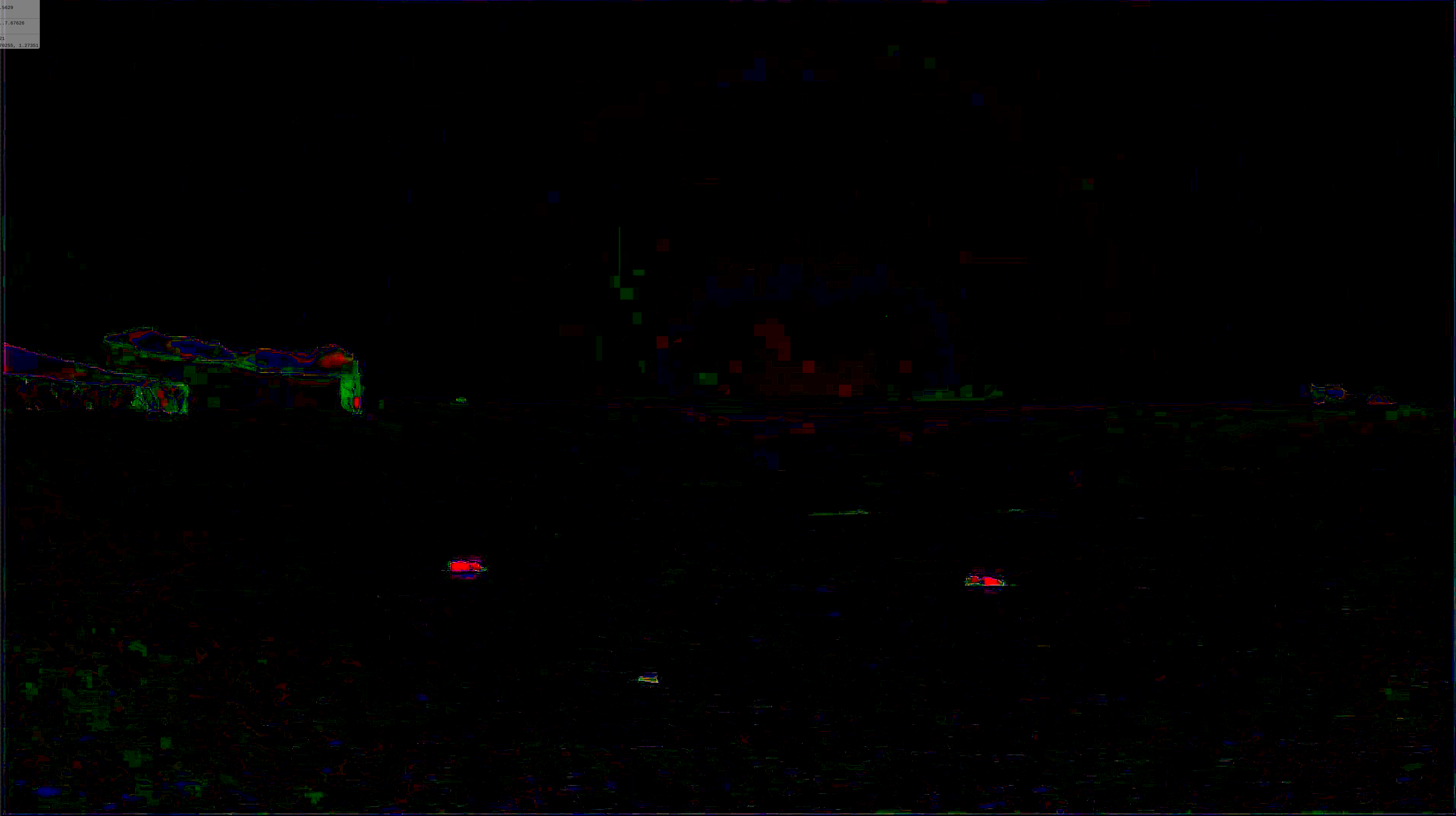}
\includegraphics[height=15mm,width=42mm]{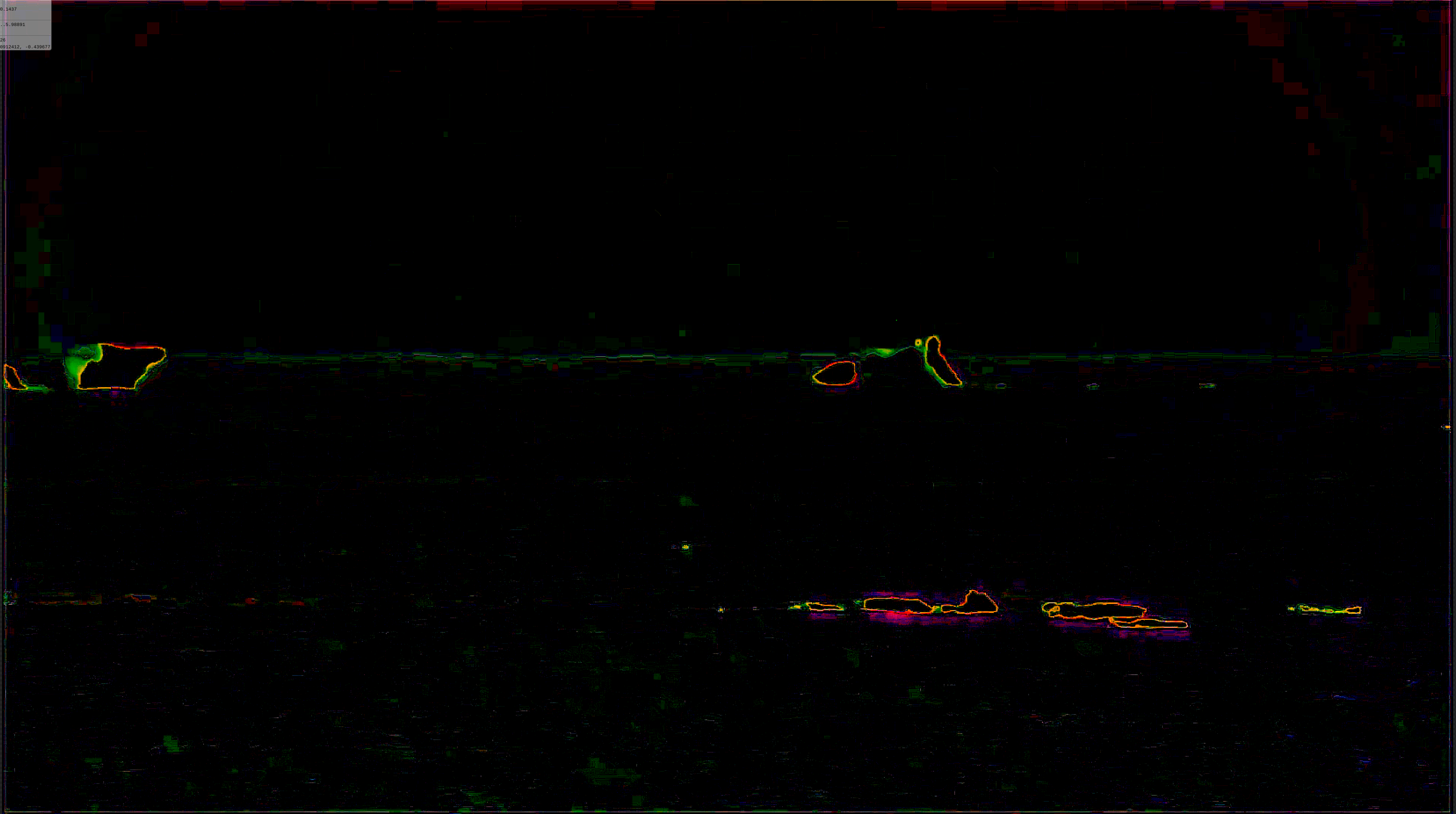}
\includegraphics[height=15mm,width=42mm]{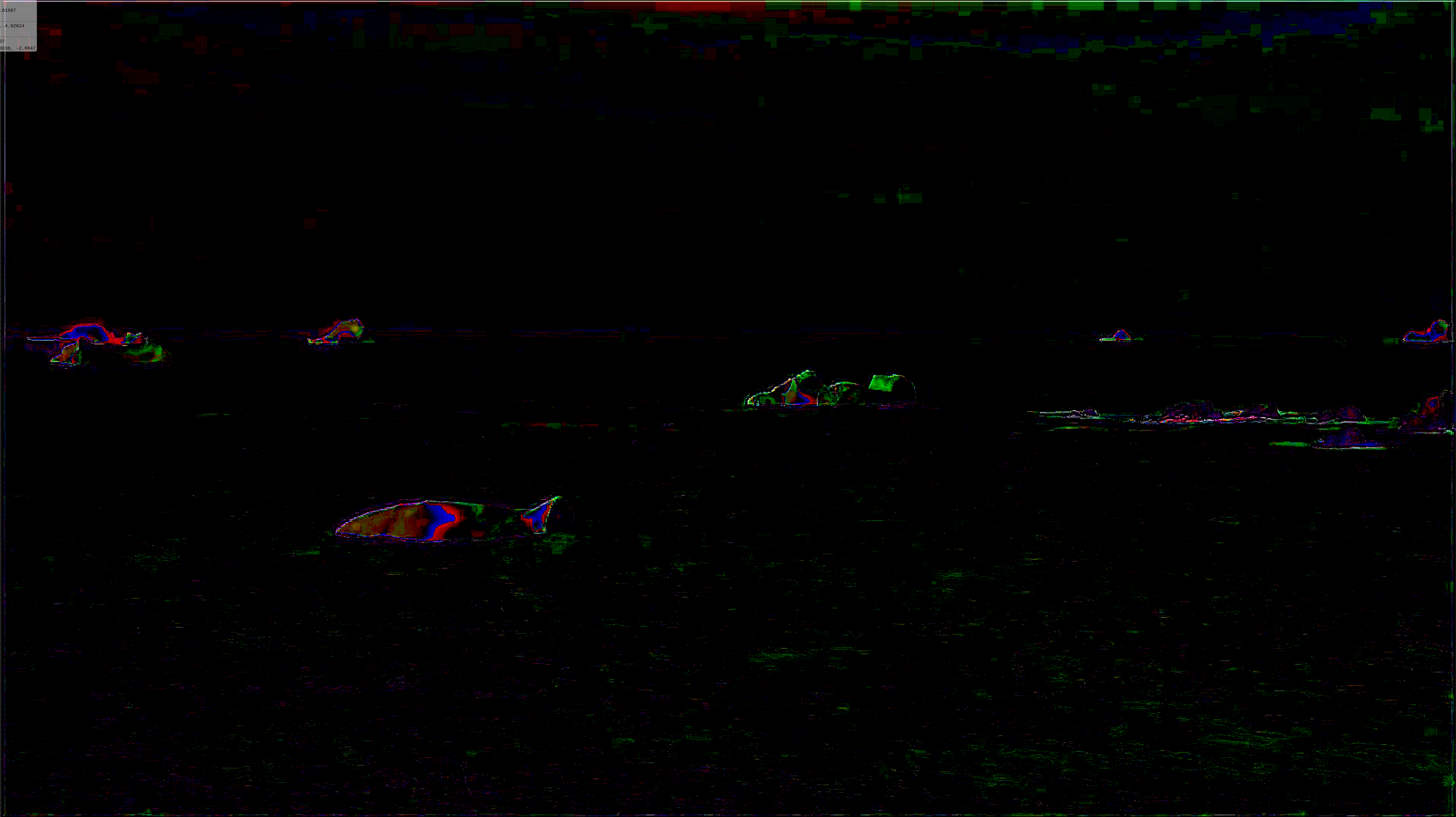}
\includegraphics[height=15mm,width=42mm]{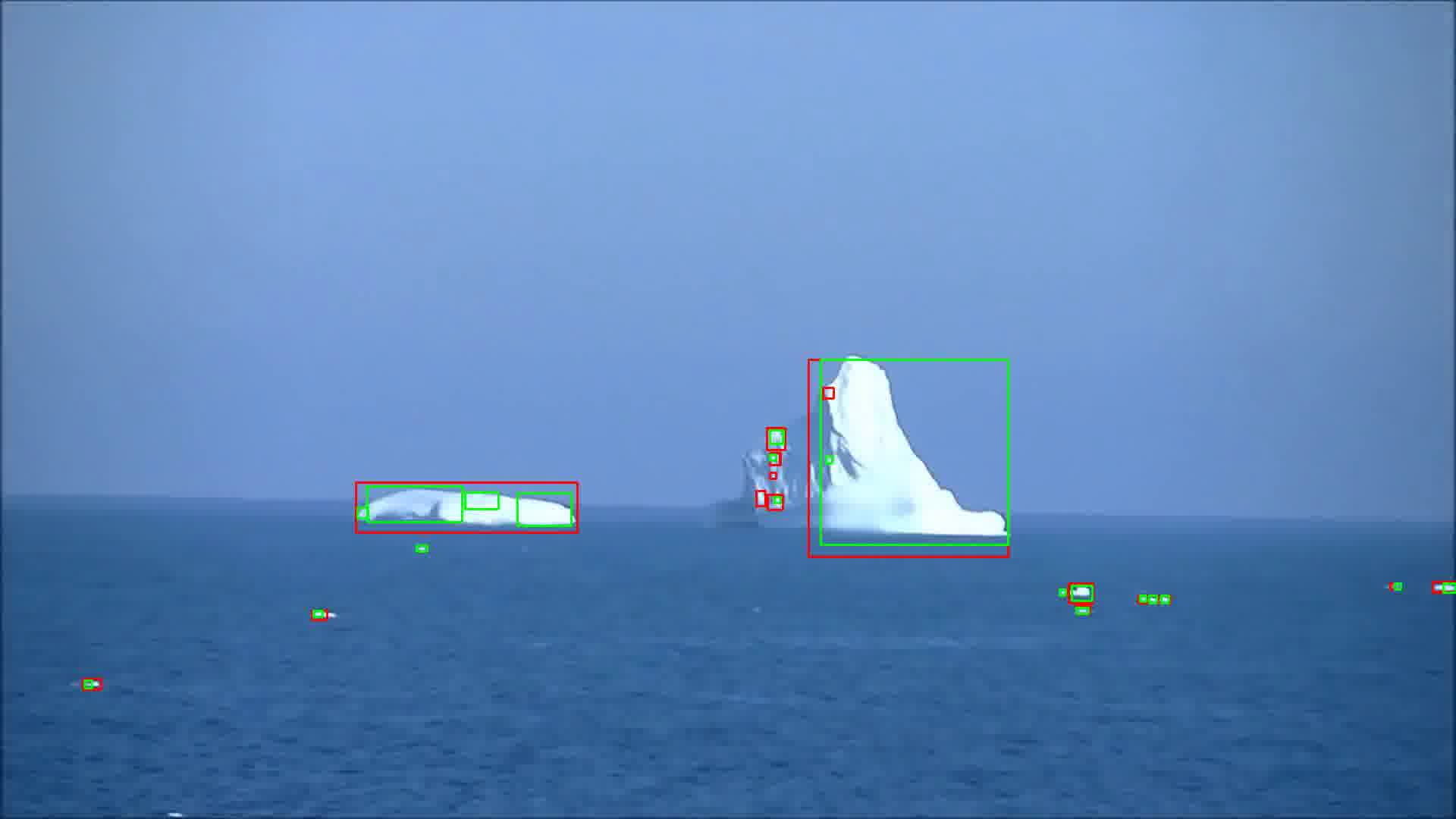}
\includegraphics[height=15mm,width=42mm]{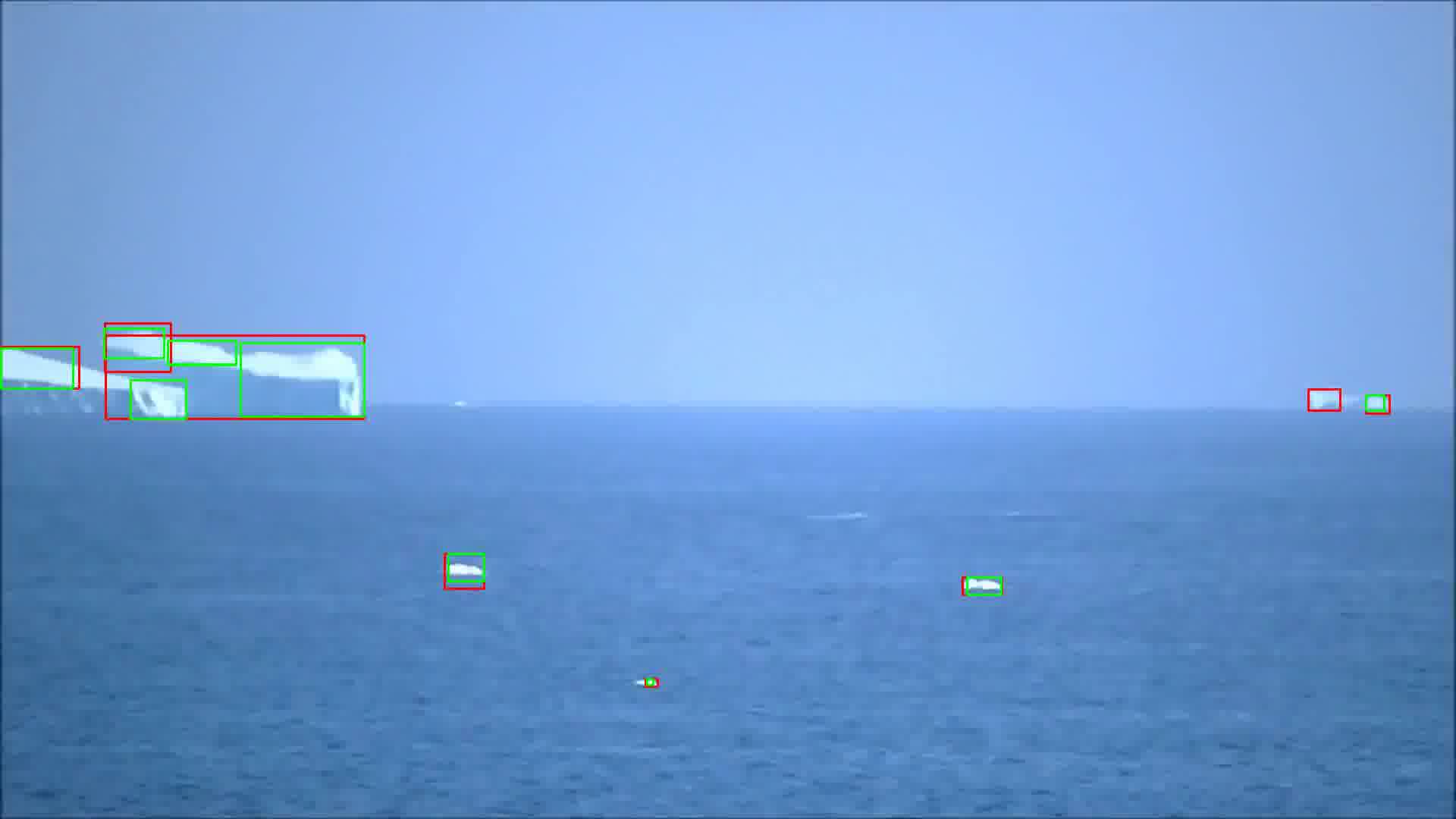}
\includegraphics[height=15mm,width=42mm]{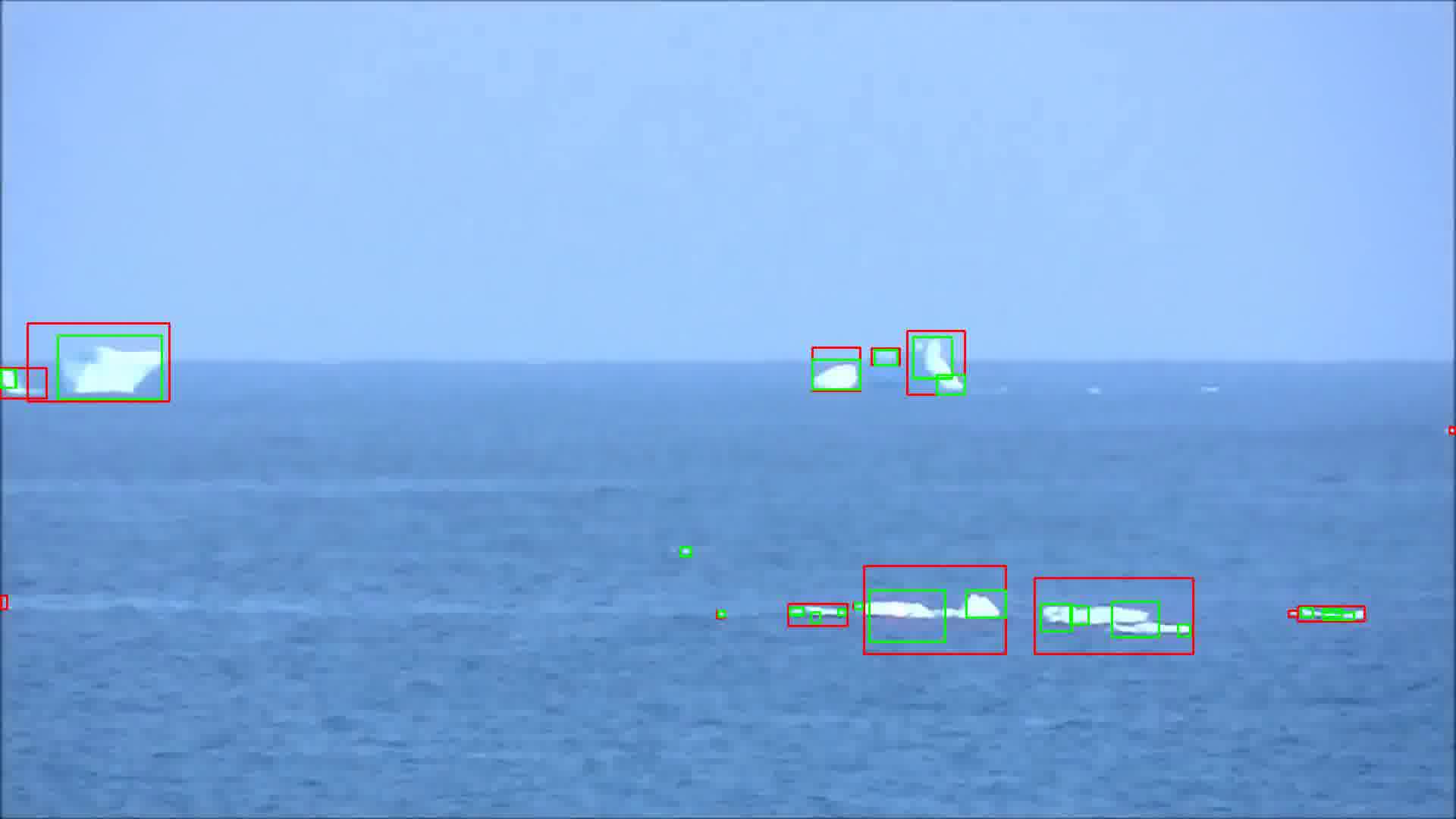}
\includegraphics[height=15mm,width=42mm]{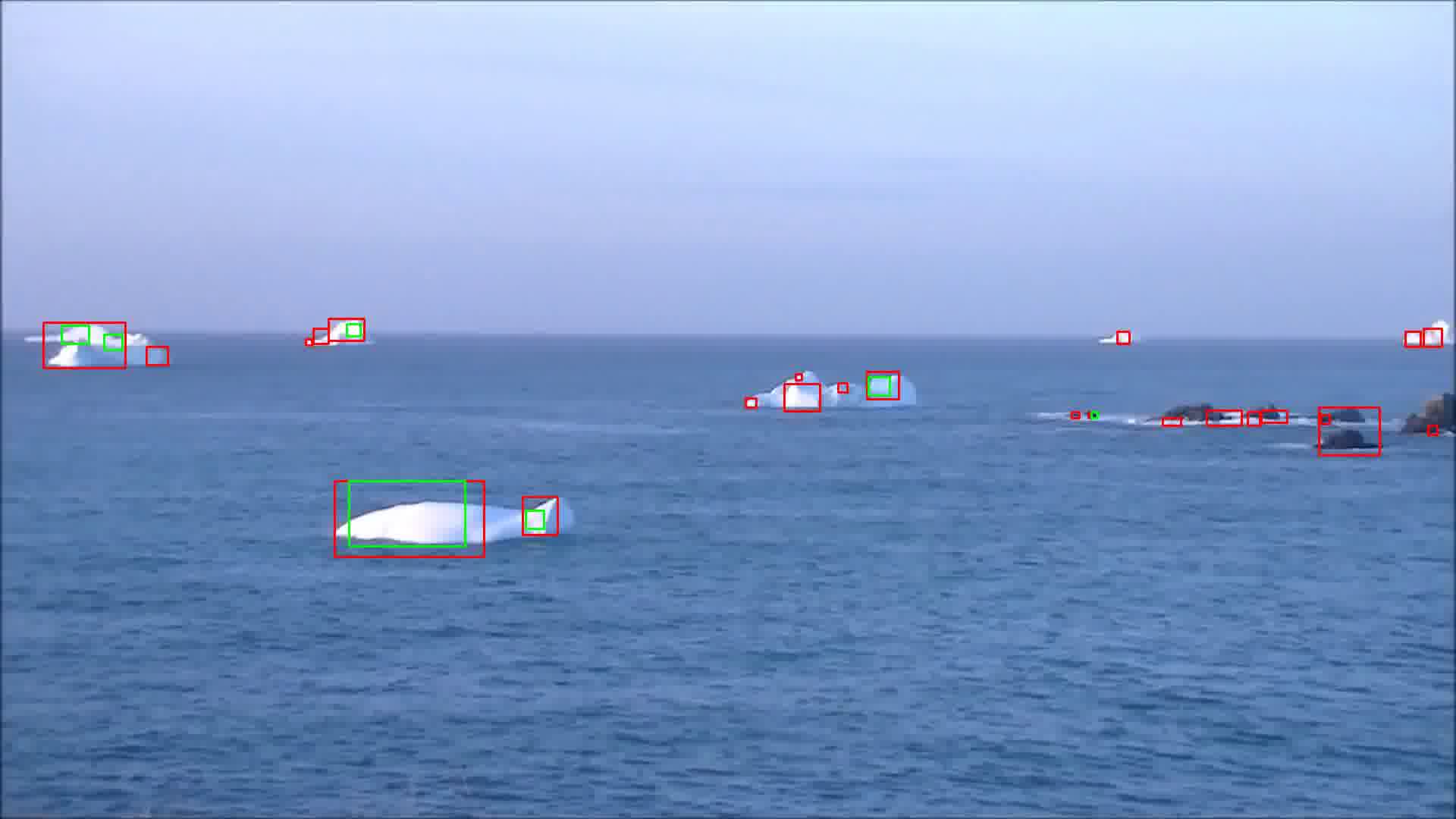}
\subfigure[Seq.B, Frame 183]{\label{fig4f}\includegraphics[height=15mm,width=42mm]{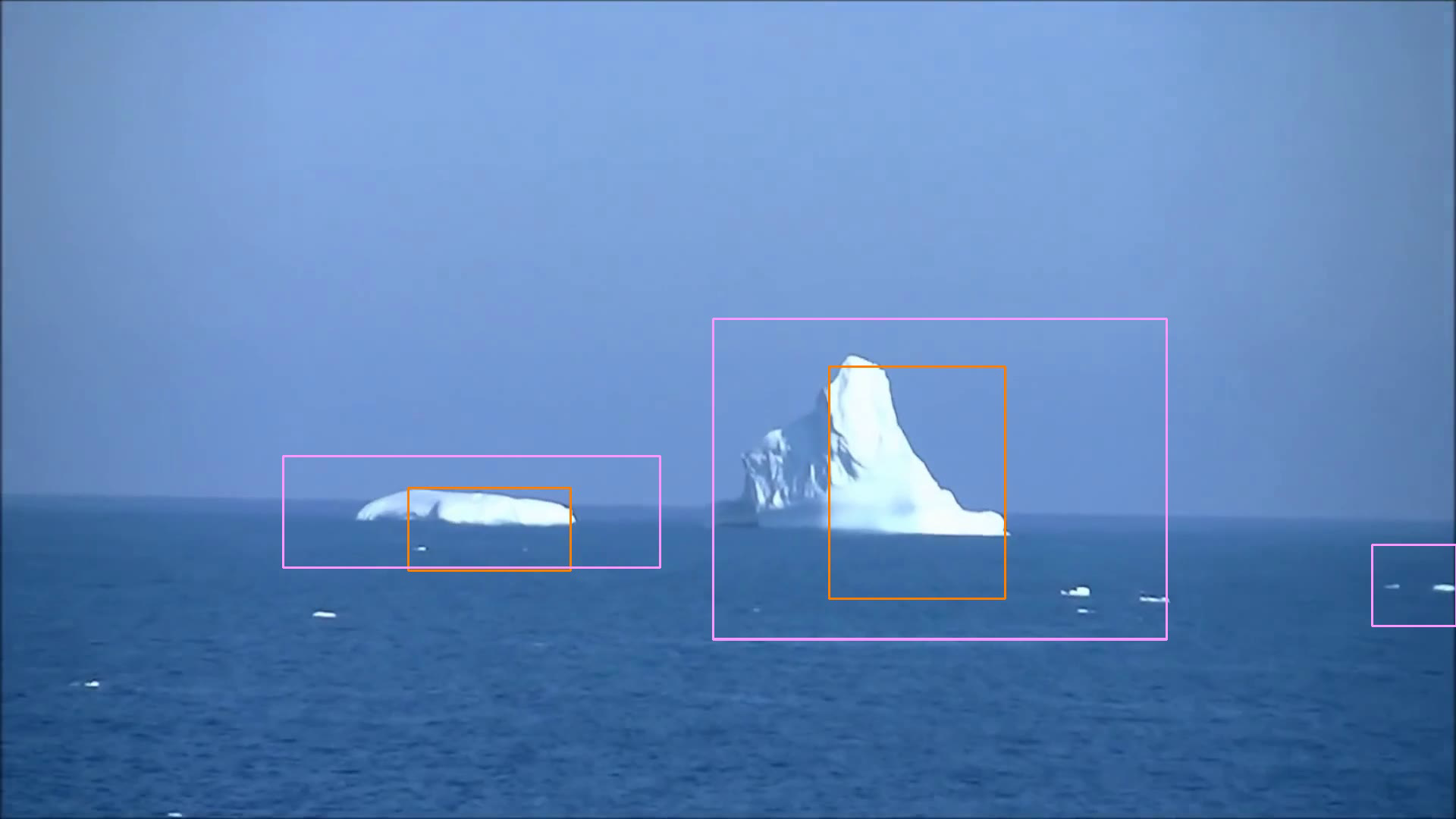}}
\subfigure[Seq.B, Frame 368]{\label{fig4f}\includegraphics[height=15mm,width=42mm]{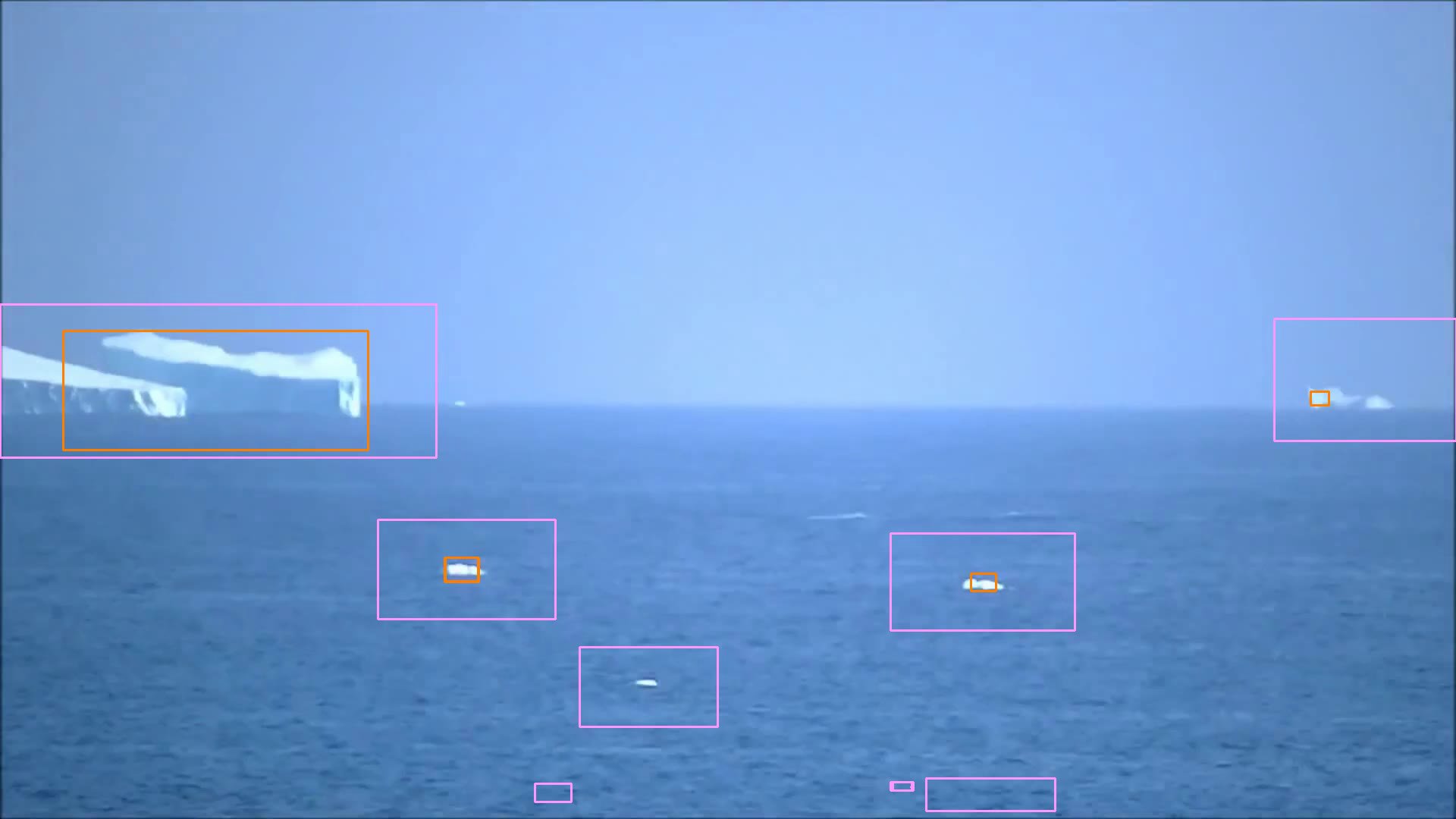}}
\subfigure[Seq.B, Frame 610]{\label{fig4f}\includegraphics[height=15mm,width=42mm]{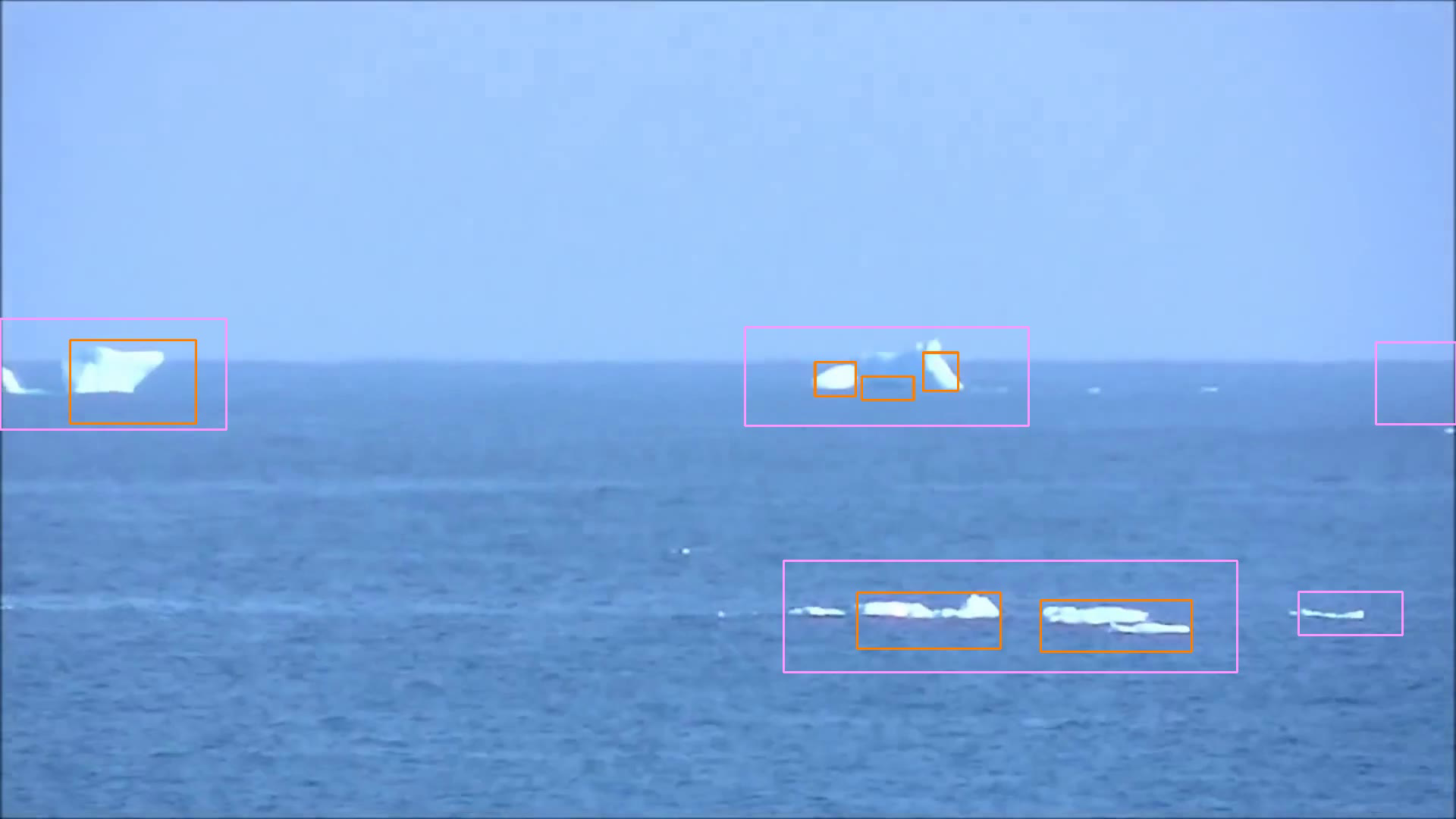}}
\subfigure[Seq.B, Frame 869]{\label{fig4f}\includegraphics[height=15mm,width=42mm]{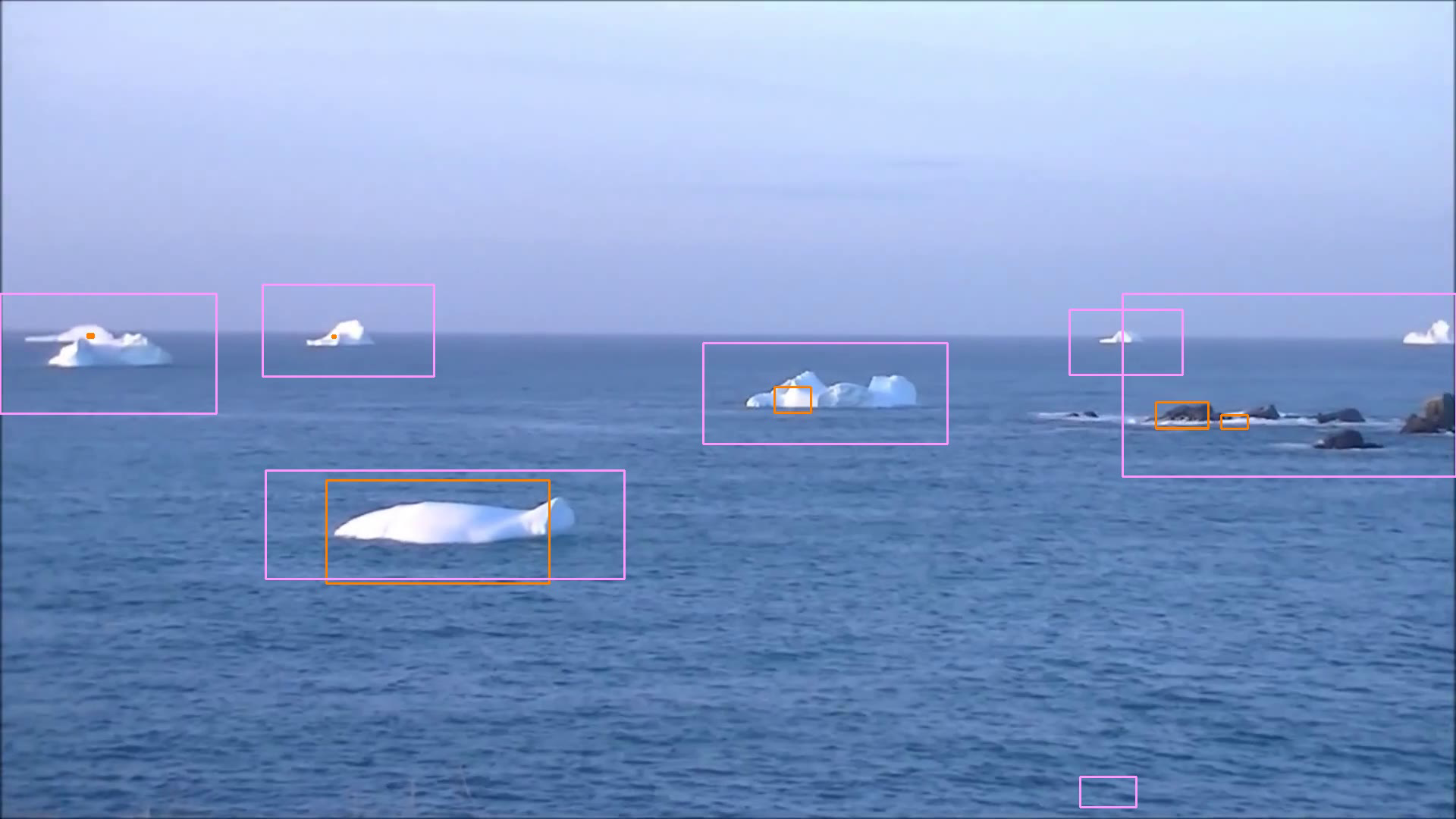}}

\includegraphics[height=15mm,width=42mm]{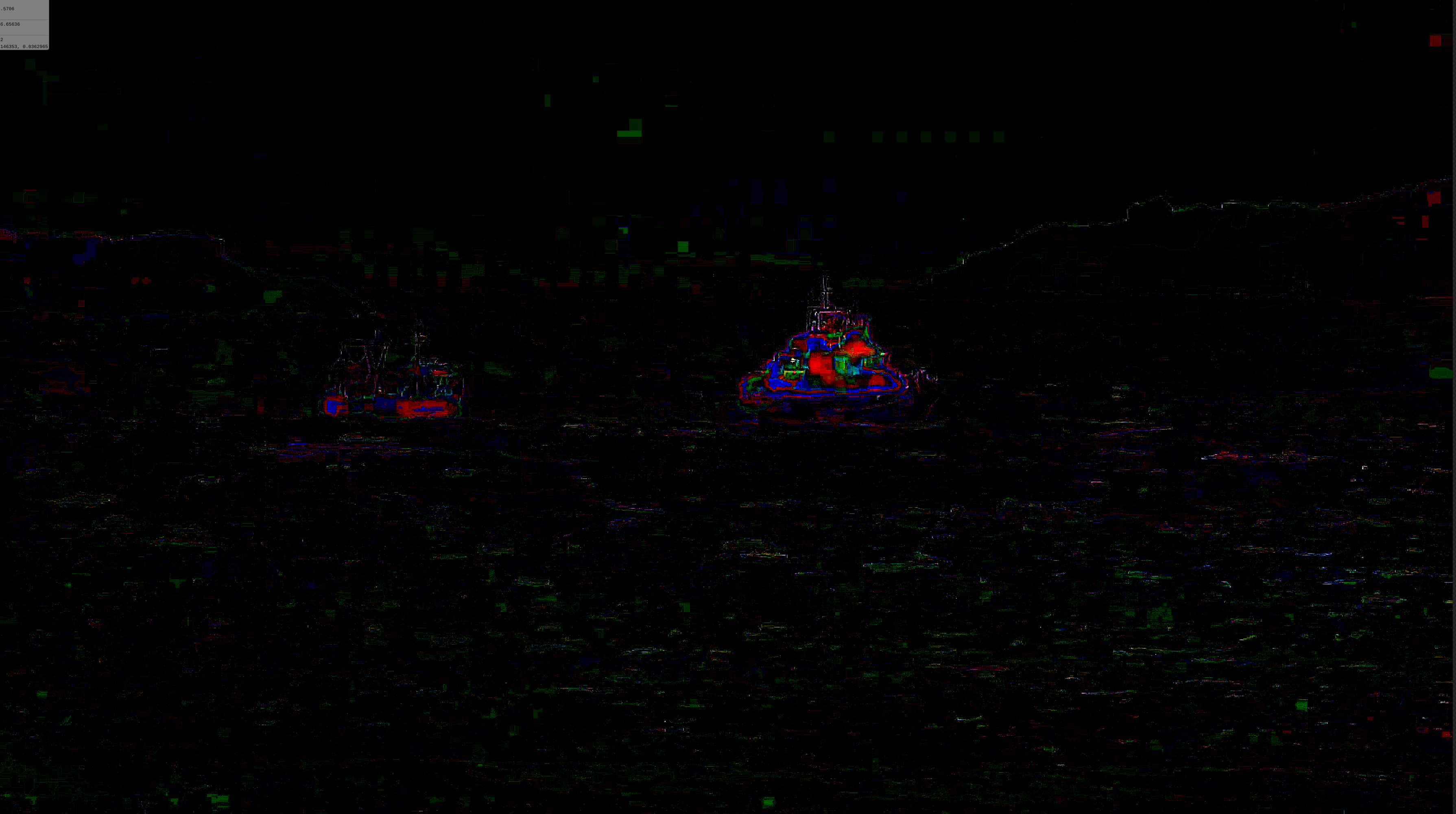}
\includegraphics[height=15mm,width=42mm]{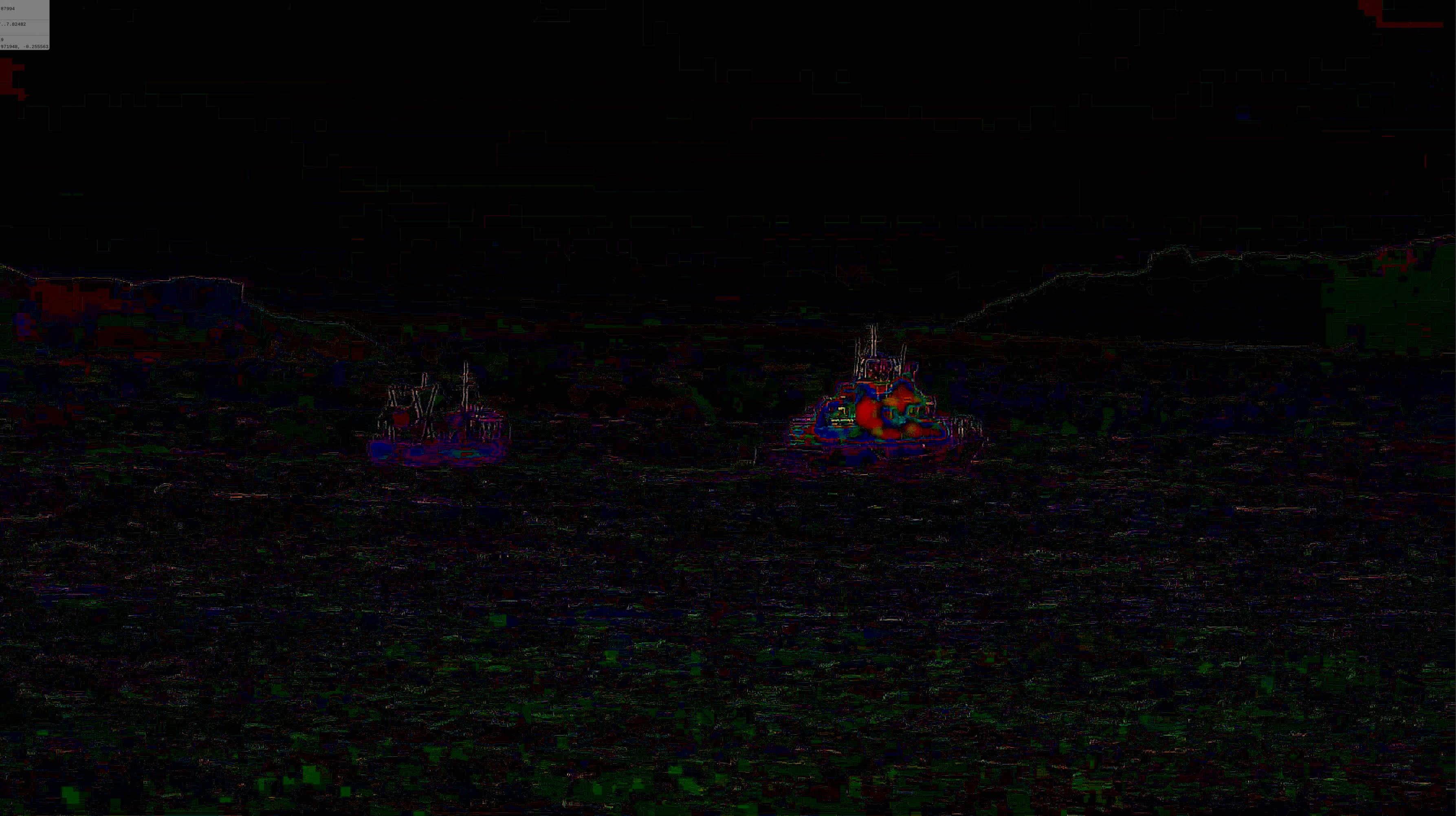}
\includegraphics[height=15mm,width=42mm]{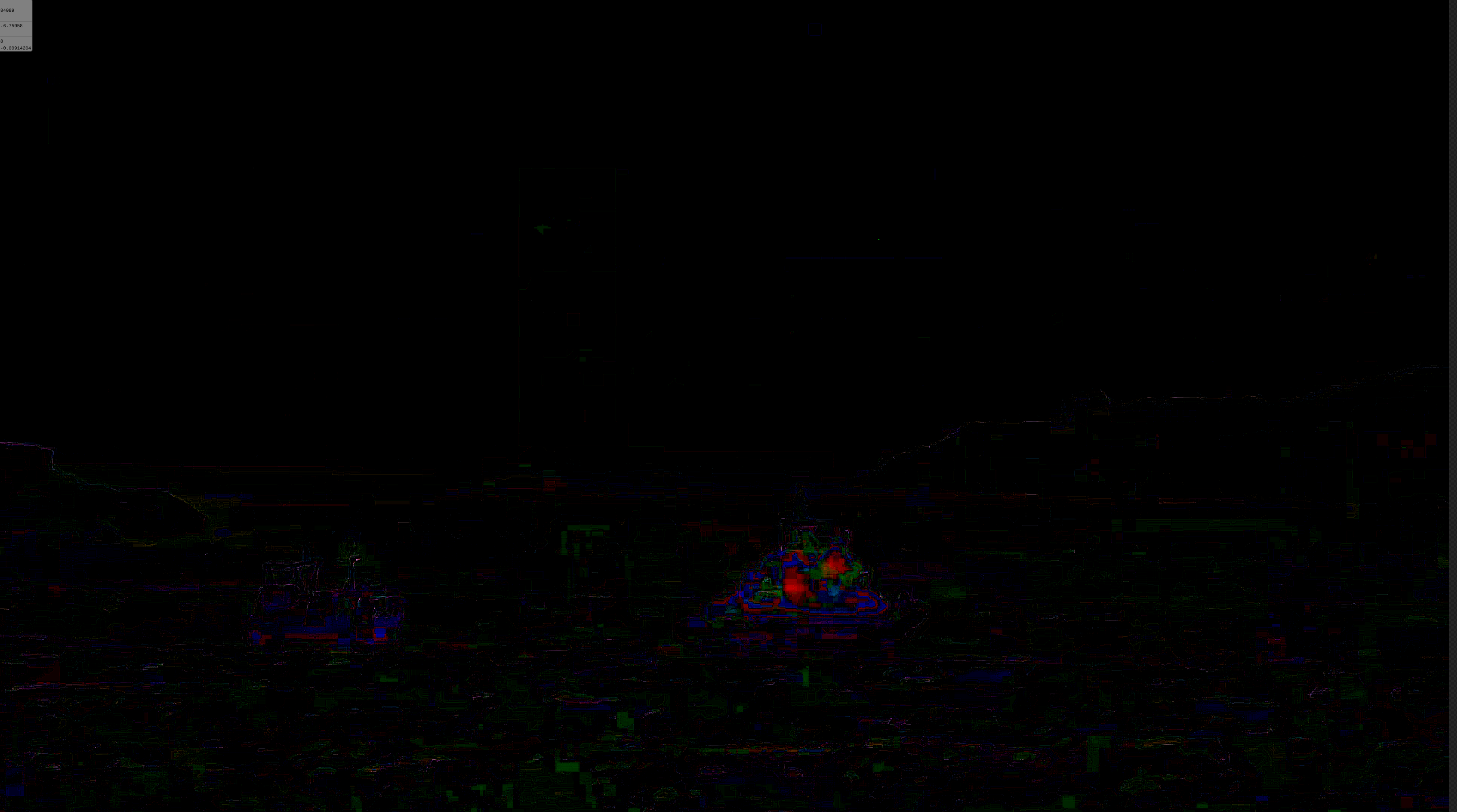}
\includegraphics[height=15mm,width=42mm]{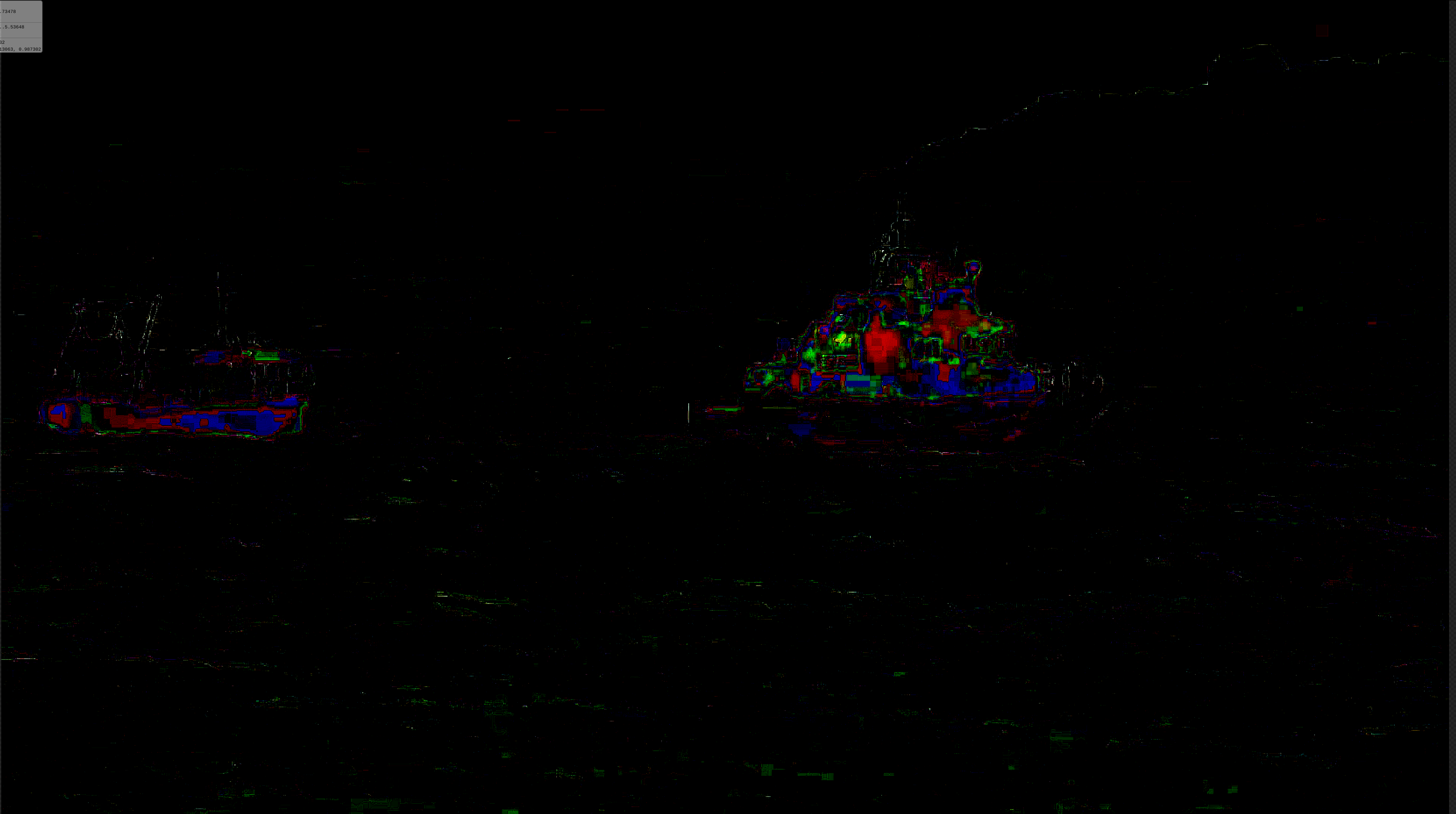}
\includegraphics[height=15mm,width=42mm]{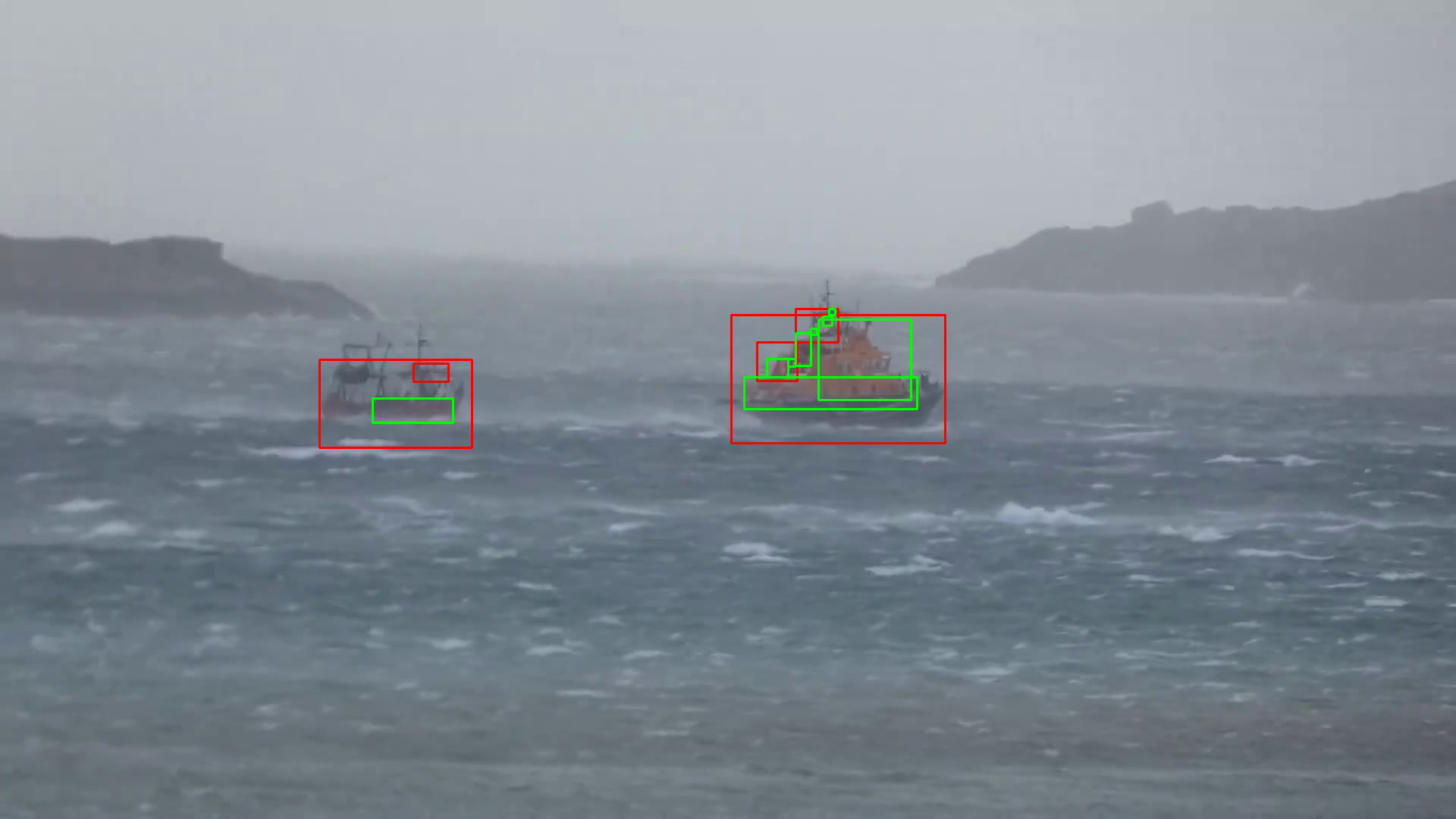}
\includegraphics[height=15mm,width=42mm]{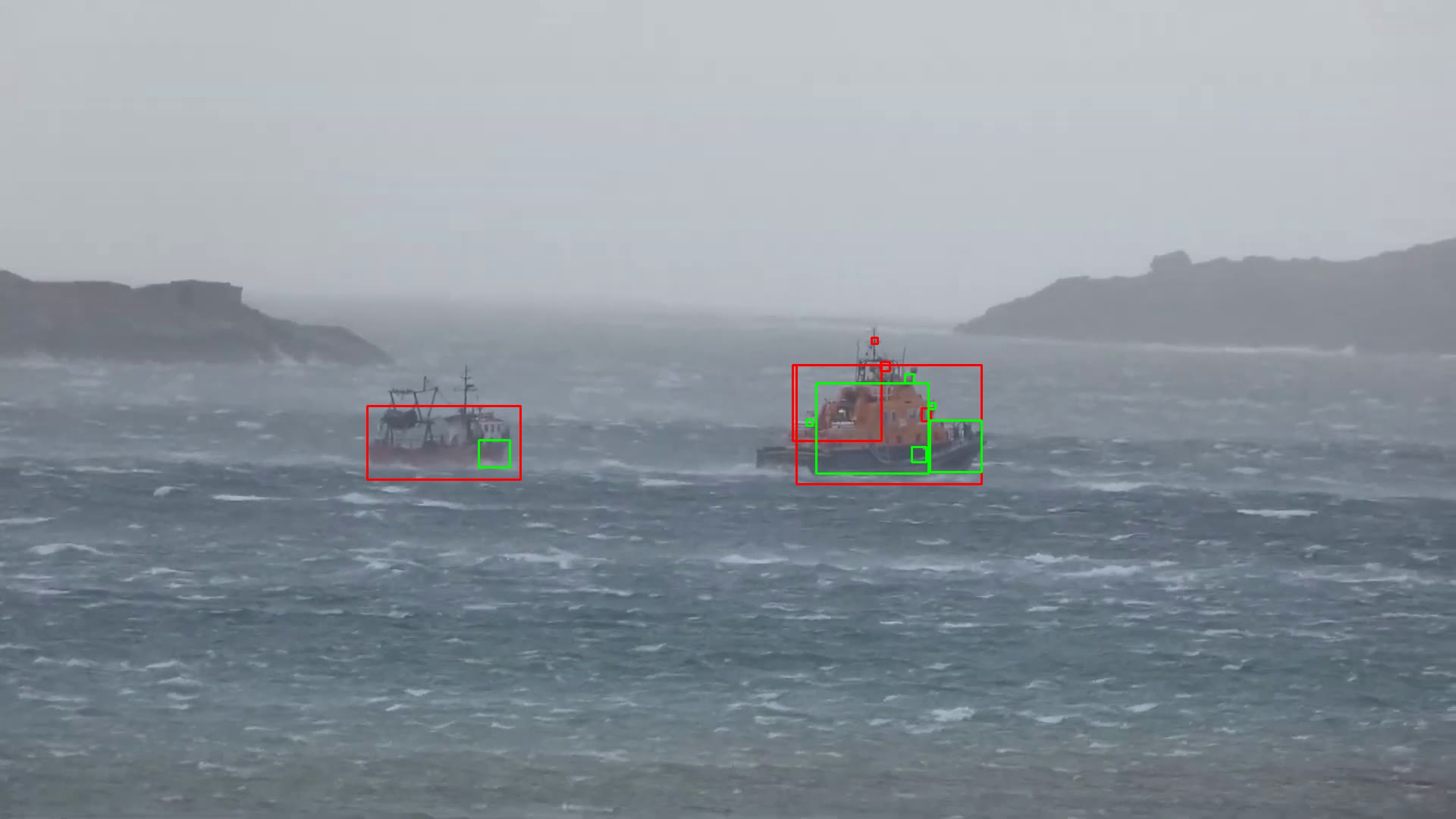}
\includegraphics[height=15mm,width=42mm]{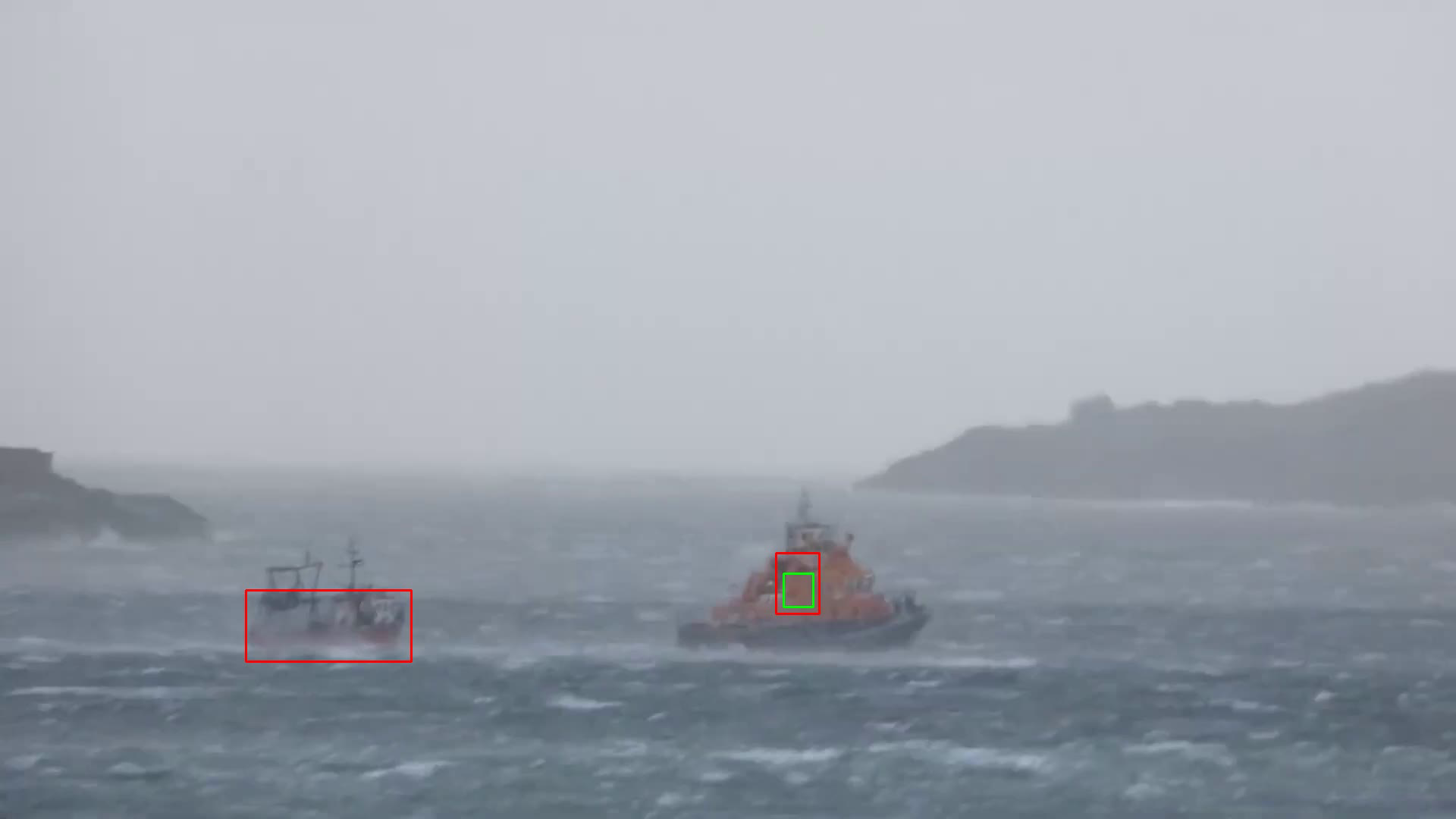}
\includegraphics[height=15mm,width=42mm]{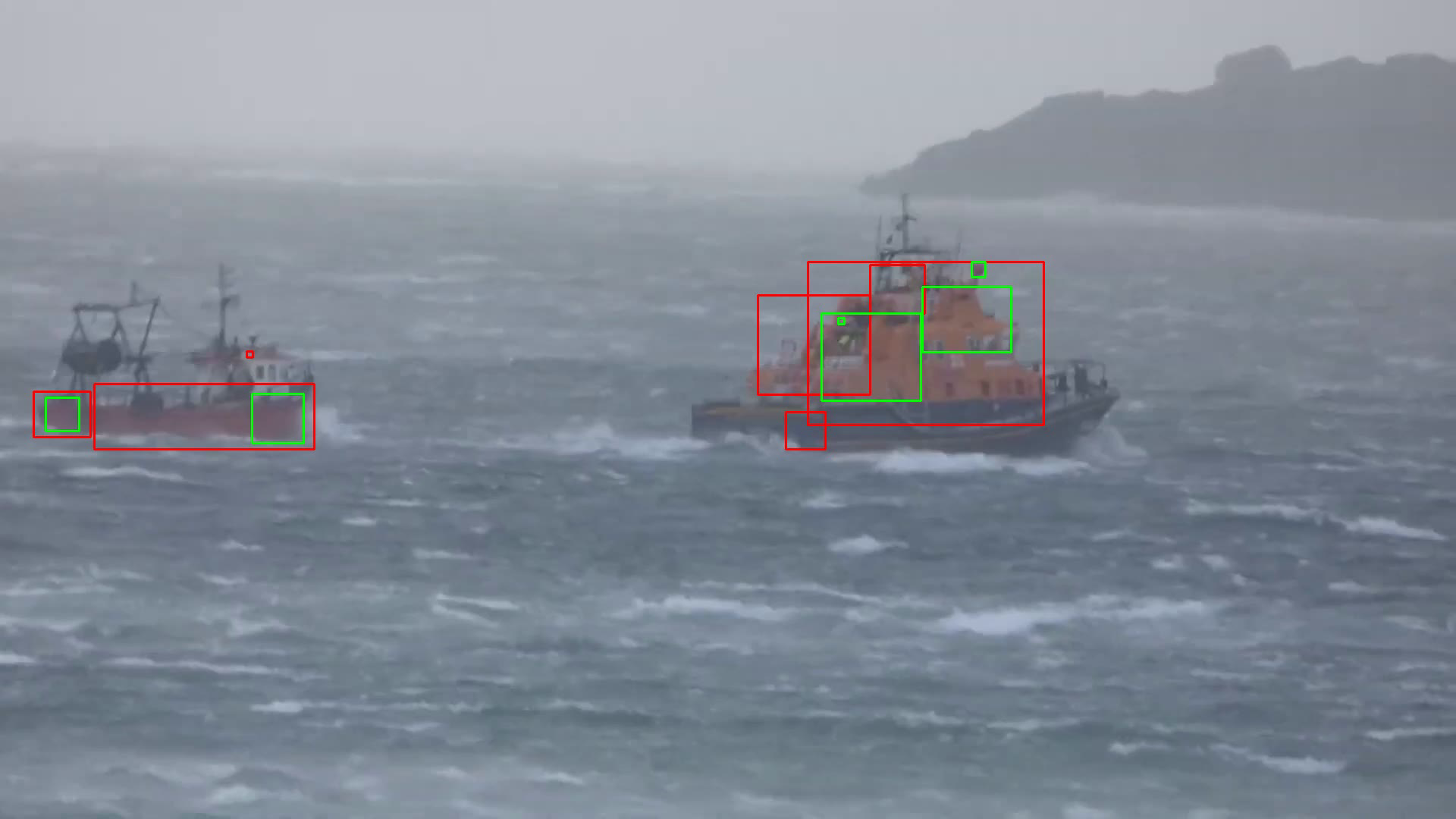}
\subfigure[Seq.C, Frame 41]{\label{fig4f}\includegraphics[height=15mm,width=42mm]{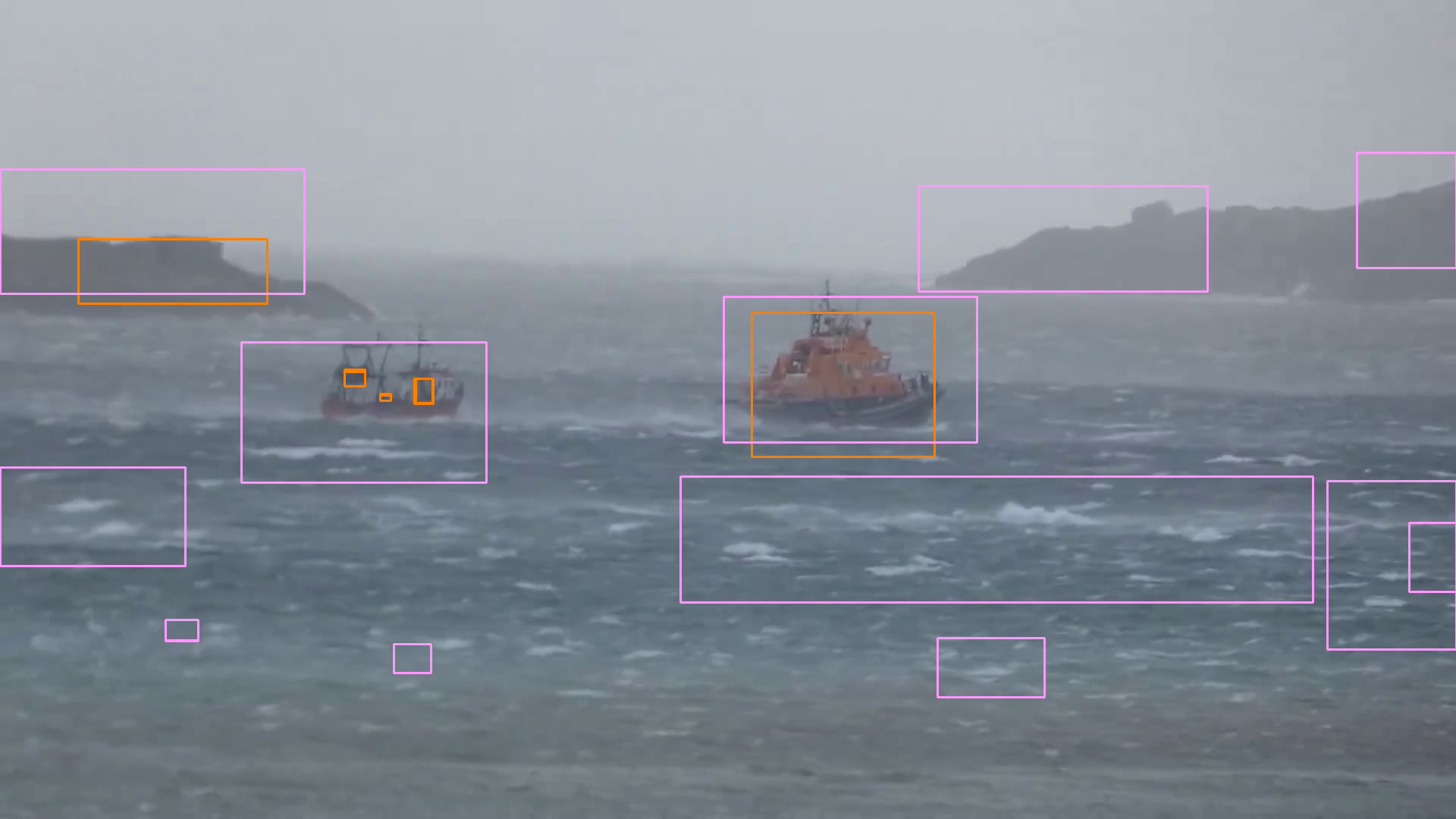}}
\subfigure[Seq.C, Frame 104]{\label{fig4f}\includegraphics[height=15mm,width=42mm]{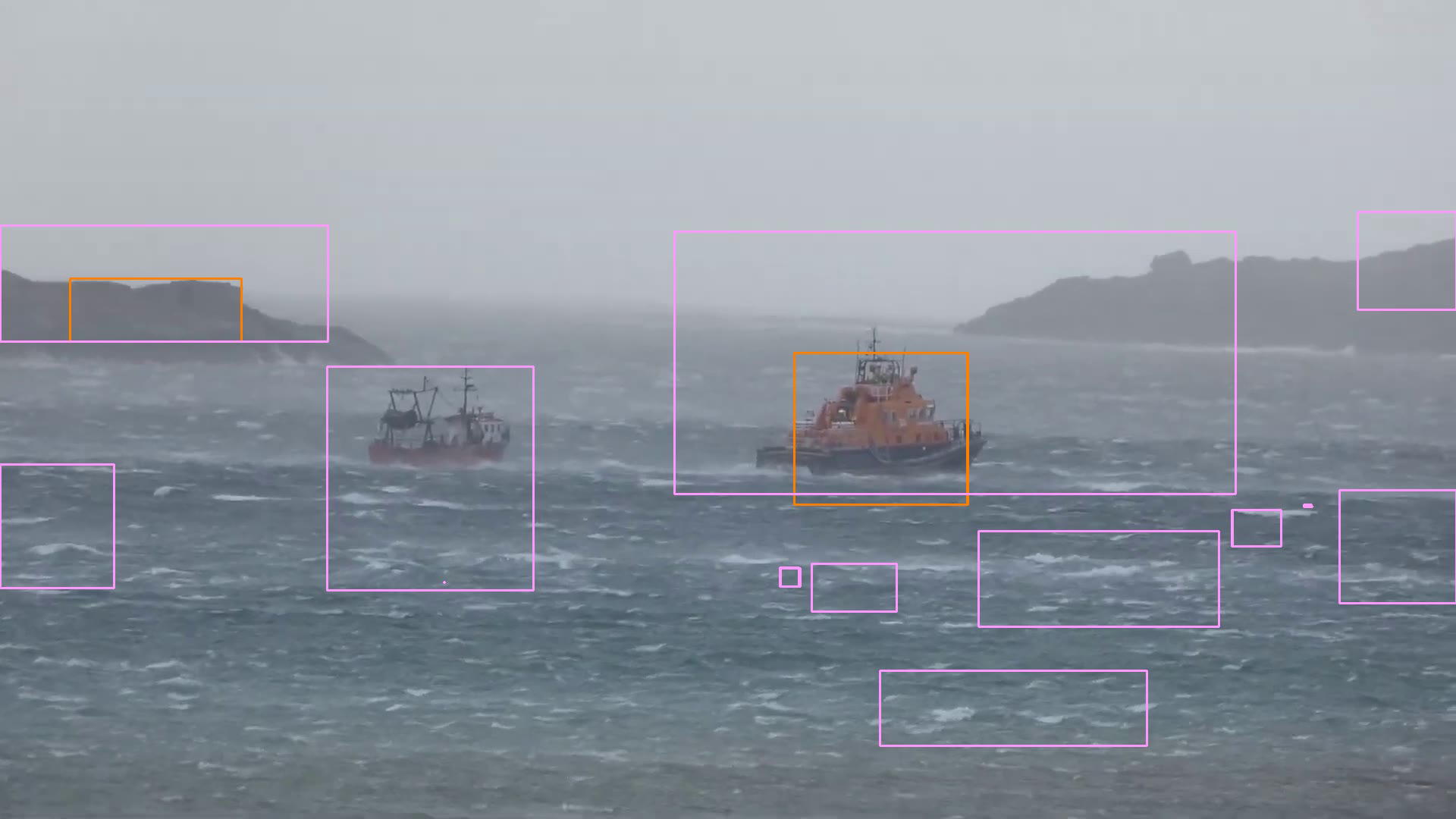}}
\subfigure[Seq.C, Frame 195]{\label{fig4f}\includegraphics[height=15mm,width=42mm]{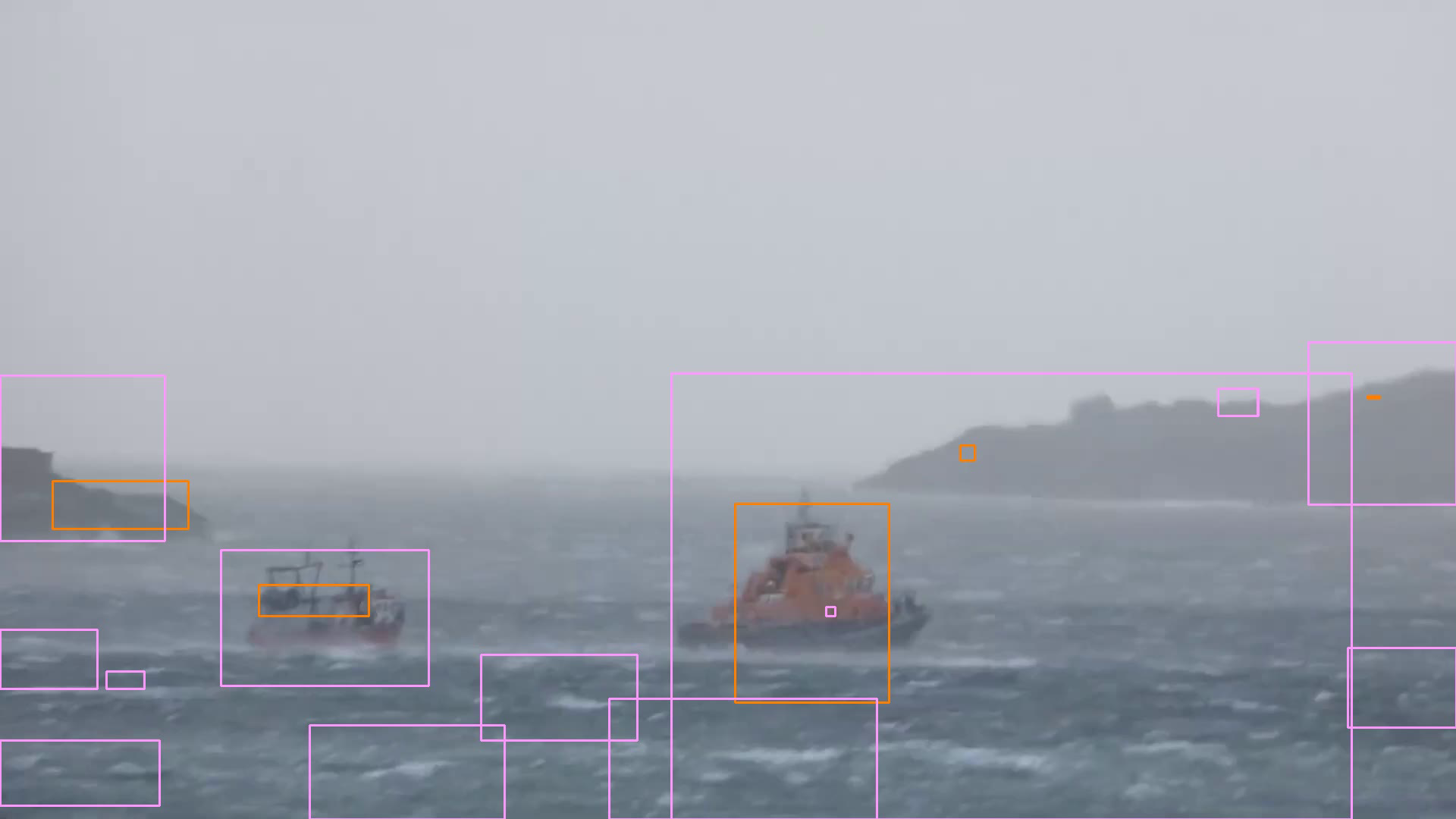}}
\subfigure[Seq.C, Frame 356]{\label{fig4f}\includegraphics[height=15mm,width=42mm]{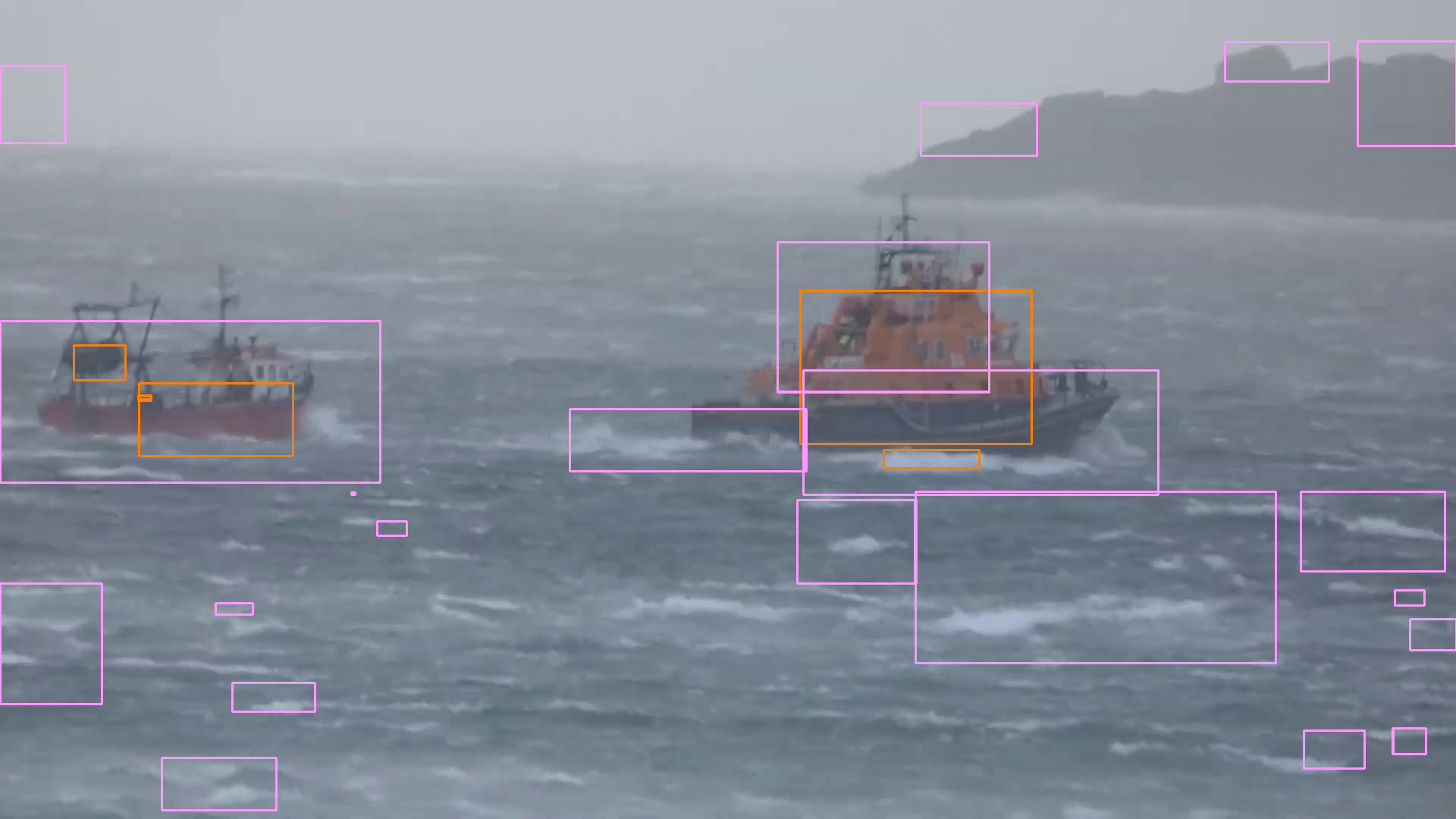}}

\includegraphics[height=15mm,width=42mm]{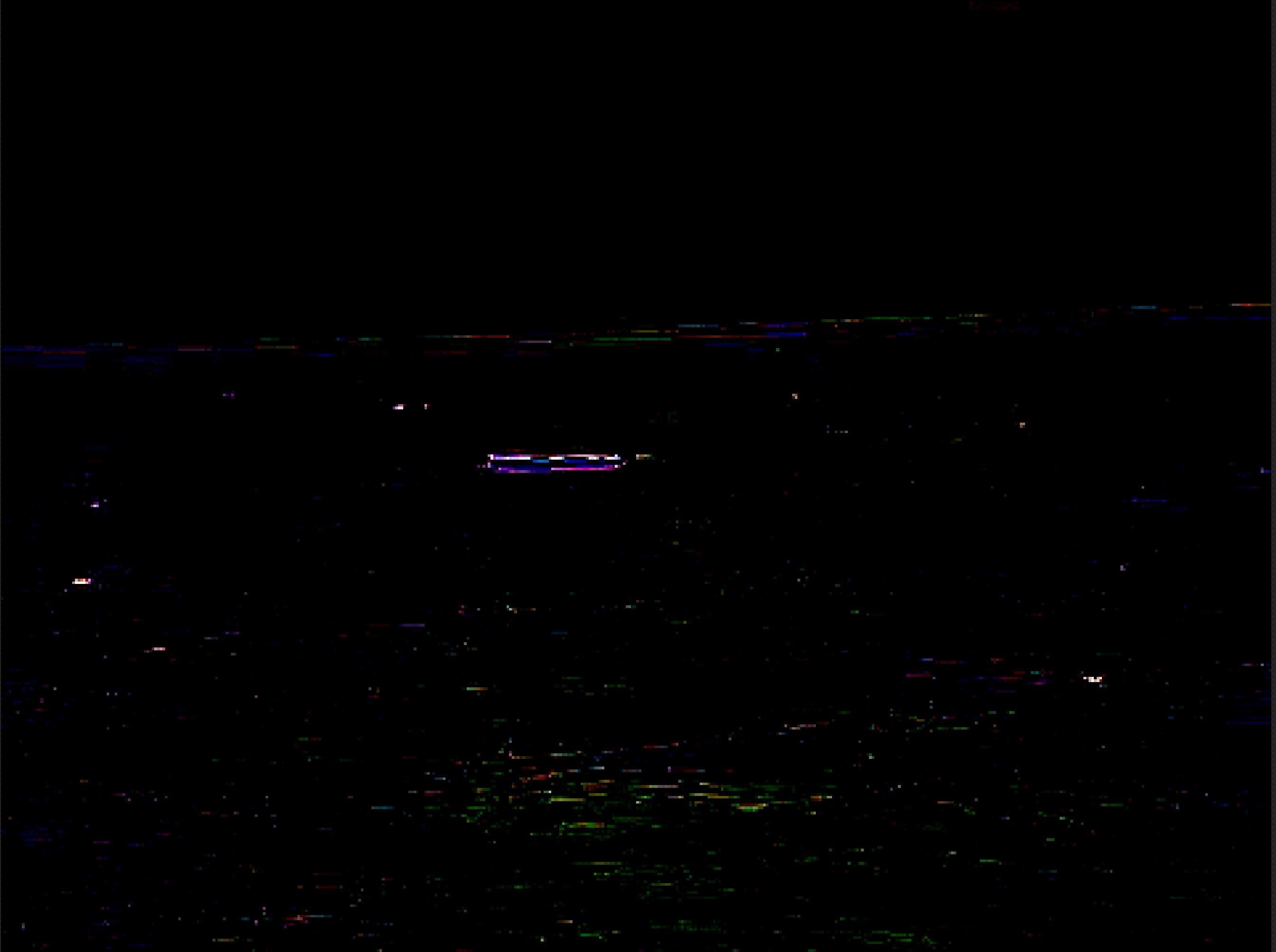}
\includegraphics[height=15mm,width=42mm]{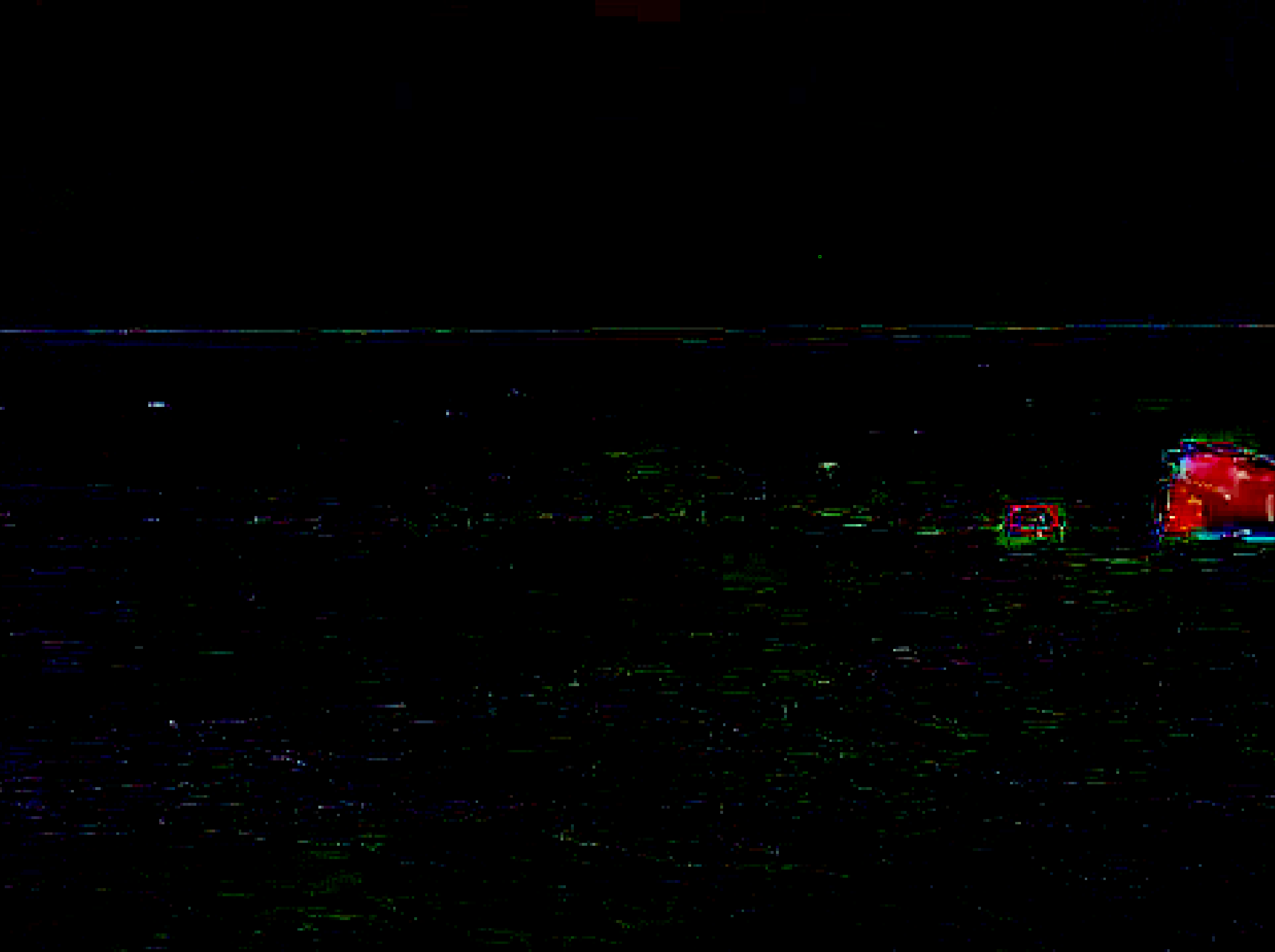}
\includegraphics[height=15mm,width=42mm]{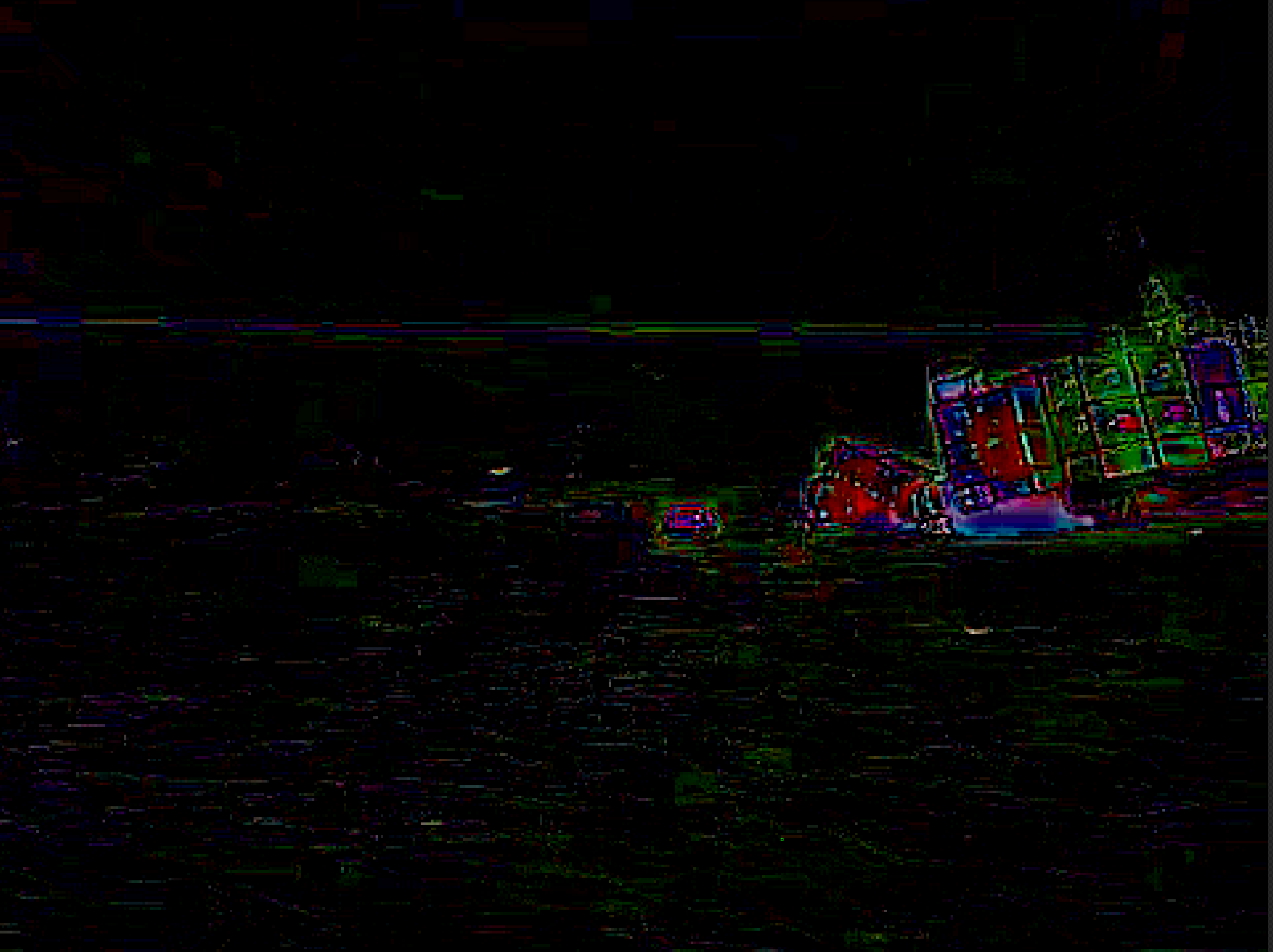}
\includegraphics[height=15mm,width=42mm]{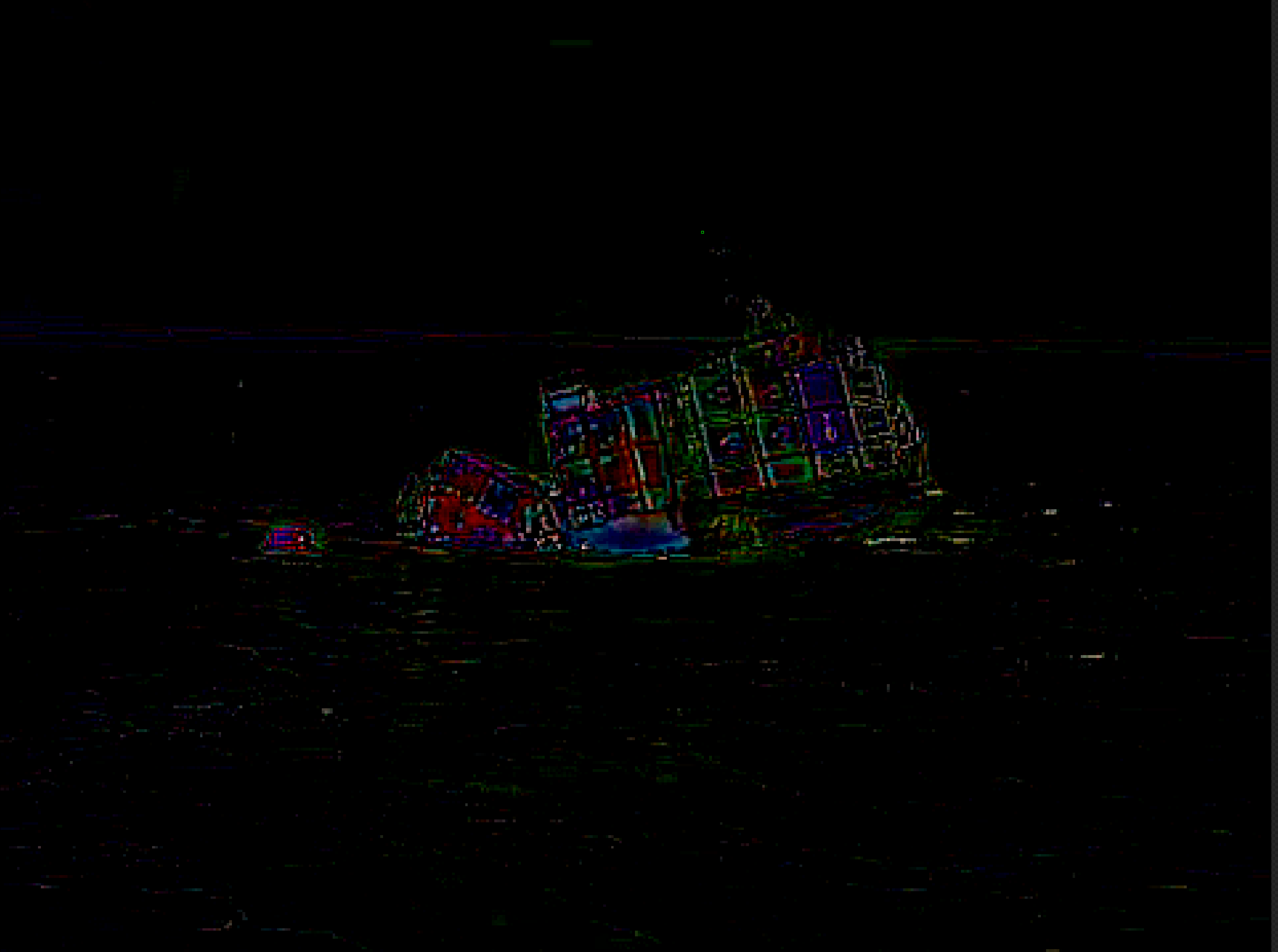}
\includegraphics[height=15mm,width=42mm]{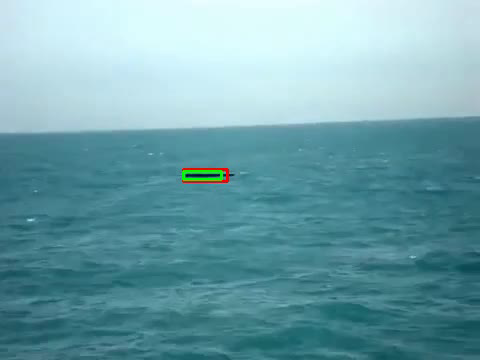}
\includegraphics[height=15mm,width=42mm]{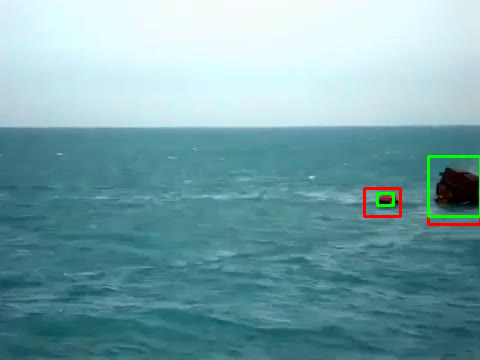}
\includegraphics[height=15mm,width=42mm]{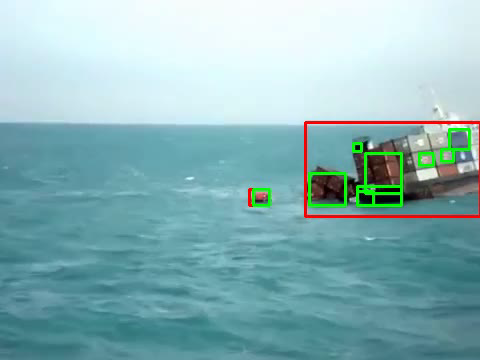}
\includegraphics[height=15mm,width=42mm]{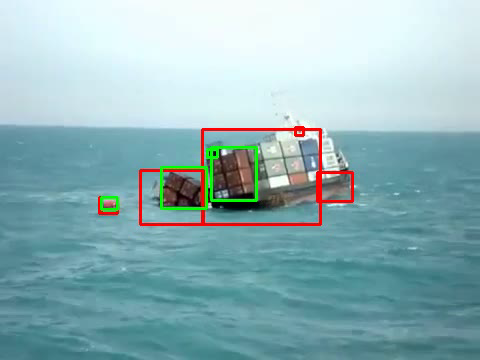}
\subfigure[Seq.D, Frame 16]{\label{fig4f}\includegraphics[height=15mm,width=42mm]{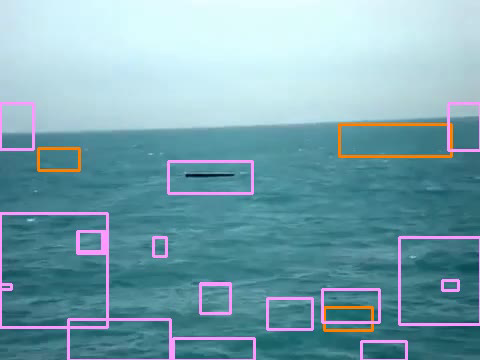}}
\subfigure[Seq.D, Frame 49]{\label{fig4f}\includegraphics[height=15mm,width=42mm]{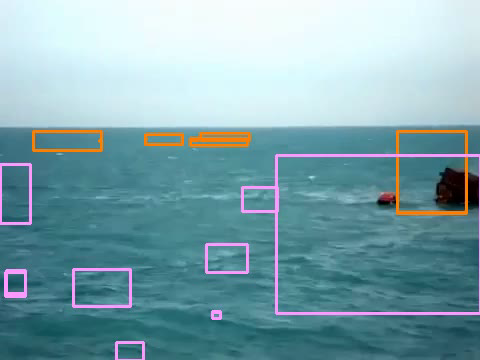}}
\subfigure[Seq.D, Frame 64]{\label{fig4f}\includegraphics[height=15mm,width=42mm]{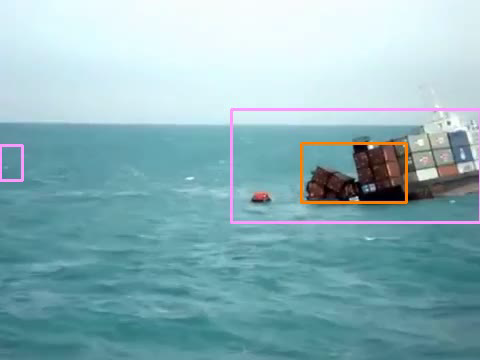}}
\subfigure[Seq.D, Frame 85]{\label{fig4f}\includegraphics[height=15mm,width=42mm]{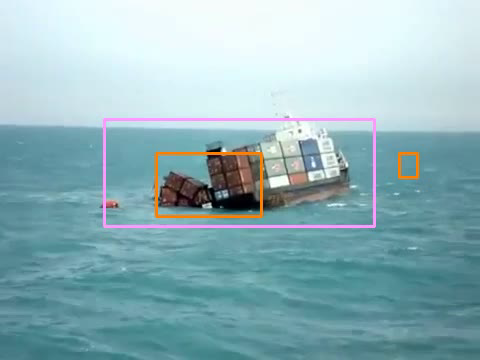}}

\caption{Qualitative results : Images in each row belongs to a single video sequence. The sequence number and frame numbers are indicated below each of the image. Top layer of each row represents the Residual image(contrast and brightness adjusted). Middle layer of each row indicates the detection made by our algorithm with logNFA=$2$ (Red BB) and logNFA=$-2$ (Green BB). Detections in the bottom layer of each row belongs to ITTI (Orange BB) and SRA (Pink BB).}
\label{fig_supp}
\end{figure*}

   The detections with logNFA = $2$ takes into account many weak detections, meaning many pixels are detected as part of salient object and this results in a red BB which is sometimes much bigger than the object itself. This can be seen in the middle layer of Fig. \ref{fig3p}, Fig. \ref{fig3q}, Fig. \ref{fig4a}, Fig. \ref{fig4e} etc. The detection with logNFA = $-2$ takes in to account only the strong detections, meaning small number of pixels are detected as part of salient object and this results in a green BB which are usually smaller than the salient object. An object can contain many of these small green BB, as seen in the middle layer of Fig. \ref{fig3a}, Fig. \ref{fig3b}, Fig. \ref{fig3g}, Fig. \ref{fig3h} etc. Thus by varying the logNFA, we can control the number of false alarms. Lower the logNFA the stronger and accurate the detection are. In some cases when the objects are occluded a single BB represents the occluded objects it can be seen in Fig. \ref{fig3p}, Fig. \ref{fig3g}, Fig. \ref{fig4k} etc . Both SRA and ITTI result in many false detections. In our experiments, varying the patch size and dictionary size increases the algorithm runtime with minor improvements in the detection results. Additional qualitative results on extra youtube video sequences can be found in Fig. \ref{fig_supp}.
   
  Quantitative evaluation is performed by calculating the True Positive TP (If a bounding box is present on the object), False Positive FP (If we detect an object when there is none) and False Negative FN (If we fail to detect the object). Using these three measures, we further calculate the Detection Rate ($DR = TP/(TP+FN)$) and False Alarm Rate ($FAR = FP/(TP+FP)$). Ideally, DR should be high, whereas the FAR should be as low as possible. Similar to \cite{FOBD_46}, for every frame in a sequence, we manually/visually check for the presence of bounding box on the object. If a BB is present on the object, then we increase the TP score by 1. If the BB is present on the background(sea, sky or land), we update the FP score by 1 and if the BB is not present on the object, we update the FN score by 1. Thus, for a given video sequence the final value of FP, FN and TP is obtained by accumulating scores across all the frames of the sequence.
   
   Table \ref{table2} tabulates the quantitative results of our experiments. From this table, it is evident that our algorithm with logNFA = $-2$ gives minimum FAR for 5 of the video sequences. As logNFA = $-2$ takes in to account only the strong detections, we can expect the FAR to be minimum, thus reducing the false alarms. This is also a reason for our algorithm with logNFA = $-2$ to have more FN compared to logNFA = $2$. For most of the video sequence, maximum DR is achieved by our algorithm with logNFA = $2$, as it takes in to account both strong and weak detections. Thus, our algorithm outperforms the other two algorithm for most of the video sequence.

Minor sun glint can be discarded using SIFT matching \cite{FOBD_35}. As, sun glint is not persistent across frames, only the objects are matched and sun glint fails to match and are rejected. In our experiments, the SIFT matching technique works well when the camera is static and fails when the camera is in motion. The SIFT matching also depends on the frame rate of the camera. If the frame rate is high, then the sun glint is persistent across many frames and can be matched, resulting in false detections. Another easy way to negate the effect of sun glint is to use polarizing lens on the camera.

\section{Conclusion and future work}
We have proposed a dictionary learning based unsupervised floating object detection algorithm specific to the maritime environment. The effectiveness of the approach was demonstrated on challenging video sequences exhibiting varying challenges of far sea maritime scenario, moving camera and small targets. The proposed algorithm exhibits good performance in detecting unidentified floating objects of varying size and shape. However, the algorithm has limited ability in the presence of strong sun glint. Future work will focus on temporal aspect, tracking of detected objects and real time (GPU) implementation of the proposed algorithm.

\end{document}